%% file: 00_main.tex
\PassOptionsToPackage{table}{xcolor}
\documentclass[]{hdsr}
\DeclareSymbolFont{matha}{OML}{txmi}{m}{it}
\DeclareMathSymbol{\varv}{\mathord}{matha}{118}
\graphicspath{{figures/}}

\newcommand{\etc}{etc\@ifnextchar.{}{.\@}}
\newcommand{\eg}{e.g.}
\newcommand{\ie}{i.e.}
\newcommand{\mathleft}{\@fleqntrue\@mathmargin0pt}

\definecolor{grey}{rgb}{0.5, 0.5, 0.5} 

\usepackage{upgreek}
\usepackage[table]{xcolor}
\usepackage{nicematrix}
\usepackage{tabularx, booktabs}
    \newcommand{\midsepremove}{\aboverulesep = 0.3mm \belowrulesep = 0.3mm}
    \midsepremove
    \newcommand{\midsepdefault}{\aboverulesep = 0.605mm \belowrulesep = 0.984mm}
    \midsepdefault
\usepackage{enumitem}
\usepackage{booktabs,arydshln}
\usepackage{soul}
\usepackage[normalem]{ulem}
\usepackage{threeparttable}
\usepackage{graphicx, caption}
\usepackage[font=scriptsize, labelfont = bf]{caption}
\usepackage{multirow, multicol}
\usepackage{mathtools}
\usepackage{amssymb}
\usepackage{bbding}
\usepackage{pifont}
\definecolor{mygreen}{rgb}{0.694, 0.816, 0.584}
\definecolor{citegreen}{rgb}{0.494, 0.856, 0.384}
\definecolor{myorange}{rgb}{0.882, 0.678, 0.522}
\definecolor{reforange}{rgb}{0.882, 0.5, 0.4}
\definecolor{mydarkgreen}{rgb}{0.369, 0.506, 0.247}
\definecolor{mydarkorange}{rgb}{0.722, 0.376, 0.161}
\newcommand{\cmark}{{\color{citegreen}\ding{51}}}%
\newcommand{\xmark}{{\color{reforange}\ding{55}}}%
\usepackage{stmaryrd}
\usepackage{tikz}
\usepackage[most]{tcolorbox}
\usepackage{hyperref}

\hypersetup{
    colorlinks=true,
    linkcolor=reforange,
    filecolor=magenta,      
    urlcolor=cyan,
}
\hypersetup{citecolor=citegreen}

\makeatletter
\newcommand{\myref}[1]{\textcolor{reforange}{\ref{#1}}}
\makeatother

\usepackage{amsthm}
\usepackage{dsfont}
\usepackage{wrapfig}
\usepackage{math}
\newtheorem{assumption}{Assumption}
\newtheorem{definition}{Definition}

\makeatletter
\def\adl@drawiv#1#2#3{%
        \hskip.5\tabcolsep
        \xleaders#3{#2.5\@tempdimb #1{1}#2.5\@tempdimb}%
                #2\z@ plus1fil minus1fil\relax
        \hskip.5\tabcolsep}
\newcommand{\cdashlinelr}[1]{%
  \noalign{\vskip\aboverulesep
           \global\let\@dashdrawstore\adl@draw
           \global\let\adl@draw\adl@drawiv}
  \cdashline{#1}
  \noalign{\global\let\adl@draw\@dashdrawstore
           \vskip\belowrulesep}}
\makeatother

\newtcolorbox{applebox}[1]{
  enhanced,
  colback=white,
  colframe=gray!20,
  fonttitle=\bfseries,
  coltitle=black,
  colbacktitle=white,
  left=6pt,
  right=6pt,
  top=2pt,
  bottom=2pt,
  boxsep=5pt,
  boxrule=1.5pt,
  arc=1.5mm,
  title=#1, %
  borderline={0pt}{0pt}{white},
  fonttitle=\small\bfseries,
  attach boxed title to top left={yshift=-2mm, xshift=3mm},
  boxed title style={boxrule=0pt, colback=gray!30, frame hidden, left=2pt, right=2pt},
}

\begin{document}

\newgeometry{bottom=1.5in}



\begin{center}

  \title{Unleashing the Power of Multi-Task Learning: A Comprehensive Survey Spanning Traditional, Deep, and Pretrained Foundation Model Eras}
  \renewcommand{\shortauthors}{Jun Yu et al.}
  \renewcommand{\shorttitle}{Unleashing the Power of MTL: A Comprehensive Survey Spanning Traditional, Deep, and PFM Eras}
  \maketitle

  \thispagestyle{empty}
  
  \vspace*{.1in}

  \begin{tabular}{cc}
    Jun Yu\upstairs{\affilone\affiltwo, $\dag$, $\ddag$}, Yutong Dai\upstairs{\affilthree}, Xiaokang Liu\upstairs{\affiltwo\affilfour}, Jin Huang\upstairs{\affilfive}, Yishan Shen\upstairs{\affiltwo}, Ke Zhang\upstairs{\affilsix},\\ Rong Zhou\upstairs{\affilone}, Eashan Adhikarla\upstairs{\affilone}, Wenxuan Ye\upstairs{\affilone}, Yixin Liu\upstairs{\affilone}, Zhaoming Kong\upstairs{\affilseven}, Kai Zhang\upstairs{\affilone}, \\ Yilong Yin\upstairs{\affilfive}, Vinod Namboodiri\upstairs{\affilone\affileight}, Brian D. Davison\upstairs{\affilone}, Jason H. Moore\upstairs{\affilnine}, Yong Chen\upstairs{\affiltwo, $\ddag$} \\[0.25ex]
   {\small \upstairs{\affilone} Department of Computer Science and Engineering, Lehigh University, USA} \\
   {\small \upstairs{\affiltwo} Department of Biostatistics, Epidemiology and Informatics, University of Pennsylvania, USA} \\
   {\small \upstairs{\affilthree} Department of Industrial and Systems Engineering, Lehigh University, USA} \\
   {\small \upstairs{\affilfour} Department of Statistics, University of Missouri, USA} \\
   {\small \upstairs{\affilfive} School of Software, Shandong University, China} \\
   {\small \upstairs{\affilsix} Department of Computer Science, University of Hong Kong, China} \\
   {\small \upstairs{\affilseven} Department of Computer Science and Engineering, South China University of Technology, China} \\
   {\small \upstairs{\affileight} Department of Community and Population Health, Lehigh University, USA} \\
   {\small \upstairs{\affilnine} Department of Computational Biomedicine, Cedars-Sinai Medical Center, USA}
  \end{tabular}

  \emails{
    \upstairs{$\dag$}This work includes efforts as a visiting student at Upenn. \\
    \upstairs{$\ddag$}Corresponding to juy220@lehigh.edu or ychen123@pennmedicine.upenn.edu.
    }
\input{tex_files/milestone}

\begin{abstract}
    Multi-Task Learning (MTL) is a learning paradigm that effectively leverages both task-specific and shared information to address multiple related tasks simultaneously. In contrast to Single-Task Learning (STL), MTL 
    offers a suite of benefits that enhance both the training process and the inference efficiency. MTL's key advantages encompass streamlined model architecture, performance enhancement, and cross-domain generalizability. Over the past twenty years, MTL has become widely recognized as a flexible and effective approach in various fields, including computer vision, natural language processing, recommendation systems, disease prognosis and diagnosis, and robotics. 
    This survey provides a comprehensive overview of the evolution of MTL, encompassing the technical aspects of cutting-edge methods from traditional approaches to deep learning and the latest trend of pretrained foundation models. Our survey methodically categorizes MTL techniques into five key areas: regularization, relationship learning, feature propagation, optimization, and pre-training. This categorization not only chronologically outlines the development of MTL but also dives into various specialized strategies within each category. Furthermore, the survey reveals how the MTL evolves from handling a fixed set of tasks to embracing a more flexible approach free from task or modality constraints. It explores the concepts of task-promptable and -agnostic training, along with the capacity for zero-shot learning, which unleashes the untapped potential of this historically coveted learning paradigm.
    Overall, we hope this survey provides the research community with a comprehensive overview of the advancements in MTL from its inception in 1997 to the present in 2023. We address present challenges and look ahead to future possibilities, shedding light on the opportunities and potential avenues for MTL research in a broad manner. This project is publicly available at \href{https://github.com/junfish/Awesome-Multitask-Learning}{https://github.com/junfish/Awesome-Multitask-Learning}.
\end{abstract}
\end{center}

\vspace*{0.15in}
\hspace{10pt}

\copyrightnotice
\restoregeometry
\newgeometry{bottom=0.5in}

\clearpage
\input{01_intro}

\input{02-1_method_traditional}

\input{02-2_method_deep}

\input{02-3_method_pfm}

\input{03_miscellaneous}

\input{04-resource}

\input{05_discussion}

\input{06_conclusion}

\subsection*{Disclosure Statement}
The authors have no conflicts of interest to declare.

\input{07_contribution}
 

\printbibliography

\end{document}

%% file: tex_files/milestone.tex
\begin{figure}[h]
    \centering
    \includegraphics[width=0.95\linewidth]{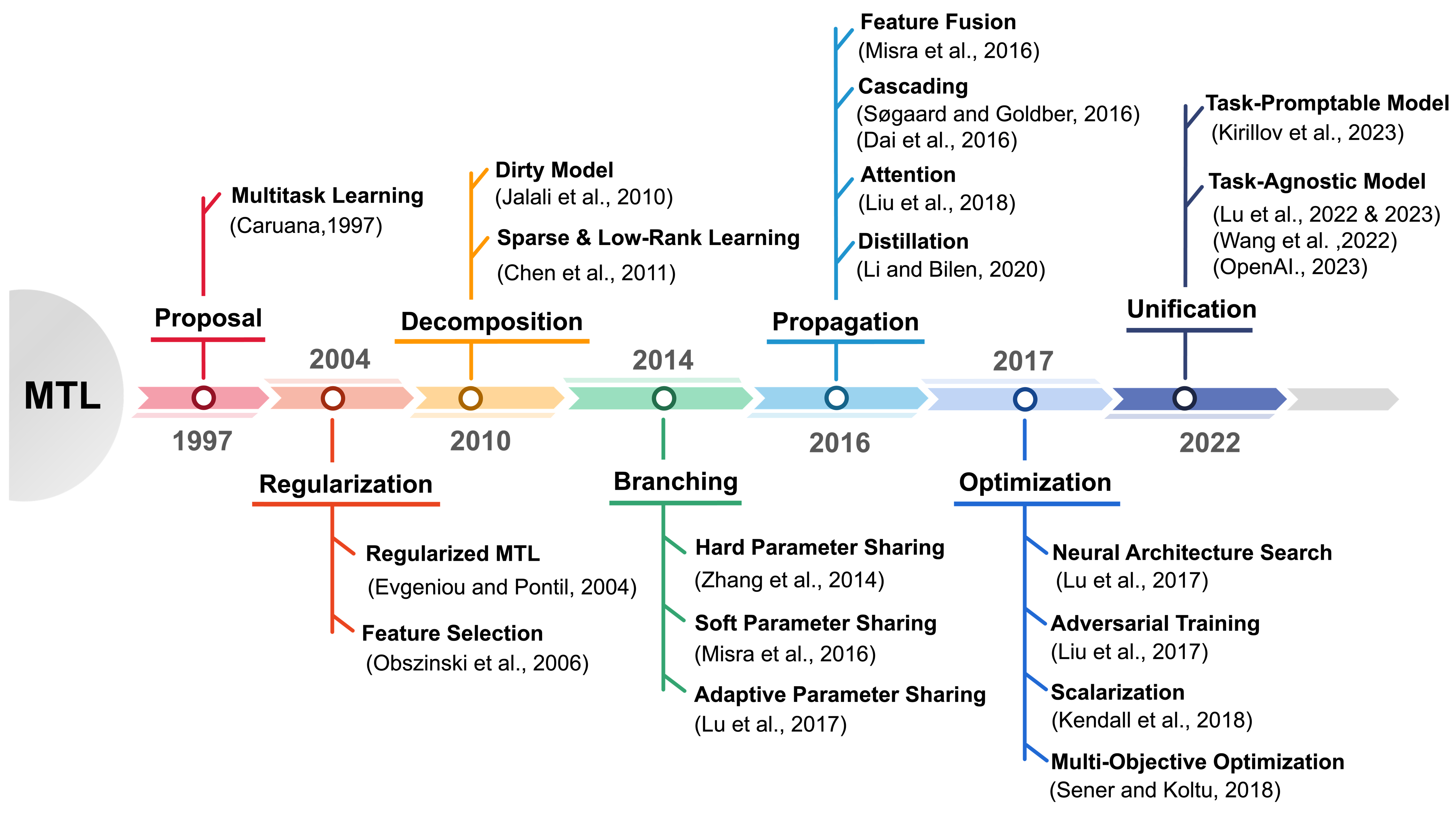}
      \caption{Significant landmarks in the evolution of Multi-Task Learning (MTL) highlighted over time.}
    \label{fig:milestone}
\end{figure}

%% file: 01_intro.tex
\input{tex_files/01_intro/outline}
\clearpage

\section{Introduction}
In the introduction, we hope to answer the following five research questions (RQs) before we overview the methodologies of Multi-task Learning (MTL):
\begin{itemize}
    \item \emph{\textbf{RQ1}}: What is the concept and definition of MTL? (See \textbf{\S}~\myref{def})
    \item \emph{\textbf{RQ2}}: How does MTL distinguish itself from other learning paradigms? (See \textbf{\S}~\ref{related})
    \item \emph{\textbf{RQ3}}: What motivates the use of MTL in learning scenarios? (See \textbf{\S}~\ref{motiv})
    \item \emph{\textbf{RQ4}}: What underlying principles does the efficacy of MTL rest on? (See \textbf{\S}~\ref{mecha})
    \item \emph{\textbf{RQ5}}: In what ways does our survey differentiate from previous studies? (See \textbf{\S}~\ref{exist})
    
\end{itemize}
In \textbf{\S}~\ref{def}, we progressively introduce Multi-Task Learning (MTL), starting with a broad sense and culminating in a formal definition. Subsequently, \textbf{\S}~\ref{related} explores the position of MTL within the Machine Learning (ML) landscape, drawing comparisons with related paradigms such as Transfer Learning (TL), Few-Shot Learning (FSL), lifelong learning, Multi-View Learning (MVL), to name a few. \textbf{\S}~\ref{motiv} delves into the motivations for employing MTL, offering insights from both explicit and subtle angles, while also addressing how MTL benefits the involved tasks. In \textbf{\S}~\ref{mecha}, we delve deeper into the fundamental mechanisms and theories underpinning MTL, specifically: 1) regularization, 2) inductive bias, and 3) feature sharing, providing an understanding of its underlying principles. Finally, \textbf{\S}~\myref{exist} reviews existing surveys on MTL, underscoring the unique contributions of our survey and laying out a structured roadmap for the remainder of this work. The structure of our survey is depicted in \textbf{Fig.}~\ref{fig:outline}. 
Before delving into this survey, readers can quickly refer to Table~\ref{tab:abbrev} for a list of acronyms not related to datasets, institutions, and newly proposed methods, while an overview of mathematical notations is provided in Table~\ref{tab:notation} and Table~\ref{tab:deep_notation}.

\input{table_files/abbrev}

\subsection{Definition}
\label{def}
{~~}\\
\vspace{-2pt}
\input{tex_files/01_intro/paper_num} 
The increasing popularity of MTL over the past few decades is evident in \textbf{Fig.}~\ref{fig:paper-num}, which displays the trend in the number of papers associated with \textit{``allintitle: `multitask learning' OR `multi-task learning' ''} as a keyword search, according to data from Google Scholar\footnote{\url{https://scholar.google.com}}.

As the name suggests, MTL is a subfield of ML where multiple tasks are jointly learned. In this manner, we hope to leverage useful information across these related tasks and break from the tradition of performing different tasks in isolation. In Single-Task Learning (STL), data specific to the task at hand is the only source to couch a learner. However, MTL can conveniently transfer extra knowledge learned from other tasks. The essence of MTL is to exploit consensual and complementary information among tasks by combining data resources and sharing knowledge. This sheds light on a better learning paradigm that can reduce memory burden and data consumption, and improve training speed and testing performance. For instance, learning the monocular depth estimation (scaling the distance to the camera)~\citep{eigen2014depth} and semantic segmentation (assigning a class label to every pixel value)~\citep{fu1981survey} simultaneously in images is beneficial since both tasks need to perceive meaningful objects. MTL has become increasingly ubiquitous as experimental and theoretical analyses continue to validate its promising results. For example, using Face ID to unlock an iPhone is a typical but imperceptible MTL application that involves simultaneously locating the user's face and identifying the user. In general, multitasking occurs when we attempt to handle two or more objectives during the optimization stage in practice. 

Consequently, MTL exists everywhere in ML, even when performing STL with regularization. This can be understood as having one target task and an additional artificial task of human preference, such as learning a constrained model via $\ell_2$ regularizer or a parsimonious model via $\ell_1$ regularizer. These hypothesis preferences can serve as an inductive bias to enhance an inductive learner~\citep{caruna1993multitask}. In the early exploration of MTL~\citep{caruana1997multitask}, the extra information that the involved tasks provide is regarded as a domain-specific inductive bias for the other tasks. Since collecting training signals from other tasks is more practical than acquiring inductive bias from model design or human expertise, we can thus empower any ML models via this MTL paradigm. 


\subsubsection{Formal Definition}
\label{prob-define}
To comprehensively understand MTL, we provide a formal definition of MTL. Suppose we have a sample dataset $\Xbold$ drawn from the feature space $\Xmcal$, and its respective ground-truth label set $\Ybold$ drawn from the label space $\Ymcal$. 
We can define \emph{experience} $\Emcal \subseteq \{\Xbold, \Ybold\}$, \emph{domain} $\Dmcal = (\Xmcal, P(\Xbold))$, and \emph{task} $\Tmcal = (\Ymcal, f)$, where $P(\Xbold)$ is the distribution of $\Xbold$ and $f$ maps a data sample $\xbold\in\Xbold$ to a prediction $\tilde{\ybold}\in\Ybold$. These predictive values consist of the predictive label set $\tilde{\Ybold}=\{\tilde{\ybold}|\tilde{\ybold} = f(\xbold), \xbold\in\Xbold\}$. Following the ML settings, we should define a \emph{measurement} $\Pmcal = (\Ybold, \tilde{\Ybold}, \Lmcal)$, where $\Lmcal$ is a function to measure the distance between any pairs of $(\ybold, \tilde{\ybold})$. More basic notations please refer to Table~\ref{tab:notation}.
Based on the definitions of four basic elements (\emph{experience}, \emph{domain}, \emph{task}, and \emph{measurement}) above, we first restate the general definition of machine learning by~\citet{mitchell1997machine} to a more exact form as follows.

\begin{definition}[Machine Learning, \citet{mitchell1997machine}]
\label{ml_define}
A computer program is said to learn from experience $\Emcal$ with respect to a set of tasks
$\{\Tmcal^{(t)}\}_{t=1}^{T}$ and performance measurement $\Pmcal$, if its performance at tasks $\{\Tmcal^{(t)}\}_{t=1}^{T}$, as measured by
$\Pmcal$, improves with experience $\Emcal$.
\end{definition}

The definition above inherently considers both single-task and multi-task scenarios during the ML process but deviates from a meticulous definition to characterize MTL that includes recent developments. Now, let us first define STL to induce the formal definition of MTL.

\begin{definition}[Single-Task Learning]
A type of machine learning specified by $\Emcal, \{\Tmcal^{(t)}\}_{t=1}^{T}$ and $\Pmcal$, where $\{\Tmcal^{(t)}\}_{t=1}^{T}$ contains only one task (i.e. $T=1$) on a specific domain $\Dmcal$.
\end{definition}

As recent developments in MTL focus more on heterogeneous tasks (e.g., regression $+$ classification) than homogeneous ones, each task should be represented by its own \emph{experience} $\Emcal$ on its corresponding \emph{domain} $\Dmcal$. Due to this diversity, we always employ distinct \emph{measurement} $\Pmcal$ to evaluate the learning performance of each task. We accordingly define the MTL as follows.

\begin{definition}[Multi-Task Learning]
A super set of STL specified by $\bigcup_{t=1}^{T}\Emcal^{(t)}, \{\Tmcal^{(t)}\}_{t=1}^{T}$ and $\{\Pmcal^{(t)}\}_{t=1}^{T}$, where experience $\Emcal^{(t)}\subseteq\{\Xbold^{(t)}, \Ybold^{(t)}\}$ is with respect to task $\Tmcal^{(t)}$ on its corresponding domain $\Dmcal^{(t)}$. Accordingly, MTL is a computer program to learn from the experience set $\bigcup_{t=1}^{T}\Emcal^{(t)}$ with respect to the task set $\{\Tmcal^{(t)}\}_{t=1}^{T}$ and the corresponding performance measurement set $\{\Pmcal^{(t)}\}_{t=1}^{T}$, if its total performance at any task $\Tmcal^{(t)}$, as measured by its corresponding $\Pmcal^{(t)}$, $t = 1,\cdots,T$, improves with experience set $\bigcup_{t=1}^{T}\Emcal^{(t)}$.
\label{def:MTL}
\end{definition}
We note that the formal MTL definition above has no conflict with the homogeneous or heterogeneous MTL.

\subsection{Related Fields}
\label{related}
{~~}\vspace{1pt}\\

Having established a formal definition of MTL grounded in fundamental ML elements, a thorough understanding can be achieved by analytically comparing it with related domains. These include Transfer Learning (TL), Meta-Learning, and In-Context Learning (ICL), among others. This comparison not only clarifies the distinct characteristics of MTL but also situates it within the broader context of these interconnected fields.

\paragraph{\emph{Transfer Learning (TL)}} TL~\citep{pan2009survey} is a prevalent learning paradigm that solves the problem of lacking labeled data when applying ML to real-world data~\citep{zhuang2020comprehensive,pan2009survey}. Specifically, TL improves the performance of a target model on target domains by transferring the knowledge in different but related source domains to the target domains. 
Such properties make TL well-appreciated in real-world applications, such as healthcare~\citep{kao2021toward,song2021transfer,perez2021transfer} and recommender systems~\citep{tl_recom_www21,liu2021leveraging,tl_recom_cikm21}. 
According to the availability of labels in the source and target domains, TL is categorized into three types, \ie, transductive TL (aka \textit{Domain Adaptation (DA)},~\citet{redko2019advances,patel2015visual}), inductive TL, and unsupervised TL~\citep{zhuang2020comprehensive,pan2009survey}. \\



\paragraph{\emph{Few-Shot Learning (FSL)}} FSL~\citep{fink2004object, fei2006one, wang2020generalizing} is a specific application case of TL. It aims at obtaining a model for the target task under a certain scenario where limited labeled samples from the target domain are available~\citep{wang2020generalizing}. FSL is well-acknowledged in tackling different real-world problems such as identifying atypical ailments~\citep{quellec2020automatic,jia2020few}, visual navigation~\citep{al2022zero,luo2021few}, and cold-start item recommendation~\citep{sun2021mfnp,zhang2021model}.\\

\paragraph{\emph{Meta-Learning}} Meta-Learning~\citep{hospedales2021meta} is an implementation approach to achieve TL. The main concept is to obtain a meta-learner (a model) that can have satisfying performance for an unseen target domain~\citep{hospedales2021meta}. Such meta-learner first extracts the meta-knowledge, \ie, the universally applicable principles, across source domains. With meta-knowledge, the meta-learner can be easily generalized to the target domain by leveraging the target samples. Meta-learning has been successfully applied in various problems such as hyper-parameter optimization~\citep{bohdal2021evograd,raghu2021meta}, algorithm selection for data mining~\citep{simchowitz2021bayesian}, and neural architecture search (NAS)~\citep{lee2021hardware,ding2022learning}.

Though TL paradigms, including FSL and meta-learning, involve multi-domain data, their ultimate goal is to obtain a model with satisfied performance or can be easily generalized to one target task. In other words, TL leverages the knowledge in different tasks to assist the model in learning a single task, which intersects with MTL according to our definition in Definition~\myref{def:MTL}. Thus, TL can bring merits to MTL, such as capturing the relations among tasks and extracting shared knowledge among involved tasks. Notably, the transfer of knowledge from pretrained foundation models (PFMs) proves beneficial for a myriad of downstream tasks in recent advancements~\citep{bommasani2021opportunities, zhou2023comprehensive}.\\

\paragraph{\emph{Lifelong Learning}} Lifelong Learning~\citep{parisi2019continual}, aka Continual Learning, Sequential Learning, or Incremental Learning, studies the problem of learning from an infinite stream of data~\citep{de2021continual}. The goal is to gradually extend the acquired knowledge and use it for future data, mitigating the occurrence of catastrophic forgetting or interference~\citep{mcclelland1995there}. With only a small portion of the input data from one or few tasks available at once, lifelong learning particularly tends to preserve the knowledge learned from the previous input when learning on new data, \ie, addressing the stability-plasticity dilemma~\citep{grossberg2012studies}. There are extensive applications of lifelong learning in solving tasks in ever-evolving systems, such as recommendations~\citep{chen2021towards,yao2021device} and anomaly detection~\citep{peng2021lime,doshi2022rethinking}.
Lifelong learning differs from MTL in the sense that its training object is a dynamic data stream, while MTL studies data from multiple tasks available at the beginning of the learning process.\\

\paragraph{\emph{Multi-View Learning (MVL)}} MVL~\citep{xu2013survey, zhao2017multi, li2018survey} studies the problem of jointly learning from multi-view data samples, whose goal is to optimize the generalization performance for the jointly learning model~\citep{li2018survey}. In real-world applications, the multi-view data indicates objects being described by multi-modal measurements, such as image+text, audio+video, and audio+articulation. Multi-Instance Multi-Label learning (MIML)~\citep{zhou2012multi} is a specific subtype of MVL, where an example is described by multiple instances and associated with multiple class labels. Due to the vast existence of multi-view data in realistic, MVL has attracted much attention in both research and industry, and the respective solutions play essential roles in cross-media retrieval~\citep{zhen2019deep, huang2020forward}, video analysis~\citep{wang2022cascade,zellers2021merlot}, recommender system~\citep{wei2022contrastive,chai2022knowledge}, etc. MVL, including MIML, can be considered a specialized form of MTL, where the input contains data from multiple domains that are handled as distinct tasks, but the output is still in one label space.\\

\paragraph{\emph{In-Context Learning (ICL)}} ICL~\citep{dong2022survey} has aroused interest as a novel learning paradigm for natural language processing (NLP) within Large Language Models (LLMs). ICL relies on templates in natural language that can demonstrate different tasks, such as solving mathematical reasoning problems~\citep{wei2022chain} and learning natural language inference (NLI)~\citep{liu2021natural}. LLMs can then make predictions by taking this demonstration and its corresponding query pair as input. While both ICL and MTL involve leveraging shared knowledge or context to enhance task generalizability, ICL is specifically tailored to the target task within a narrower scope in real-world applications. However, recent large PFMs, like GPT-4~\citep{openai2023gpt4}, are inherently task-agnostic, accommodating various tasks owing to the diversity of demonstration templates encountered during their large-scale training stage.
\subsection{Motivation and Benefit}
\label{motiv}
{~~}\vspace{2pt}\\
MTL can be motivated from the following five perspectives with different benefits: cognitive/social psychology, data augmentation, learning efficacy, real-world scenarios, and learning theory.
\begin{itemize}
    \item Psychologically, humans are inherent with flexible adaptability to new problems and settings, as the human learning process can transfer knowledge from one experience to another~\citep{national2000people}.
    Therefore, MTL is inspired by simulating this process to empower a model with the potentiality of multitasking. 
    Coincidentally, another example of this knowledge transfer happens among organizations~\citep{argote2000knowledge}. It is proved that organizations with more effective knowledge transfer are more productive and likely to survive than those with less. 
    These prior successes of transfers or mutualizations in other areas encourage the joint learning of tasks in ML~\citep{caruana1997multitask}.
    \item In the pre-big data era, real-world problems were usually represented by small but high-dimensional datasets ($\#$~samples$<\#$~features). This data bottleneck forces early methods to learn a sparse-structured model, which always leads to a parsimonious solution to a problem with insufficient data.
    However, the MTL emerged to aggregate labeled data from different domains or tasks to enlarge the training dataset against overfitting. 
    \item The pursuit of efficiency and effectiveness is also one of the motivations. MTL can aggregate data from different sources together, and the joint training process of multiple tasks can save both computation and storage resources. In addition, the potential of performance enhancement makes it popular in research communities. In brief, universal representations for any tasks can be learned from multi-source data, and benefit all tasks in terms of both the learning cost and performance.
    \item Motivated by the majority of real-world problems naturally being multimodal or multitasking, MTL is proposed to remedy the suboptimal achieved by STL that only models parts of the whole problem separately. 
    For example, predicting the progression of Alzheimer's Disease (AD) biomarkers for Mild Cognitive Impairment (MCI) risk and clinical diagnosis is simultaneously based on multimodal data such as computed tomography (CT), Magnetic Resonance Imaging (MRI), and Positron Emission Tomography (PET)~\citep{jie2015manifold, kwak2018multi, chen2022machine}. 
    Autonomous driving, another example, also involves multiple subtasks to calculate the final prediction~\citep{yang2018end, chowdhuri2019multinet}, including the recognition of surrounding objects, adjustments to the fastest route according to the traffic conditions, the balance between efficiency and safety, etc.
    \item From the perspective of learning theory, bias-free learning is proved to be impossible~\citep{mitchell1980need}, so we can motivate the MTL by using the extra training signals for related tasks. Generally, MTL is one of the ways to achieve inductive transfer via multitasking assistance, which improves both learning speed and generalization. 
    Specifically, during the process of the combined training of multiple tasks, some tasks can be provided inductive bias from other related tasks, and these stronger inductive biases (compared with universal regularizers, e.g., $\ell_2$) enable the knowledge transfer and yield more generalization abilities on a fixed training dataset.
    In other words, task-related biases make a learner prefer hypotheses that can explain more than one task and prevent specific task from overfitting.
\end{itemize}

\subsection{Mechanism and Explanation}
\label{mecha}
{~~}\vspace{2pt}\\
In this section, we explore three key mechanisms -- regularization, inductive bias, and feature sharing -- shedding light on how MTL operates to achieve enhanced performance across multiple tasks. 
\paragraph{\emph{Regularization}}
In MTL, the total loss function is a combination of multiple loss terms with respect to each task. The related tasks play a role as regularizers, enhancing the generalizability across them. The hypothesis space of an MTL model is confined to a more limited scope as it tackles multiple tasks simultaneously. Consequently, this constraint on the hypothesis space reduces model complexity, mitigating the risk of overfitting.

\paragraph{\emph{Inductive Bias}}
The training signals from co-training tasks act as mutual inductive biases due to their shared domain information. These biases facilitate cross-task knowledge transfer during training, guiding the model to favor task-related concepts rather than the tasks themselves. Consequently, this broadens the model's horizons beyond a singular task, enhancing its generalization capabilities for unseen out-of-distribution (OOD) data.

\paragraph{\emph{Feature Sharing}} 
MTL can enable feature sharing across related tasks. One approach involves selecting overlapping features and maximizing their utility across all tasks. This is referred to as ``eavesdropping''~\citep{ruder2017overview}, considering that some features may be unavailable for specific tasks but can be substituted by that learned from related tasks. Another way is to concatenate all the features extracted by different tasks together; these features can be holistically used across tasks via linear combination or nonlinear transformation. 

Overall, MTL can be an efficient and effective way to boost the performance of the ML model on multiple tasks by regularization, inductive transfer, and feature sharing.


\subsection{Contributions and Highlights}
\label{exist}
{~~}\vspace{7pt}\\
\emph{Existing Surveys.}
\citet{ruder2017overview} is a pioneering survey in MTL, offering a broad overview of MTL and focusing on advances in deep neural networks from 2015 to 2017.~\citet{thung2018brief} reviews MTL methods from a taxonomy perspective of input-output variants, mainly concentrating on traditional MTL prior to 2016. These two reviews can be complementary materials to each other. \citet{vafaeikia2020brief} is an incomplete survey that briefly reviews recent deep MTL approaches, particularly focusing on the selection of auxiliary tasks for enhanced learning performance.~\citet{crawshaw2020multi} presents the well-established and advanced MTL methods before 2020 from the perspective of applications. \citet{vandenhende2021multi} 
provides a comprehensive review of deep MTL in dense prediction tasks, which generate pixel-level predictions such as in semantic segmentation and monocular depth estimation. \citet{zhang2021survey} first give a comprehensive overview of MTL models from the taxonomy of feature-based and parameter-based approaches, but with limited inclusion of deep learning (DL) methods. Notably, all these surveys overlook the development of MTL in the last three or four years, named the era of large PFMs~\citep{bommasani2021opportunities, zhou2023comprehensive}, exemplified by the GPT-series models~\citep{radford2018improving, radford2019language, brown2020language, openai2023gpt4}.

\emph{Roadmap.} This survey adopts a well-organized structure, distinguishing it from its predecessors, to demonstrate the evolutionary journey of MTL from traditional methods to DL and the innovative paradigm shift introduced by PFMs, as shown in \textbf{Fig.~\ref{fig:milestone}}. In \textbf{{\S}~\ref{traditional-era}}, we provide a comprehensive summary of traditional MTL techniques, including feature selection, feature transformation, decomposition, low-rank factorization, priori sharing, and task clustering. 
Moving forward, \textbf{{\S}~\myref{deep-era}} is devoted to exploring the critical dimensions of deep MTL methodologies, encompassing feature fusion, cascading, knowledge distillation, cross-task attention, scalarization, multi-objective optimization (MOO), adversarial training, Mixture-of-Experts (MoE), graph-based methods, and NAS. The recent advancements in PFMs are introduced in \textbf{{\S}~\myref{fm-era}}, categorized based on task-generalizable fine-tuning, task promptable engineering, as well as task-agnostic unification. Additionally, we provide a concise overview of the miscellaneous aspects of MTL in \textbf{{\S}~\ref{sec:miscellaneous}}. \textbf{{\S}~\ref{benchmark}} provides valuable resources and tools to enhance the engagement of researchers and practitioners with MTL. 
Our discussions and future directions are presented in \textbf{{\S}~\ref{sec:discuss}}, followed by our conclusion in \textbf{{\S}~\ref{sec:conclud}}.
The goal of this review is threefold: 1) to provide a comprehensive understanding of MTL for newcomers; 2) to function as a toolbox or handbook for engineering practitioners; and 3) to inspire experts by providing insights into the future directions and potentials of MTL.



%% file: tex_files/01_intro/outline.tex
\begin{figure}[t]
    \centering
    \includegraphics[width=0.85\linewidth]{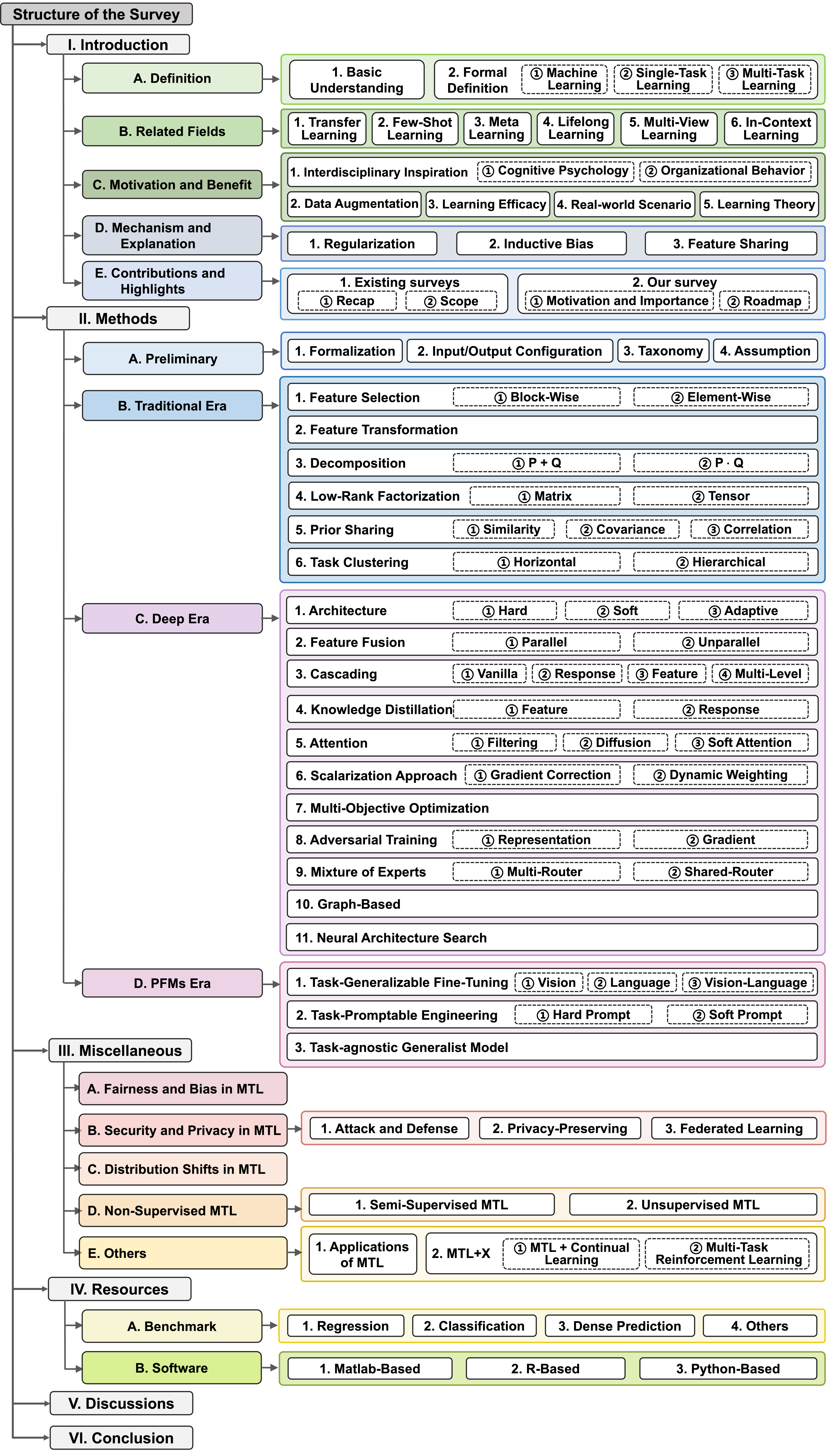}
      \caption{The structure of this survey.}
    \label{fig:outline}
\end{figure}

%% file: table_files/abbrev.tex
\begin{table*}[t]
    \centering
    \tiny
    \caption{Alphabetically sorted index table of acronyms.}
    \midsepremove
\scalebox{1.05}{
    \begin{tabular}{ll||ll}
    \rowcolor{gray!40}
    \toprule
         Abbreviation & Expanded Form & Abbreviation & Expanded Form \\
    \cmidrule(lr){1-2}\cmidrule(lr){3-4}
    
        AD & \textbf{A}lzheimer’s \textbf{D}isease & 
        AGM & \textbf{A}ccelerated \textbf{G}radient \textbf{M}ethod \\

    \rowcolor{gray!20}
        APM & \textbf{A}ccelerated \textbf{P}roximal \textbf{M}ethod &
        CE & \textbf{C}ross-\textbf{E}ntropy \\
        
        CNN & \textbf{C}onvolutional \textbf{N}eural \textbf{N}etwork &
        CT & \textbf{C}omputed \textbf{T}omography \\

    \rowcolor{gray!20}
        CV & \textbf{C}omputer \textbf{V}ision &
        DA & \textbf{D}omain \textbf{A}daptation \\

        DL & \textbf{D}eep \textbf{L}earning &
        DNN & \textbf{D}eep \textbf{N}eural \textbf{N}etwork \\

    \rowcolor{gray!20}
        FCN & \textbf{F}ully \textbf{C}onvolutional \textbf{N}etwork &
        FNN & \textbf{F}eedforward \textbf{N}eural \textbf{N}etwork \\
        
        FSL & \textbf{F}ew \textbf{S}hot \textbf{L}earning &
        GAN & \textbf{G}enerative \textbf{A}dversarial \textbf{N}etwork \\

    \rowcolor{gray!20}
        GCN & \textbf{G}raph \textbf{C}onvolutional \textbf{N}etwork &
        GNN & \textbf{G}raph \textbf{N}eural \textbf{N}etwork \\
        
        GP & \textbf{G}aussian \textbf{P}rocess &
        GPT & \textbf{G}enerative \textbf{P}retrained \textbf{T}ransformer \\

    \rowcolor{gray!20}
        GPU & \textbf{G}raphics \textbf{P}rocessing \textbf{U}nit &
        GRL & \textbf{G}radient \textbf{R}eversal \textbf{L}ayer \\
        
        I/O & \textbf{I}nput/\textbf{O}utput &
        KD & \textbf{K}nowledge \textbf{D}istillation \\

    \rowcolor{gray!20}
        LLM & \textbf{L}arge \textbf{L}anguage \textbf{M}odel &
        LSTM & \textbf{L}ong \textbf{S}hort-\textbf{T}erm \textbf{M}emory \\
        
        MAP & \textbf{M}aximum \textbf{A} \textbf{P}osteriori &
        MCI & \textbf{M}ild \textbf{C}ognitive \textbf{I}mpairment \\
        
    \rowcolor{gray!20}
        MDP & \textbf{M}arkov \textbf{D}ecision \textbf{P}rocess &
        MIM & \textbf{M}asked \textbf{I}mage \textbf{M}odeling \\
        
        MIML & \textbf{M}ulti-\textbf{I}nstance \textbf{M}ulti-\textbf{L}abel learning &
        MIMO & \textbf{M}ulti-\textbf{I}nput \textbf{M}ulti-\textbf{O}utput \\

\rowcolor{gray!20}
        MISO & \textbf{M}ulti-\textbf{I}nput \textbf{S}ingle-\textbf{O}utput &
        ML & \textbf{M}achine \textbf{L}earning \\
        
        MLM & \textbf{M}asked \textbf{L}anguage \textbf{M}odeling &
        MLP & \textbf{M}ulti-\textbf{L}ayer \textbf{P}erceptron \\

\rowcolor{gray!20}
        MoE & \textbf{M}ixture-\textbf{o}f-\textbf{E}xperts &
        MOO & \textbf{M}ulti-\textbf{O}bjective \textbf{O}ptimization \\

        MRI & \textbf{M}agnetic \textbf{R}esonance \textbf{I}maging &
        MSE & \textbf{M}ean \textbf{S}quared \textbf{E}rror \\
        
\rowcolor{gray!20}
        MTL & \textbf{M}ulti-\textbf{T}ask \textbf{L}earning &
        MTRL & \textbf{M}ulti-\textbf{T}ask \textbf{R}einforcement \textbf{L}earning \\
        
        MVL & \textbf{M}ulti-\textbf{V}iew \textbf{L}earning &
        NAS & \textbf{N}eural \textbf{A}rchitecture \textbf{S}earch \\
        
\rowcolor{gray!20}   
        NLI & \textbf{N}atural \textbf{L}anguage \textbf{I}nference &
        NLP & \textbf{N}atural \textbf{L}anguage \textbf{P}rocessing \\

        OCR & \textbf{O}ptical \textbf{C}haracter \textbf{R}ecognition &
        OOD & \textbf{O}ut-\textbf{O}f-\textbf{D}istribution \\

\rowcolor{gray!20}
        PET & \textbf{P}ositron \textbf{E}mission \textbf{T}omography &
        PFM & \textbf{P}retrained \textbf{F}oundation \textbf{M}odel \\
        
        PSD & \textbf{P}ositive \textbf{S}emi-\textbf{D}efinite &
        RL & \textbf{R}einforcement \textbf{L}earning \\

\rowcolor{gray!20}
        RNN & \textbf{R}ecurrent \textbf{N}eural \textbf{N}etwork &
        seq2seq & \textbf{seq}uence \textbf{to} \textbf{seq}uence \\
        
        SIMO & \textbf{S}ingle-\textbf{I}nput \textbf{M}ulti-\textbf{O}utput &
        SNP & \textbf{S}ingle \textbf{N}ucleotide \textbf{P}olymorphism \\
        
\rowcolor{gray!20}
        SGD & \textbf{S}tochastic \textbf{G}radient \textbf{D}escent &
        SSL & \textbf{S}elf-\textbf{S}upervised \textbf{L}earning \\
        
        SOTA & \textbf{S}tate-\textbf{O}f-\textbf{T}he-\textbf{A}rt &
        STL & \textbf{S}ingle-\textbf{T}ask \textbf{L}earning \\

\rowcolor{gray!20}
        SVD & \textbf{S}ingular \textbf{V}alue \textbf{D}ecomposition &
        SVM & \textbf{S}upport \textbf{V}ector \textbf{M}achine \\

        TL & \textbf{T}ransfer \textbf{L}earning &
        TPU & \textbf{T}ensor \textbf{P}rocessing \textbf{U}nit \\

\rowcolor{gray!20}
        VLM & \textbf{V}ision-\textbf{L}anguage \textbf{M}odel &
        VQA & \textbf{V}isual \textbf{Q}uestion \textbf{A}nswering \\
        ZSL & \textbf{Z}ero-\textbf{S}hot \textbf{L}earning &
        & \\

    \bottomrule
    
    \end{tabular}
    \label{tab:abbrev}
}
    \raggedright
    \scalebox{0.9}{
    \footnotesize
    This table excludes abbreviations pertaining to datasets, institutions, and newly proposed methods.
    }
\end{table*}

%% file: tex_files/01_intro/paper_num.tex
\begin{wrapfigure}[19]{r}{10.0cm}
    \centering
    \includegraphics[width=0.9\linewidth]{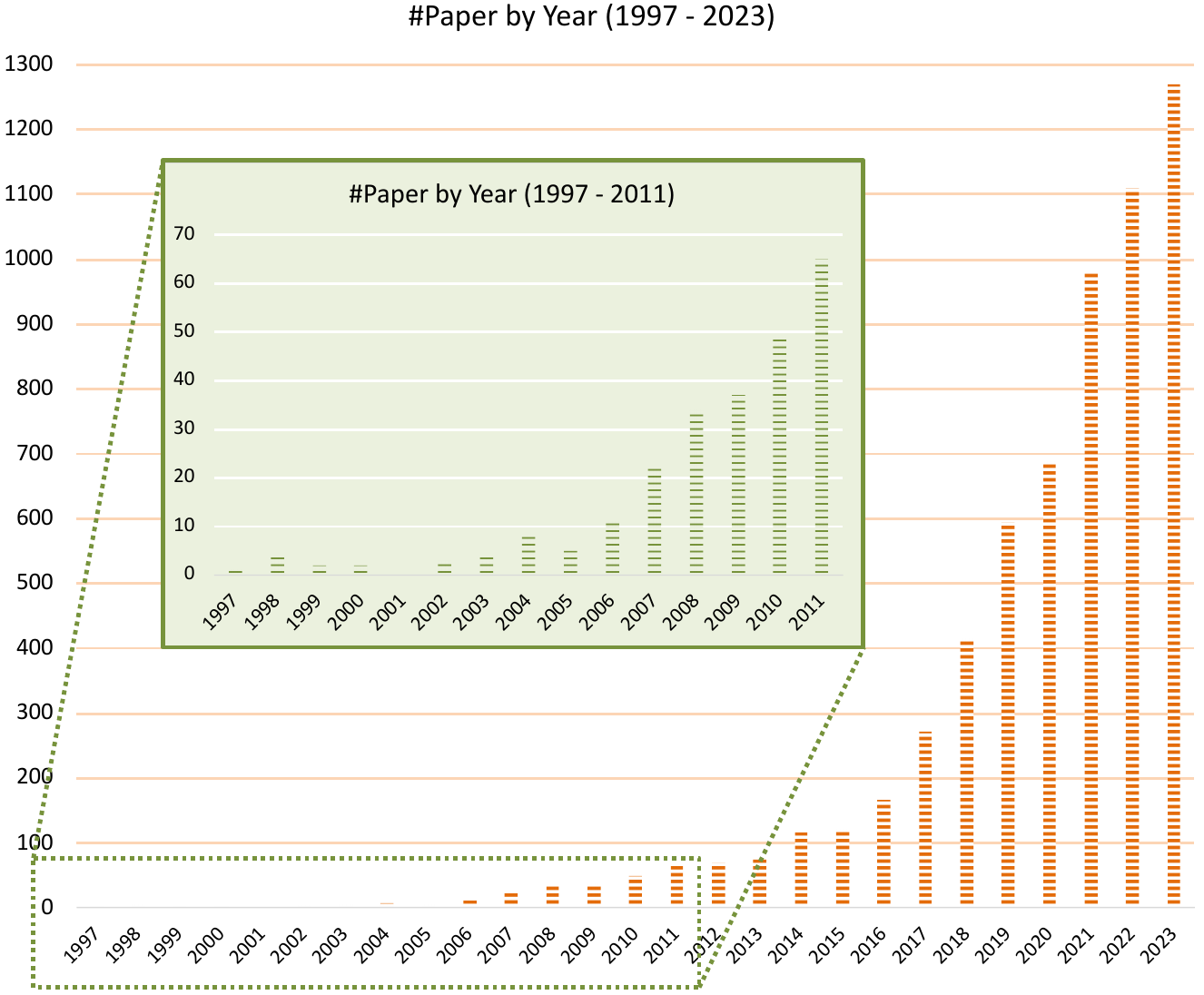}
    \caption{The total number of published papers ($y$-axis) has surged for the MTL topic from 1997 to 2023 ($x$-axis).}
    \label{fig:paper-num}
\end{wrapfigure}

%% file: 02-1_method_traditional.tex
\section{MTL Models}
\label{sec:mtlmodel}
\input{tex_files/02-1/stl-mtl-framework}
\paragraph{\emph{Formalization}} In machine learning, no matter the problem (discriminative, generative, adversarial, etc.), we hope to learn a predictive model by minimizing the regularized empirical loss as
\begin{equation}\label{eq:main.prob}
 \min\limits_{\Wbold}\Lmcal(f_{\Wbold}(\Xbold), \Ybold) + \lambda\Omega(\Wbold),
\end{equation}
where $(\Xbold, \Ybold)$ is data pairs sampled from a single task, and $\Wbold$ includes weights of learning model $f(\cdot)$. In general, $\Lmcal$ measures the distance between the predictions and ground-truth, and $\Omega$ adds constraints to the learning model, e.g., sparsity. The trade-off parameter $\lambda$ controls the balance between the loss and penalty. Fig.~\ref{stl} shows the detailed framework of STL. In comparison, as shown in {Fig.}~\ref{mtl}, the optimization in MTL is conducted on the multiple loss functions to achieve joint learning, and each task can maintain a specific loss function. Accordingly, MTL considers the problem in the following:
\begin{equation}
    \min\limits_{\{\boldsymbol{W}^{(t)}\}_{t=1}^T}\sum_{t=1}^{T}\Lmcal^{(t)}\left(f_{\Wbold^{(t)}}(\Xbold^{(t)}), \Ybold^{(t)}\right) + \lambda\Omega\left(\Wbold^{(1)}, \cdots, \Wbold^{(T)}\right),
    \label{general-mtl}
\end{equation}
    where $T$ denotes the number of tasks, and $f(\cdot)$ is the MTL model to be learned. In MTL, $f(\cdot)$ always encodes both task-specific and -shared representations, and $\Omega(\cdot)$ builds task relatedness and reciprocity; both contribute to the effectiveness and efficiency of MTL.
\input{tex_files/02-1/io-config}
\paragraph{\emph{I/O Configurations}} To accommodate data in Eq.~(\ref{general-mtl}), it is necessary to consider various input/output (I/O) configurations that may impose constraints on the MTL modeling process. For instance, tasks such as semantic segmentation and depth estimation can utilize the same input images, and the applications are always developed using datasets where each image is attached with dense prediction labels for both segmentation and depth. On the other hand, when dealing with a digital recognition problem involving multiple domains (e.g., handwritten digits and license plate digits), different inputs are mapped to the same output space. We refer the former as a single-input multi-output (SIMO) configuration and the latter as a multi-input single-output (MISO) configuration. In MTL, the most prevalent scenarios reside in multi-input multi-output (MIMO) configuration where each task maintains its own set of samples and the labels are omnivorous, e.g., autonomous driving that involves pedestrian detection and traffic sign recognition. Let us denote the data input space and its corresponding label space for the $t$-th task $(t = 1, \cdots, T)$ by $\mathcal{X}^{(t)}$ and $\mathcal{Y}^{(t)}$, respectively. We classify the MTL problems into three cases: SIMO, MISO, and MIMO. Fig.~\ref{io-config} shows the illustration of these three configurations. It is worth noting that the I/O configurations do not significantly impact the taxonomy of methods in MTL. As indicated in Table~\ref{tab:summary}, there are numerous shared practices of applying different methods to these I/O configurations, as well as various data modalities and task types.
\input{table_files/summary}

\paragraph{\emph{Taxonomy}} MTL has seen significant advancement prior to the DL era~\citep{caruna1993multitask, caruana1997multitask, bakker2003task, ando2005framework, obozinski2006multi, zhang2006a}. Initially, there was a strong focus on weight/parameter regularization, including sparse learning for cross-task feature selection, low-rank learning to uncover underlying factors, and decomposition methods to capture informative components.
These approaches, while innovative in integrating intuitive variations from existing methods (e.g., the $\ell_{2,1}$ regularizer derived from the classic $\ell_1$ regularizer), still face limitations in practical applications due to the idealistic assumptions and a lack of consideration for task relationships. The emergence of methods like task clustering, priori sharing, graph-based learning, and MoE marked a shift towards more effective task relationship modeling. With the transition to the DL era, the abundance of features learned from architectures like convolutional neural networks (CNNs)~\citep{fukushima1980neocognitron, lecun1998gradient}, recurrent neural networks (RNNs)~\citep{werbos1988generalization, hochreiter1997long} and Transformers~\citep{vaswani2017attention, dosovitskiy2020image} spurred the exploration of feature propagation methods, such as feature fusion, cascading, knowledge distillation (KD), and cross-task attention, all crucial for leveraging multi-source features. Alternatively, optimization-based methods, including scalarization, MOO, adversarial training and NAS, focused on gradients to harmonize optimization directions across tasks. These methods, while not restricted by I/O configurations, are constrainted on the pre-defined number of tasks and the use of heterogeneous architectures. Pre-training techniques, which leverages TL, markes a significant advancement towards unified and versatile multitasking, breaking limitations related to data modalities, dimensions, task numbers, model architectures, etc. 
The only cost is the large computation resources to train a really large model that can accommodate multi-task distributions. The MTL models are accordingly organized into five categories: regularization, relationship learning, feature propagation, optimization, and pre-training. 
Each contains a series of topics arranged chronologically in \textbf{\S}~\ref{traditional-era} (traditional ML era), \textbf{\S}~\ref{deep-era} (DL era), and \textbf{\S}~\ref{fm-era} (PFM era). All of these topics can be inferred from three self-evident assumptions (but have been extensively validated by empirical evidence) as below:
\begin{assumption}[Parameter Relatedness]\label{assump:parameter} 
Under the same hypothesis space, models learned to perform related tasks can exhibit similarities.
\end{assumption}
\begin{assumption}[Feature Richness]\label{assump:feature} 
Given the same level of experience, expanding the number of tasks to be learned can enhance the richness of features.
\end{assumption}
\begin{assumption}[Optimization Consistency]\label{assump:optimization} 
Learning multiple related tasks jointly in a single model can ensure consistency in optimization directions for each task.
\end{assumption}

We acknowledge that the presented taxonomy is not exhaustive, and certain methods may be classified differently when viewed from a different perspective. For example, Task Tree (TAT)~\citep{han2015learning}, a clustering MTL method, establishes task hierarchy by decomposing the parameter matrix into different component matrices for each tree layer; we discuss it within the context of clustering MTL (see \textbf{\S}~\ref{sec:clustering}).
We also acknowledge that some methods that may be of interest to readers may not be included in this survey due to similarities or oversight. We welcome paper recommendations and will update the survey on our project page accordingly.\footnote{\url{https://github.com/junfish/Awesome-Multitask-Learning}}.
In Table~\ref{tab:summary}, we summarize their assumptions, common practice, and technical constraints of these topics in terms of I/O configuration, data modality, and task type.

\subsection{Traditional Era: Provable but Restrictive}
\label{traditional-era}
\input{table_files/notation}
{~~}\vspace{2pt}\\
To establish a unified formulation, we start the review of traditional methods by defining a common framework. The notations for subsequent discussions are summarized in Table~\ref{tab:notation}. Building upon this, we initiate our discussion with multiple standard regression models for each task as a paradigm. The weights of these homogeneous models can be arranged into one weight matrix, catalyzing a series of MTL studies through matrix regularization techniques in the traditional era. We denote by $\{(\boldsymbol{X}^{(t)},\boldsymbol{y}^{(t)})\}_{t=1}^T$ our dataset across $T$ tasks. For each task indexed by $t={1,2,\cdots,T}$, we are given $N_t$ samples with $D$ features, i.e., $\boldsymbol{X}^{(t)}\in \mathbb{R}^{N_t\times D}$, and the corresponding response values $\boldsymbol{y}^{(t)}\in\mathbb{R}^{N_t}$.

The single-task setting of these multiple linear regression problems is
\begin{equation}
    \boldsymbol{y}^{(t)} = {\boldsymbol{X}^{(t)}}\boldsymbol{w}^{(t)} + \epsilon^{(t)}, t = 1, \cdots, T,
\end{equation}

where $\boldsymbol{w}^{(t)}\in\mathbb{R}^D$ for any $t\in\{1,\cdots,T\}$, $\epsilon^{(t)}\sim \mathcal{N}(0,\sigma_{t}^{2}\bold{I})$ is the error term independent of $\boldsymbol{X}^{(t)}$, and $\sigma_{t}$ is determined by the system state for $t$-th task.
Each model is separately learned from independent samples $\{({\boldsymbol{x}_1^{(t)}}^\top, y_1^{(t)}), \cdots, ({\boldsymbol{x}_{N_t}^{(t)}}^\top, y_{N_t}^{(t)})\}$.

A trivial simplification of the above linear regressions is that all tasks maintain the same feature size $D$, thus leading to a natural idea of stacking weight vectors for these tasks: $\boldsymbol{W}=[\boldsymbol{w}^{(1)}, \cdots, \boldsymbol{w}^{(T)}]\in\mathbb{R}^{D\times T}$, where the matrix-based regularizers come into play. To estimate as $\boldsymbol{W}$, the MTL method minimizes the objective function:
\begin{equation}
    \min\limits_{\boldsymbol{W}}\sum\limits_{t=1}^{T}\frac{1}{n_t}\mathcal{L}^{(t)}\left({\boldsymbol{X}^{(t)}}\boldsymbol{w}^{t}, \boldsymbol{y}^{(t)}\right) + \lambda\Omega(\boldsymbol{W}) \label{eq:2.4},
\end{equation}
where we consider the weights of multiple models, i.e., $\boldsymbol{W}$, as a union, and denote by
$\wbold^t$ the $t$-th column of $\Wbold$. Normally, an identical loss function, e.g., mean squared error (MSE), is always applied to $\{\Lmcal^{(t)}\}_{t=1}^T$, which originates from the $i.i.d.$ assumption of $\{\epsilon^{(t)}\}_{t=1}^T$. To capture task relatedness from the Assumption~\ref{assump:parameter} that multiple models are similar to each other, $\Omega$ is designed to take various regularization forms in traditional MTL. The overview of regularization techniques used in the traditional ML era for MTL (will be discussed in the following subsections) is presented in Table \ref{tab:regularizer}.

\input{table_files/regularizer}

\subsubsection{Feature Selection}
\label{subsubsec:selection}
The high-dimensional scaling~\citep{negahban2008joint} where the number of model weights is much larger than that of the observations/features, i.e., $D\gg N$, arises in many real-world problems, leading it costly and arduous to seek effective predictor variables. Sparse learning with an $\ell_1$ regularizer that aims to identify a structure characterized by a reduced number of non-zero elements. This parsimonious solution ensures the retention and selection of the most effective and efficient subset of features tailored to the target task~\citep{tibshirani1996regression}. 
In MTL, Assumption~\ref{assump:parameter} underpins the development of all sparse learning models. Under the settings of sparse learning, this assumption posits that \textit{similar sparsity patterns in model parameters suggest the relatedness between tasks.}
As a result, sparsity patterns subtly represent task relatedness, underscoring a subset of common features derived from these limited samples. More benefits and efficacy of employing sparsity in MTL have been thoroughly assessed and discussed in~\citet{lounici2009taking}. In this section, our discussion of feature selection in MTL encompasses both the block-wise ($\ell_{2,1}$) and element-wise ($\ell_{1,1}$) approaches. Each approach maintains both shared and task-specific features, optimizing performance across all tasks. In the block-wise approach, tasks can differentiate themselves from others' priorities by attributing distinct weights to the commonly selected features. Conversely, the element-wise approach allows tasks to highlight their distinct preferences on predictors by opting for specific features in addition to the shared ones.

\paragraph{\underline{Block-Wise Sparsity}}Multi-Task Feature Selection~\citep{obozinski2006multi} is the first method to address the problem of joint feature selection across a group of related tasks. This method extends the $\ell_1$ regularization for STL to the $\ell_{2,1}$ regularization for MTL. The assumption for $\ell_{2,1}$ regularization scheme is that multiple related tasks have a similar preference for a few common features, which encourages a solution to share the sparsity pattern.
Therefore, $\ell_{2,1}$ imposes a sparse penalty on the $\ell_2$ norms of the $T$-dimensional weight vectors associated with each feature across tasks (i.e., row vectors of the weight matrix $\boldsymbol{W}^{D\times T}$). This is formulated as follows:
\begin{equation}\label{Multi-Task Feature Selection}
    \min\limits_{\boldsymbol{W}}\frac{1}{2}\sum\limits_{t=1}^{T}\frac{1}{N_t}\|{\boldsymbol{X}^{(t)}}\boldsymbol{w}^{t} - \boldsymbol{y}^{(t)}\|^2_2 + \lambda\sum\limits_{d=1}^{D}{\|\boldsymbol{w}_d\|}_2,
\end{equation}
which selects features globally via encouraging several feature-wise weight vectors $\wbold_d$ across all tasks to be $\vec{\boldsymbol{0}}$. The $\ell_2$ norm imposed on feature-wise weight vectors (i.e., $\wbold_d$) before $\ell_1$ norm here is a magnitude measurement, which could be substituted by any other $\ell_p$ ($p\geq 1$) norm~\citep{obozinski2006multi}. This penalty term can be seen as a generalization of $\ell_1$ regularization when task number $T=1$.
To solve the problem~(\ref{Multi-Task Feature Selection}),~\citet{obozinski2006multi} offers a block-coordinate descent optimization method to update the block of weight vector associated with each feature.~\citet{liu2012multi} proposes an accelerated algorithm by reformulating it as two equivalent smooth convex optimization problems.

Multi-Task Lasso~\citep{zhang2006a} extends the efficient $\ell_{p,1}$ regularizers via imposing $\ell_{\infty}$ norm to each feature-wise weight vector $\wbold_d$. Based on the assumption that the number of effective predictor features is much smaller than the total features, Multi-task Lasso learns a sparser structure by
\begin{equation}
    \min\limits_{\boldsymbol{W}}\frac{1}{2}\sum\limits_{t=1}^{T}\frac{1}{N_t}\|{\boldsymbol{X}^{(t)}}\boldsymbol{w}^{t} - \boldsymbol{y}^{(t)}\|^2_2 + \lambda\sum\limits_{d=1}^{D}{\|\boldsymbol{w}_d\|}_\infty.
\end{equation}
The use of $\ell_\infty$ enforces the procedure to take the maximum value of each feature-wise vector $\wbold_d$ across all tasks. This is appropriate if relevant features are not shared by every task, and this situation frequently happens as the number of tasks grows.~\citet{zhang2006a} proves that this $\ell_{\infty, 1}$ problem can be solved by an efficient convex optimization technique. Furthermore, a full spectrum of $\ell_{p,1}$ regularization ($\ell_{1,1}$, especially) suitable for MTL is investigated and discussed. However,~\citet{negahban2008joint} prove that the use of $\ell_{1,\infty}$ can improve learning efficiency only if the overlap of feature entries across tasks is large enough ($>2/3$), as compared to the situation where each task learns Lasso problem separately.

Temporal Group Lasso~\citep{zhou2011multi} is an MTL formulation for predicting the disease progression, which considers $t$ time points of disease progression as related tasks. They first admit the limitation of task independence for the analytical solution $\boldsymbol{W} = ({\boldsymbol{X}^\top\boldsymbol{X}+\lambda_1 \boldsymbol{I}})^{-1}\boldsymbol{X}^\top\boldsymbol{Y}$ to the ridge regression problem $\min_{\boldsymbol{W}}{\|\boldsymbol{X}\boldsymbol{W}-\boldsymbol{Y}\|}_F^2 + \lambda_1{\|\boldsymbol{W}\|}_F^2$, where $\boldsymbol{X}$ is identical and $\boldsymbol{Y}=[\boldsymbol{y}^{(1)}, \cdots,\boldsymbol{y}^{(T)}]$ denotes the progression of disease across $T$ tasks (time points). To capture the temporal smoothness for the adjacent time points, Temporal Group Lasso adds the temporal smoothness term and feature selector term to form the formalization as
\begin{align}
    \min\limits_{\boldsymbol{W}} \frac{1}{2}\|S\odot({\boldsymbol{X}}\boldsymbol{W} - \boldsymbol{Y})\|^2_F + \lambda_1\sum\limits_{d=1}^D\|\boldsymbol{w}_d\|_2^2 + \lambda_2\sum\limits_{t=1}^{T-1}\|\boldsymbol{w}^{t}-\boldsymbol{w}^{t+1}\|_2^2 + \lambda_3\sum\limits_{d=1}^D\|\boldsymbol{w}_d\|_2,
\end{align} 
where $S\in\mathbb{R}^{N\times T}$ is the indication matrix for the incomplete data, i.e., for any $n\in\{1,\cdots,N\}, t\in\{1,\cdots,T\}, s_{n,t}=0$ if the target value of sample $n$ at the $t$-th time point is missing and $s_{n,t}=1$ otherwise. It is noted that this problem can be easily solved by accelerated gradient method (AGM)~\citep{nesterov2013gradient} using SLEP~\citep{liu2009slep}. However, to avoid the shrinkage of relevant features that would result in sub-optimal performance,~\citet{zhou2011multi} proposed a standard two-stage procedure to relax the $\ell_1$ regularization.

Adaptive Multi-Task Elastic-Net~\citep{chen2012adaptive} aims to address the problem of collinearity existing in the multi-task feature selection method. Inspired by elastic-net~\citep{zou2005regularization}, a natural thought is to add another quadratic penalty $\sum_{d=1}^D\|\boldsymbol{w}_d\|_2^2$ to the sparse multi-task constraint $\sum_{d=1}^D\|\boldsymbol{w}_d\|_2$, which forms the corresponding multi-task elastic-net problem as
\begin{equation}
    \label{amten}
    \min\limits_{\boldsymbol{W}}\frac{1}{2}\sum\limits_{t=1}^{T}\frac{1}{N_t}\|{\boldsymbol{X}^{(t)}}\boldsymbol{w}^{t} - \boldsymbol{y}^{(t)}\|^2_2 + \lambda_1\sum\limits_{d=1}^{D}{\|\boldsymbol{w}_d\|}_2 +  \lambda_2\sum\limits_{d=1}^D\|\boldsymbol{w}_d\|_2^2,
\end{equation}
where the traditional $\ell_{2,1}$ mixed norm learns the same amount of regularization across all features. As discussed below in the adaptive sparse multi-task lasso~\citep{lee2010adaptive}, it is promising to learn different regularization weights $\{\boldsymbol{w}_d\}_{d=1}^{D}$ for each feature. However, unlike the application of eQTL detection~\citep{lee2010adaptive} where features on single nucleotide polymorphisms (SNPs) make it easier to incorporate prior knowledge for each feature (see Eq.~(\ref{snp_prior})), the priors scaling the importance of adaptive weights for each feature are always unavailable in many real-world problems.~\citet{chen2012adaptive} proposes a three-stage algorithm to estimate the adaptive weights $\boldsymbol{w}_d$ via using a data-driven method: (1) estimate the initial regression weights $\{\hat{\boldsymbol{w}}_d\}_{d=1}^{D}$ with uniform weight for each feature; (2) construct adaptive scaling weights $\{\hat{\lambda}_d\}_{d=1}^{D}, \hat{\lambda}_d = (\|\hat{\boldsymbol{w}}_d\|_2)^{-\gamma}$ according to the weights estimated in the first step, where $\gamma$ is a fixed constant; (3) compute the final estimated parameters via the multi-task elastic-net
with the adaptive scaling weights, i.e., $\hat{\boldsymbol{W}} = \min\limits_{\boldsymbol{W}}\frac{1}{2}\sum_{t=1}^{T}\frac{1}{N_t}\|{\boldsymbol{X}^{(t)}}\boldsymbol{w}^{t} - \boldsymbol{y}^{(t)}\|^2_2 + \lambda_1\sum_{d=1}^{D}\hat{\lambda}_d{\|\boldsymbol{w}_d\|}_2 +  \lambda_2\sum_{d=1}^D\|\boldsymbol{w}_d\|_2^2$.

\paragraph{\underline{Element-Wise Sparsity}}Sparse Multi-Task Lasso~\citep{lee2010adaptive} allows feature-specific penalty magnitude by incorporating a set of priors 
with fixed scaling parameters. This method also generalizes the sparse group Lasso penalty~\citep{simon2013sparse} by suing both the $\ell_{2,1}$ and $\ell_{1,1}$ norms to perform joint block-wise and element-wise feature selection. Specifically, sparse multi-task Lasso proposes

\begin{equation}\label{SMTL}
    \min\limits_{\boldsymbol{W}}\frac{1}{2}\sum\limits_{t=1}^{T}\frac{1}{N_t}\|{\boldsymbol{X}^{(t)}}\boldsymbol{w}^{t} - \boldsymbol{y}^{(t)}\|^2_2 + \lambda_1\sum\limits_{d=1}^{D}\rho_d{\|\boldsymbol{w}_d\|}_2 + \lambda_2\sum\limits_{d=1}^D\theta_d{\|\boldsymbol{w}_d\|}_1,
\end{equation}

where $\boldsymbol{\rho}=[\rho_1,\cdots,\rho_D]^\top$ and $\boldsymbol{\theta}=[\theta_1,\cdots,\theta_D]^\top$ are the scaling weights for the $\ell_{2,1}$ and $\ell_{1,1}$ regularizers, respectively. There exist two advantages of this method: (1) Unlike previous work by~\citet{obozinski2006multi, zhang2006a}, which considers $\ell_{p,1}~(p>1)$ norm that learns block-wise sparsity well but overlooks element-wise sparsity within each feature group, sparse multi-task Lasso balances the $\ell_{2,1}$ and $\ell_{1,1}$ regularizers via $\lambda_1$ and $\lambda_2$ to achieve both simultaneously. (2) Unlike~\citet{obozinski2006multi, zhang2006a}, which treats every feature-wise weight vectors ($\{\boldsymbol{w}_d\}_{d=1}^D$) equally, i.e., $\rho_d=\theta_d=1, d\in\{1,\cdots,D\}$, the two scaling vectors in \citet{lee2010adaptive} can be automatically learned from data. Furthermore, \citet{maurer2013sparse} uses the $\ell_1$ regularizer on data preprocessed by a linear mapping function and provides bounds on the generalization error for both MTL and TL settings.

\input{tex_files/02-1/bayesian}
Adaptive Sparse Multi-Task Lasso~\citep{lee2010adaptive} is induced as a super-problem from above. This method adaptively incorporates prior knowledge on SNPs~\citep{brookes1999essence} to learn two scaling vectors $\boldsymbol{\rho}$ and $\boldsymbol{\theta}$, which are defined as the mixtures of features on the $j$-th SNP

\begin{minipage}{0.6\textwidth}
\begin{align}
    & \rho_d = \sum\limits_i \varv_i f_i^d~\text{and}~\theta_d = \sum\limits_i\omega_i f_i^d, d = 1,\cdots,D, \nonumber \\
    & \text{s.t.}\quad\sum\limits_i \varv_i = \sum\limits_i \omega_i = 1,
    \label{snp_prior}
\end{align}
\end{minipage}

where $f_i^d$ is the $i$-th feature of the $d$-th SNP. Here, the component $x_{n_t, d}\in\{0,1,2\}$ of $\boldsymbol{X}^{(t)}\in\mathbb{R}^{N_t\times D}$ in {Eq.}~(\ref{SMTL}) denotes the number of minor alleles at the $d$-th SNP of the $n_t$-th sample.~\citet{lee2010adaptive} uses a directed graphical model as an elegant Bayesian tool to find the maximum a posteriori (MAP) estimate of all the above learnable weights, shown in Fig.~\ref{bayesian}. Then the conditional probability of weight matrix $\boldsymbol{W}$ given $\boldsymbol{\rho}$ and $\boldsymbol{\theta}$ is
\begin{equation}
    P(\boldsymbol{W}|\boldsymbol{\rho}, \boldsymbol{\theta}) = \frac{1}{Z(\boldsymbol{\rho}, \boldsymbol{\theta})}\prod\limits_{d=1}^D\prod\limits_{t=1}^T\exp (-\theta_d\lvert w_{d,t}\rvert)\times\prod\limits_{d=1}^D\exp(-\rho_d\|\mathbf{w}_d\|_2),
\end{equation}
where the normalization factor $Z(\boldsymbol{\rho}, \boldsymbol{\theta})$ is upper-bounded by the inference of high dimensional multivariate Laplace distribution~\citep{gomez1998multivariate}. Accordingly,~\citet{lee2010adaptive} proposes an alternating minimization approach that iteratively optimizes one of $(\boldsymbol{\varv},\boldsymbol{\omega})$ and $\boldsymbol{W}$ by fixing another until convergence.

Convex Fused Sparse Group Lasso (cFSGL)~\citep{zhou2012modeling} considers a formulation that additionally allows the element-wise feature selection compared to the temporal group Lasso~\citep{zhou2011multi}. cFSGL encourages the sparsity for joint feature selection across tasks and specific feature selection within a task. The formulation can be written as
\begin{align}
    \min\limits_{\boldsymbol{W}} \frac{1}{2}{\|S\odot(\boldsymbol{X}\boldsymbol{W} - \boldsymbol{Y})\|}^2_F + \lambda_1\sum\limits_{d=1}^D\|\boldsymbol{w}_d\|_1 + \lambda_2\sum\limits_{t=1}^{T-1}\|\boldsymbol{w}^t-\boldsymbol{w}^{t+1}\|_1 + \lambda_3\sum\limits_{d=1}^D\|\boldsymbol{w}_d\|_2,
\end{align}
where $\sum_{t=1}^{T-1}\|\boldsymbol{w}^t-\boldsymbol{w}^{t+1}\|_1$ is the fused Lasso penalty, and the combination of $\ell_{1,1}$ and $\ell_{2,1}$ is also known as the sparse group Lasso penalty~\citep{simon2013sparse}. Thus, this problem with three non-smooth regularization terms can be solved by AGM via computing the decoupled proximal operator.

Multi-Stage Multi-Task Feature Learning~\citep{gong2012multi} represents a pioneering approach to address the sub-optimal solutions observed in prior convex sparse regularization problems. This sub-optimality can be attributed to the challenges in approximating $\ell_0$ regularization. In response to this limitation, the method introduces a non-convex formulation utilizing capped $\ell_{1,1}$ regularization for MTL:
\begin{equation}
    \label{msmtfl}
    \min\limits_{\boldsymbol{W}}\frac{1}{2}\sum\limits_{t=1}^{T}\frac{1}{N_t}\|{\boldsymbol{X}^{(t)}}\boldsymbol{w}^{t} - \boldsymbol{y}^{(t)}\|^2_2 + \lambda\sum\limits_{d=1}^D\min\{\|\boldsymbol{w}_d\|_1, \tau\},
\end{equation}
where $\tau$ is a threshold to tailor the $\ell_1$ norm of weight vectors, i.e., $\{\|\boldsymbol{w}_d\|\}_{d=1}^D$ corresponding to each feature, and the term $\sum_{d=1}^D\min\{\|\boldsymbol{w}_d\|_1, \tau\}$ is a natural generalization of capped $\ell_1$ norm in~\citet{zhang2010analysis, zhang2013multi}. To solve this non-convex problem (\ref{msmtfl}), \citet{gong2012multi} proposed an efficient algorithm and investigated the estimation error bound of the resulting estimator.

\begin{applebox}{Remarks}
\begin{enumerate}[leftmargin=0.4cm, label=(\roman*)]
    \item Feature selection can highlight task relatedness, especially in scenarios with limited data availability ($\#$feature $>$ $\#$data).
    \item The $\ell_1$-series regularization easily facilitates feature selection, offering broad generalizability across various parametric models in MTL.
    \item In MTL contexts with plenty of training resources, feature selection might compromise performance; however, it enhances interpretability through the selected features.
    \item In situations with limited data, certain feature selection techniques may become vulnerable to minor data variations, which can potentially impact the stability of the learning process.
\end{enumerate}
\end{applebox}

\input{tex_files/02-1/naive-mtl}
\subsubsection{Feature Transformation}
Unlike the sparse learning methods discussed in $\S$\ref{subsubsec:selection}, which assume direct use of observed features, feature transformation methods aim to combine and transform--rather than simply select--the raw features into new representations. This approach enables handling coarse-grained input data. Sparse learning in MTL builds task relatedness into model $f(\cdot)$ through sharing similar weight structure across multiple tasks, however, feature learning in MTL makes tasks be related to each other via enforcing a common underlying representation~\citep{argyriou2006multi}. For example,~\citet{yu2019towards} points out that two tasks of aesthetic quality assessment and emotional recognition in digital image analysis share similar feature representations. Another example from~\citet{caruna1993multitask, caruana1997multitask}, as shown in Fig.~\ref{feedforward}, reveals that different tasks can synchronously learn from the same feature encodings in feedforward neural networks (FNNs).

Multi-Task Feature Learning~\citep{argyriou2006multi} linearly combines observations/features via introducing a transformation matrix $\boldsymbol{U}\in\boldsymbol{O}^D$, which can be extended to nonlinear combinations by using kernel methods. As formulated in the following,
\begin{equation}
\min\limits_{\boldsymbol{U}, \boldsymbol{W}}\frac{1}{2}\sum\limits_{t=1}^{T}\frac{1}{N_t}\|({\boldsymbol{X}^{(t)}}\boldsymbol{U})\boldsymbol{w}^{t} - \boldsymbol{y}^{(t)}\|^2_2 + \lambda(\sum\limits_{d=1}^{D}{\|\boldsymbol{w}_d\|}_2)^2,\quad s.t.~\boldsymbol{U}\in\boldsymbol{O}^D,
\label{mtfl}
\end{equation}
we need to estimate $\boldsymbol{U}$ and $\boldsymbol{W}$ from the data. The $\ell_{2,1}$ norm imposed on $\boldsymbol{W}$ ensures that the transformed features, i.e., ${\boldsymbol{X}^{(t)}}\boldsymbol{U}$, with a fixed $\boldsymbol{U}$, would be collectively selected across tasks. To learn the transformed features, \citet{argyriou2006multi} fixed $\boldsymbol{W}$ to minimize the objective function (\ref{mtfl}) over $\boldsymbol{U}$ under the orthogonal constraints. Even with this two-step iterated optimization algorithm to solve for $\boldsymbol{W}$ and $\boldsymbol{U}$, solving the problem (\ref{mtfl}) is still a non-convex problem. Accordingly, it is transformed into an equivalent convex problem\footnote{It is also known as convex multi-task feature learning~\citep{argyriou2006multi, argyriou2008convex}, which is mentioned in~\citet{argyriou2006multi} and further discussed in~\citet{argyriou2008convex} with the learning of non-linear features using kernel methods.} as follows.
\begin{align}
    \min\limits_{\boldsymbol{V}, \boldsymbol{W}} & \frac{1}{2}\sum\limits_{t=1}^{T}\frac{1}{N_t}\|{\boldsymbol{X}^{(t)}}\boldsymbol{w}^{t} - \boldsymbol{y}^{(t)}\|^2_2 + \lambda\sum\limits_{t=1}^T{\boldsymbol{w}^t}^\top\boldsymbol{V}^{+}\boldsymbol{w}^t,\nonumber \\ 
    ~s.t.~ & \boldsymbol{V}\in\boldsymbol{S}_+^D, \text{tr}(\boldsymbol{V})\leq1, \text{col}(\boldsymbol{W})\subseteq\text{col}(\boldsymbol{V}).
\end{align}

\citet{dong2015multi} first extends the neural machine translation to an MTL framework which shares a bidirectional recurrent representation with forward and backward sequence information, as shown in Fig.~\ref{recurrent}. Suppose we have $T$ different language pairs $\{(\mathbf{x}^{(t)}, \ybold^{(t)})\}_{t=1}^T$, for instance, from English to many other languages like French, Spanish, Dutch, and Portuguese, the probability of generating each translated word at time step $i$ is
\begin{equation}
    p(y_i^{(t)}|y_1^{(t)}, \cdots, y_{i-1}^{(t)}, \xbold^{(t)}) = f(y_{i-1}^{(t)},  s_{i}^{(t)},  c_{i}^{(t)}), t = 1, \cdots, T,
\end{equation}
where $f$ is parameterized by a FNN, $s_{i}^{(t)}$ is the hidden state of a recurrent neural network at time step $i$, and $c_{i}^{(t)}$ is a context vector calculated from a sequence of annotations $(h_1, \cdots, h_{L_{\boldsymbol{x}}})$, which is mapped from the original sentence $\boldsymbol{x}$ by an encoder. More details of bidirectional sequence learning please refer to~\citet{dong2015multi}. After that, all annotations $h_j~(j = 1,\cdots,L_{\boldsymbol{x}})$ are collectively transformed by soft alignment parameters $A^{(t)}~(t = 1, \cdots, T)$ for each encoder-decoder to achieve cross-task communications.

\begin{applebox}{Remarks}
\begin{enumerate}[leftmargin=0.4cm, label=(\roman*)]
    \item Feature transformation can facilitate multiple tasks to share the same underlying representations.
    \item The features from different tasks can interact with each other, providing mutual benefits across all tasks.
\end{enumerate}
\end{applebox}

\subsubsection{Low-Rank Factorization} 
\label{subsec:low-rank}

In MTL, as discussed before, information sharing among multiple tasks can be achieved by assuming that all the tasks are impacted by the same small subset of predictors. On the other hand, low-rank structures imposed on the coefficient matrices or tensors can induce a different type of information sharing among tasks, i.e., the tasks are affected by the predictors through a shared small set of latent variables or directions, which are extracted from the original feature space and are the most relevant subspace to the outcomes. Depending on the way of indexing multiple learning tasks, one can choose to organize the coefficient vectors from multiple learning tasks into a matrix of dimension $D\times T$ or a tensor with a more delicate structure. In general, the multi-dimensional indices of tasks commonly imply that there are multi-layer relationships among multiple tasks, and the tensor form can help keep this inherent structure which allows leveraging information from different dimensions of task similarities.

\paragraph{\underline{Matrix Factorization}}
The most commonly seen situation is when we organize the coefficient vectors from multiple tasks into a matrix $\Wbold$, and the rank penalized problem can be formulated as 
\begin{equation}
\label{low-rank}
    \min_{\boldsymbol{W}} \sum\limits_{t=1}^T\mathcal{L}^{(t)}\left(f(\boldsymbol{X}^{(t)}, \boldsymbol{w}^{t}), \boldsymbol{y}^{(t)}\right)+\lambda~\text{rank}(\boldsymbol{W}).
\end{equation}
However, to minimize the rank of a matrix is NP-hard~\citep{vandenberghe1996semidefinite} due to the combinatorial nature of the rank function~\citep{ji2009accelerated, han2016multi}. An alternative is to substitute the rank penalty with the trace of the rank for the symmetric positive semidefinite matrix~\citep{mesbahi1999semi}, but it excludes non-symmetric or even non-square matrices in real-world applications.~\citet{fazel2001rank} generalized the trace heuristic to any matrix by introducing the trace norm (a.k.a, nuclear norm or Ky-Fun k-norm)~\citep{horn2012matrix}, which is defined as the sum of a matrix's all singular values (See Table~\ref{tab:notation}). 

Low Rank Multi-Task Learning~\citep{ji2009accelerated} first introduces the trace norm optimization problem into MTL, which yields a low-rank solution that maps to a low-dimensional feature subspace. The problem can be written as 
\begin{equation}
    \min\limits_{\boldsymbol{W}}\frac{1}{2}\sum\limits_{t=1}^{T}\frac{1}{N_t}\|{\boldsymbol{X}^{(t)}}\boldsymbol{w}^{t} - \boldsymbol{y}^{(t)}\|^2_2 + \lambda\|\boldsymbol{W}\|_*,
    \label{eq:lrmtl}
\end{equation}
where $\|\cdot\|_*$ denotes the trace norm of the weight matrix $\boldsymbol{W}$. The technical challenge for the problem above is the non-smooth nature of the trace norm, which makes it converge slowly ($O(\frac{1}{\sqrt{k}}), k$ is the iterations).~\citet{ji2009accelerated} developed an accelerated gradient method that boosts the learning process of trace norm minimization from $O(\frac{1}{\sqrt{k}})$ to $O(\frac{1}{k})$, even to $O(\frac{1}{k^2})$ with the help of Nesterov’s method~\citep{nesterov1983method}. It is noticed that a dual reformation~\citep{pong2010trace} of problem (\ref{eq:lrmtl}) can make it more solvable.
In fact, both the rank penalty and the trace norm can be written in a more general form $\sum_{r=1}^{\min(D, T)} \rho( \sigma_r(\boldsymbol{W}) ) $ where $\rho(\cdot)$ is a penalty function and $\sigma_r(\Wbold)$ is the $r$-th largest singular value of $\Wbold$. When  $\rho(\sigma_r(\boldsymbol{W})) = I(\sigma_r(\boldsymbol{W}) \neq 0 )$, where $I(\cdot)$ is the indicator function, we get the rank penalty which is also the $\ell_0$ norm of the singular values. When $\rho(\sigma_r(\boldsymbol{W})) = \sigma_r(\boldsymbol{W})$, we get the nuclear norm penalty, i.e., the $\ell_1$ norm of the singular values. For $ 0\leq h \leq 1 $, the properties of the $\ell_h$ norm of the singular values, i.e., the Schatten-$h$ quasi-norm penalty, have been investigated in~\citet{rohde2011estimation}. 

Instead of using different power functions of singular values as penalty functions, there are some other variants of the nuclear norm penalty that can lead to more delicate learning of a low-rank matrix. 

The rank of a matrix is defined by the count of its non-zero singular values, meaning that a lower rank corresponds to fewer non-zero singular values. Unlike penalizing all singular values, which the trace norm avoids, it is more desirable and reasonable. This is because the trace norm specifically shrinks only small singular values toward zero, contributing to a more focused and effective regularization approach.
To leave the larger singular values un-penalized, Reduced Rank Multi-Stage Multi-Task Learning (RAMUSA)~\citep{han2016multi} considers the objective function with truncated trace norm~\citep{zhang2012matrix} as 
\begin{equation}
    \min\limits_{\boldsymbol{W}}\frac{1}{2}\sum\limits_{t=1}^{T}\frac{1}{N_t}\|{\boldsymbol{X}^{(t)}}\boldsymbol{w}^{t} - \boldsymbol{y}^{(t)}\|^2_2 + \lambda\sum\limits_{r=1}^{\min(D, T)}\min\{\sigma_r(\boldsymbol{W}), \tau\}.
    \label{ramusa}
\end{equation}
The parameter $\tau$ serves as a threshold of the singular value magnitude, and only those singular values smaller than $\tau$ will get penalized. When $\tau\rightarrow\infty$, problem (\ref{ramusa}) is reduced to the low-rank MTL problem~(\ref{eq:lrmtl}). 
To address this non-convex problem,~\citet{han2016multi} introduce a multi-stage algorithm designed to learn a surrogate upper-bound function. Theoretical proofs affirm its capability for shrinkage, making it an effective approach to tackle the non-convex optimization challenge.

An alternative to the truncated trace norm to relieve the shrinkage on large singular values is the adaptive nuclear norm penalization 
$ \lambda \sum_{r=1}^{R=\min(D, T)} \alpha_r \sigma_r(\boldsymbol{W})$ proposed by~\citet{chen2013reduced}. The weights $\{ \alpha_r \}_{r=1}^R$ are used to adjust for the level of penalization on each singular value, which should be non-negative values and satisfy $ \alpha_1 \leq \ldots \leq \alpha_R$. The explanation is straightforward, i.e., the larger weights on the smaller singular values ensure a greater shrinkage towards 0, while the smaller weights on the larger singular values are helpful in reducing the shrinkage magnitude.

Low-rank methods are useful to achieve dimension reduction by learning a small set of latent variables. However, low-rank methods alone cannot identify which variables are truly predictive of the outcomes. To obtain a more interpretable model, one can assume that not all predictors are affecting the outcomes by adding a sparsity-inducing penalty in addition to a low-rank restriction. In the field of statistics, this line of research has received lots of attention, and variable selection can be achieved by adding a row-wise penalization on the coefficient matrix in a rank-restricted model. For example,~\citet{chen2012sparse} apply a group-lasso type penalty on the rows of the coefficient matrix. Similar work include~\citet{bunea2012joint} and~\citet{she2017selective}. One of the other forms of sparsity structure considered in low-rank models is sparse SVD discussed in~\citet{chen2012reduced} and~\citet{uematsu2019sofar}. Sparse SVD achieves predictor and response selection simultaneously. With a rank $r$, SVD dissects the correlation between responses and predictors, i.e., the coefficient matrix, into $r$ orthogonal channels. The importance of each channel is measured by a singular value, and within each channel, the weights on predictors (responses) are in the corresponding right (left) singular vectors. The sparse SVD can achieve both SVD layer-specific sparsity pattern, by imposing sparsity on elements of each singular vector to find different subsets of predictors/responses that are making effects in each correlation pathway~\citep{chen2012reduced}, and global variable selection, by shrinking all weights related to a certain variable contained in singular vectors to be zeroes~\citep{uematsu2019sofar}.

\paragraph{\underline{Tensor Factorization}}
When we have multiple learning tasks that can be indexed by multi-dimensional indices, instead of stacking all the weight vectors into a matrix of dimension features $\times$ tasks, keeping the structure of the index of tasks by saving the weight vectors into a tensor leads to MultiLinear Multi-Task learning (MLMT)~\citep{wimalawarne2014multitask}. 
MLMT brings us with several advantages compared with the conventional MTL. Firstly, it allows us to keep the inherent structure of the learning tasks so that different dimensions of task similarities can be learned, and the higher-order structures among tasks can be recovered as well. What's more, task imputation (i.e., TL) is made available with MLMT for tasks with no training data~\citep{wimalawarne2014multitask}. The learning problem can be written as
\begin{equation}
\min\limits_{\boldsymbol{\mathcal{W}}}\sum_{t=1}^{T}\frac{1}{N_t}\|{\boldsymbol{X}^{(t)}}\boldsymbol{w}^{t} - \boldsymbol{y}^{(t)}\|^2_2
    \label{tensor1}
\end{equation}
where $\boldsymbol{\mathcal{W}}\in \mathbb{R}^{D \times I_2 \times \cdots \times I_N}$ is a tensor consisting of learning weights $\boldsymbol{w}^{t} \in \mathbb{R}^D$, and the total number of tasks $T= \prod_{j=2}^{N} I_j$.

To exploit task similarities at each dimension, similar to low-rank matrix-based MTL, a multilinear rank restriction can be imposed on the weight tensor. In~\citet{romera2013multilinear}, the authors directly incorporated the rank restriction into the learning task by using a low-rank Tucker decomposition~\citep{kolda2009tensor} of the weight tensor, and the Frobenius norms of Tucker decomposition components are added as regularizations to reduce overfitting. This optimization problem is solved by alternating minimization. 

Alternatively, tensor trace norms are commonly used as a convex approximation to rank restrictions. However, not like the matrix rank, since a tensor rank has no unique definition, various trace norms are developed to fulfill different analysis demands for different anticipated information sharing mechanisms among tasks~\citep{zhang2022learning}. With $R(\boldsymbol{\mathcal{W}})$ denoting a tensor trace norm, the learning task is
\begin{equation}\min\limits_{\boldsymbol{\mathcal{W}}}\frac{1}{2}\sum_{t=1}^{T}\frac{1}{N_t}\|{\boldsymbol{X}^{(t)}}\boldsymbol{w}^{t} - \boldsymbol{y}^{(t)}\|^2_2 + \lambda R(\boldsymbol{\mathcal{W}})
\label{tensor2}
\end{equation}
where $\lambda$ is the tuning parameter to control the magnitude of penalization. 

In general, in the sense of Tucker decomposition or multi-linear SVD~\citep{tomioka2013convex, kolda2009tensor}, tensor trace norms include two categories: the overlapped tensor trace norms and the latent tensor trace norms. The latent trace norm \citep{tomioka2013convex, wimalawarne2014multitask} can be written as 
\begin{equation}
    \|\boldsymbol{\mathcal{W}}\|_{*,latent} = \inf_{\boldsymbol{\mathcal{W}}^{(1)} + \cdots + \boldsymbol{\mathcal{W}}^{(N)}=\boldsymbol{\mathcal{W}}} \sum_{k=1}^N \| \boldsymbol{W}_{(k)}^{(k)} \|_*
\end{equation}
where $\boldsymbol{\mathcal{W}}^{(k)}$ are latent tensors of $\boldsymbol{\mathcal{W}}$ and $\boldsymbol{W}_{(k)}^{(k)}$ denotes a flattened tensor $\boldsymbol{\mathcal{W}}^{(k)}$ along its $k$th axis. Thus, the latent trace norm is the infimum of the summation of the matrix trace norm of flattened latent tensors of $\boldsymbol{\mathcal{W}}$.  To account for the heterogenous multilinear rank and dimensions,~\citet{wimalawarne2014multitask} propose a scaled latent trace norm by adding a weight $I_k^{-1/2}$ to each component $\| \boldsymbol{W}_{(k)}^{(k)} \|_*$. It can identify the dimension with the lowest rank $r_k$ relative to its dimensionality $I_k$. 
The overlapped tensor trace norm~\citep{romera2013multilinear} of a tensor is defined as the weighted sum of nuclear norm of its flattened tensors. 
With different ways of tensor flattening, the overlapped tensor trace norms have different forms, including the Tucker trace norm~\citep{romera2013multilinear} that is a convex combination of matrix trace norms of tensor flattening along each axis in the tensor and the Tensor-Train trace norm~\citep{oseledets2011tensor} that conducts tensor flattening along successive axes starting from the first axis. Given that the feature representation can be factorized into semantic basis vectors and linear coefficients mapping the basis vector space to the original feature vector space, \citet{yang2016deep} introduce the utilization of low-rank tensors in MTL through deep representation learning.

Since most of the overlapped tensor trace norms only make use a subset of all possible flattening of a tensor that reflect different beliefs of the information sharing mechanism among tasks, to search for all the low-rank structures in a weight tensor and unify various overlapped tensor trace norms,~\citet{zhang2022learning} propose a Generalized Tensor Trace Norm (GTTN) which is the convex sum of matrix trace norms of all possible tensor flattening. The combination weights of matrix trace norms of tensor flattenings are treated as unknown variables in the optimization problem to accommodate different levels of importance of each flattening. 

When nonlinear low-rank structures among tasks are expected to achieve better learning performance,~\citet{zhang2022learning} propose the nonlinear GTTN that firstly transforms the rows or columns of each flattened tensor nonlinearly via a neural network and then performs GTTN on the transformed parameters to capture the nonlinear low-rank structure among all the tasks. For models that are nonlinear in the data,~\citet{signoretto2013learning} also provide a kernel-based method for MLMT. 


\begin{applebox}{Remarks}
\begin{enumerate}[leftmargin=0.4cm, label=(\roman*)]
     \item Low-rank structures can achieve both information sharing among tasks and dimension reduction by enforcing all the tasks being affected by the same small set of latent variables extracted from the original feature space.
    \item Sparsity-inducing penalties can be added in addition to the rank restriction to achieve variable selection. 
    \item Keeping the multi-dimensional indices of multiple tasks by saving the weight vectors into a tensor allows us to keep the inherent structure of the learning tasks so that: a. different dimensions of task similarities can be learned; b. the higher-order structures among tasks can be recovered; c. task imputation is made available for tasks with no training data.
\end{enumerate}
\end{applebox}

\subsubsection{Decomposition} Task-relatedness can be learned based on the assumption that similar tasks share the same non-zero elements, and these tasks can acquire richer representations through transformation or low-rank regularization. The decomposition methods discussed in this section aim to capture multiple aspects of task-relatedness, such as sparsity and low-rankness, by decomposing model weights into a sum or product of distinct components. These components not only capture shared information but also task-specific information that benefits each task.
The flexibility of decomposition techniques provides deeper insights into the nature of multitasking, enabling exploration of various combinations of regularizers suitable for different types of multitasking, including the incorporation of irrelevant or outlier tasks. However, decomposition methods have a limitation. The regularization applied to complex components may lead to non-smooth optimization problems involving a large number of variables, which can pose challenges in efficiently solving the devised decomposition problem.
In the MTL setting, the general formalization of decomposition problems can be expressed as
\begin{align}
\label{decomp}
    \min_{\boldsymbol{W}} & \sum\limits_{t=1}^T\mathcal{L}^{(t)}\left(f(\boldsymbol{X}^{(t)}, \boldsymbol{w}^{t}), \boldsymbol{y}^{(t)}\right)+\lambda_1~\text{reg}_1(\boldsymbol{P})+\lambda_2~\text{reg}_2(\boldsymbol{Q}),\nonumber \\
    s.t.~ & \boldsymbol{W}=\boldsymbol{P} + \boldsymbol{Q}~\text{or}~\boldsymbol{W}=\boldsymbol{P}\cdot\boldsymbol{Q},
\end{align}
where the $\text{reg}_1$ and $\text{reg}_2$ are regularizers for the learning of different task-relatedness.

\paragraph{\underline{Form ``$\Pbold+\Qbold$''}}The Dirty Block-Sparse Model~\citep{jalali2010dirty} is introduced by recognizing that block-sparsity regularizers ($\ell_{p,1}$) are influenced by the degree of feature overlap among tasks. Acknowledging the prevalence of dirty high-dimensional data\footnote{It refers to data that are not only high-dimensional (containing a large number of features or attributes) but also contain errors, inaccuracies, or misleading information.} in many multi-task scenarios, this model adeptly addresses the challenges posed by explicitly permitting the decomposition of the weight matrix into element-wise sparse and block-sparse components:
\begin{align}
    \min\limits_{\boldsymbol{W}} \frac{1}{2}\sum\limits_{t=1}^{T}\frac{1}{N_t}\|{\boldsymbol{X}^{(t)}}(\boldsymbol{s}^{t} + \boldsymbol{b}^{t}) - \boldsymbol{y}^{(t)}\|^2_2 + \lambda_1\sum\limits_{d=1}^{D}{\|\boldsymbol{s}_d\|}_1 + \lambda_2\sum\limits_{d=1}^{D}{\|\boldsymbol{b}_d\|}_\infty, \quad s.t.~\boldsymbol{W}=\boldsymbol{S}+\boldsymbol{B},
    \label{eq:dirty_block_sparse}
\end{align}
where the $\boldsymbol{s}^t$ and $\boldsymbol{b}^t$ are the $t$-th columns of $\boldsymbol{S}$ and $\boldsymbol{B}$, respectively. The $\ell_{1,1}$ norm learns an uneven sparse structure~\citep{obozinski2006multi, zhang2006a} while $\ell_{\infty,1}$ norm guarantees features that admit block-wise sparsity to be learned collectively across tasks~\citep{zhang2006a}.~\citet{jalali2010dirty} proves that Eq.~(\ref{eq:dirty_block_sparse}) can match Lasso ($\ell_1$) for no-sharing STL and $\ell_{\infty,1}$ for fully-sharing MTL, and it strictly outperforms both methods elsewhere, including the dirty setting.

Robust Multi-Task Feature Learning (rMTFL)~\citep{gong2012robust} can capture the task-shared features among relevant
tasks and identify outlier tasks simultaneously. Specifically, the weight matrix for all tasks is first decomposed into two components. And then, \citet{gong2012robust} impose the well-known $\ell_{2,1}$ penalty on the first component and the $\ell_{1,2}$ penalty on the second component. Formally, the proposed rMTFL can be formulated as
\begin{align}
    \min\limits_{\boldsymbol{W}} \frac{1}{2}\sum\limits_{t=1}^{T}\frac{1}{N_t}\|{\boldsymbol{X}^{(t)}}\boldsymbol{w}^{t} - \boldsymbol{y}^{(t)}\|^2_2 + \lambda_1\sum\limits_{d=1}^D\|\boldsymbol{p}_d\|_2 + \lambda_2\sqrt{\sum\limits_{d=1}^D\|\boldsymbol{q}_d\|_1^2}, \quad s.t.~\boldsymbol{W} = \boldsymbol{P} + \boldsymbol{Q},
\end{align}
where the penalty applied to the rows of the weight matrices captures shared information, as it selects the same non-zero elements across all tasks. Simultaneously, the penalty on the columns enforces the weights for outlier tasks to be constrained to zero. In \citet{gong2012robust}, a theoretical bound is established to quantify the approximation accuracy of the optimization in relation to the true evaluation. Additionally, error bounds between the estimated weights of rMTFL and the underlying true weights are provided. It is important to note that this method is specifically applicable to MTL settings where some of the tasks are considered outliers.

Robust Multi-Task Learning (RMTL)~\citep{chen2011integrating} addresses real-world applications where certain tasks are irrelevant to other aggregated groups in MTL, impacting the learning performance of different tasks. RMTL is designed to capture task relatedness by learning a low-rank structure while identifying outlier tasks. This approach draws inspiration from previous research on group sparsity~\citep{obozinski2006multi, lee2010adaptive}.
It is formulated as a non-smooth convex optimization problem as
\begin{align}
    \min\limits_{\boldsymbol{W}} \frac{1}{2}\sum\limits_{t=1}^T\frac{1}{N_t}\|{\boldsymbol{X}^{(t)}}(\boldsymbol{p}^t+\boldsymbol{q}^t)-\boldsymbol{y}^{(t)}\|_2^2 + \lambda_1\|\boldsymbol{P}\|_* + \lambda_2\sum\limits_{t=1}^T\|\boldsymbol{q}_t\|_2,\quad s.t.~\boldsymbol{W}=\boldsymbol{P}+\boldsymbol{Q}.
\end{align}
Different from feature selection techniques, $\ell_{2,1}$ norm here is imposed on the columns of the weight matrix. This penalty aims to learn group sparsity of different tasks across all features. It enforces that the weights associated with outlier tasks are constrained to approach zero, thereby diminishing the negative influence of outlier tasks. The low-rank structure encoded in RMTL encapsulates the positive effectiveness, mitigating the impact of outlier tasks. This differs from \citet{hsu2010robust} that focuses on learning both low-rank and sparse structures and provides a theoretically established and unique decomposition. RMTL, on the other hand, simultaneously learns both the low-rank and task-wise sparse structures through an accelerated proximal method (APM)~\citep{nemirovski1994efficient, nesterov1998introductory}. The performance bound of this integrated approach is also proven.


Sparse and Low-Rank Multi-Task Learning~\citep{chen2012learning} also decomposes the weight matrix into a low-rank component and a sparse component. Unlike~\citet{chen2011integrating} that jointly optimizes both structures in the objective function,~\citet{chen2012learning} uses a trace norm constraint to implicitly encourage the low-rank structure. The formulation is
\begin{align}
    \min\limits_{\boldsymbol{W}} \frac{1}{2}\sum\limits_{t=1}^{T}\frac{1}{N_t}\|{\boldsymbol{X}^{(t)}}\boldsymbol{w}^{t} - \boldsymbol{y}^{(t)}\|^2_2 + \lambda\sum\limits_{d=1}^D\|\boldsymbol{p}_d\|_{1}, \quad s.t.~\boldsymbol{W}=\boldsymbol{P}+\boldsymbol{Q}, \|\boldsymbol{Q}\|_*\leq\tau.
    \label{slrmtl}
\end{align}
It is proved to be the tightest convex surrogate function to the non-convex NP-hard problem with a cardinality regularization term ($\ell_0$ norm) and a low-rank constraint. A general projected gradient scheme~\citep{boyd2004convex} is applied to solve this relaxed convex problem (\ref{slrmtl}), which can also be accelerated using Nesterov's method~\citep{nesterov1998introductory}.



\paragraph{\underline{Form ``$\Pbold \cdot \Qbold$''}}
Alternating Structure Optimization (ASO)~\citep{ando2005framework} aims to facilitate structural learning from multiple tasks. By introducing an auxiliary variable $\ubold^{(t)}$ for each task $t$ such that $\ubold^{(t)} = \wbold^{(t)} + \Theta^\top\vbold^{(t)}$, the problem is formulated as
\begin{align}
    \min\limits_{\{\boldsymbol{W}, \boldsymbol{V}\}, \Theta}& \frac{1}{2}\sum_{t=1}^T\frac{1}{N_t}\|{\boldsymbol{X}^{(t)}}\ubold^{(t)} - \boldsymbol{y}^{(t)}\|_2^2 + \lambda\sum_{d=1}^D \|\boldsymbol{w}_d\|_2^2,\nonumber\\
    ~~\text{s.t.}~~& \Theta\Theta^\top=\boldsymbol{I}
    \label{eq:ASO}
\end{align}
The solution process for problem (\ref{eq:ASO}) comprises two steps: fixing $(\Theta, \vbold)$ and then $\ubold$. The first step involves a convex problem, easily addressed by classic optimization methods such as stochastic gradient descent (SGD). The second step can be tackled using singular value decomposition (SVD) along with a series of linear algebra transformations. However, it is important to note that the non-convex ASO algorithm is not guaranteed to converge to a global optimum and may encounter challenges like getting stuck in local optima.

Convex ASO (cASO)~\citep{chen2009convex} investigates the use of convex relaxations to improve the convergence properties of the algorithm and can converge to a global optimum. Firstly, an improved ASO (iASO)  formulation is proposed as an initial non-convex problem
\begin{align}
    \sum_{t=1}^T \frac{1}{N_t} \max & \{\boldsymbol{0}, 1 - (\Xbold^{(t)}\ubold^{(t)})\cdot\ybold^{(t)}\} + \lambda_1 \|\ubold^{(t)} - \Theta^\top\vbold^{(t)}\|^2 + \lambda_2 \|\ubold^{(t)}\|^2, \nonumber \\
    ~~\text{s.t.}~~& \Theta\Theta^\top=\boldsymbol{I}, 
    \label{eq:iASO}
\end{align}
where the intercept is omitted in SVM learner for simplicity. In Eq.~(\ref{eq:iASO}), the constraint terms effectively manage both task relatedness and model complexity. It is noteworthy that the traditional ASO formulation, represented Eq.~(\ref{eq:ASO}), serves as a special case of iASO, irrespective of the loss function choices.

To address the non-convex iASO problem (\ref{eq:iASO}), based on the observation that $\ubold^{(t)}=\Theta^\top\vbold^{(t)}$ minimizes the constraint terms, the formulation of the constraint term can be restructured as
\begin{equation}
    \Gbold(\Ubold,\Theta) = \lambda_1\eta(1-\eta)\text{tr}(\boldsymbol{U}^\top(\eta\boldsymbol{I} + \Theta^\top\Theta)^{-1}\boldsymbol{U}),
\end{equation}
where $\eta = \lambda_2/\lambda_1>0$ and $\Ubold = [\ubold^{(1)}, \cdots, \ubold^{(T)}]$. Thus, the convex ASO formulation can be written as
\begin{align}
    \sum_{t=1}^T & \frac{1}{N_t} \max\{\boldsymbol{0}, 1 - (\Xbold^{(t)}\ubold^{(t)})\cdot\ybold^{(t)}\} + \Gbold(\Ubold, \Theta), \nonumber \\
    ~~\text{s.t.}~~& \Theta\Theta^\top=\boldsymbol{I}.
    \label{eq:cASO}
\end{align}
The convex optimization procedures contain the alternating steps of the estimation of $\Ubold$ with the fixed $\Theta^\top\Theta$ and the estimation of $\Theta^\top\Theta$ with a fixed $\Ubold$. Via the convergence analysis, it is proved that cASO (\ref{eq:cASO}) can converge to a global optimum~\citep{chen2009convex}.

Multi-level Lasso, introduced by \citet{lozano2012multi}, is an approach that relies on the decomposition of the regression coefficients into two components—one shared across all tasks and another designed to capture task-specific features. Specifically,~\citet{lozano2012multi} suppose that the ``global'' sparsity would be controlled by a part of the ``main effect'' variables. Thus, an alternative decomposition is proposed to satisfy the desired property by rewriting $\boldsymbol{w}^{t}$ as
\begin{equation}
\boldsymbol{w}^t_d = \theta_d\boldsymbol{\gamma}^{(t)}_d, d = 1,\cdots,D,
\end{equation}
where $\theta_d$ indicates the ``effect'' from the $d$-th feature, and $\boldsymbol{\gamma}_d^{(t)}$ reflects task specificity. Accordingly, the optimization problem can be written as
\begin{align}
    \min\limits_{\boldsymbol{W}} \frac{1}{2}\sum\limits_{t=1}^{T}\frac{1}{N_t}\|{\boldsymbol{X}^{(t)}}\boldsymbol{w}^{t} - \boldsymbol{y}^{(t)}\|^2_2 + \lambda_1\sum\limits_{d=1}^{D}\theta_d +  \lambda_2\sum\limits_{d=1}^D\|\boldsymbol{\boldsymbol{\gamma}}_d\|_1,\quad s.t.~\boldsymbol{W}=\vec{\boldsymbol{\theta}}\boldsymbol{\Lambda}\boldsymbol{\Gamma}, \vec{\boldsymbol{\theta}}\geq\boldsymbol{0}.
\end{align}

This model accommodates variations in support across multiple tasks while preserving common structures. The optimization process involves iteratively solving for either $\theta$ or $\gamma$ while keeping the other fixed, which is proved to be converged in \citet{lozano2012multi}. The limitation is associated with the alternate optimization procedure of Multi-level Lasso. When learning $\gamma$ while fixing $\theta$, this process essentially becomes a classical Lasso problem, which is relatively easy to solve. However, obtaining the solution for the global problem can be time-consuming, as pointed out in \citet{friedman2007pathwise}.

\begin{applebox}{Remarks}
\begin{enumerate}[leftmargin=0.4cm, label=(\roman*)]
    \item Decomposition methods facilitate the learning of additional task relatedness via imposing different regularizations on the weight components from the decomposition.
    \item Regularizations applied to different components can indeed introduce new challenges in the optimization process when solving the problem.
\end{enumerate}
\end{applebox}

\subsubsection{Priori Sharing}
\label{subsec:prior}
Multi-task priori sharing focuses on understanding and exploiting the relationships between different tasks to improve learning efficiency and performance. This approach is predicated on the idea that tasks, especially those that are related, can provide complementary information that enhances learning when approached collectively rather than in isolation. By identifying and leveraging the priori interconnections among tasks, priori sharing aims to achieve better generalization, more robust models, and improved predictions for each task. 

The typical formulation of priori sharing in MTL is given in the same form as equation (\ref{eq:2.4})
This optimization objective function seeks to minimize a cumulative loss function over $T$ tasks, which is a summation of individual losses for each task's predictions against its true values, adjusted by a global regularization term. The regularization term, $\lambda  \Omega(\boldsymbol{W})$ is then applied to the combined weight vector $\boldsymbol{W}$ which concatenates all task-specific weights $\boldsymbol{w}^{(t)}$, thereby incorporating shared information across tasks into the model. It is designed based on a priori knowledge of task interrelations and enforces certain structure of constraints on $\boldsymbol{W}$ to reflect the assumed relationships between tasks within the model. This formulation allows for the integration of similarities and differences across tasks to inform the learning process, aiming to improve the generalization of the model by leveraging shared patterns and task-specific peculiarities. The categorization of multi-task prior sharing can be broadly understood in the following ways: 

\underline{Task similarity.}
There is compelling evidence supporting the advantages of learning information from multiple task domains compared to single-task data. In earlier studies, such as \citet{evgeniou2004regularized}, and \citet{parameswaran2010large}, the formulation proposed by multi-task relationship learning was all generated based on prior assumptions of task relatedness. Specifically, \citet{evgeniou2004regularized}, and \citet{parameswaran2010large} assumed that the learning tasks are similar to each other and employed task-coupling parameters to model the target average task. In Regularized MTL~\citep{evgeniou2004regularized}, task-coupling parameters were utilized to model the relationships between tasks and extend existing kernel-based single-task methods like support vector machine (SVM) through a novel kernel function. Their formulation is
\begin{align}
    &\min_{\wbold_0, \vbold_0, \xi_{it}} \big\{\sum_{t=1}^T\sum_{i=1}^m \xi_{it} + \frac{\lambda_1}{T} \sum_{t=1}^T \|\vbold_t\|_2^2 + \lambda_2 \|\boldsymbol{w}_0\|_2^2 \big\}, \nonumber\\
    & s.t. \quad y_{it}(\wbold_0 + \vbold_t) \cdot \xbold_{it} \ge 1-\xi_{it}, \,\, \xi_{it} \ge 0, \forall i\in \{1,2,\dots, m\} \,\text{and}\,\, t \in \{1,2,\dots, T\}
    \label{eq:svm_regularized_mtl}
\end{align}
where $m$ represents sample size of data points for each task, $\xi_{it}$ represents the error for each estimation of parameter $\boldsymbol{w}_0+\boldsymbol{v}_t$ generated from the data distribution. They followed the formulation from Hierarchical Bayes \citep{allenby1998marketing, arora1998hierarchical, heskes2000empirical} and described the target T functions as hyperplanes $f_t (x)=\boldsymbol{w}_t \cdot \boldsymbol{x}$, where $\boldsymbol{w}_t=\boldsymbol{w}_0+\boldsymbol{v}_t$ denotes each corresponding target model. In their approach, the authors assume that when learning from tasks that are similar to each other, the discrepancies between different tasks $\boldsymbol{v}_t$ are small, and the task relationships are linked to a common model $\boldsymbol{w}_0$. Additionally, \citet{evgeniou2005learning} and \citet{kato2007multi} provide prior information on the similarities between pairs of tasks and incorporate regularization terms to adjust the learning of multiple tasks in a manner that aligns the distance between model parameters with the distance between tasks. Furthermore, \citet{gornitz2011hierarchical} describes the relationship between tasks using a tree structure, and the model parameters learn the similarity from their parent nodes.

\underline{Task correlation.}
Nevertheless, simply assuming the relationship among tasks without evidence support is somewhat detrimental and may extrapolate the results. By proposing a model that learns task relatedness directly from the data, Bayesian models like \citet{bonilla2007multi} defines prior information over all the unobserved functions for each task and adapts the model parameters regarding the task identities as well as observed information without giving much model assumptions. Particularly, they use multi-task Gaussian Process (GP) prediction techniques to model the correlation among tasks, the formulation is
\begin{align}
    &<f_l(\boldsymbol{x})f_k(\boldsymbol{x}^\top)> = K_{lk}^f k^x <\boldsymbol{x},\boldsymbol{x}^\top>, y_{il} \sim \mathcal{N} (f_l(x_i), \sigma_l^2), l,k \in \{1,\dots,T\}, i \in \{1, \dots N\} \nonumber\\
    &\min_{\boldsymbol{\theta}_X} \bigg(N \log|<F^T(\boldsymbol{K}^x(\boldsymbol{\theta}_x))^{-1} F>| + T\log|\boldsymbol{K}^x(\boldsymbol{\theta}_x)| \bigg),
\end{align}
where they approach this problem by placing a GP prior over the latent functions $\{f_l\}$ to directly induce correlations between tasks, $\boldsymbol{K}^f$ denotes the inter-task dependency via a positive semi-definite (PSD) matrix, $k^x$ denotes the covariance between input data points, and $\sigma_l^2$ refers to the random noise of the $l$-th task, $\boldsymbol{F}$ is the vector of function values corresponding to $\boldsymbol{Y}$. \citet{bonilla2007multi} introduces a novel approach that employs a common covariance function for input features and a 'free-form' covariance matrix for different tasks, offering significant flexibility in modeling diverse data forms and task relationship. Furthermore, the utilization of this 'free-form' covariance matrix mitigates the need for extensive observed data, enhancing the efficiency of the method. To address the overfitting concern stemming from the point estimation approach in \citet{bonilla2007multi}, \citet{zhang2010multi} extended multi-task GP to a weight-space view for the multi-task $t$ process, incorporating an inverse-Wishart prior to modeling the covariance matrix. This adaptation helps mitigate overfitting and enhances the robustness of the method.

\underline{Task covariance.}
In addition to learning through task correlation and task similarities, \citet{zhang2012convex, zhang2014regularization} introduced the concept of Multi-Task Relationship Learning (MTRL) by utilizing a task covariance matrix to capture task relatedness. Within the regularization framework, they derived a convex formulation for multi-task learning, enabling simultaneous learning of model parameters and task relationship. Their innovation lies in the application of a matrix-variate normal prior on the weight matrix $\boldsymbol{W}$, lending a structured prior, alongside certain likelihood functions, to guide the formulation of an objective function that seeks for a posterior solution maximizing the likelihood. The objective function they employed is
\begin{align}
    &\min_{\boldsymbol{W},\boldsymbol{\Omega}} \Lmcal(\boldsymbol{W}) + \lambda_1||\boldsymbol{W}||_F^2 + \lambda_2 tr(\boldsymbol{W} \boldsymbol{\Omega}^{-1}\boldsymbol{W}^T) \notag\\
    &s.t. \quad \boldsymbol{\Omega} \succ 0, tr(\boldsymbol{\Omega}) \le 1, 
\end{align}
where the optimization target they proposed can be expressed as the minimization of a loss function $\Lmcal(\boldsymbol{W})$ augmented by a regularization term scaled by $\lambda_1$ that penalizes the Frobenius norm of $\boldsymbol{W}$, and an additional term scaled by $\lambda_2$ involving the trace of $\boldsymbol{W} \boldsymbol{\Omega}^{-1}\boldsymbol{W}^T$, reflecting the matrix-variate normal prior. Here, $\boldsymbol{\Omega}$ denotes a positive definite matrix capturing task covariance, and its complexity is controlled through constraints ensuring its positive definiteness and bounded trace. This formulation has been established as jointly convex in $\boldsymbol{W},\boldsymbol{\Omega}$, allowing for simultaneous optimization of model parameters and task covariance matrix. 

In essence, their approach extends the principles of single-task learning with regularization while incorporating alternative optimization techniques to achieve a convex objective function. Further developments have extended this framework to enhance multi-task boosting \citep{zhang2012multi} and multi-label learning \citep{zhang2013multilabel}, illustrating its adaptability and potential for a broad spectrum of applications. The approach also offers an interpretative angle from the viewpoint of reproducing kernel Hilbert spaces for vector-valued functions \citep{ciliberto2015learning, jawanpuria2015efficient}, showcasing its theoretical elegance and practical utility. Also, in the context of MTL with a considerable number of tasks, it becomes evident that not all tasks are equally interrelated; many display a tendency toward sparsity in their inter-task relationships. Recognizing that a task may not contribute meaningfully to every other task and that sparse task relationships can mitigate overfitting issues more effectively than dense relationships, there is a growing interest in models that can capture these sparse patterns. \citet{zhang2017learning} pays attention to the elucidation of such sparse task relationships, and the objective function can be written as
\begin{align}
    \min_{\Wbold, \boldsymbol{\Omega} \geq 0} \sum_{t=1}^{T} \frac{1}{N_t} \sum_{j=1}^{N_t} \Lmcal(\boldsymbol{w}_t^\top \boldsymbol{\phi}(x_j), y_j) + \frac{\lambda_1}{2} \text{tr}(\Wbold \boldsymbol{\Omega}^{-1}\boldsymbol{W}^\top) + \lambda_2 \|\boldsymbol{\Omega}\|_1,
\end{align}
where $\boldsymbol{\phi}(\cdot)$ corresponds to the feature mapping, and the learning task refers to $f_t(\boldsymbol{x}) = \boldsymbol{w}_t^\top \phi(\boldsymbol{x})$. 
By adding an $l_1$ regularization on the covariance matrix $\boldsymbol{\Omega}$, their proposed approach, termed the SParse covAriance based mulTi-taSk (SPATS) model, is designed to determine a sparse task covariance structure. This method embraces the $l_1$ regularization, renowned for promoting sparsity, within a regularization framework tailored for MTL. The convex nature of the SPATS model's objective function facilitates the development of an efficient alternating optimization strategy to find the solution.

\begin{applebox}{Remarks}
\begin{enumerate}[leftmargin=0.4cm, label=(\roman*)]
    \item In environments where tasks are interdependent and data is limited or imbalanced, the ability to discern and exploit the latent task interrelations becomes crucial.
    \item Overestimating task similarity can lead to negative transfer, where learning one task may adversely affect the performance of another. Task similarities might change dynamically during training, requiring adaptive models that can adjust to these changes.
    \item Models that heavily rely on task covariances are at risk of overfitting to the specific relations present in the training data, reducing their generalization capabilities.
\end{enumerate}
\end{applebox}
  
\subsubsection{Task Clustering/Grouping}
\label{sec:clustering}
Task relationships can be elucidated through the clustering or grouping of associated tasks, whereby tasks within the same cluster exhibit greater similarities. Executing clustering algorithms at the task level proves particularly advantageous in scenarios with numerous tasks. Typically, task clustering requires leveraging shared structural information across tasks, such as task similarity or distance. These are termed horizontal methods contrasting with hierarchical methods that harness inherent task structures, such as tree formations, to achieve MTL. Task priori sharing and clustering are closely related as both share the commonness across tasks, but clustered structure is an unknown priori that needs to be learned. For example, the problem defined in Eq.~(\ref{eq:svm_regularized_mtl}) could also be equivalent to solving the following optimization problem (See proof in~\citet[Page 3]{evgeniou2004regularized}):
\begin{align}
    &\min_{\wbold_t, \xi_{it}} \big\{\sum_{t=1}^T\sum_{i=1}^m \xi_{it} + \frac{\lambda_1\lambda_2}{T(\lambda_1 + \lambda_2)} \sum_{t=1}^T \|\wbold_t\|^2 + \frac{\lambda_1^2}{T(\lambda_1 + \lambda_2)} \sum_{t=1}^T\|\wbold_t - \frac{1}{T}\sum_{s=1}^T\wbold_s\|^2 \big\}, \nonumber\\
    & s.t. \quad y_{it}\cdot\wbold_t \cdot \xbold_{it} \ge 1-\xi_{it}, \,\, \xi_{it} \ge 0,
    \label{eq:svm_regularized_mtl_2}
\end{align}
where $\wbold_t = \wbold_0 + \vbold_t$ (see Eq.~(\ref{eq:svm_regularized_mtl})). The second regularization term in Eq.~(\ref{eq:svm_regularized_mtl_2}) implies that all tasks are clustered into a single group, and the parameters across all tasks are constrained to exhibit maximum similarity. This special case shows that all tasks are clustered into one group. In practice, however, it is worth noting that certain related tasks might frequently be clustered into different groups.

\paragraph{\underline{Horizontal Methods}} Clustered Multi-Task Learning (CMTL)~\citep{zhou2011clustered} assumes that multiple tasks in the same cluster are similar to each other, and provides the insights of inherent relationships between ASO~\citep{ando2005framework} and CMTL. Specifically, the CMTL is non-convex, and the proposed convex relaxation of CMTL is equivalent to an existing convex relaxation of ASO. The objective function of CMTL can be formulated as
\begin{align}
    \min\limits_{\boldsymbol{W}, \boldsymbol{F}} & \frac{1}{2}\sum_{t=1}^T\frac{1}{N_t}\|{\boldsymbol{X}^{(t)}}\boldsymbol{w}^t - \boldsymbol{y}^t\|_2^2 + \lambda_1(\text{tr}(\boldsymbol{W}^\top\boldsymbol{W}) - \text{tr}(\boldsymbol{F}^\top\boldsymbol{W}^\top\boldsymbol{W}\boldsymbol{F})) + \lambda_2\sum_{t=1}^{T}{\|\boldsymbol{w}^t\|}^2_2, \nonumber\\
    \text{s.t.}~~& \boldsymbol{F}_{t,j}=1/\sqrt{n_j}~\text{if}~t\in\mathcal{C}_j~\text{otherwise}~0, t = 1, \cdots, T,
\end{align}
    
where $n_j$ is the \#task in the $j$-th cluster $\mathbf{\mathcal{C}}_j$.

\paragraph{\underline{Hierarchical Methods}}

TAsk Tree (TAT)~\citep{han2015learning} model is the first method for MTL to learn the tree structure under the regularization framework. By specifying the number of tree layers as $H$, \citet{han2015learning} utilizes matrix decomposition to learn model weights for each layer, i.e., $\{\Wbold_h\}_{h=1}^H$. TAT devises sequential constraints on the distance between the consecutive weight matrices over tree layers. By combining the loss functions, its learning objective can be shown as:
\begin{align}
    \min\limits_{\boldsymbol{W}} & \frac{1}{2}\sum_{t=1}^T\frac{1}{N_t}\|{\boldsymbol{X}^{(t)}}\sum_{h=1}^H\boldsymbol{w}_h^t - \boldsymbol{y}^t\|_2^2 + \sum_{h=1}^H\lambda_h\sum_{i<j}^T\|\wbold_{h}^i - \wbold_{h}^j\|^2_2, \nonumber \\
    \text{s.t.} & |\wbold_{h-1}^i - \wbold_{h-1}^j| \succeq |\wbold_{h}^i - \wbold_{h}^j|, \forall h \geq 2, \forall i<j,
\end{align}
where the hyperparameters $\{\lambda_h\}_{h=1}^H$ indicate the importance of different tree layers, and $|\cdot|$ and $\succeq$ denotes the elementwise operation.
This sequential constraint encourages a non-increasing order for the pair distance between tasks from bottom to top.

\begin{applebox}{Remarks}
\begin{enumerate}[leftmargin=0.4cm, label=(\roman*)]
    \item Task clustering methods are scalable with respect to the number of tasks in MTL.
    \item Both clustering and priori sharing methods in MTL carry similar underlying meanings as they inherently decipher task relationships.
    \item Task clustering complements other MTL strategies, as any MTL approach can be implemented within the task clusters.
    \item Solutions in this section tend to be suboptimal, given that task clustering is not exclusive.
\end{enumerate}
\end{applebox}

%% file: tex_files/02-1/stl-mtl-framework.tex
\begin{figure*}[t]
    \centering
    \begin{subfigure}{0.79\textwidth}
        \includegraphics[width=1\textwidth]{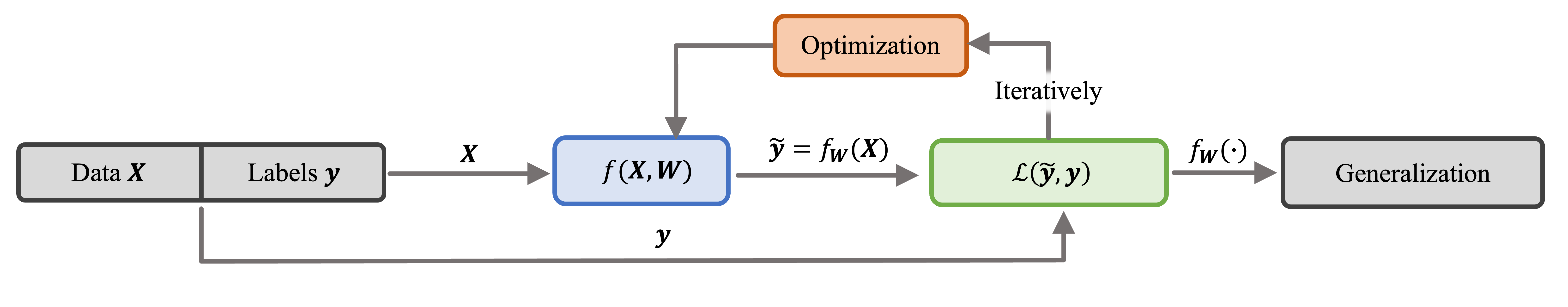}
        \caption{Single-Task Learning (STL).}
        \label{stl}
    \end{subfigure}
    \begin{subfigure}{0.79\textwidth}
        \includegraphics[width=1\textwidth]{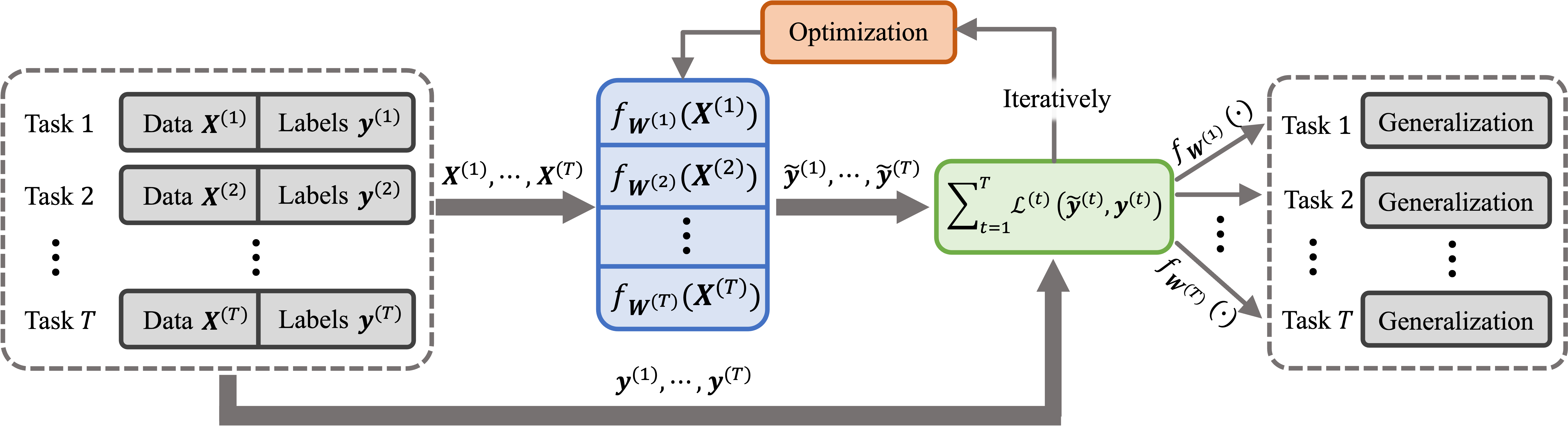}
        \caption{Multi-Task Learning (MTL).}
        \label{mtl}
    \end{subfigure}
    \caption{The comparison of general framework between STL and MTL. (a) In STL, the learning function $f$ is trained on a single dataset ${(\Xbold, \Ybold)}$, where $\Xbold$ represents the input data and $\ybold$ represents the corresponding labels. The function $f$ is parametrized by $\Wbold$, and is trained to minimize a predefined loss function $\Lmcal(\Tilde{\Ybold}, \Ybold)$, where $\Tilde{\Ybold}$ is the prediction value. Once $f$ is trained, it can be used to generalize to unseen data. (b) In MTL, the learning pipeline is similar to STL, but instead of training on a single dataset, multiple datasets are combined for different tasks. The multiple tasks are learned jointly by optimizing multiple loss functions $\Lmcal^{(1)}(\Tilde{\Ybold}^{(1)}, \Ybold^{(1)}), ..., \Lmcal^{(T)}(\Tilde{\Ybold}^{(T)}, \Ybold^{(T)})$ simultaneously. It should be noted that although multiple tasks are learned jointly, the generalization of each task can still be performed independently.}
    \label{stl-mtl}
\end{figure*}

%% file: tex_files/02-1/io-config.tex
\begin{figure*}[t]
    \centering
    \begin{subfigure}{0.26\textwidth}
        \includegraphics[width=1\textwidth]{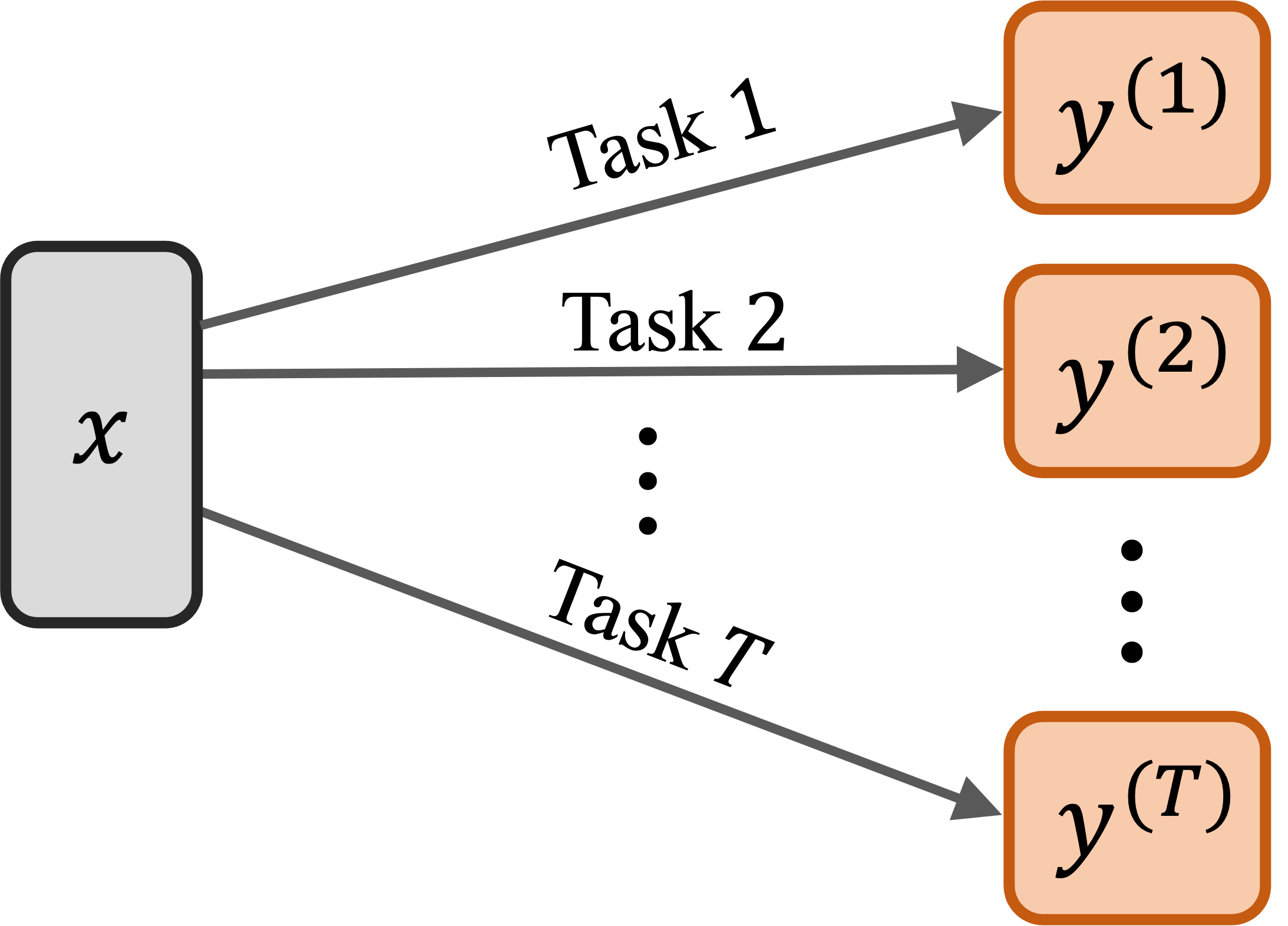}
        \caption{SIMO.}
        \label{simo}
    \end{subfigure}\quad\quad
    \begin{subfigure}{0.26\textwidth}
        \includegraphics[width=1\textwidth]{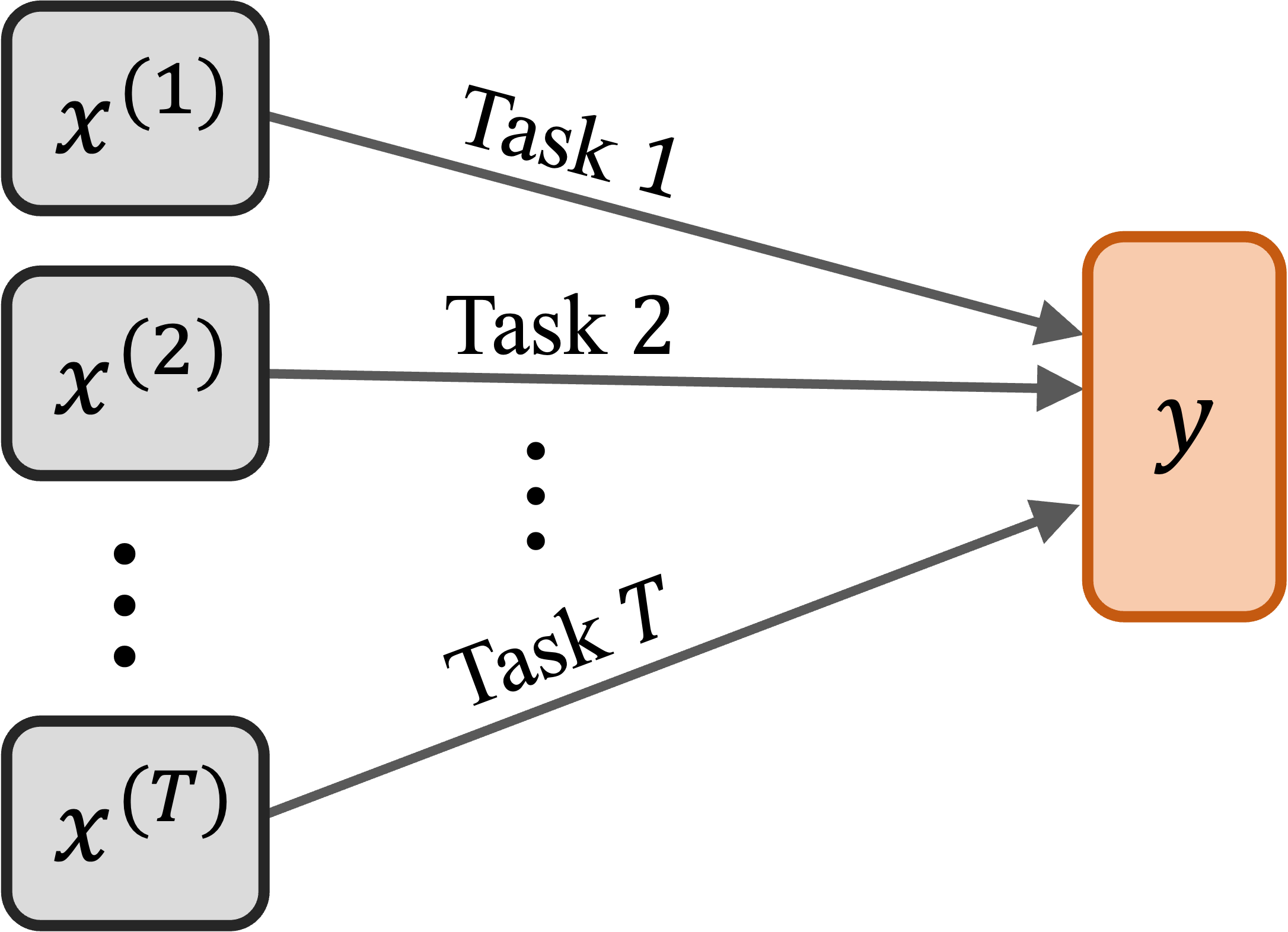}
        \caption{MISO.}
        \label{miso}
    \end{subfigure}\quad\quad
    \begin{subfigure}{0.26\textwidth}
        \includegraphics[width=1\textwidth]{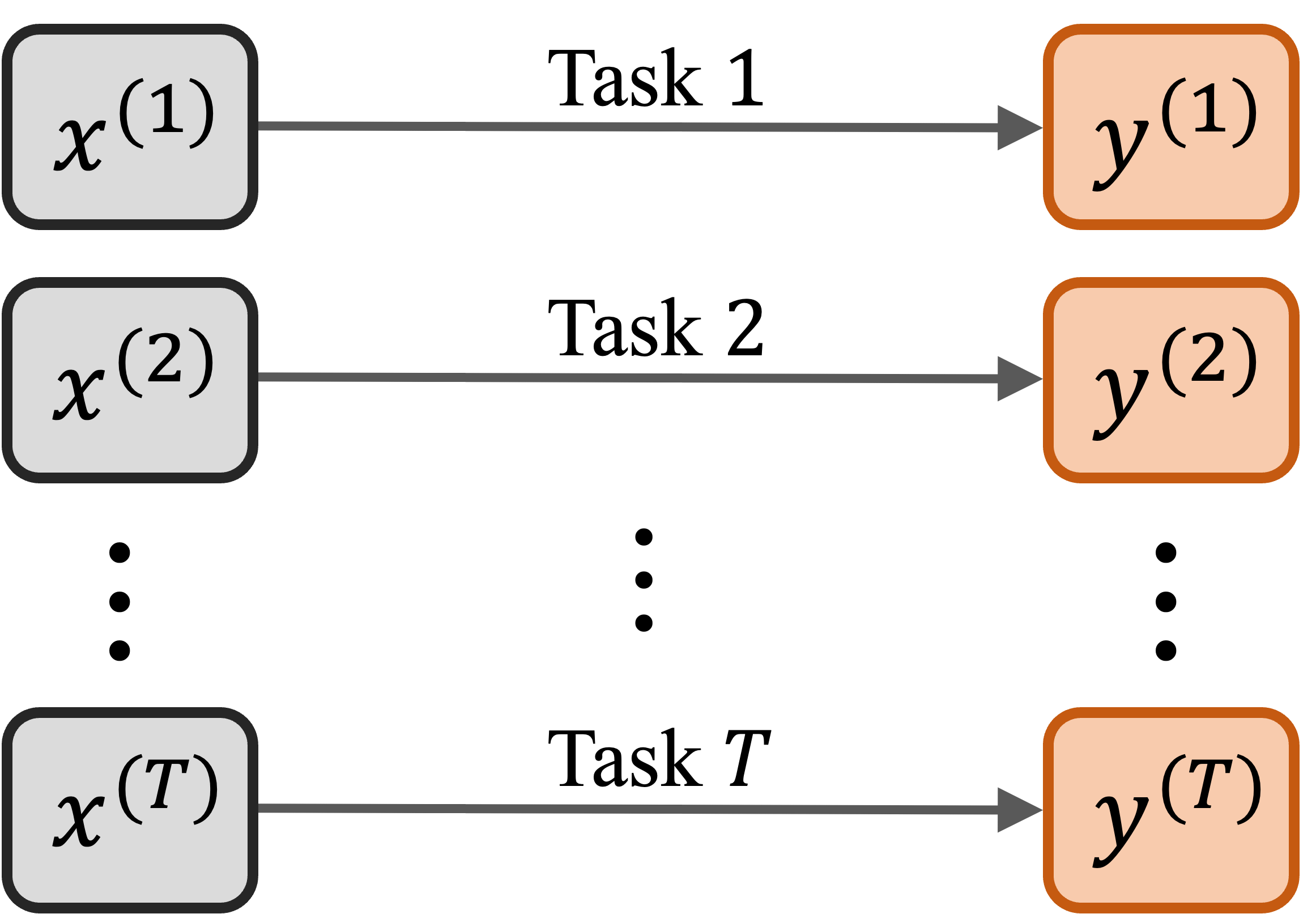}
        \caption{MIMO.}
        \label{mimo}
    \end{subfigure}
    \caption{The classification of MTL problems into three different input/output configurations: (a) single-input multi-output (MISO), (b) multi-input single-output (MISO), and (c) multi-input multi-output (MISO).}
    \label{io-config}
\end{figure*}

%% file: table_files/summary.tex
\begin{table*}[t]
    \centering
    \tiny
    \caption{Summary of MTL methods discussed in \textbf{\S}~\ref{sec:mtlmodel}.}
    \scalebox{0.67}{
    \midsepremove
    \begin{tabular}{ccccccccccccc}
    \hline
    \rowcolor[HTML]{D0D0D0} 
    \multicolumn{2}{c}{\cellcolor[HTML]{D0D0D0}} &
      \cellcolor[HTML]{D0D0D0} &
      \multicolumn{3}{c}{\cellcolor[HTML]{D0D0D0}I/O} &
      \multicolumn{4}{c}{\cellcolor[HTML]{D0D0D0}Data Modality} &
      \multicolumn{3}{c}{\cellcolor[HTML]{D0D0D0}Task Type} \\ \cline{4-13} 
    \rowcolor[HTML]{D0D0D0} 
    \multicolumn{2}{c}{\multirow{-2}{*}{\cellcolor[HTML]{D0D0D0}MTL Strategy}} &
      \multirow{-2}{*}{\cellcolor[HTML]{D0D0D0}Assumption} &
      SIMO &
      MISO &
      MIMO &
      Table &
      Image &
      Text &
      Graph &
      Regression &
      Classification &
      Dense Prediction \\ \toprule
                                                                    & Feature Selection            & \ref{assump:parameter} & \cmark & & & \cmark & & & & \cmark & & \xmark \\
                                                                    & Decomposition                & \ref{assump:parameter} & \cmark & & & \cmark & & & & \cmark & \cmark & \\
    \multirow{-3}{*}{Regularization}                                & Low-Rank Factorization       & \ref{assump:parameter} & \cmark & & & \cmark & \cmark & & & \cmark & \cmark & \\ \midrule

    \rowcolor[HTML]{F0F0F0} 
    \cellcolor[HTML]{F0F0F0}                                        & Priori Sharing               & \ref{assump:parameter} & \cmark & \cmark & & \cmark & \cmark & & & \cmark & \cmark & \\
    \rowcolor[HTML]{F0F0F0} 
    \cellcolor[HTML]{F0F0F0}                                        & Task Clustering/Grouping     & \ref{assump:parameter} & \cmark & & & \cmark & & &  & \cmark & & \\
    \rowcolor[HTML]{F0F0F0} 
    \cellcolor[HTML]{F0F0F0}                                        & Group-Based Learning         & \ref{assump:parameter} & \cmark & \xmark & \xmark & & \cmark & & & \cmark & \cmark & \\
    \rowcolor[HTML]{F0F0F0} 
    \multirow{-4}{*}{\cellcolor[HTML]{F0F0F0}Relationship Learning} & Mixture-of-Experts           & \ref{assump:parameter} & \cmark & & & \cmark & \cmark & & & \cmark & \cmark & \\ \midrule

                                                                    & Feature Fusion               & \ref{assump:feature} & \cmark & \xmark & \xmark & & \cmark & \cmark & & \cmark & \cmark & \cmark \\
                                                                    & Cascading                    & \ref{assump:feature} & \cmark & \xmark & \xmark & & \cmark & \cmark & & \cmark & \cmark & \cmark \\
                                                                    & Knowledge Distillation       & \ref{assump:feature} & \cmark & \xmark & \xmark & \cmark & \cmark & & & \cmark & \cmark & \cmark \\
    \multirow{-4}{*}{Feature Propagation}                           & Cross-Task Attention         & \ref{assump:feature} & \cmark & \xmark & \xmark & \xmark & \cmark & & & \cmark & \cmark & \cmark \\ \midrule

    \rowcolor[HTML]{F0F0F0} 
    \cellcolor[HTML]{F0F0F0}                                        & Scalarization                & \ref{assump:optimization} & \cmark  & &  &  \cmark & \cmark  & \cmark & \cmark  & \cmark  & \cmark  & \cmark \\
    \rowcolor[HTML]{F0F0F0} 
    \cellcolor[HTML]{F0F0F0}                                        & Multi-Objective Optimization & \ref{assump:optimization} & \cmark  &  &  &  \cmark & \cmark  & \cmark & \cmark   & \cmark   & \cmark &  \cmark\\
    \rowcolor[HTML]{F0F0F0} 
    \cellcolor[HTML]{F0F0F0}                                        & Adversarial Training         & \ref{assump:optimization} & \cmark & \cmark & & & \cmark & \cmark & & \cmark & \cmark & \cmark \\
    \rowcolor[HTML]{F0F0F0}
    \multirow{-4}{*}{\cellcolor[HTML]{F0F0F0}Optimization}          & Neural Architecture Search   & \ref{assump:parameter} & \cmark & & & & \cmark & & & \cmark & \cmark & \cmark \\ \midrule

                                                                    & Downstream Fine-tuning       & \ref{assump:parameter} & \cmark & \xmark & \xmark & \cmark & \cmark & \cmark & \cmark & \cmark & \cmark & \cmark\\
                                                                    & Task Prompting               & \ref{assump:parameter} & \cmark & \cmark & \cmark & \cmark & \cmark & \cmark &  & \cmark & \cmark & \cmark \\
    \multirow{-3}{*}{Pre-training}                                  & Multi-Modal Unification      & \ref{assump:parameter} & \cmark & \cmark & \cmark & \cmark & \cmark & \cmark & & \cmark & \cmark & \cmark \\ \bottomrule
    \end{tabular}}
    \label{tab:summary}
    \raggedright
    \scalebox{0.6}{
        \footnotesize
        \cmark \color{black}~indicates \color{mydarkgreen}common practice \color{black}in the research community.
        \xmark \color{black}~indicates \color{mydarkorange}not applicable \color{black}due to technical constraints.
    }
\end{table*}

%% file: table_files/notation.tex
\begin{table*}[t]
    \centering
    \tiny
    \caption{Summary of basic notations used in this paper.}
    \midsepremove
\scalebox{0.80}{
    \begin{tabular}{ll}
    \toprule
    \rowcolor{gray!40}
        Notation & Description \\
    \midrule
        $n, N \in \mathbb{R}$ & Scalars are denoted by plain lowercase or uppercase letters.\\
    \midrule
    \rowcolor{gray!20}
        \#object & The number of object, e.g. \#task denoting the number of task.\\
    \midrule
        $\boldsymbol{x}$ or $\vec{\boldsymbol{x}} \in \mathbb{R}^{N}$ & A vector $\boldsymbol{x}$ with $N$ entries, denoted by bold lowercase letters.\\
    \midrule
    \rowcolor{gray!20}
        $\boldsymbol{X} \in \mathbb{R}^{M \times N}$ & A matrix $\boldsymbol{X}$ with size $M \times N$, denoted by bold uppercase letters. \\
    \midrule
        $\boldsymbol{\mathcal{X}} \in \mathbb{R}^{I_1 \times \cdots \times I_N}$ & A tensor $\boldsymbol{\mathcal{X}}$ with size $\mathbb{R}^{I_1 \times \cdots \times I_N}$, denoted by bold calligraphic letters. \\
    \midrule
    \rowcolor{gray!20}
        $\{\star^{(i)}\}_{i=1}^N$ & A set contains $\star^{(1)}, \cdots, \star^{(N)}$, where $\star$ could be anything, e.g., scalar, vector, data pair, learner, etc. \\
    \midrule
        $x_n \in \mathbb{R}$ & The $n$-th entry for vector $\boldsymbol{x} \in \mathbb{R}^{N}, n \in \{1, 2, \cdots, N\}$.\\
    \midrule
    \rowcolor{gray!20}
        $x_{m,n}$ or $[\boldsymbol{X}]_{m,n} \in \mathbb{R}$ & The $(m, n)$-th entry of matrix $\boldsymbol{X} \in \mathbb{R}^{M \times N}, m \in \{1, 2, \cdots, M\}, n \in \{1, 2, \cdots, N\}$.\\
    \midrule
        $\boldsymbol{X} \odot \boldsymbol{Y} \in\mathbb{R}^{M \times N}$ & Element-wise product of $\boldsymbol{X}\in\mathbb{R}^{M \times N}$ and $\boldsymbol{Y}\in\mathbb{R}^{M \times N}$, which means the $(m,n)$-th entry of $\boldsymbol{X}\odot\boldsymbol{Y}$ is $x_{m,n}y_{m,n}$. \\
    \midrule
    \rowcolor{gray!20}
        $\boldsymbol{x}^n \in \mathbb{R}^{M}$ & The $n$-th column vector of matrix $\boldsymbol{X} \in \mathbb{R}^{M \times N}, n \in \{1, 2, \cdots, N\}$.\\
    \midrule
        $\boldsymbol{x}_m \in \mathbb{R}^{N}$ & The $m$-th row vector of matrix $\boldsymbol{X} \in \mathbb{R}^{M \times N}, m \in \{1, 2, \cdots, M\}$.\\
    \midrule
    \rowcolor{gray!20}
        $\boldsymbol{I}_{N\times N}\in\mathbb{R}^{N\times N}$ & The identity matrix of size $N\times N$, which has ones on the diagonal and zeros elsewhere.\\
    \midrule
        tr$(\boldsymbol{X}) \in \mathbb{R}$ & The trace of a matrix $\boldsymbol{X}\in\mathbb{R}^{N\times N}$, defined as the sum of its $N$ components on the diagonal.\\
    \midrule
    \rowcolor{gray!20}
        col$(\boldsymbol{X}) \subseteq \mathbb{R}^M$ & The column space of a matrix $\boldsymbol{X}\in\mathbb{R}^{M\times N}$, which consists of all linear combinations of its column vectors.\\
    \midrule
        rank$(\boldsymbol{X}) \in \mathbb{R}$ & The rank of matrix $\boldsymbol{X}$, defined as the maximum number of linearly independent column (or row) vectors of $\boldsymbol{X}$.\\
    \midrule
    \rowcolor{gray!20}
        vec$(\boldsymbol{X}) \in \mathbb{R}^{MN} $ & The vectorization of the matrix $\boldsymbol{X} \in \mathbb{R}^{M\times N}$ in the row-by-row stacking way. \\
    \midrule
        $\boldsymbol{D}^{+} \in \mathbb{R}^{N\times M}$ & The pseudoinverse of a matrix $\boldsymbol{D}\in\mathbb{R}^{M\times N}$.\\
    \midrule
    \rowcolor{gray!20}
        $\boldsymbol{O}^N\subset\mathbb{R}^{N\times N}$ & The set of $N\times N$ orthogonal matrices.\\
    \midrule
        $\boldsymbol{X}\in\boldsymbol{O}^N$ & The column vectors $\boldsymbol{x}^1, \cdots, \boldsymbol{x}^N$ of matrix $\boldsymbol{X}$ are orthogonal.\\
    \midrule
    \rowcolor{gray!20}
        $\boldsymbol{S}^N\subset\mathbb{R}^{N\times N}$ & The set of $N\times N$ real symmetric matrices.\\
    \midrule
        $\boldsymbol{S}_{+}^N\subset\boldsymbol{S}^N$ & The subset of $\boldsymbol{S}^N$ that contains positive semidefinit matrices.\\
    \midrule
    \rowcolor{gray!20}
        $\|\boldsymbol{w}\|_1$ & The $\ell_1$ norm of a vector, calculated as the sum of the absolute vector values.\\
    \midrule
        $\|\boldsymbol{w}\|_2$ & The $\ell_2$ norm of a vector, calculated as the square root of the sum of the squared vector values.\\
    \midrule
    \rowcolor{gray!20}
        $\|\boldsymbol{w}\|_{\infty}$ & The $\ell_{\infty}$ norm of a vector, calculated as the maximum of the absolute vector values.\\
    \midrule
         $\|\boldsymbol{W}\|_{0}$ & The $\ell_0$ norm, i.e., cardinality of a matrix, defined as the number of nonzero components.\\
    \midrule
    \rowcolor{gray!20}
        $\|\boldsymbol{W}\|_{1}$ & The $\ell_1$ norm of a matrix, calculated as the maximum of the $\ell_1$ norm of the column vectors.\\
    \midrule
        $\|\boldsymbol{W}\|_{2}$ & The $\ell_2$ norm of a matrix, calculated as its maximum singular value.\\
    \midrule
    \rowcolor{gray!20}
        $\|\boldsymbol{W}\|_{F}$ & The Frobenius norm of a matrix, calculated as the  square root of the sum of the squared matrix values.\\
    \midrule
        $\{\sigma_r(\boldsymbol{W})\}_{r=1}^R$ & The set of non-increasing ordered singular values of matrix $\boldsymbol{W}$.\\
    \midrule
    \rowcolor{gray!20}
        $\|\boldsymbol{W}\|_{*}$ & The trace norm of a matrix, defined as the sum of its singular values, i.e., $\sum_{r=1}^R\sigma_r(\boldsymbol{W})$.\\
    \midrule
        $\|\boldsymbol{W}\|_{\infty}$ & The $\ell_\infty$ norm of a matrix, calculated as the maximum of the $\ell_1$ norm of the row vectors.\\
    \midrule
    \rowcolor{gray!20}
        $\|\boldsymbol{W}\|_{p,q}$ & The $\ell_{p, q}$ norm of a matrix, defined as the $q$-norm of the vector whose components are $p$-norm of $~\boldsymbol{W}$'s row vectors. \\
    \midrule
        $\|\boldsymbol{W}\|_{1,1}$ & The $\ell_{1, 1}$ norm of a matrix, defined as the sum of the absolute matrix components.\\
    \midrule
    \rowcolor{gray!20}
        $\|\boldsymbol{W}\|_{1,2}$ & The $\ell_{1,2}$ norm of a matrix, calculated as the $\ell_2$ norm of the vector whose components are $\ell_1$ norm of the row vectors. \\
    \midrule
        $\|\boldsymbol{W}\|_{2,1}$ & The $\ell_{2,1}$ norm of a matrix, calculated as the sum of the $\ell_2$ norm of the row vectors. \\ 
    \bottomrule
    \end{tabular}
}
    \label{tab:notation}
\end{table*}

%% file: table_files/regularizer.tex
\begin{table*}[t]
    \centering
    \scriptsize
    \caption{Summary of regularization technique used in MTL.}
    \label{tab:regularizer}
    \midsepremove
    \scalebox{0.4}{
    \begin{threeparttable}
    \begin{tabular}{llllll}
    \toprule
    \rowcolor{gray!40}
        Model Name & Origin & Year & Type & Matrix Regularizer & Vector Formalization  \\
    \toprule
        Regularized MTL & KDD & \citeyear{evgeniou2004regularized} & Group regularization & Frobenius norm & $\min\limits_{\boldsymbol{W}}\frac{1}{2}\sum_{t=1}^{T}\frac{1}{N_t}\|{\boldsymbol{X}^{(t)}}\boldsymbol{w}^{t} - \boldsymbol{y}^{(t)}\|^2_2 + \lambda_1\sum_{t=1}^{T}{\|\boldsymbol{w}^t - \frac{1}{T}\sum_{t=1}^T\boldsymbol{w}^t\|}^2_2 + \lambda_2\sum_{t=1}^{T}{\|\boldsymbol{w}^t\|}^2_2$ \\
    \midrule
    \rowcolor{gray!20}
        Learning Multiple Tasks with Kernel Methods & JMLR & \citeyear{evgeniou2005learning} & Priori Sharing & Adaptive penalty & $\min\limits_{\boldsymbol{V}, \boldsymbol{W}}\frac{1}{2}\sum_{t=1}^{T}\frac{1}{N_t}\|{\boldsymbol{X}^{(t)}}\boldsymbol{w}^{t} - \boldsymbol{y}^{(t)}\|^2_2 + \lambda\sum_{t=1}^T{\boldsymbol{w}^t}^\top\boldsymbol{V}^{+}\boldsymbol{w}^t,$~~s.t.~$\boldsymbol{V}\in\boldsymbol{S}_+^D,$ $\boldsymbol{V}\in\boldsymbol{S}^D$ \\
    \midrule
        Alternating structure optimization & JMLR & \citeyear{ando2005framework} & Decomposition & Frobenius norm & $\min\limits_{\{\boldsymbol{W}, \boldsymbol{V}\}, \Theta}\frac{1}{2}\sum_{t=1}^T\frac{1}{N_t}\|{\boldsymbol{X}^{(t)}}(\boldsymbol{w}^t+\Theta^\top\boldsymbol{v}^t) - \boldsymbol{y}^t\|_2^2 + \lambda\sum_{d=1}^D \|\boldsymbol{w}_d\|_2^2$,~~s.t.~$\Theta\Theta^\top=\boldsymbol{I}_{h\times h}$ \\
    \midrule
    \rowcolor{gray!20}
        Multi-task feature selection & Tech. Rep.\tnote{1} & \citeyear{obozinski2006multi} & Group-sparse learning & $\ell_{2,1}$ norm & $\min\limits_{\boldsymbol{W}}\frac{1}{2}\sum_{t=1}^{T}\frac{1}{N_t}\|{\boldsymbol{X}^{(t)}}\boldsymbol{w}^{t} - \boldsymbol{y}^{(t)}\|^2_2 + \lambda\sum_{d=1}^{D}{\|\boldsymbol{w}_d\|}_2$  \\
    \midrule
        Multi-task Lasso & Thesis\tnote{2} & \citeyear{zhang2006a} & Group-sparse learning & $\ell_{\infty,1}$ norm & $\min\limits_{\boldsymbol{W}}\frac{1}{2}\sum_{t=1}^{T}\frac{1}{N_t}\|{\boldsymbol{X}^{(t)}}\boldsymbol{w}^{t} - \boldsymbol{y}^{(t)}\|^2_2 + \lambda\sum_{d=1}^{D}{\|\boldsymbol{w}_d\|}_\infty$ \\ 
    \midrule
    \rowcolor{gray!20}
        Multi-task feature learning & NeurIPS & \citeyear{argyriou2006multi} & Group-sparse learning, feature learning & $\ell_{2,1}$ norm & $\min\limits_{\boldsymbol{U}, \boldsymbol{W}}\frac{1}{2}\sum_{t=1}^{T}\frac{1}{N_t}\|({\boldsymbol{X}^{(t)}}\boldsymbol{U})\boldsymbol{w}^{t} - \boldsymbol{y}^{(t)}\|^2_2 + \lambda(\sum_{d=1}^{D}{\|\boldsymbol{w}_d\|}_2)^2$,~~s.t.~$\boldsymbol{U}\in\boldsymbol{O}^D$ \\
    \midrule
        Convex multi-task feature learning & Mach. Lea. & \citeyear{argyriou2008convex} & Feature learning & Adaptive penalty & $\min\limits_{\boldsymbol{V}, \boldsymbol{W}}\frac{1}{2}\sum_{t=1}^{T}\frac{1}{N_t}\|{\boldsymbol{X}^{(t)}}\boldsymbol{w}^{t} - \boldsymbol{y}^{(t)}\|^2_2 + \lambda\sum_{t=1}^T{\boldsymbol{w}^t}^\top\boldsymbol{V}^{+}\boldsymbol{w}^t,$~~s.t.~$\boldsymbol{V}\in\boldsymbol{S}_+^D,$ tr$(\boldsymbol{V})\leq1$, col$(\boldsymbol{W})\subseteq$col$(\boldsymbol{V})$ \\
    \midrule
    \rowcolor{gray!20}
        Low rank MTL & ICML & \citeyear{ji2009accelerated} & Low-rank learning & Trace norm & $\min\limits_{\boldsymbol{W}}\frac{1}{2}\sum_{t=1}^{T}\frac{1}{N_t}\|{\boldsymbol{X}^{(t)}}\boldsymbol{w}^{t} - \boldsymbol{y}^{(t)}\|^2_2 + \lambda\|\boldsymbol{W}\|_*$ \\
    \midrule
        Convex ASO & ICML & \citeyear{chen2009convex} & --- & --- & $\min\limits_{\boldsymbol{U}, \Theta}\frac{1}{2}\sum_{t=1}^T\frac{1}{N_t}\|{\boldsymbol{X}^{(t)}}\boldsymbol{u}^t - \boldsymbol{y}^t\|_2^2 + \lambda\eta(1-\eta)\text{tr}(\boldsymbol{U}^\top(\eta\boldsymbol{I} + \Theta^\top\Theta)^{-1}\boldsymbol{U}),~~s.t.~\Theta\Theta^\top=\boldsymbol{I}_{h\times h}$ \\
    \midrule
    \rowcolor{gray!20}
        Dirty block-sparse model & NeurIPS & \citeyear{jalali2010dirty} & Group-sparse learning, decomposition & $\ell_{\infty, 1}$ norm $+$ $\ell_{1,1}$ norm & $\min\limits_{\boldsymbol{W}}\frac{1}{2}\sum_{t=1}^{T}\frac{1}{N_t}\|{\boldsymbol{X}^{(t)}}(\boldsymbol{s}^{t} + \boldsymbol{b}^{t}) - \boldsymbol{y}^{(t)}\|^2_2 + \lambda_1\sum_{d=1}^{D}{\|\boldsymbol{s}_d\|}_1 + \lambda_2\sum_{d=1}^{D}{\|\boldsymbol{b}_d\|}_\infty$,~~s.t.~$\boldsymbol{W}=\boldsymbol{S}+\boldsymbol{B}$ \\
    \midrule
        Sparse multi-task Lasso & NeurIPS & \citeyear{lee2010adaptive} & Group-sparse learning & Weighted $\ell_{2,1}$ norm $+$ weighted $\ell_{1,1}$ norm & $\min\limits_{\boldsymbol{W}}\frac{1}{2}\sum_{t=1}^{T}\frac{1}{N_t}\|{\boldsymbol{X}^{(t)}}\boldsymbol{w}^{t} - \boldsymbol{y}^{(t)}\|^2_2 + \lambda_1\sum_{d=1}^{D}\rho_d{\|\boldsymbol{w}_d\|}_2 + \lambda_2\sum_{d=1}^D\theta_d{\|\boldsymbol{w}_d\|}_1$\\
    \cdashlinelr{1-6} 
        & & & & Weighted $\ell_{2,1}$ norm $+$ weighted $\ell_{1,1}$ norm & $\min\limits_{\boldsymbol{W}}\frac{1}{2}\sum_{t=1}^{T}\frac{1}{N_t}\|{\boldsymbol{X}^{(t)}}\boldsymbol{w}^{t} - \boldsymbol{y}^{(t)}\|^2_2 + \lambda_1\sum_{d=1}^{D}\rho_d{\|\boldsymbol{w}_d\|}_2 + \lambda_2\sum_{d=1}^D\theta_d{\|\boldsymbol{w}_d\|}_1 + \log Z(\boldsymbol{\rho}, \boldsymbol{\theta})$,\\
        \multirow{-2}{*}{Adaptive multi-task Lasso} & \multirow{-2}{*}{NeurIPS} & \multirow{-2}{*}{\citeyear{lee2010adaptive}} & \multirow{-2}{*}{Group-sparse learning} & $+$ adaptive penalty & $P(\boldsymbol{W}|\boldsymbol{\rho}, \boldsymbol{\theta}) = \frac{1}{Z(\boldsymbol{\rho}, \boldsymbol{\theta})}\prod_{d=1}^D\prod_{t=1}^T\exp (-\theta_d\lvert w_{n,t}\rvert)\times\prod_{d=1}^D\exp(-\rho_d\|\mathbf{w}_d\|_2)$ \\
    \midrule
    \rowcolor{gray!20}
         & & & & & $\min\limits_{\mathbf{M}_0,\ldots,\mathbf{M}_T} \gamma_0 \|\mathbf{M}_0 - \mathbf{I}\|_F^2 + \sum\nolimits_{t=1}^{T} \left[ \gamma_t \|\mathbf{M}_t\|_F^2 + \sum\nolimits_{(i,j) \in J_t, j \neq i} d_t^2(\mathbf{x}_i, \mathbf{x}_j) + \sum\nolimits_{(i,j,k) \in S_t} \xi_{ijk} \right]$ \\
    \rowcolor{gray!20}
        \multirow{-2}{*}{Large margin multi-task metric learning} & \multirow{-2}{*}{NeurIPS} & \multirow{-2}{*}{\citeyear{parameswaran2010large}} & \multirow{-2}{*}{Priori Sharing} & \multirow{-2}{*}{Frobenius norm} & 
        \text{s.t.} $\forall t, \forall (i, j, k) \in S_t\colon \quad d_t^2(\mathbf{x}_i, \mathbf{x}_k) - d_t^2(\mathbf{x}_i, \mathbf{x}_j) \geq 1 - \xi_{ijk}; \xi_{ijk} \geq 0; \mathbf{M}_0, \mathbf{M}_1, \ldots, \mathbf{M}_T \geq 0$ \\
    \midrule
        Hierarchical multitask structured output learning & NeurIPS & \citeyear{gornitz2011hierarchical} & Priori Sharing & Frobenius norm & $\min\limits_{\boldsymbol{W}}\frac{1}{2}\sum_{t=1}^{T}\frac{1}{N_t}\|{\boldsymbol{X}^{(t)}}\boldsymbol{w}^{t} - \boldsymbol{y}^{(t)}\|^2_2 + \frac{1}{2}\sum_{t=1}^T ||\boldsymbol{w}||_2^2 - \lambda \boldsymbol{w}^T \boldsymbol{w}_p$, where $p$ is the parent node. \\     
    \midrule
    \rowcolor{gray!20}
         & & & low-rank learning & & \\
    \rowcolor{gray!20}
        \multirow{-2}{*}{Robust MTL} & \multirow{-2}{*}{KDD} & \multirow{-2}{*}{\citeyear{chen2011integrating}} & Decomposition, group-sparse learning, & \multirow{-2}{*}{Trace norm + $\ell_{2,1}$ norm} & \multirow{-2}{*}{$\min\limits_{\boldsymbol{W}}\frac{1}{2}\sum_{t=1}^T\|{\boldsymbol{X}^{(t)}}(\boldsymbol{l}^t+\boldsymbol{s}^t)-\boldsymbol{y}^{(t)}\|_2^2 + \lambda_1\|\boldsymbol{L}\|_* + \lambda_2\sum_{t=1}^T\|\boldsymbol{s}_t\|_2$,~~s.t.~$\boldsymbol{W}=\boldsymbol{L}+\boldsymbol{S}$} \\
    \midrule
        Temporal group Lasso & KDD & \citeyear{zhou2011multi} & Group-sparse learning & Frobenius norm + $\ell_{2,1}$ norm & $\min\limits_{\boldsymbol{W}}\frac{1}{2}\sum_{t=1}^{T}\frac{1}{N_t}\|{\boldsymbol{X}^{(t)}}\boldsymbol{w}^{t} - \boldsymbol{y}^{(t)}\|^2_2 + \lambda_1\sum_{d=1}^D\|\boldsymbol{w}_d\|_2^2 + \lambda_2\sum_{t=1}^{T-1}\|\boldsymbol{w}^{t}-\boldsymbol{w}^{t+1}\|_2^2 + \lambda_3\sum_{d=1}^D\|\boldsymbol{w}_d\|_2$ \\
    \midrule
    \rowcolor{gray!20}
        \multirow{2}{*}{Clustered MTL} & \multirow{2}{*}{NeurIPS} & \multirow{2}{*}{\citeyear{zhou2011clustered}} & \multirow{2}{*}{task clustering} & \multirow{2}{*}{Clustering penalty + $\ell_{2,2}$ norm} & $\min\limits_{\boldsymbol{W}, \boldsymbol{F}}\frac{1}{2}\sum_{t=1}^T\frac{1}{N_t}\|{\boldsymbol{X}^{(t)}}\boldsymbol{w}^t - \boldsymbol{y}^t\|_2^2 + \lambda_1(\text{tr}(\boldsymbol{W}^\top\boldsymbol{W}) - \text{tr}(\boldsymbol{F}^\top\boldsymbol{W}^\top\boldsymbol{W}\boldsymbol{F})) + \lambda_2\sum_{t=1}^{T}{\|\boldsymbol{w}^t\|}^2_2,$\\
        \rowcolor{gray!20}
         & & & & & $~~\text{s.t.}~\boldsymbol{F}_{t,j}=1/\sqrt{n_j}~\text{if}~t\in\mathcal{C}_j~\text{otherwise}~0,$ $t=1,\cdots,T$, where $n_j$ is the \#task in the $j$-th cluster $\mathbf{\mathcal{C}}_j$. \\
         
    \midrule
        & & & Decomposition, sparse learning, & & \\
        \multirow{-2}{*}{Sparse and low rank MTL} & \multirow{-2}{*}{TKDD} & \multirow{-2}{*}{\citeyear{chen2012learning}} & low-rank learning & \multirow{-2}{*}{$\ell_{1, 1}$ norm + trace norm} & \multirow{-2}{*}{$\min\limits_{\boldsymbol{W}}\frac{1}{2}\sum_{t=1}^{T}\frac{1}{N_t}\|{\boldsymbol{X}^{(t)}}\boldsymbol{w}^{t} - \boldsymbol{y}^{(t)}\|^2_2 + \lambda\sum_{d=1}^D\|\boldsymbol{p}_d\|_{1}$,~~s.t.~$\boldsymbol{W}=\boldsymbol{P}+\boldsymbol{Q}, \|\boldsymbol{Q}\|_*\leq\tau$} \\
    \midrule
    \rowcolor{gray!20}
        Convex fused sparse group Lasso & KDD & \citeyear{zhou2012modeling} & Group-sparse learning & $\ell_{1,1}$ norm $+$ $\ell_{2,1}$ norm & $\min\limits_{\boldsymbol{W}}\frac{1}{2}\sum_{t=1}^{T}\frac{1}{N_t}\|{\boldsymbol{X}^{(t)}}\boldsymbol{w}^{t} - \boldsymbol{y}^{(t)}\|^2_2 + \lambda_1\sum_{d=1}^D\|\boldsymbol{w}_d\|_1 + \lambda_2\sum_{t=1}^{T-1}\|\boldsymbol{w}^t-\boldsymbol{w}^{t+1}\|_1 + \lambda_3\sum_{d=1}^D\|\boldsymbol{w}_d\|_2$ \\
    \midrule
        Adaptive multi-task elastic-net & SDM & \citeyear{chen2012adaptive} & Group-sparse learning & $\ell_{2,1}$ norm $+$ Frobenius norm & $\min\limits_{\boldsymbol{W}}\frac{1}{2}\sum_{t=1}^{T}\frac{1}{N_t}\|{\boldsymbol{X}^{(t)}}\boldsymbol{w}^{t} - \boldsymbol{y}^{(t)}\|^2_2 + \lambda_1\sum_{d=1}^{D}{\|\boldsymbol{w}_d\|}_2 +  \lambda_2\sum_{d=1}^D\|\boldsymbol{w}_d\|_2^2$ \\
    \midrule
    \rowcolor{gray!20}
        Multi-level Lasso & ICML & \citeyear{lozano2012multi} & Decomposition, sparse learning & $\ell_{1,1}$ norm + adaptive penalty & $\min\limits_{\boldsymbol{W}}\frac{1}{2}\sum_{t=1}^{T}\frac{1}{N_t}\|{\boldsymbol{X}^{(t)}}\boldsymbol{w}^{t} - \boldsymbol{y}^{(t)}\|^2_2 + \lambda_1\sum_{d=1}^{D}\theta_d +  \lambda_2\sum_{d=1}^D\|\boldsymbol{\boldsymbol{\gamma}}_d\|_1$,~~s.t.~$\boldsymbol{W}=\vec{\boldsymbol{\theta}}\boldsymbol{\Lambda}\boldsymbol{\Gamma}, \vec{\boldsymbol{\theta}}\geq\boldsymbol{0}$  \\
    \midrule
        Robust multi-task feature learning & KDD & \citeyear{gong2012robust} & Decomposition, group-sparse learning & $\ell_{2,1}$ norm + $\ell_{1,2}$ norm & $\min\limits_{\boldsymbol{W}}\frac{1}{2}\sum_{t=1}^{T}\frac{1}{N_t}\|{\boldsymbol{X}^{(t)}}\boldsymbol{w}^{t} - \boldsymbol{y}^{(t)}\|^2_2 + \lambda_1\sum_{d=1}^D\|\boldsymbol{p}_d\|_2 + \lambda_2\sqrt{\sum_{d=1}^D\|\boldsymbol{q}_d\|_1^2}$,~~s.t.~$\boldsymbol{W} = \boldsymbol{P} + \boldsymbol{Q}$ \\
    \midrule
    \rowcolor{gray!20}
        Multi-stage multi-task feature learning & NeurIPS & \citeyear{gong2012multi} & Sparse learning & Capped $\ell_1$ norm~\citep{zhang2010analysis}& $\min\limits_{\boldsymbol{W}}\frac{1}{2}\sum_{t=1}^{T}\frac{1}{N_t}\|{\boldsymbol{X}^{(t)}}\boldsymbol{w}^{t} - \boldsymbol{y}^{(t)}\|^2_2 + \lambda\sum_{d=1}^R\min\{\|\boldsymbol{w}_d\|_1, \tau\}$ \\
    \midrule
        Convex formulation for MTL & IJCAI & \citeyear{zhang2012convex} & Priori sharing & Clustering penalty & $\min\limits_{\boldsymbol{W}}\frac{1}{2}\sum_{t=1}^{T}\frac{1}{N_t}\|{\boldsymbol{X}^{(t)}}\boldsymbol{w}^{t} - \boldsymbol{y}^{(t)}\|^2_2 + \frac{\lambda_1}{2}$tr$(\boldsymbol{W}\boldsymbol{W}^T)+\frac{\lambda_2}{2}$tr$(\boldsymbol{W}\boldsymbol{\Omega}^{-1}\boldsymbol{W}^T)$~~s.t.~$\boldsymbol{\Omega}\in\boldsymbol{S}_+^D$, tr$\boldsymbol{\Omega}=1$\\
        \midrule
        \rowcolor{gray!20}
    Multi-linear multi-task learning & ICML & \citeyear{romera2013multilinear} & Low-rank learning &  Overlapped tensor trace norm & $\min\limits_{\boldsymbol{\mathcal{W}}}\frac{1}{2}\sum_{t=1}^{T}\frac{1}{N_t}\|{\boldsymbol{X}^{(t)}}\boldsymbol{w}^{t} - \boldsymbol{y}^{(t)}\|^2_2 + \lambda\sum_{k=1}^N \| \boldsymbol{W}_{(k)} \|_*$ where $\boldsymbol{W}_{(k)}$ is the mode-$k$ unfolding of tensor $\boldsymbol{\mathcal{W}}\in \mathbb{R}^{D \times I_2 \times \cdots \times I_N}$.\\
    \midrule
        Regularization approach to learn MTL & TKDD & \citeyear{zhang2014regularization} & Priori sharing & Clustering penalty + $\ell_{2,2}$ norm & $\min\limits_{\boldsymbol{V}, \boldsymbol{W}}\frac{1}{2}\sum_{t=1}^{T}\frac{1}{N_t}\|{\boldsymbol{X}^{(t)}}\boldsymbol{w}^{t} - \boldsymbol{y}^{(t)}\|^2_2 + \frac{\lambda}{2}\sum_{t=1}^T||\boldsymbol{w}^t||_2^2 + $tr$(\boldsymbol{W}\boldsymbol{\Omega}^{-1}\boldsymbol{W}^T) + d$ln$\boldsymbol{\Omega}$~~s.t.~$\boldsymbol{\Omega}\in\boldsymbol{S}_+^D$\\
    \midrule
    \rowcolor{gray!20}
    Multi-linear multi-task learning & NeurIPS & \citeyear{wimalawarne2014multitask} & Low-rank learning & Scaled latent tensor trace norm & $\min\limits_{\boldsymbol{\mathcal{W}}}\frac{1}{2}\sum_{t=1}^{T}\frac{1}{N_t}\|{\boldsymbol{X}^{(t)}}\boldsymbol{w}^{t} - \boldsymbol{y}^{(t)}\|^2_2 + \inf_{\boldsymbol{\mathcal{W}}^{(1)} + \cdots + \boldsymbol{\mathcal{W}}^{(N)}=\boldsymbol{\mathcal{W}}} \lambda \sum_{k=1}^N I_k^{-1/2} \| \boldsymbol{W}_{(k)}^{(k)} \|_*$ where $\boldsymbol{\mathcal{W}}\in \mathbb{R}^{D \times I_2 \times \cdots \times I_N}$ is a tensor.\\
    \midrule
        Task Tree model & KDD & \citeyear{han2015learning} & task clustering & $\ell_{2,2}$ norm & $\min\limits_{\boldsymbol{W}} \frac{1}{2}\sum_{t=1}^T\frac{1}{N_t}\|{\boldsymbol{X}^{(t)}}\sum_{h=1}^H\boldsymbol{w}_h^t - \boldsymbol{y}^t\|_2^2 + \sum_{h=1}^H\lambda_h\sum_{i<j}^T\|\wbold_{h}^i - \wbold_{h}^j\|^2_2, \text{s.t.} |\wbold_{h-1}^i - \wbold_{h-1}^j| \succeq |\wbold_{h}^i - \wbold_{h}^j|, \forall h \geq 2, \forall i<j$ \\
    \midrule
    \rowcolor{gray!20}
        Reduced rank multi-stage MTL & AAAI & \citeyear{han2016multi} & Low-rank learning & Capped trace norm~\citep{sun2013robust} & $\min\limits_{\boldsymbol{W}}\frac{1}{2}\sum_{t=1}^{T}\frac{1}{N_t}\|{\boldsymbol{X}^{(t)}}\boldsymbol{w}^{t} - \boldsymbol{y}^{(t)}\|^2_2 + \lambda\sum_{r=1}^R\min\{\sigma_r(\boldsymbol{W}), \tau\}$ \\
    \bottomrule
    \end{tabular}
    \begin{tablenotes}
    \footnotesize
    \item[1] This work is published in Technical Report, the Department of Statistics, UC Berkeley.
    \item[2] This work is published in Jian Zhang's Ph.D. Thesis, CMU Technical Report CMU-LTI-06-006, 2006.
    \end{tablenotes}
    \end{threeparttable}
    }
\end{table*}


%% file: tex_files/02-1/bayesian.tex
\begin{wrapfigure}[10]{r}{5.6cm}
    \centering
    \includegraphics[width=0.95\linewidth]{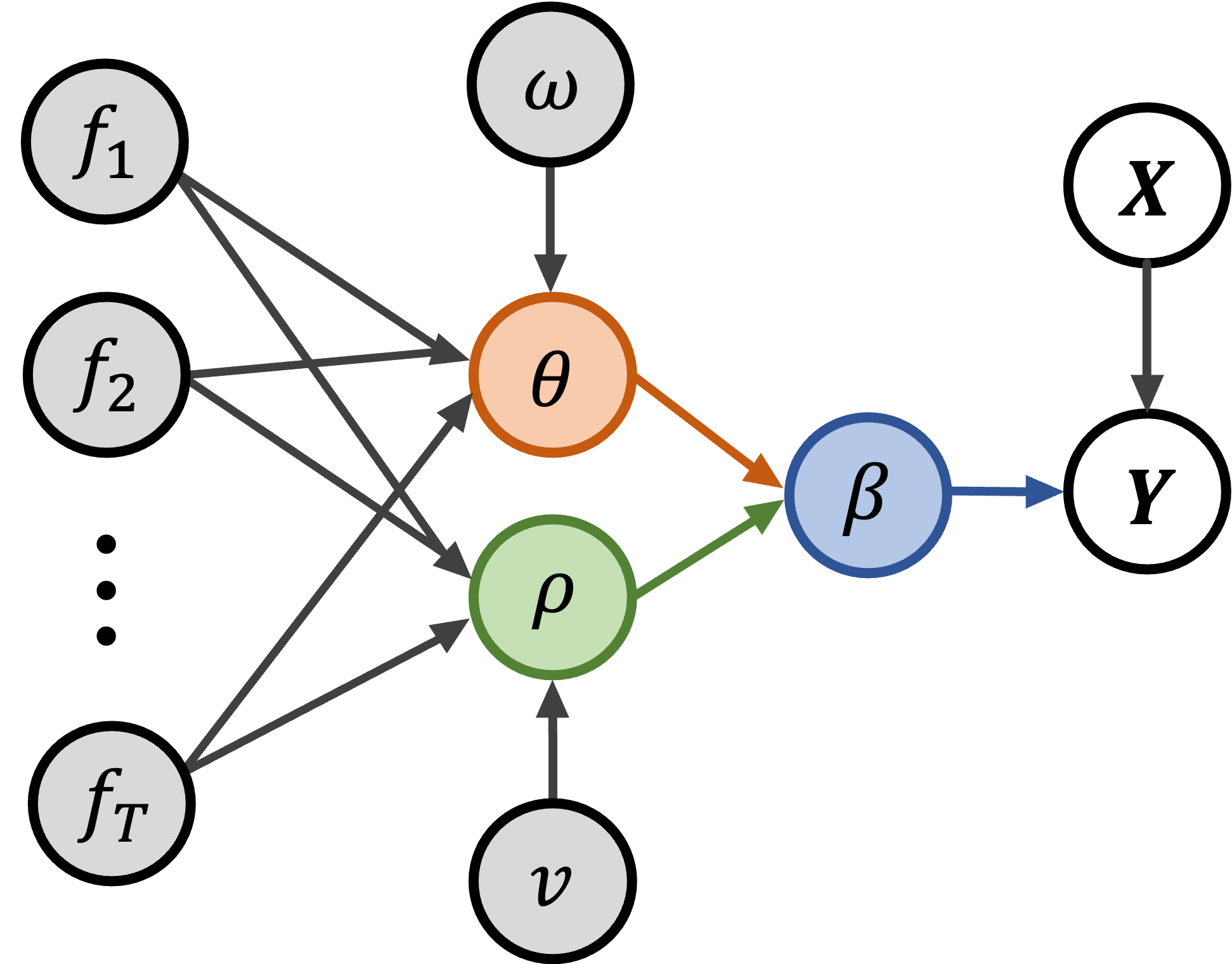}
    \caption{The Bayesian graph for adaptive sparse multi-task Lasso model.}
    \label{bayesian}
\end{wrapfigure}

%% file: tex_files/02-1/naive-mtl.tex
\begin{figure*}[t]
    \centering

     \begin{subfigure}{0.41\textwidth}
        \includegraphics[width=1\textwidth]{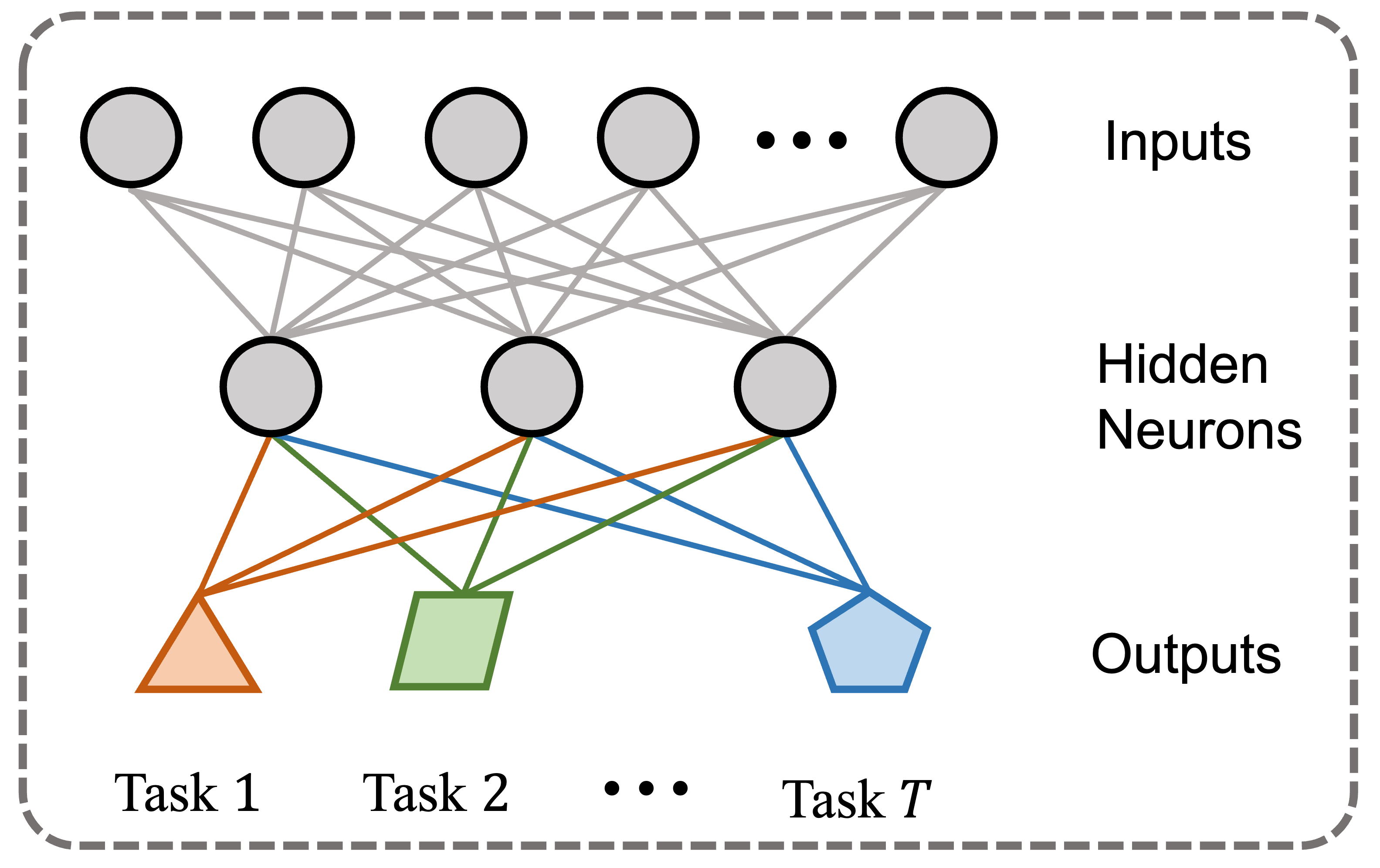}
        \caption{Feedforward Neural Networks.}
        \label{feedforward}
    \end{subfigure} \quad\quad
    \begin{subfigure}{0.41\textwidth}
        \includegraphics[width=1\textwidth]{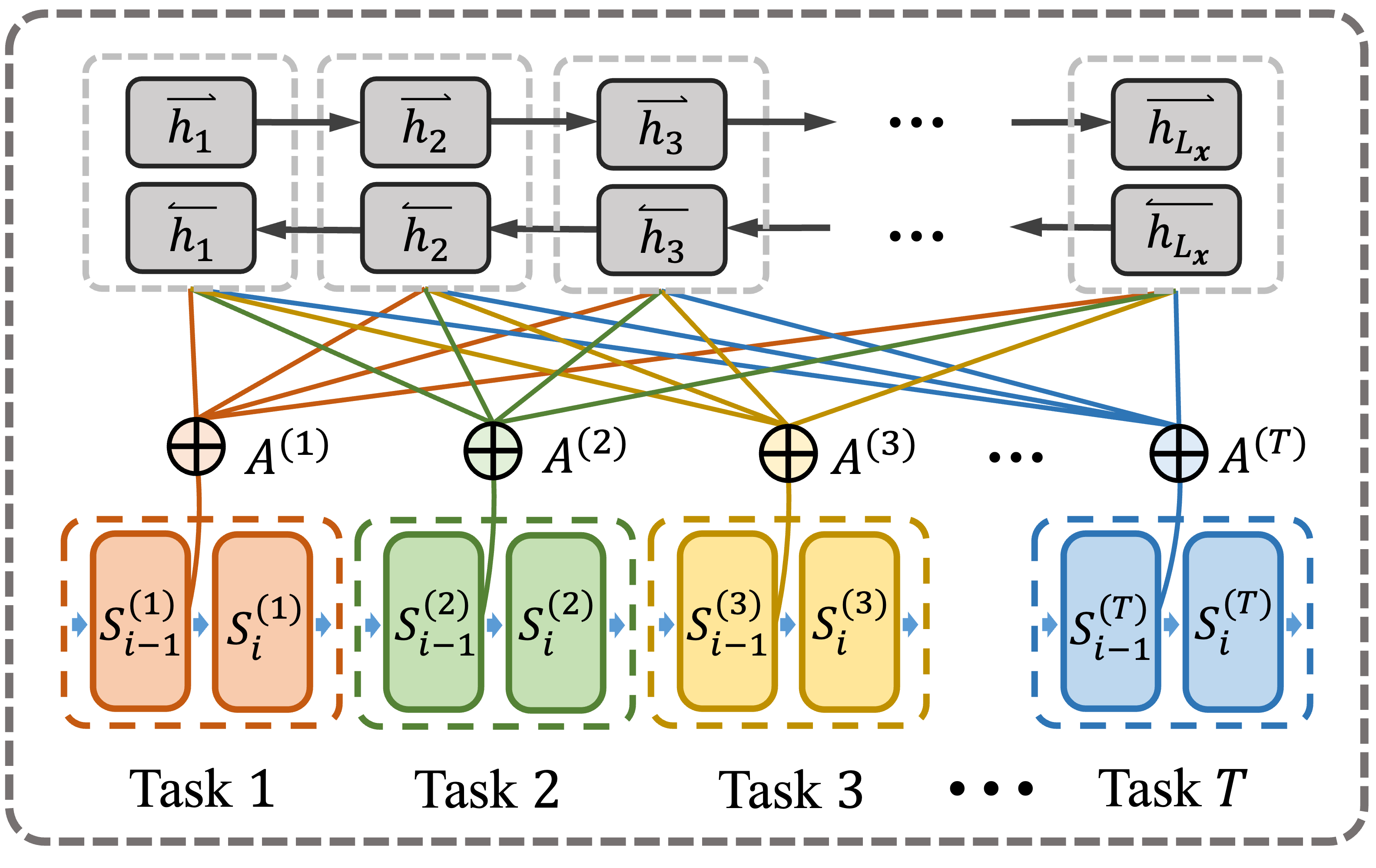}
        \caption{Recurrent Neural Networks.}
        \label{recurrent}
    \end{subfigure}
    
    \caption{Hard-parameter sharing in FNNs and RNNs. (a) The most early version of hard parameters sharing. The connections between inputs and hidden neurons jointly transform features, which are then utilized for Task 1 to Task $T$. (b) A modern-day RNN used for multiple-target language translation, which jointly transforms features from shared sequence-based representations. $(h_1, \cdots,  h_{L_{\boldsymbol{x}}})$ represent the sequence of bidirectional recurrent representations, where $L_{\boldsymbol{x}}$ is the number of tokens for the source sentence $\boldsymbol{x}$. $s_i^{(t)}$ is is a recurrent neural network hidden state at time $i$ for the $t$-th task, which is estimated based on the combination of $(h_1, \cdots, h_{L_{\boldsymbol{x}}})$ weighted by $A^{(t)}$.}
    \label{naive-mtl}
\end{figure*}

%% file: 02-2_method_deep.tex
\input{table_files/deep_summary}

\subsection{DL Era: Effective and Diversified}
\label{deep-era}
{~~}\vspace{2pt}\\
With the advent of DL, more powerful computational units and more effective memory bandwidth, e.g., Graphics Processing Units (GPUs) and Tensor Processing Units (TPUs), have made it possible to learn richer features for challenging tasks. Deep MTL methods, unlike traditional MTL methods imposing parameter regularizations or decompositions, can handle large-scale parameter sharing, feature propagation, NAS, task balancing, and optimization intervention, to name a few. The traditional techniques often involve complicated mathematical analysis but fail to learn a satisfactory performance in the real-world scenario with noise-polluted data or loosely-related tasks. However, deep MTL methods can overcome these issues by (1) directly extracting features in raw data and gradually elevating features layer-by-layer from low-level textures to mid-level semantics to high-level responses; and (2) progressively learning activations by stochastic gradients descent (SGD)~\citep{robbins1951stochastic, lecun2002efficient} that is provably efficient and practical in obtaining an expressive networks~\citep{livni2014computational}. In this manner, hierarchical features can be efficiently communicated at different levels for jointly learning of multi-task objectives.

This section begins with a discussion of the architecture taxonomy commonly adopted in deep MTL, which serves as the backbone for the rest of the method overview. In the following, we summarize the feature propagation techniques that include feature fusion (see \textbf{\S}~\ref{sec:fusion}), cascading (see \textbf{\S}~\ref{sec:cascading}), distillation (see \textbf{\S}~\ref{sec:distill}), and cross-task attention (see \textbf{\S}~\ref{sec:attention}). These techniques encourage networks to automatically combine the features learned from different tasks, addressing the crucial challenge of effectively and efficiently utilizing the rich features enabled by DL. \textbf{\S}~\ref{subsec:scalar} presents an overview of task balancing techniques in deep MTL, incorporating the linear combination of different tasks through three essential factors: gradient, loss, and learning speed. The comparison and recalibration of these factors aim to coordinate diverse tasks during the model weight update process. We will discuss this section from the point of gradient correction and dynamic weighting. In contrast, \textbf{\S}~\ref{subsec:multi-obj-opt} explores MOO in the context of MTL, which aims to simultaneously optimize potentially conflicting objective functions. Other promising topics covered include adversarial multi-task training (see \textbf{\S}~\ref{sec:adversarial}), MoE (see \textbf{\S}~\ref{sec:moe}), GCN-based MTL (see \textbf{\S}~\ref{sec:graph}), and NAS for MTL (see \textbf{\S}~\ref{sec:nas}). The summary of deep MTL models is presented in Table~\ref{tab:deep-sum}, and representative DL frameworks in MTL are illustrated in Fig.~\ref{fig:deep_all}.
\\

\input{tex_files/02-2/all_archi}

\input{tex_files/02-2/hard-soft}
\paragraph{\emph{Architecture Taxonomy}} The remarkable success of deep MTL can be attributed to the rich extracted representations and their efficient sharing. Multi-task sharing relies on the basic splitting ways of architectures among involved tasks.~\citet{liu2016recurrent} first discuss three different sharing mechanisms based on text classification in Recurrent Neural Networks (RNNs): uniform-, coupled-, and shared-layer architectures.~\citet{ruder2017overview} first organize it into two categories: hard parameter sharing and soft parameter sharing. According to this taxonomy, the uniform-layer architecture falls under hard-parameter sharing, while coupled- and shared-layer architectures are considered soft-parameter sharing. In general,~\citet{ruder2017overview}'s taxonomy has been widely accepted by the research community~\citep{vandenhende2021multi}.
We carry forward this taxonomy and enrich it with more details. 

In hard parameter sharing, as shown in {Fig.}~\ref{hard}, different tasks can share identical parameters in shallow layers and maintain their own specific parameters in the splitting heads. As shown in {Fig.}~\ref{feedforward}, this idea can be dated back to 1990s~\citep{bromley1993signature, caruanamultitask, caruana1997multitask} when high-related tasks are introduced into a shared FNNs to serve as inductive bias for each other. {Fig.}~\ref{recurrent} shows this idea used in RNNs in a modern way~\citep{dong2015multi}. 
CNNs can also adopt hard parameter sharing to perform multiple related tasks. As shown in {Fig.}~\ref{fig:hard-example}, TCDCN~\citep{zhang2014facial} and Fast RCNN~\citep{girshick2014rich, girshick2015fast} are the earliest practice of this idea in computer vision. From a representation learning perspective, shallow layers are typically shared as a feature encoder that extracts common features such as edges and textures. By enriching these common features with more related tasks, deeper layers can help enable multitasking on task-specific heads. 

~\citet{misra2016cross} argue that there is no principled way of architecture splitting in hard parameter sharing, and conducted the first empirical study to investigate the performance trade-offs amongst varieties of involved tasks and splitting ways in CNNs. 
The dependence between involved tasks and the splitting ways of architecture motivates the exploration of an architecture that can capture all possible splittings and thus learn an optimal combination of task-shared and task-specific representations, i.e., soft parameter sharing shown in {Fig.}~\ref{soft}. 
While hard-parameter sharing requires shallow layers to be identical across tasks, soft-parameter sharing encourages each task to maintain its own shallow layers and leverage features from related tasks during the propagation to capture similarities. These feature propagation techniques include but are not limited to fusion, aggregation, attention, etc. 
However, whether employing hard or soft parameter sharing, exploring the MTL architecture space still remains error-prone.. First of all, this space for deep neural architectures grows exponentially with depth, and incorporating more tasks significantly expands the range of optimal solutions. On the other hand, while hard parameter sharing compresses the model size, leading to a sub-optimal solution, soft parameter sharing ensures advancement by maintaining the maximum total model size, allowing each task to learn a specific architecture in contrast to STL. An adaptive architecture search in a greedy manner during the neural network training process shows promise. As shown in {Fig.}~\ref{adaptive} the adaptive parameter sharing, each path from the different layers of different tasks is active before training. The connections vanish with the pursuit of model compression in the process of multi-task optimization, and usually, a thin network is finalized after this dynamic branching procedure.
\input{table_files/notation_deep}

Unless explicitly stated otherwise, we employ the notation provided in Tab.~\ref{tab:deep_notation} within the context of DL settings to expand upon and complement the information presented in Tab.~\ref{tab:notation}.

\input{tex_files/02-2/hard-example}

\subsubsection{Feature fusion}
\label{sec:fusion}
Feature fusing is a common technique used in MTL to fuse features extracted under the supervision of different tasks, which can leverage shared and private knowledge across tasks. This technique allows each network to better exploit the relationships between tasks and thus improve overall performance. 
In general, feature fusion in MTL involves weighted summation, concatenation, or a combination of both. We categorize the feature fusion methods into two classes: parallel sharing, where the feature fusion happens at the same position of layers between tasks, and Non-parallel sharing, in which the permutation of sharing layers may exist.
The representative works in the line of parallel sharing include Cross-Stitch Networks~\citep{misra2016cross}, Sluice Networks~\citep{ruder2019latent}, and Neural Discriminative Dimensionality Reduction in Convolutional Neural Networks (NDDR-CNN)~\citep{gao2019nddr}. As research in this direction progresses, an increasing number of learnable parameters are being used to control the fusion process. For example, Cross-Stitch Networks utilize four task-aware parameters, Sluice Networks capture latent subspaces of features via extra parameters, and NDDR-CNN models layer-wise fusion by using $1\times 1$ convolutions. However, expecting task feature hierarchies to align perfectly, even among closely related tasks, is unreasonable. Imposing parallel sharing in these unmatched layers could lead to negative transfer. To remedy this dilemma, Soft Order~\citep{meyerson2018beyond} uses a more flexible ordering of shared layers to assemble them in different ways for different tasks.



\underline{Parallel sharing.} Cross-Stitch Networks~\citep{misra2016cross} is a soft parameter-sharing architecture that can learn an optimal combination of task-shared and task-specific representations via four learnable parameters, which is named cross-stitch unit. As shown in {Fig.}~\ref{cross-stitch-unit}, the activations from different tasks are linear combined via four parameters $(\alpha_{11}, \alpha_{12}, \alpha_{21}, \alpha_{22})$. We denote by $\Xmcal_l^i$ the feature maps in the $l$-th layer of task $i$. Then the formalization of the Cross-Stitch unit is
\begin{equation}
    \begin{bmatrix}
        \Xmcal_{l+1}^1  \\
        \Xmcal_{l+1}^2
    \end{bmatrix} = 
    \begin{bmatrix}
        \alpha_{11} & \alpha_{12}\\ \alpha_{21} & \alpha_{22}
    \end{bmatrix} 
    \begin{bmatrix}
        \Xmcal_l^1  \\
        \Xmcal_l^2
    \end{bmatrix}
\end{equation}
Specifically, the extreme setting of $(\alpha_{11}, \alpha_{12}, \alpha_{21}, \alpha_{22}) = (1, 0, 0, 1)$ can make certain layers to be non-sharing. From this perspective, the separate STL is a special case of cross-stitch combinations. By varying $\alpha_{\cdot 1}$ and $\alpha_{\cdot 2}$ values, this proposed unit can move between task-shared and -specific representations, and even choose a middle ground if necessary.

Sluice Networks~\citep{ruder2019latent} learns shared parameters between two BiLSTM-based sequence labeling networks~\citep{plank2016multilingual}. This work aims to model loosely related tasks with non-overlapping datasets. As shown in {Fig.}~\ref{sluice-block} a sluice meta-network with two tasks, of which each layer is partitioned into two orthogonal subspaces $\Gbold$ and $\Gbold^\perp$. Accordingly, the activations in the $l$-th layer of task $i$ are also partitioned into $\Xmcal_{l}^i$ and $\Xmcal_{l}^{i^\perp}$, thus leading to a matrix in $\Rmbb^{4\times4}$ to combine activations from two tasks:
\begin{equation}
    \begin{bmatrix}
        \Xmcal_{l+1}^1  \\
        \Xmcal_{l+1}^{1^\perp}\\
        \Xmcal_{l+1}^2\\
        \Xmcal_{l+1}^{2^\perp}
    \end{bmatrix} = 
    \begin{bmatrix}
        \alpha_{11} & \alpha_{11^\perp} & \alpha_{12} & \alpha_{12^\perp}\\ 
        \alpha_{1^\perp1} & \alpha_{1^\perp1^\perp} & \alpha_{1^\perp2} & \alpha_{1^\perp2^\perp}\\
        \alpha_{21} & \alpha_{21^\perp} & \alpha_{22} & \alpha_{22^\perp}\\
        \alpha_{2^\perp1} & \alpha_{2^\perp1^\perp} & \alpha_{2^\perp2} & \alpha_{2^\perp2^\perp}
    \end{bmatrix} 
    \begin{bmatrix}
        \Xmcal_l^1  \\
        \Xmcal_l^{1^\perp}  \\
        \Xmcal_l^2 \\
        \Xmcal_l^{2^\perp}
    \end{bmatrix}
\end{equation}
Inspired by Cross-stitch networks, these $\alpha$ values are learnable to control how much to share for task-shared information and how much to preserve for task-specific information. Finally, $\beta$ parameter (see Fig.~\ref{sluice-block}), through the skip-connections, linearly summarizes the multi-task representations at various levels of the network architecture.

Neural Discriminative Dimensionality Reduction in Convolutional Neural Networks (NDDR-CNN)~\citep{gao2019nddr} further concatenates feature maps from different tasks in a channel-wise manner. This NDDR, as shown in {Fig.}~\ref{nddr-unit}, can be fulfilled by using simple $1\times 1$ convolutional layer plus batch nomalization layer, and be extended to any end-to-end training CNN in a ``plug-and-play'' fashion. 
Considering the number of tasks being $T$, we can denote $1\times 1$ convolution by $\mathcal{W}\in\mathbb{R}^{1\times 1\times TC\times TC}$, where $TC$ is the depth of combined feature maps from all tasks. We concatenate feature maps according to the channel dimension and divide $1\times 1$ convolution according to the output dimension by $T$ tasks as follows:
\begin{equation}
    \Xmcal_l = [\Xmcal_l^1, \cdots, \Xmcal_l^T], \Wmcal = [\Wmcal^1, \cdots, \Wmcal^T], \nonumber
\end{equation}
where $\Xmcal_l\in\mathbb{R}^{H\times W\times TC}$ and $\Wmcal^t\in\mathbb{R}^{1\times 1 \times TC \times C}$.
Then, the output feature maps at the $(l+1)$-th layer for the $t$-th task can be calculated as
\begin{equation}
    \label{nddr-conv}
    \Xmcal_{l+1}^t = CONV_{\Wmcal^t}(\Xmcal_l), t = 1, \cdots, T.
\end{equation}
The NDDR layer defined by {Eq.}~(\ref{nddr-conv}) is a standard $1\times 1$ convolution operation in CNNs. To avoid a trivial solution on $\Wmcal$ and the noise directions of learned features, the batch normalization layer is followed after each NDDR layer, and the $\ell_2$ weight decay is applied on the weights of the NDDR layer, respectively.

\underline{Unparallel sharing.} Soft Order~\citep{meyerson2018beyond} learns how shared layers are assembled in permuted ways for different tasks. Specifically, a learnable tensor of scalars $S\in\Rmbb^{L\times L\times T}$, is used to implement the soft ordering, where $L$ is \#layer and $T$ is \#task. For simplicity, consider a hard sharing network with $L$ shared layers $\{f_{\Wmcal_l}\}_{l=1}^L$ ($f$ can be $CONV$ or Linear function), then the soft ordering of this hard sharing for the $t$-th task is:
\begin{equation}
    \Xmcal_l^t = \sum\nolimits_{j=1}^L s_{t,j,l}f_{\Wmcal_j}(\Xmcal_{l-1}^t), l = 1, \cdots, L, t = 1, \cdots, T, \quad\text{s.t.} \sum\nolimits_{j=1}^L s_{t,j,l} = 1~\text{with}~\forall (t, l),
\end{equation}
where $s_{t,j,l}$ is the $(t,j,l)$-th entry of the tensor $S$. Fig.~\ref{soft-order} visualizes this layer permutation operation. It is noticed that the constraint on $s_{t,j,l} = 1$ for $\forall (t, l)$ can be easily implemented via a softmax function. In practice, a dropout operation is beneficial to increasing the generalization capacity of shared representations.

\begin{applebox}{Remarks}
\begin{enumerate}[leftmargin=0.4cm, label=(\roman*)]
    \item Feature fusion enables the exploration of multi-task interactions in a "plug and play" manner, making it a general-purpose MTL solution that can be generalized to any backbones.
    \item The feature-level relationships between tasks can be investigated by examining the introduced learnable parameters after training.
    \item Feature fusion cannot reveal what information is propagated during the multitasking process, highlighting the need for design guidelines that go beyond common practices.
    \item Feature fusion inherently imposes constraints on the SIMO setting, as it allows features to be fused only within the same context.
    \item Feature fusion creates task-specific branches that also need to learn shared features across tasks, which can hinder task-awareness compared to STL, which focuses on capturing representations specific to the target task.
\end{enumerate}
\end{applebox}

\input{tex_files/02-2/cascading_framework}
\subsubsection{Cascading} 
\label{sec:cascading}
Having supervision from all tasks at the outermost level is shown to be sub-optimal, another avenue of investigation for mitigating this parallel sharing is through the implementation of multi-task cascaded learning~\citep{sogaard-goldberg-2016-deep}. This field of study involves supervising tasks at different levels within their respective layers, facilitating higher-level tasks to effectively leverage the shared representation derived from lower-level tasks. In practice, multi-task cascading can be applied to 1) the complicated task that can be decomposed into several sub-tasks, e.g., instance-aware semantic segmentation decomposed into differentiating instances, estimating masks and categorizing objects in CV~\citep{dai2016instance}, and 2) a group of hierarchical tasks, e.g., part-of-speech (POS) tagging (word-level), dependency parsing (syntactic-level) and question answering (QA) (semantic-level) in NLP~\citep{sogaard-goldberg-2016-deep, hashimoto-jmt:2017:EMNLP2017}. In this line of research, early work~\citep{sogaard-goldberg-2016-deep} realize cascading by having low-level tasks supervised at shallow layers, and then reusing representations from shallow layers for higher-level tasks. The Joint Many-Task (JMT) model~\citep{hashimoto-jmt:2017:EMNLP2017} adds shortcut connections from each lower-level task prediction to higher-level tasks, which can further reflect task hierarchies. Furthermore, shortcut connections in Multi-task Network Cascades (MNCs)~\citep{dai2016instance} and Deep Cascade Multi-Task Learning (DCMTL)~\citep{gong2019deep} come from both cascade connection (predictions) and residual connection (features). Hierarchical MTL (HMTL)~\citep{sanh2019hierarchical} introduces more semantic tasks to share both common embeddings and encoders in a hierarchical cascading architecture.

Vanilla Cascading~\citep{sogaard-goldberg-2016-deep} first presents a multi-task learning architecture that utilizes bi-directional RNNs. This architecture enables the supervision of different tasks at various layers, as shown in Fig.~\ref{cascade_1}. In this study, the POS task is supervised at the innermost layer, and the syntactic chunking and Combinatory Categorical Grammar (CCG) supertagging join at the outermost layer to utilize the shared representation of the lower-level tasks via a hard parameter sharing. In this case, the incorporation of lower-level task supervision affects the shallow layer parameter updating, which is beneficial to all involved tasks in MTL.

Multi-task Network Cascades (MNCs)~\citep{dai2016instance} performs three sub-tasks of the instance-aware semantic segmentation at the different stages and reuses the features of these tasks at different layers. Each of the three stages involves its own predictions of box-level instance proposals, mask-level instance regression, and instance categorization, respectively, and the later task learning relies on previous prediction output. As shown in Fig.~\ref{cascade_2}, the innermost features are utilized by all sub-tasks, which is beneficial to both the accuracy and speed in an end-to-end training manner. 

Joint Many-Task (JMT) Model~\citep{hashimoto-jmt:2017:EMNLP2017} is another cascading model to predict NLP tasks with different linguistic levels of morphology, syntax, and semantics. JMT shares a similar architecture with MNCs, as shown in Fig.~\ref{cascade_3}, but each higher-level task contains the shortcut connections from the predictions of all lower-level tasks. In addition, the na\"ive $\ell_2$ regularization term is imposed on model weights to allow the improvement of one task without exhibiting catastrophic interference with the other tasks.

Deep Cascade Multi-Task Learning (DCMTL)~\citep{gong2019deep} first incorporates both cascade and residual connections. As shown in Fig.~\ref{cascade_4}, the cascade connections transmit predictions from lower tasks, while the residual connections transmit inputs from lower layers. It has been validated that these skip connections are effective for strictly ordering tasks. The cascading structure alone proves inadequate for high-level tasks that heavily rely on low-level tasks. In addition, DCMTL can outperform previous SOTA methods and has been deployed on the online shopping assistant of a dominant Chinese E-commerce platform. 

Hierarchical Multi-Task Learning (HMTL)~\citep{sanh2019hierarchical} is a parallel method trained in a hierarchical fashion. This model can supervise a set of low-level tasks at the bottom layers and more complex tasks at the top layers. Similar to MNCs~\citep{dai2016instance}, representations extracted at the very beginning are fed into all the successive encoders for different tasks, which is beneficial to the training stability and acceleration. Also shown in Fig.~\ref{cascade_4}, HMTL is a variation that parallels high-level tasks could exist, e.g., Coreference Resolution (CR) and Relation Extraction (RE), and more types of word representations like pre-trained GloVe~\citep{pennington2014glove} and ELMo~\citep{peters-etal-2018-deep} embeddings, are combined to achieve the best performance.

\begin{applebox}{Remarks}
\begin{enumerate}[leftmargin=0.4cm, label=(\roman*)]
    \item Cascading facilitates feature communication across different layers.
    \item Cascading enhances the utility of features for tasks at different levels.
\end{enumerate}
\end{applebox}

\subsubsection{Knowledge Distillation (KD)}
\label{sec:distill}
Motivated by KD~\citep{44873} where a teacher model can guide a student model via passing meaningful knowledge (e.g., soft labels), separate models in MTL for different tasks can utilize definite information. Specifically, a teacher model can be trained on multiple tasks that are of interest and then serves as an expert in performing those tasks and possessing versatile knowledge. The knowledge from the teacher model is then transferred to a student model. This can be done by training the student model to mimic the behavior of the teacher model, e.g., the student model learns to predict the outputs or pattern structures of the teacher model on the shared tasks. On the other hand, the student model can be trained jointly on multiple tasks, using both the labeled data for each task and the guidance from the teacher model. The shared information and generalizable representations learned from the teacher model can benefit the student model's performance on all the tasks. In this manner, the teacher model performs auxiliary tasks to assist the student model in target tasks. For example, the depth prediction from a customized CNN can help the segmentation task via multi-modal distillation (i.e., train with RGB-Depth data instead of RGB data), while the depth prediction is an intermediate auxiliary task to the target segmentation task~\citep{xu2018pad}. The research in this subfield can be classified into two categories that correspond to the knowledge encompassed within a teacher model: feature-level and response-level. 
KD4MTL~\citep{li2020knowledge} carries forward FitNets~\citep{romero2014fitnets} via optimizing the distance between the features of the offline task-specific networks and the online multi-task network. MuST~\citep{ghiasi2021multi} and OKD-MTL~\citep{jacob2023online} distill the knowledge (i.e., pseudo labels) from pre-trained specialized teachers to general-purpose students. MuST~\citep{ghiasi2021multi} pretrains several specialized teachers capable of generating multi-task labels for the target dataset. CrossDistil~\citep{yang2022cross} distills the responses of item preference across different tasks in the recommender system.

\underline{Feature-Level.} Knowledge Distillation for Multi-task Learning (KD4MTL)~\citep{li2020knowledge}, as shown in {Fig.}~\ref{kd4mtl-pipeline}, first trains an offline task-specific network for each task, and then learns the multi-task network via adding the loss to minimize the distance between the task-specific network and the multi-task network. As the multi-task purpose network is capable of multiple tasks while the task-specific network is more professional at its own task, the two output features cannot be completely matched. Instead, the feature map from multi-task network, denoted by $\mathcal{X}\in\mathbb{R}^{H\times W\times C}$, is transformed via an adaptor $\phi^{(t)}: \mathbb{R}^{H\times W\times C} \xrightarrow{1\times 1\times C\times C~CONV} \mathbb{R}^{H\times W\times C}, t = 1, \cdots, T$. These adaptors are jointly learned with the multi-task network via the loss function defined as
\begin{equation}
    \mathcal{L}^d = \sum\nolimits_{t=1}^T\ell^d(\phi^{(t)}(\mathcal{X}), \Tilde{\mathcal{X}}^{(t)}),
\end{equation}
where $\Tilde{\mathcal{X}}^{(t)}$ is the feature map from an offline single network corresponding to the task $t$, and $\ell^d$ is defined as the Euclidean distance between the two feature maps that is $\ell_2$ normalized.

Online KD for MTL (OKD-MTL)~\citep{jacob2023online} proposes an online knowledge distillation method to mitigate negative transfer across tasks. The adaptive feature distillation (AFD) loss with an online task weighting (OTW) scheme is designed to selectively train layers for each task. As shown in {Fig.}~\ref{okd-pipeline}, the critical component AFD is an online weighted knowledge distillation performed on intermediate features from the shared ViT backbone of MTL, and the distilled features are from the teacher model that performs STL on each task. We denote by $L$ the total number of layers of the ViT encoder backbone and let $T$ denote the number of tasks. Then the AFD loss is defined as
\begin{equation}
\label{afd-loss}
    \mathcal{L}_{\text{AFD}} = \sum\nolimits_{l=1}^L \|\Bar{\mathcal{X}_l} - \sum\nolimits_{t=1}^T w^t_l\mathcal{X}_l^t \|^2_2
\end{equation}
where $w^{t}_l$ denotes the learnable parameters for the $t$-th task in the $l$-th layer, which balances the multiple tasks. $\Bar{\mathcal{X}_l}$ is the shared features learned from the teacher model at $l$-th layer. The shared features can be distilled for each task features $\mathcal{X}^t_l$ through {Eq.}~(\ref{afd-loss}) above. In the framework of OKD-MTL, the STL teacher and MTL students are trained in an end-to-end manner through the total loss
\begin{equation}
    \mathcal{L}_{total} = \mathcal{L}_{AFD} + \sum_{t=1}^{T}(\mathcal{L}_{STL}^t + \lambda_t\mathcal{L}_{MTL}^t).
\end{equation}
To mitigate the gap between the MTL and STL losses, OTW adjusts the task weight $\lambda_t$ for the $t$-th task at iteration $i$ as follows:
\begin{equation}
    \lambda_{t}(i) = T\frac{\exp{(r^{t}(i)}/k)}{\sum_{s=1}^T \exp{(r^s(i)/k)}}, r^{t}(i) = \frac{\mathcal{L}_{MTL}^{t}(i)}{\mathcal{L}_{STL}^{t}(i)}, t = 1, \cdots, T,
\end{equation}
where $k$ serves as the temperature hyperparameter to control this task weighting process, and $i$ represents the iteration index.

\underline{Response-Level.} Multi-Task Self-Training (MuST)~\citep{ghiasi2021multi} first trains\footnote{Pre-trained checkpoints are also recommended to alleviate computational burdens.} the classification, detection, and segmentation teacher models from scratch on ImageNet~\citep{deng2009imagenet, russakovsky2015imagenet}/JFT-300M~\citep{sun2017revisiting}, Objects365~\citep{shao2019objects365}, and COCO~\citep{kirillov2019panoptic}, respectively. The knowledge is then transferred from these specialized teachers to a general-purpose student model via pseudo-labeling. {Fig.}~\ref{must-pipeline} shows us as overview of MuST, every image in the shared dataset has supervision for all tasks, either supervised or pseudo labels. To balance these loss functions are tricky (See \textbf{\S}~\ref{subsec:scalar}) and MuST adopts $w_i = b^s{lr}_i^t / (b^t{lr}^s)$~\citep{goyal2017accurate} for ImageNet experiments, where $b$ denotes the batch size, $lr$ denotes the learning rate, the superscript indicates the student or teacher, and the total loss of MTL is defined as $\mathcal{L}_{total} = \sum_i w_i \mathcal{L}_i$. For  JFT300M, the algorithm in~\citet{kendall2018multi} was used to learn $w_i$ for each task. For depth loss, the weight $w_i$ was chosen by a parameter sweep. It has been validated that MuST can both rival supervised STL and enhance transfer learning performance.

CrossDistil (Cross-Task Knowledge Distillation)~\citep{yang2022cross} proposes a recommender framework that can transfer the fine-grained ranking knowledge about user’s preference towards items, as shown in {Fig.}~\ref{crossdistil}. To facilitate fine-grained ranking, the training samples are divided into multiple subsets, taking into account all possible combinations of the tasks. For instance, in a recommender system where two tasks involve predicting ``Buy'' and ``Like'' for an item, the potential task combinations include ``Buy:1, Like:1'', ``Buy:1, Like:0'', ``Buy:0, Like:1'', and ``Buy:0, Like:0''. For simplicity, the division of multiple subsets on two tasks are:
\begin{equation}
    \begin{cases}
    \mathcal{D}^{++} = \{(\boldsymbol{x}_i, y_i^{(1)}, y_i^{(2)})\in\mathcal{D}|y_i^{(1)} = 1, y_i^{(2)} = 1\}, \mathcal{D}^{+-} = \{(\boldsymbol{x}_i, y_i^{(1)}, y_i^{(2)})\in\mathcal{D}|y_i^{(1)} = 1, y_i^{(2)} = 0\}, \\
    \mathcal{D}^{-+} = \{(\boldsymbol{x}_i, y_i^{(1)}, y_i^{(2)})\in\mathcal{D}|y_i^{(1)} = 0, y_i^{(2)} = 1\}, \mathcal{D}^{--} = \{(\boldsymbol{x}_i, y_i^{(1)}, y_i^{(2)})\in\mathcal{D}|y_i^{(1)} = 0, y_i^{(2)} = 0\}, \\
    \mathcal{D}^{+\cdot} = \mathcal{D}^{++}\cup\mathcal{D}^{+-}, \mathcal{D}^{-\cdot} = \mathcal{D}^{-+}\cup\mathcal{D}^{--}, \mathcal{D}^{\cdot+} = \mathcal{D}^{++}\cup\mathcal{D}^{-+}, \mathcal{D}^{\cdot-} = \mathcal{D}^{+-}\cup\mathcal{D}^{--},
    \end{cases}
\end{equation}
where $\boldsymbol{x}$ represents the input feature vector from the whole dataset $\mathcal{D}$.

We denote by $\boldsymbol{x}^{++}\in\mathcal{D}^{++}$ and so forth.
The fine-grained ranking considers the corresponding multipartite
order $\boldsymbol{x}^{++}\succ\boldsymbol{x}^{+-}\succ\boldsymbol{x}^{-+}\succ\boldsymbol{x}^{--}$ instead of bipartite orders, e.g., $\boldsymbol{x}^{+\cdot}\succ\boldsymbol{x}^{-\cdot}$ or $\boldsymbol{x}^{\cdot+}\succ\boldsymbol{x}^{\cdot-}$, which may be contradictory among different tasks. Based on the fine-grained ranking, an augmented loss is introduced for each task as
\begin{equation}
    \label{ranking_teacher}
    \mathcal{L}_{\text{aug}} = -\sum\nolimits_{(\boldsymbol{x}^{++}, \boldsymbol{x}^{+-}, \boldsymbol{x}^{-+}, \boldsymbol{x}^{--})} [\beta_1\ln{\sigma{(\hat{r}_{++\succ+-})}} + \beta_2\ln{\sigma{(\hat{r}_{-+\succ--})}}] - \sum\nolimits_{(\boldsymbol{x}_{+\cdot},\boldsymbol{x}_{-\cdot})}\ln{\sigma(\hat{r}_{+\cdot\succ-\cdot})},
\end{equation}
where $\beta_1$ and $\beta_2$ are two hyper-parameters to balance the importance of pair-wise ranking relations and $\hat{r}$ is the logit value before the sigmoid function $\sigma$. Additionally, $\hat{r}_{++\succ--}=\hat{r}_{++} - \hat{r}_{--}$ and so forth. In contrast, the original regression-based loss function for each task is
\begin{equation}
    \label{regression_student}
    \mathcal{L}_{\text{CE}} = -\sum\nolimits_{\boldsymbol{x}_i\in\mathcal{D}}[y_i\ln \sigma(\hat{r_i}) + (1-y_i)\ln (1-\sigma(\hat{r_i}))].
\end{equation}
Based on Eqs. (\ref{ranking_teacher}) and (\ref{regression_student}), CrossDistil regards the learning task of augmented loss as teachers and the learning task of regression-based loss as students, the distillation loss for each of task is
\begin{equation}
    \mathcal{L}_{\text{KD}} = -\sum\nolimits_{\boldsymbol{x}_i\in\mathcal{D}}[\sigma(\Tilde{{r_i}}/\tau)\ln \sigma(\hat{r_i}/\tau) + (1-\sigma(\Tilde{r_i}/\tau))\ln (1-\sigma(\hat{r_i}/\tau))],
\end{equation}
where $\Tilde{r_i}$ is learned and calibrated from {Eq.}~(\ref{ranking_teacher}), and an error correction mechanism is applied to ensure its alignment with the hard label  $y_i$. The original regression loss and knowledge distillation loss contribute to the learning of students for multiple tasks as
\begin{equation}
    \mathcal{L}_{\text{MT}} = \sum\nolimits_{t=1}^T [(1-\alpha^{(t)})\mathcal{L}_{\text{CE}}^{(t)} + \alpha^{(t)}\mathcal{L}_{\text{KD}}^{(t)}],
\end{equation}
where $\alpha^{(t)}, t = 1, \cdots, T\in[0, 1]$ is a hyper-parameter to balance two loss functions. In this manner, by distilling the fine-grained ranking of task combinations, cross-task knowledge is effectively transferred.

\subsubsection{Cross-Task Attention}
\label{sec:attention}

Attention mechanism~\citep{niu2021review, brauwers2021general, guo2022attention} has been one of the most crucial concepts in RNNs, CNNs, and Transformers over the past decade in DL. Generally, attention is an information aggregation technique inspired by a human recognition system that tends to prioritize part of local regions over others when processing rich information. Under MTL settings, features from different tasks are more abundant than in STL, thus leading to a natural integration of the attention mechanism. 
Cross-task attention~\citep{bruggemann2021exploring}, encoding task-aware features into cross-task queries, can perform task-association via refinement of multi-source features. Unlike feature fusion methods~\citep{misra2016cross, ruder2019latent, gao2019nddr} that propagate task-shared information among different task-specific branches, cross-task attention calculates what/how to share based on cross-task comparison between source tasks and target task. Considering the "morphological" aspect, the hard compartmentalization effect caused by a block-structured communication matrix in feature fusion methods could preserve the interference of features in some cases for tasks. This dilemma could be alleviated with a soft, learnable form of task-aware feature attention.
Early works~\citep{xu2018pad, liu2019end, zhang2019pattern, zhou2020pattern, bruggemann2021exploring} build na\"ive attention modules (e.g., sigmoid function or inner product) to refine feature affinity or capture relational contexts across tasks, and then locate/diffuse features according to the attention map. PAD-Net~\citep{xu2018pad} and MTAN~\citep{liu2019end} select attentive features via an attention mask after the sigmoid activation. PAP~\citep{zhang2019pattern} and PSD~\citep{zhou2020pattern} iteratively diffuse features based on a cross-task affinity matrix. MTI-Net~\citep{vandenhende2020mti} first considers task interactions at multiple scales using both Sigmoid function and squeeze-and-excitation block~\citep{hu2018squeeze}.

Transformer-based works exploit long-range dependencies using self-attention mechanisms.

\begin{applebox}{Remarks}
    \begin{enumerate}[leftmargin=0.4cm, label=(\roman*)]
    \item Cross-task attention allows the model to focus on features that are more relevant to each specific task. This targeted attention helps in better feature extraction and can lead to improved task-specific performance, especially when tasks are related but not identical.
    \item The attention mechanism can adaptively weigh the contribution of each task during training, allowing for flexible balancing based on task difficulty or the amount of available data.
    \item Cross-task attention is a lightweight module that can leverage source-target pairwise similarity to refine task-specific features.
    \item Compared with direct feature fusion, the addition of attention mechanisms can lead to over-parameterization if not managed carefully, where the model has more parameters than necessary, complicating the learning process and increasing the risk of overfitting on the tasks with limited data.
    \end{enumerate}
\end{applebox}

\underline{Feature Filtering.} Multi-Task Guided Prediction-And-Distillation Network (PAD-Net)~\citep{xu2018pad} utilizes the predictions from hierarchical auxiliary tasks as multi-modal inputs to distill knowledge for the final tasks. As shown in {Fig.}~\ref{pad-module}, the framework of PAD-Net, a hard parameter sharing-based encoder, extracts common feature maps that can be used for different tasks, and then the decoder for each auxiliary task generates intermediate predictions for the usage of multi-modal distillation. The source paper proposes three distillation modules to incorporate useful multi-modal information for the final tasks. Suppose the feature maps from $s$-th task at $l$-th layer is denoted as $\mathcal{X}^s_l\in\mathbb{R}^{H\times W\times C}, s=1,\cdots,T$, which are transformed from predictions of $s$-th task via convolutional layers. The output feature maps for the usage of $t$-th task after the multi-modal distillation is represented as $\mathbf{\mathcal{X}}^{o,t}_{l+1}$. 

The first way to perform cross-modal distillation is a na\"ive concatenation via $\mathbf{\mathcal{X}}^{o}_{l+1}=[\mathbf{\mathcal{X}}^1_l, \cdots, \mathbf{\mathcal{X}}^T_l]$
$\in\mathbb{R}^{H\times W\times TC}$, which is then fed into the separate decoders for each task. Differently, the second way refines feature $\mathbf{\mathcal{X}}^t_l$ via passing knowledge from other tasks as below:
\begin{equation}
    \mathbf{\mathcal{X}}^{o,t}_{l+1} = \mathbf{\mathcal{X}}^t_l + \sum\nolimits_{s\neq t}^T CONV_{\Wmcal^{s\rightarrow t}}(\mathcal{X}^s_l), t = 1, \cdots, T,
\end{equation}
where $\Wmcal^{s\rightarrow t}$ denotes the weight tensor of convolutions that maps the $s$-th task to the $t$-th task. Furthermore, the third way utilizes the sigmoid function to filter the passing knowledge, which learns an attention map $\mathbf{G}^t$ for the $t$-th task as follows:
\begin{equation}
    \mathbf{G}^t = \sigma(CONV_{\Wmcal^t}(\mathcal{X}^t_l)), t = 1, \cdots, T.
\end{equation}
Then the knowledge is filtered via this attention map as follows:
\begin{equation}
    \mathbf{\mathcal{X}}^{o,t}_{l+1} = \mathbf{\mathcal{X}}^t_l + \sum\nolimits_{s\neq t}^T \mathbf{G}^t \odot CONV_{\Wmcal^{s\rightarrow t}}(\mathcal{X}^s_l), t = 1, \cdots, T.
\end{equation}
After the multi-modal distillation, the distilled feature
maps are up-sampled for the final pixel-level prediction tasks.

Multi-Task Attention Network (MTAN)~\citep{liu2019end} presents a novel MTL architecture based on task-specific feature-wise attention, while global features are shared across different tasks. Suppose the shared global features are denoted by $\mathcal{X}_l$ at the $l$-th layer, and the features learned from task $t$ are denoted by $\mathcal{X}_l^t$. Then the feature-wise attention on the global feature pool is computed as follows:
\begin{equation}
    \mathcal{X}_{l+1}^t = \sigma(\mathcal{X}_l^t)\odot\mathcal{X}_l,
\end{equation}
where $\mathcal{X}_{l+1}^t$ is then concatenated with the features from the global pool again and fed into the task-specific convolution blocks. The attention map $\sigma(\mathcal{X}_l^t)$ is learned in an end-to-end fashion as a parameter-free activation function.

To make the learning process more balanced between different tasks,~\citet{liu2019end} also suggests a simple yet effective Dynamic Weight Average (DWA) strategy (See \textbf{\S}~\ref{subsec:scalar}) to adjust losses according to their magnitudes in different epochs.

Multi-Scale Task Interaction Networks (MTI-Net)~\citep{vandenhende2020mti} aggregates multi-modal features at different scales from the decoder. As shown in {Fig.}~\ref{mti-net}, features at each scale are transformed and distilled by the feature propagation module and multi-modal distillation, respectively. This allows the model to capture task interactions at multiple scales. As the higher resolution scales have a limited receptive field, low-quality task-related features are presented. Simple upsampling and passing of task-related features from lower scales to higher scales~\citep{ronneberger2015u} inspire the design of the Feature Propagation Module (FPM). In this manner, features from different tasks at each scale are harmonized via the traditional convolutions and activation functions. To obtain the task-attentive features, a Sigmoid function along the task dimension is inserted to generate a task attention mask. To remedy the negative transfer among unrelated tasks, a per-task channel gating mechanism (SE, i.e. Squeeze-And-Excitation module~\citep{hu2018squeeze}) is used to refine the shared representations. 

Furthermore, suppose the feature maps for the task $s$ at scale $l~(\in\{1/4, 1/8, 1/16, 1/32\})$ represented by $\mathcal{X}^s_l, s=1,\cdots,T$, then the per-scale multi-modal distillation process for task $t$ is repeated as follows:
\begin{equation}
\label{multi-scale-multi-modal-attention}
    \mathcal{X}^{t}_l = \mathcal{X}^{t}_l + \sum\nolimits_{s\neq t} \sigma(CONV_{\Wmcal^{s\rightarrow t}_l}(\mathcal{X}^{s}_l))CONV_{\hat{\Wmcal}^{s\rightarrow t}_l}(\mathcal{X}^{s}_l), t = 1, \cdots, T,
\end{equation}
where the Sigmoid function $\sigma$ produces a spatial-wise attention mask to filter the features at different scales. $\Wmcal^{s\rightarrow t}_l$ and $\hat{\Wmcal}^{s\rightarrow t}_l$ denote the weights to map features before attention. The FPM and multi-scale multi-modal distillation result in distilled cross-task features at every scale, which are then fed into the final aggregation module. The predictions are based on decoding these final representations via a task-specific head for each task.

\underline{Feature Diffusion.}
Pattern-Affinitive Propagation (PAP)~\citep{zhang2019pattern} builds a cross-task affinity matrix based on a spatial-wise attention mechanism and then iteratively diffuses features on each of the tasks to refine affinitive patterns among tasks. The detailed architecture is shown in {Fig.}~\ref{pap-module}. Suppose the feature maps before the computing of task-specific affinity matrix are denoted by $\mathcal{X}^t_l\in\mathbb{R}^{H\times W\times C}$, the affinity matrix for each task is computed using the inner product between each pair of spatial-wise feature vector with the length of $C$:
\begin{equation}
    \Xbold^t_l = RESHAPE(\mathcal{X}^t_l)\in\mathbb{R}^{HW\times C}, \Mbold^t = \Xbold^t_l{\Xbold^t_l}^\top\in\mathbb{R}^{HW\times HW}, t = 1, \cdots, T,
\end{equation}
where $RESHAPE(\cdot)$ is used to preserve the channel dimension. If the affinity matrix of each task is weighted by a learnable parameter $\alpha_t (t=1,\cdots,T, \text{and} \sum\nolimits_{t=1}^T\alpha_t=1)$, then the final affinity matrix for the task $t$ can be adaptively combined as follows:
\begin{equation}
    \hat{\Mbold}^s = \sum\nolimits_{t=1}^T\alpha_t^s\Mbold^t, s = 1, \cdots, T,
\end{equation}
which is an adaptive combination process that can propagate the cross-task affinitive patterns for the target $s$-th task. Furthermore, the cross-task affinitive patterns are used to iteratively diffuse features for each task:
\begin{equation}
    \Xbold^t_{l}(i+1) = \hat{\Mbold}^t\cdot \Xbold^t_{l}(i), t = 1,\cdots,T, i = 0, 1, \cdots, i_{\text{max}},
\end{equation}
where $i$ denotes the diffusion step. In general, the multi-step iterative diffusion process propagates the affinity information best. Suppose the maximum of step is $i_{\text{max}}$, finally the feature maps in the next layer are computed as follows:
\begin{equation}
    \Xbold^t_{l+1} = \beta\cdot\Xbold^t_{l}(i_{\text{max}}) + (1 - \beta)\cdot\Xbold^t_{l}(0), \mathcal{X}^t_{l+1} = RESHAPE(\Xbold^t_{l+1})\in\mathbb{R}^{H\times W\times C}, t=1, \cdots, T,
\end{equation}
where $\beta$ is a hyperparameter to control the feature consistency.

Pattern-Structure Diffusion (PSD)~\citep{zhou2020pattern} utilizes a shared CNN encoder to extract feature maps that can be fed into the task-specific decoders, where the pattern structures are distilled within intra-task and across inter-task. As shown in {Fig.}~\ref{psd-module}, the intra-task PSD is used to transmit pattern structure within each task to enhance the task-specific patterns and then connect with inter-task PSD to correlate relations of pattern structures across different tasks. Without loss of generality, we assume a $l\times l$ patch cropped at each position of feature maps $\mathcal{X}\in \mathbb{R}^{H\times W\times C}$ as $\mathcal{X}_{P_i} \in \mathbb{R}^{l\times l\times C}$, where $P_i$ means the pattern at position $i$. Then the pattern structure can be defined from the KNN graph on $l\times l$ points within $\mathcal{X}_{P_i}$ as follows:
\begin{equation}
    [\Abold_{P_i}]_{j,k} = \exp{\{- \|RESHAPE(\mathcal{X}_{P_i})_j - RESHAPE(\mathcal{X}_{P_i})_k\|_2^2/\tau^2\}},  i = 1, \cdots, HW, j,k = 1, \cdots, l^2,
\end{equation}
where $\tau$ is a fixed hyper-parameter set by user. To make pattern structure at different scale comparable, $\Abold_{P_i}$ is further normalized as follows:
\begin{equation}
    \Abold_{P_i} \leftarrow \Abold_{P_i} / (\boldsymbol{1}^\top \Abold_{P_i} \boldsymbol{1}).
\end{equation}
Then the intra-task PSD can be formulated as a recursive process:
\begin{equation}
    [RESHAPE(\mathcal{X}^{i+1})]_j = [RESHAPE(\mathcal{X}^i)]_j + \beta\sum\nolimits_{k\in\mathcal{N}(v_j)} \Abold_{j,k}\times[RESHAPE(\mathcal{X}^i)]_k,
\end{equation}
where $\Abold$ denotes the pattern structure of the whole feature map, $\mathcal{N}(v_j)$ is the neighbor set of the target pixel $v_j$, and $\beta$ is a fixed hyper-parameter to control the residual connection. The iteration above contains multiple steps to guarantee that each local pattern is spread into distant regions, which is a diffused process.

To achieve cross-task pattern-structure propagation, inter-task PSD transfers the patterns from other tasks as follows:
\begin{align}
    [RESHAPE(\mathcal{X}^{(t)})]_j = &[RESHAPE(\mathcal{X}^{(t)}]_j + \sum\nolimits_{s\neq t}\sum\nolimits_{k\in\mathcal{N}(v_j)} \beta_{s\rightarrow t} \Abold_{j,k}^{s\rightarrow t}\times[RESHAPE(\mathcal{X}^{(t)})]_k, \nonumber \\
    \text{s.t.} \quad & \Abold_{P_i}^{s\rightarrow t} = \Abold_{P_i}^{(t)} \odot \Abold_{P_i}^{(s)} / [\boldsymbol{1}^\top(\Abold_{P_i}^{(t)} \odot \Abold_{P_i}^{(s)})\boldsymbol{1}], s,t = 1,\cdots,T,
\end{align}
where $\{\Abold_{P_i}^{ts}\}_{s\neq t}$ represent the
transferred pattern-structures from task $s$ to the target task $t$. In this manner, the PSD method distills feature similarity across different tasks.

\underline{Soft Attention.} Attentive Single-Tasking of Multiple Tasks~(ASTMT)~\citep{maninis2019attentive} argues the dilemma that the critical information from one task to another could be a nuisance while inferring multiple tasks together. ASTMT addresses it by single-tasking, a strategy that executes one task at a time instead of inferring all of them simultaneously. Technically, every task shares a backbone network in a hard manner but adapts its specificity with residual adapter (RA) branches, which is shown in {Fig.}~\ref{astmt}. Suppose the RA operation is represented by $RA_t$ for the $t$-th task, and its original residual skip connection is $R$. Then the single-tasking process by RA is calculated as below:
\begin{equation}
    \mathcal{X}_{l+1}^t = \mathcal{X}_{l}^t + R(\mathcal{X}_{l}^t) + RA_t(\mathcal{X}_{l}^t), t = 1, \cdots, T,
\end{equation} 

where $R$ denotes the residual connection that is not influenced by the task. $RA_t$ can be na\"ive bottleneck convolutions or transformed to an attentive block $SE_t$ (e.g. SE-ResNet block~\citep{hu2018squeeze}). In order to address the limitation of this adaptation failing to disentangle the shared and task-specific space, a GRadiEnt Adversarial Training (GREAT) process~\citep{sinha2018gradient} is introduced for different tasks to ensure that the shared backbone learns the shared representations and maintains this quality during the single-tasking process. More details of multi-task adversarial training are shown in \textbf{\S}~\ref{sec:adversarial}.

\input{tex_files/02-2/context-pooling}

Adaptive Task-Relational Context (ATRC) module~\citep{bruggemann2021exploring} enables global cross-task and local spatial-wise attention mechanisms to refine each task prediction, which is a general module that can be applied to any backbones across any supervised dense prediction tasks. The ATRC refinements begin with a hard-parameter sharing encoder, of which each task head can generate task-specific features $\mathcal{X}_t$ and auxiliary predictions $\mathcal{P}_t$, where $t=1,\cdots,T$. Specifically, the features $\mathcal{X}_t$ of each target task $\mathcal{T}_t$ is refined by attending to the features $\mathcal{X}_s$ of every available task $\mathcal{T}_s, s\in\{1, \cdots, T\}$ within a separate Context Pooling (CP) block. As shown in {Fig.}~\ref{atrc-module}, the original features $\mathcal{X}_t$ and refined features $\{\mathcal{X}_{s\rightarrow t}\}_{s=1}^T$ are combined to predict the target task $\mathcal{T}_t$.

There are three categories of context information (global context, local context, and label context) to be learned via refining features from the source task to the target task. The detailed illustration can be observed in Fig.~\ref{context_pooling} positioned to the right. Each CP block accepts the features $\mathcal{X}_s, \mathcal{X}_t$ and predictions $\mathcal{P}_s, \mathcal{P}_t$ from the source task and target task, respectively. $\mathcal{X}_t$ and $\mathcal{X}_s$ are transformed into queries $\Qbold$, keys $\Kbold$ and values $\Vbold$ (flattening along the spatial dimension and preserving channel dimension) as below:

\begin{align}
    & \Qbold = RESHAPE(CONV_{\Wmcal_q}(\mathcal{X}_t)), \Kbold = RESHAPE(CONV_{\Wmcal_k}(\mathcal{X}_s)), \nonumber\\
    & \Vbold = RESHAPE(CONV_{\Wmcal_v}(\mathcal{X}_s)),
\end{align}

where $CONV_*(\cdot)$ is a $1\times 1$ CONV-BN-ReLU operation, and $\Qbold, \Kbold, \Vbold\in\Rmbb^{HW\times C}$. 
In the attention of global context, a target feature value $v_i$ at position $i$ is substituted with
\begin{equation}
    \vbold_i^\prime = \sum\nolimits_{j=1}^L\text{sim}(\qbold_i,\kbold_j)\vbold_j / \sum\nolimits_{j=1}^L\text{sim}(\qbold_i,\kbold_j), i = 1, \cdots, L,
\end{equation}
where $L$ denotes the number of total pixels (i.e. feature values) and sim$(\cdot,\cdot)$ denotes an arbitrary similarity function. For the local context attention, let us denote by $\mathcal{N}_p(i)$ the 2D spatial neighborhood of target pixel at position $i$ with the patch extent $p$, then the spatial-wise local attention is formulated as below:
\begin{equation}
    \vbold_i^\prime = \sum\nolimits_{j\in\mathcal{N}_p(i)}\text{softmax}(\qbold_i \kbold_j/\sqrt{C})\vbold_j, i = 1, \cdots, L,
\end{equation}
where $C$ is the channel dimension of $\Kbold$. For the $T$-label context and $S$-label context defined in the label space that is partitioned into a set of disjoint label regions. The aim is to find a prototypical representation for each pixel. Suppose $\mathcal{P}_t\in HW \times R_t$, where each entry of the last dimension indicates the degree that a pixel belongs to a label region $r\in\{1,\cdots, R_t\}$. For the $T$-label context, the keys $\Kbold$ and values $\Vbold$ are calculated via the the region prototypes as below:
\begin{equation}
    \Kbold = CONV_{\Wmcal_k}({\hat{\mathcal{P}}_t}^\top RESHAPE(\mathcal{X}_s)), \Vbold = CONV_{\Wmcal_v}(\hat{\mathcal{P}}_t^\top RESHAPE(\mathcal{X}_s)), 
\end{equation}
where $\hat{\mathcal{P}}_t$ denotes the softmax normalization over the spatial dimension, and the matrix $\hat{\mathcal{P}}^\top$ $RESHAPE(\mathcal{X}_s)\in\mathbb{R}^{R_t\times C}$ represents the region prototypes. Alternatively, $\mathcal{P}_t$ is substituted with the source task prediction maps $\mathcal{P}_s$ in the $S$-label context. The outputs of both are attention-weighted combinations of features $\vbold$:
\begin{equation}
    \vbold^\prime = \text{softmax}(\qbold \kbold^\top/\sqrt{C})\vbold.
\end{equation}


Deformable Mixer Transformers (DeMT)~\citep{zhang2023demt} is an encoder-decoder architecture that combines the merits of deformable CNNs~\citep{dai2017deformable, zhu2019deformable} and attention-based ViT~\citep{dosovitskiy2021an} to model multiple tasks, the details are shown in Fig.~\ref{demt-block}. The encoder, aka the deformable mixer in~\citet{zhang2023demt}, is aware of feature mixing across channels through $1\times 1$ convlutions and captures the deformable spatial features through learnable offsets. After task-specific features are learned by the encoder part, the task-aware transformer decoder first applies the task interactions based on the attention mechanism (MHSA + MLP) and then constructs the task query block to decode the task awareness features for each task. Suppose the transformer operator inside the task interaction block can be abstracted as 
\begin{equation}
    \Xmcal_{l+1} = MHSA_{inter}(q = LN(\Xmcal_{l}), k = LN(\Xmcal_{l}), v = LN(\Xmcal_{l})),
\end{equation}
where $LN$ denotes the layer norm on fused feature $\Xmcal_{l}$, and the subscripts $l$ and $l+1$ denote the feature index before and after the task interaction block, respectively. To decode task awareness in the task query block, another transformer involves task-specific query before $MHSA_{inter}$ (i.e., $\Xmcal_{l}^t$):
\begin{equation}
    \Xmcal_{l+2}^t = MHSA_{query}(q = LN(\Xmcal_{l}^t), k = LN(\Xmcal_{l+1}), v = LN(\Xmcal_{l+1})), t = 1, \cdots, T,
\end{equation}
where the subscript $l+2$ denotes the feature index after the task query block.

\begin{applebox}{Remarks}
    \begin{enumerate}[leftmargin=0.4cm, label=(\roman*)]
    \item Knowledge distillation can utilize and transfer interpretable patterns across multiple tasks, resulting in meaningful principles that can provide guidance for architectural design.
    \item Knowledge distillation has the capability to aggregate refined features from multiple tasks at various scales, thereby enhancing task generalization ability and significantly improving performance.
    \item Knowledge distillation allows for the creation of smaller and more efficient student models on target tasks. The distilled knowledge helps compress the complex teacher model into a more lightweight student model while retaining a comparable level of performance.
    \item Knowledge distillation enables the transfer of knowledge across tasks, even if they are different or loosely related. This flexibility allows for leveraging insights from related tasks to enhance the learning process, resulting in better performance on each individual task.
    \item The overall performance heavily depends on the quality and capabilities of the teacher model. If the teacher model is not well-trained or lacks expertise in the specific tasks, the knowledge distillation process may not be effective, limiting the potential benefits.
    \item Implementing knowledge distillation adds extra computational complexity that often involves the processes of training, transferring, and fine-tuning, thus inevitably being time-consuming and resource-intensive.
    \end{enumerate}
\end{applebox}

\subsubsection{Scalarization Approach.}
\label{subsec:scalar}
One of the most popular methods to solve multi-task learning problems is the scalarization approach, which formulates the problem as a linear combination of loss functions of different tasks~\citep{kendall2018multi, liu2019end, chen2018gradnorm, Senushkin_2023_CVPR} as
\begin{equation}\label{eq:MTK.sclarization}
    \min_{\Wbold} \Lmcal_{\text{total}}(\Wbold) = \sum_{t=1}^{T} \alpha^{(t)}\Lmcal^{(t)}(\Wbold)
\end{equation}
where $\{\alpha^{(t)}\}_{t=1}^{T}\subset\R{}_{+}$ are the tasks' weights and are used to encode preferences over different tasks. $\Wbold$ is the model parameter and $\{\Lmcal^{(t)}\}_{t=1}^{T}$ are loss functions for different tasks. In each loss function $\Lmcal^{(t)}$, we drop the dependency on training samples $\{\Xbold^{t},\ybold^{t}\}$ to avoid cluttered notations.

Gradient-based methods are perhaps the most popular choices to solve Eq.~(\ref{eq:MTK.sclarization}), whose update rule of $\Wbold$ takes the form of $\Wbold\gets \Wbold + \eta \dbold$, where $\eta>0$ is the learning rate and $\dbold$ is the search direction. $\dbold$ is a function of $\{\alpha^{(t)}\grad, \Lmcal^{(t)}\}_{t=1}^{T}$, for example, $\dbold=-\sum_{t=1}^{T}\alpha^{(t)}\grad \Lmcal^{(t)}(\Wbold)$.  Aside from the challenge of choosing a proper learning rate $\eta$, there are two additional challenges, \textit{dominant gradients}  and \textit{conflicting gradients}, see Fig.~\ref{fig:mtl.gradient.issues} for an illustration. Dominating gradient issue occurs when the norm of gradients of some tasks' losses are significantly larger than the others, hence the updating direction $\dbold$ are biased towards to tasks with larger gradient norm. Conflicting gradients issue arises when one makes progress in one task, the performance of another task is degraded. 

\input{tex_files/02-2/gradients_issues}

In the remainder of this section, we review some works with different philosophies to address dominant and conflicting gradients' challenges. These methods can be roughly characterized as \textit{gradient correction} approach, where transformations are made to gradients to address the conflicting gradients issue and \textit{dynamic weighting}, where $\{\alpha^{(t)}\}$ are updated in each iteration to address the dominant gradients issue. 

\input{tex_files/02-2/PCGrad}
\underline{\emph{Gradient Correction.}}  Projecting Conflicting Gradients ({PCGrad})~\citep{yu2020gradient} proposes to mitigate the conflicting gradients issue by projecting the conflicting gradients in the orthogonal subspace. Formally, {PCGrad}~\citep{yu2020gradient} defines two gradients $(g_i,g_j)$ to be conflicting if $g_i^Tg_j<0$. To address this issue, instead of forming the search direction as $\dbold=-(g_i+g_j)$, {PCGrad} suggested using $\dbold=-(\textbf{Proj}_{n_j}(g_i)+\textbf{Proj}_{n_i}(g_i))$, where $n_i^Tg_i=0$ and $n_j^Tg_j=0$ and \textbf{Proj} is the Euclidean projection operator. See Fig.~\ref{fig:PCGrad} for an illustration. This method, from the perspective of multi-objective optimization perspective (which will be discussed in the next section), is a particular choice of choosing a common descent direction. 
Gradient sign Dropout ({GradDrop})\citep{chen2020just} attributed conflicts to the differences in the signs of gradients along each coordinate direction. Motivated by the dropout, a probabilistic masking procedure is proposed to keep only gradients consistent in signs in each update. 
Conflict-Averse Gradient descent ({CAGrad})~\citep{liu2021conflictaverse} proposes to mitigate gradient conflicts
by solving the problem
\begin{equation}\label{eq:cagrad}
\max_{\dbold}\min_{t\in[T]} \grad \Lmcal^{(t)}(\Wbold)^T (-\dbold) \text{ s.t. } \norm{\dbold - \grad \Lmcal_{\text{total}}(\Wbold)} \leq c \norm{ \grad \Lmcal_{\text{total}}(\Wbold)},
\end{equation}
where $c>0$ is a prescribed parameter. The intuition is that $-\min_{t\in[T]} \grad \Lmcal^{(t)}(\Wbold)^T d$ can be used as the approximated evaluation of the conflict among objectives, and one wants to find the direction $\dbold$ that minimizes such a conflict while stays close to the original negative gradient of $\Lmcal_{\text{total}}(\Wbold)$. 

Reducing conflicting gradient ({Recon})~\citep{shi2023recon} empirically observes that {PCGrad}, {GradDrop}, and {CAGrad}~\citep{yu2020gradient, chen2020just,liu2021conflictaverse} can only slightly reduce the occurrence of conflicting gradients (compared to joint-training\footnote{The joint-training refers to the case that $\alpha^{(t)}=1$ for all $t\in[T]$ in Eq.~\eqref{eq:MTK.sclarization}.}) in some cases, and in some other cases they even increase the occurrence. Therefore, {Recon} proposed to analyze parameters in a layer-wise fashion to pinpoint the shared parameters that are most likely to incur conflicting gradients. Concretely, let $(g_i^{k}, g_j^{k})$ be the gradients of the $(i,j)$ task pair with respect to the $k$th layer's parameters. $(g_i^{k}, g_j^{k})$ is said to be $S$-conflicting if $s^{k}:=\frac{\langle g_i^{k}, g_j^{k}\rangle}{ \norm{g_i^{k}}\norm{g_j^{k}} }<S$ for any $s\in[-1,0)$. {Recon} first trained the models via any gradient-based method with $E$ epochs, e.g., {PCGrad}, {GradDrop}, and {CAGrad}, and then
derived the conflicting scores for each layer over $E$ epochs to identify the top $K$ layers with the highest (most negative) conflicting scores. Finally, {Recon} turned these $K$ layers' parameters into task-specific parameters and retrained the network from scratch. As pointed out in \citet{shi2023recon}, while {Recon} is sensitive to the parameters $K$ and $S$, one only needs to tune them once for a given network architecture.

\underline{\emph{Dynamic weighting.}}
{GradNorm} proposed in ~\citet{chen2018gradnorm} suggests to mitigate the dominant gradient issue so that gradients for each task have the proper magnitude. The strategy to adjust $\{\alpha^{(t)}\}$ is based on the average gradient norm of each task and the relative progress achieved for each task. With this information, {GradNorm} constructs a reference point at each iteration, $\{\alpha^{(t)}\}$ was then selected to minimize the $\ell_1$ distance between the actual gradient of each task and the reference point. Concretely, let $GN_{\Wbold}^{(t)}(i) = \|\grad_{\Wbold}\alpha^{(t)}(i)\mathcal{L}^{(t)}(i)\|_2$ be the measure of $\ell_2$ norm of the $t$th task's weighted gradient at iteration $i$\footnote{We add addition index $i$ to indicate their dependence on the iteration counter $i$.}. Next, the averaged gradient norm across all tasks was calculated as $\overline{GN}_{\Wbold}(i) = \mathbb{E}_{t}[GN_{\Wbold}^{(t)}(i)]$. To measure the training progress of each task, $\tilde\Lmcal^{(t)}(i) =\frac{\mathcal{L}^{(t)}(i)}{\mathcal{L}^{(t)}(0)}$ was introduced, which inversely proportional to the training rate. Lastly, the relative inverse training rate for task $t$ can be formulated as $r^{(t)}(i) = \frac{\mathcal{L}^{(t)}(i)}{\mathcal{L}^{(t)}(0)}/\mathbb{E}_{t}[\tilde\Lmcal^{(t)}(i)]$. The higher value of $r^{(t)}(i)$ indicates a higher gradient magnitude for task $t$ at iteration $i$, which encourages task $t$ to learn more quickly. Finally, the weight $\alpha^{t+1}$ was determined by solving the following problem
\begin{equation}
    \min_{\{\alpha^{(t)}\}_{t=1}^{T}}\sum\nolimits_{t=1}^{T}\|GN_{\Wbold}^{(t)}(i) - \overline{GN}_{\Wbold}(i)\cdot{[r^{(t)}(i)]}^\zeta\|_1,
\end{equation}
where $\zeta$ is introduced to avoid dramatically different learning dynamics between tasks caused by various task complexity. Inspired by {GradNorm}, Dynamic Weight Averaging~({DWA}) is another strategy proposed in~\citet{liu2019end} to balance the task-specific losses. The updating process of $\alpha^{(t)}(i)$ is defined as $
    \alpha^{(t)}(i) = \frac{\sum_t\alpha^{(t)}(i) e^{r^{(t)}(i-1)/T}}{\sum_{t=1}^T e^{r^{(t)}(i-1)/T}} \text{ and } r^{(t)}(i-1) = \frac{\mathcal{L}^{(t)}(i-1)}{\mathcal{L}^{(t)}(i-2)},$
where $r^{(t)}(i)$ is the relative progress for the task $t$ at the iteration $i$. Reinforced MTL ({RMTL})~\citep[Chapter 3]{liu2018exploration} adjusts $\{\alpha^{(t)}\}$ using the reinforcement learning strategy and Loss-Balanced Task Weighting. {LBTW}~\citep{liu2019loss} combines {GradNorm} and {RMTL} in a way such that the weights $\{\alpha^{(t)}\}$ were adapted to both samples and tasks. Impartial MTL ({IMTL})~\citep{liu2021towards} proposes to update $\{\alpha^{(t)}\}$ in each iteration such that the aggregated gradient $\sum_{t=1}^{T}\alpha^{(t)}\grad \Lmcal^{(t)}(\Wbold)$ has equal projections onto the raw gradients of individual tasks. It achieves this goal by solving the following linear system (with respect to $\{\alpha^{(t)}\}$)
\begin{align*}
&\left(\sum_{t=1}^{T}\alpha^{(t)}\grad \Lmcal^{(t)}(\Wbold)\right)^T\frac{\grad \Lmcal^{(t)}(\Wbold)}{\norm{\grad \Lmcal^{(t)}(\Wbold)}} = \left(\sum_{t=1}^{T}\alpha^{(t)}\grad \Lmcal^{(t)}(\Wbold)\right)^T\frac{\grad \Lmcal^{(1)}(\Wbold)}{\norm{\grad \Lmcal^{(1)}(\Wbold)}}, \text{ for } t\in\{2, \cdots, T\} \\
&\sum_{t=1}^{T}\alpha^{(t)} = 1.
\end{align*}
Before solving for $\{\alpha^{(t)}\}_{t=1}^{T}$, {IMTL} also proposes a heuristic to scale $\{\Lmcal^{(t)}(\Wbold)\}$ such that all losses are in the similar scales, which essentially is another scaling of the  $\{\grad \Lmcal^{(t)}(\Wbold)\}$. {Achievement-based MTL} \citep{yun2023achievement} suggests defining the weights for each task by measuring the training progress as $\alpha^{(t)}=(1-\text{acc}_t/(m\cdot\text{maxacc}_t))^{\gamma}$ where $\gamma>0$, $m>1$, $\text{acc}_t$ and $\text{maxacc}_t$ are the current training accuracy (trained in the multitask setting) for the task $t$ and the max training accuracy (trained in the single setting), respectively. And Achievement-based MTL considers using the geometric mean instead of arithmetic mean to define the loss function; namely, it solves $\min_{\Wbold} \left(\prod_{t=1}^{T}(L^{(t)}(\Wbold))^{\alpha^{(t)}}\right)^{1/T}$.

Uncertainty Weighting~\citep{kendall2018multi} takes a different perspective from the above dynamic weighting approaches. This work assumes there are underlying distributions for different tasks' labels, and different tasks are independent. The final loss function, deriving from the likelihood perspective, takes the same form as Eq. \eqref{eq:MTK.sclarization} with $\{\alpha_k^{(t)}\}$ being specified as the reciprocal of the variance of each distribution used to modeling each task and loss function. Instead of just optimizing over the parameter $\Wbold$, \citet{kendall2018multi} optimizes $\Wbold$ and $\{\alpha^{(t)}\}$ simultaneously
\begin{equation}\label{eq:MTK.sclarization.ext}
    \min_{\Wbold, \{\alpha^{(t)}\}_{t=1}^{T}} \Lmcal_{\text{total}}(\Wbold) = \sum_{t=1}^{T} \alpha^{(t)}\Lmcal^{(t)}(\Wbold). 
\end{equation}

At this point, one can observe that all aforementioned works under the dynamic weighting category, excluding \citet{kendall2018multi}, do not necessarily respect optimization problem formulation in Eq.~\eqref{eq:MTK.sclarization} even though they empirically work well in producing useful solutions. Nonetheless, one can also regard the dynamic weighting approach as either solving Eq.~\eqref{eq:MTK.sclarization.ext} using different rule-based strategies to update $\{\alpha^{(t)}\}_{t=1}^{T}$ or using gradient-based methods to inexactly solve a sequence of problems in the form of Eq. \eqref{eq:MTK.sclarization}.

To conclude this section, we point out that there are some works that try to address two issues simultaneously~\citep{javaloy2022rotograd, Senushkin_2023_CVPR}. For example, Alignment for MTL (Aligned-MTL~\citep{Senushkin_2023_CVPR} considers 
the condition number of the linear system $\dbold = \Gbold\boldsymbol{\alpha}$ as a measure of the degree of the severeness of both gradient dominance and conflict, where $\Gbold=[-\grad \Lmcal^{(1)}, \ldots, -\grad \Lmcal^{(T)}]$ and $\boldsymbol{\alpha}=(\alpha^{(1)}, \ldots, \alpha^{(T)})^T$. Therefore, the authors propose to find well-conditioned $\hat \Gbold$ to approximate $\Gbold$ and, therefore, obtain a refined update direction $\hat \dbold$. Concretely, the author proposed to solve $\min_{\hat\Gbold}\norm{\hat\Gbold - \Gbold} \quad  \text{s.t.} \quad \hat\Gbold^T\hat\Gbold = I$,
by singular value decomposition (SVD)
and use the refined direction $\hat\dbold = \hat \Gbold \boldsymbol{\alpha}$ instead of $\dbold = \Gbold\boldsymbol{\alpha}$. 
The convergence rate of the proposed algorithm is established under the assumption that all loss functions are Lipschitz smooth and bounded below.
Although the numerical results are promising, one should be aware of the computation cost of the SVD despite the existence of efficient algorithms \citep{bondhugula2006fast}.

\begin{applebox}{Remarks}
    \begin{enumerate}[leftmargin=0.4cm, label=(\roman*)]
    \item Scalarization approach features in its simplicity as it transforms a multi-objective problem into a single-objective one. Hence, it is easy to implement, and many off-shelf optimizers can be applied.
    \item Generally, the scalarization approach has computational efficiency advantages over multi-objective optimization approaches, as will be discussed in the next section.
    \item The solution found by the scalarization approach might lack diversity as it could be biased to a solution depending on prescribed weights\footnote{We characterize the diversity through the Pareto Front, which will be discussed in the next section.}. Also, it is hard to conduct convergence analysis, especially for the dynamic weighting approach, since it attempts to solve a sequence of problems inexactly.
    \end{enumerate}
\end{applebox}

\subsubsection{Multi-objective Optimization (MOO).}
\label{subsec:multi-obj-opt}
In contrast to the scalarization approach, which converts different objective functions $\{\Lmcal^{(1)}, \ldots, \Lmcal^{(T)}\}$ into one aggregated objective function $\Lmcal_\text{total}$ and then optimizes it, MOO, 
aims to \textit{simultaneously} optimizing several objective functions (potentially conflicting). Concretely, MOO aims to solve the following problem
\begin{equation}\label{prob:moo}
    \min_{\Wbold}\Lmcal(\Wbold)=(\Lmcal^{(1)}(\Wbold), \ldots, \Lmcal^{(T)}(\Wbold))^T \quad \text{s.t.}\quad \Wbold\in\Fmcal,
\end{equation}
where $\Fmcal$ is the feasible domain for $\Wbold$ (examples will be given shortly). For a comprehensive background on the MOO topic, we refer readers to~\citet{ehrgott2005multicriteria}; for readers who prefer a quick overview of this subject, we recommend~\citet{liu2020review}. Below, we just provide the minimum backgrounds required to make the exposition accessible to readers with backgrounds in single objective optimization.

We begin with a few concepts that help readers understand the type of solutions that MOO algorithms can normally obtain.
\begin{definition}
\item 
\begin{enumerate}
    \item $\Wbold^*$ is called a \textbf{weak Pareto minimizer} of $\Lmcal$ over $\Fmcal$ if there is no $\Wbold\in\Fmcal$ such that $\Lmcal(\Wbold)<\Lmcal(\Wbold^*)$. Here, $<$ is the element-wise comparison. The set $PF(\Lmcal)=\{\Lmcal(\Wbold^*)~|~\Wbold^*$ is a weak Pareto minimizer$\}$ is called the \textbf{Pareto front}.
    \item $\Wbold^*$ is called a \textbf{strict Pareto minimizer} of $\Lmcal$ over $\Fmcal$ if there is no $\Wbold\in\Fmcal$ such that $\Lmcal(\Wbold)\leq\Lmcal(\Wbold^*)$ and $\Wbold\neq \Wbold^*$.Here, $\leq$ is the element-wise comparison. 
    \item 
    $\Wbold^*$ is called a \textbf{Pareto stationary point} of $\Lmcal$ over $\Fmcal$ if  $\max_{t=1,\cdots,T}(\Wbold-\Wbold^*)^T\grad \Lmcal^{(t)}(\Wbold^*)\geq 0$ for all $\Wbold\in\Fmcal$. Intuitively, this definition implies that for the objective function, there exists at least one such that there does not exist any feasible direction $d:=\Wbold-\Wbold^*$ to further decrease it. 
\end{enumerate}
\end{definition}

We give a graphical illustration of all these Pareto-related points in Fig.~\ref{fig:Pareto}. In Fig.~\ref{fig:Pareto}, the $\Wbold^*$s that correspond to circles and crosses are Pareto stationary points. However, when $\{\Lmcal^{(t)}(\Wbold)\}_{t=1}^{T}$ are not convex, the Pareto stationary points can generate $\{\Lmcal^{t}(\Wbold^*)\}$ that are NOT sit on the Pareto font. An analogy for this phenomenon in single objective optimization would be that a stationary point of a nonconvex objective function may not be the global minimum. Due to the nonconvexity nature of neural networks, algorithms considered here (when the convergence analysis is provided), if not all,  can only guarantee to find the \textbf{Pareto stationary point} instead of the weak/strict Pareto minimizers. However, if additional assumptions like (strong) convexity are assumed, then one can obtain solutions whose objective values are on the Pareto front. In the sequel, we review some works with different strategies to generate the a (set of) Pareto stationary point(s).

\input{tex_files/02-2/pareto}

The first line of works, e.g.,~\citet{sener2018multi,lin2019pareto,navon2022multi} were built upon and extended the seminal work, Multiple-Gradient Descent Algorithm (MGDA)~\citep{fliege2000steepest} to the neural network settings. The essence of MGDA is, at each iteration, to find a common descent direction $\dbold$ that decreases all objective functions $\{\Lmcal^{(t)}\}$ simultaneously. If no such direction exists, the algorithm terminates and returns a (set of)  Pareto stationary point(s). MGDA constructs the \textbf{common descent direction} $\dbold$ by solving the following optimization problem\footnote{For simplicity, we now only consider the unconstrained case $\Fmcal=\R{n}$; we will discuss the constrained case $\Fmcal\subset\R{n}$ shortly.}

\begin{minipage}{0.6\textwidth}
\begin{equation}\label{prob:mgda.primal}
    \max_{\dbold\in\R{n}}\min_{t=1, \ldots, T} \left(-\grad \Lmcal(\Wbold)^{(t)}\right)^{T} d+\frac{1}{2}\norm{d}^2.
\end{equation}
\end{minipage}

In problem \eqref{prob:mgda.primal}, if we drop the second order term $\frac{1}{2}\norm{\dbold}^2$, it intuitively tries to find the search direction $\dbold$ that can maximize the minimal progress\footnote{The progress is measured by the difference between of $\Lmcal^{(t)}(\Wbold)$ and the first order Taylor approximation of $\Lmcal^{(t)}$ at $\Wbold+\dbold$.} can be made. The second order term is added to guarantee the uniqueness of the solution of problem~\eqref{prob:mgda.primal}. The solution $\dbold$ is known as the \textit{steepest common descent direction} in the optimization literature. In deep neural network applications, however,  $n$ can be of the billion scale, so it is very challenging to solve problem~\eqref{prob:mgda.primal} directly. Instead of solving ~\eqref{prob:mgda.primal}, MGDA-MTL~\citep{sener2018multi} considers to the solve the dual problem  
\begin{equation}\label{prob:mgda.dual}
    \min_{\beta\in\R{T}} \frac{1}{2}\norm{\sum_{t=1}^{T}[\beta]_{t}\grad \Lmcal^{(t)}(\Wbold)}^2 \quad \text{s.t.} \quad \sum_{t=1}^{T}[\beta]_{t}=1 \text{ and } [\beta]_t\geq 0 \text{ for all } t,
\end{equation}
where $[\beta]_t$ is the $t$-th element of the vector $\beta$. One can see that the dual problem's dimension reduces to $T$, which is usually smaller than $n$ in several orders of magnitude and can be solved efficiently, e.g., Frank-Wolfe algorithm~\citep{jaggi2013revisiting} as is used in \citet{sener2018multi}. The solution $\dbold^*$ to the problem \eqref{prob:mgda.primal} can be recovered by the solution to the problem \eqref{prob:mgda.dual} $\beta^*$ as $\dbold^*=-\sum_{t=1}^{T}[\beta]_{t}^*\grad \Lmcal^{(t)}(\Wbold)$ and the model parameter is updated as $\Wbold\gets \Wbold + \eta \dbold^*$ with $\eta\geq 0$. With proper assumption, iterates or a subsequence of the iterates converge to a Pareto stationary point. If all $\{\Lmcal^{(t)}\}$ are convex, then the point that the iterates converge to is not only a Pareto stationary point but also is a weak Pareto minimizer, meaning its corresponding function value vector is on the Pareto front. {MGDA-MTL} further developed an efficient variant of MGDA when the neural network's parameters can be decoupled as $\Wbold=(\Wbold^{\text{shared}}, \Wbold^{(1)}, \ldots, \Wbold^{(T)})$, and the common descent direction only needs to be found with respect to the $\Wbold^{\text{shared}}$ part. Another work, Nash-MTL~\citep{navon2022multi}, formulates the problem of finding the common descent direction as a bargain game. Concretely, the common descent direction $\dbold$ is obtained as $\dbold=G\beta$ where $G=[\grad\Lmcal^{(1)}(\Wbold), \cdots, \grad\Lmcal^{(T)}(\Wbold)]$ and $\beta$ is a solution to the linear system\footnote{ $1/\beta$ is the element-wise reciprocal.} $G^TG\beta=1/\beta$. 

 One potential issue with 
MGDA-MTL and Nash-MTL, more generally, MGDA-type methods are the algorithms that can only produce one Pareto stationary point instead of a set of Pareto stationary points. Producing a set of solutions has the advantage of allowing practitioners to choose one solution that best fits their needs. To address this issue, {Pareto-MTL}~\citep{lin2019pareto} considers restricting the solution  $\Wbold^*$ produced by one run of MGDA in a certain domain such that $\{\Lmcal^{(t)}(\Wbold^*)\}$ is on a restricted region of the Pareto front \footnote{This is realizable only if the solution is a weak Pareto minimizer.}. By carefully crafting the regions, the algorithm can generate $K$ well-separated solutions on the Pareto front. Specifically,  assuming \textbf{$\Lmcal(\Wbold)\geq 0$ for all $\Wbold$} and that a set of $K$ preference vectors $\{\ubold_k\}_{k=1}^{K}$ are given, Pareto-MTL considered to solve the $K$ problems in parallel, where the $k$th problem is 
\begin{equation}\label{prob:moo.subregion}
    \min_{\Wbold}\Lmcal(\Wbold)=(\Lmcal^{(1)}(\Wbold), \ldots, \Lmcal^{(T)}(\Wbold))^T \quad \text{s.t.}\quad u_{i}^T\Lmcal(\Wbold)\leq u_{k}^T\Lmcal(\Wbold) \text{ for all } i\in[K]\setminus\{k\}.
\end{equation}
where $[K]=\{1,\ldots, K\}$. Intuitively, the constraints in Eq. \eqref{prob:moo.subregion} force the solution $\Lmcal(\Wbold)$ to stay close to $u_k$ in the angular space. The problem \eqref{prob:moo.subregion} is more challenging than problem \eqref{prob:moo} since it has $K-1$ nonlinear inequality constraints. Consequently, problem \eqref{prob:mgda.primal} is changed to account for these additional $K-1$ constraints. For more details, we refer readers to Eq. (14) in \citet{lin2019pareto}. 
However, as pointed out in exact Pareto Optimal Search (EPO search)~\citep{mahapatra2020multi}, Pareto-MTL does not guarantee that the solution matches the exact preference, and $K$ needs to grow exponentially fast as $T$ increases. Therefore {EPO search} re-designs the constraints and develops a new algorithm to search for the exact solution that matches the preference. Formally, {EPO search} proposes to solve
\begin{equation}\label{prob:moo.epo}
    \min_{\Wbold}\Lmcal(\Wbold)=(\Lmcal^{(1)}(\Wbold), \ldots, \Lmcal^{(T)}(\Wbold))^T \quad \text{ s.t. } \Lmcal^{(1)}(\Wbold) [u]_1 = \cdots = \Lmcal^{(T)}(\Wbold) [u]_T,
\end{equation}
where $u\in\R{T}$ is the user-specified preference vector, $[\cdot]_i$ takes the $i$th elements, and $\Lmcal^{(t)}(\Wbold)$ is non-negative for all $t\in[T]$. Geometrically, this constraint enforces the solution $\Wbold^*$ in a way such that the ray $(1/u_1, \cdots, 1/u_T)$ intersects with the Pareto front at $\Lmcal(\Wbold^*)$. Given an iterate $\Wbold$, {EPO search}  forms a search direction that tries to balance the constraint violation (the new iterate can ``better" satisfy the constraint) and decrease all objective functions. Formally, the paper borrows the uniformity to measure the constraint violation by defining the non-uniformity measure
$\mu(\Wbold) = \sum_{t=1}^{T}\hat\Lmcal^{(t)}(\Wbold)\log\left(\frac{\hat\Lmcal^{(t)}(\Wbold)}{1/T}\right)=\textbf{KL}(\hat \Lmcal(\Wbold) | \frac{\textbf{1}}{T})$ with $\hat \Lmcal^{(t)}(\Wbold)=\frac{[u]_t\Lmcal^{(t)}(\Wbold)}{\sum_{t'=1}^{m}[u]_{t'}\Lmcal^{(t')}(\Wbold)}$. One can easily check that $\mu(\Wbold)=0$ if and only if $\Wbold$ satisfies the constraints. {EPO search} shows that taking a step along the direction $\dbold_1=\sum_{t=1}^{T}\grad \Lmcal^{(t)}(\Wbold)[u_k]\left(\log(\hat\Lmcal(\Wbold)/(1/m)) - \mu(\Wbold)\right)$ can reduce the non-uniformity (constraint violation). Meanwhile, the common descent direction $\dbold_2$ that reduces the all objective functions, if there exists any, takes form of $G\beta$, where $G=[\grad \Lmcal^{(1)}(\Wbold), \cdots, \Lmcal^{(T)}(\Wbold)]$ and $\beta_t\geq0$ for all $t\in[T]$ and $\mathbf{1}^T\beta=1$. Then  {EPO search} designs a linear programming problem to find a search direction $\dbold$ that balances reducing constraint violation and reducing the loss functions guided by $(\dbold_1, \dbold_2)$. For more details, please refer to \citet[Equation 24]{mahapatra2020multi}. Built upon {EPO search}, {PHN} (Pareto hyperNetworks)~\citep{navon2021learning} proposes to use hypernetwork, which takes the preference vector $\ubold$ as the input and outputs the neural network weights for the multi-tasking, to attempt to learn the whole Pareto-front. Although the training is more challenging, if the hypernetwork could be properly trained, then at the inference time, the user can supply any preference vector $\ubold$, and the hypernetwork can output a Pareto stationary solution that closely aligns with the preference vector $\ubold$ without requiring any additional efforts.

All aforementioned algorithms, despite their actual implementation, assume access to true gradients $\{\grad \Lmcal{(t)}(\Wbold)\}_{t=1}^{T}$. This assumption might fail when in deep neural network settings. {MoCo} (multi-objective gradient correction )~\citep{fernando2023mitigating} is proposed 
to address this issue. It extends {MGDA} to the stochastic setting, providing convergence rates for both convex and non-convex cases. The most notable challenge with extending {MGDA} to the stochastic setting lies in the noise of stochastic gradient estimators of true gradients 
$\{\grad \Lmcal{(t)}(\Wbold)\}_{t=1}^{T}$. The standard way to address the issue is through the variance reduction technique. Unlike the seminar work of~\citet{liu2021stochastic}, which achieves the variance reduction via increasing batch sizes, {MoCo}~\citep{fernando2023mitigating} reduces the variance via the momentum-based method, which has the advantage of keeping the batch size as small as one while still guarantee the convergence (under proper assumptions). Concretely, at the $k$th iteration, instead of solving problem \eqref{prob:mgda.dual}, {MoCo} solves 
\begin{equation}\label{prob:mgda.dual.stochastic}
    \min_{\beta\in\R{T}} \frac{1}{2}\norm{\sum_{t=1}^{T}[\beta]_{t}d_k^{(t)}}^2 \quad \text{s.t.} \quad \sum_{t=1}^{T}[\beta]_{t}=1 \text{ and } [\beta]_t\geq 0 \text{ for all } t,
\end{equation}
where $d_{k+1}^{(t)}=\textbf{Proj}_{L_t}[d_{k}^{(t)}-\zeta_t(d_k^{(t)} - \grad \tilde\Lmcal^{(t)}(\Wbold_k))]$, where $\textbf{Proj}_{L_t}$ projects vector to a ball centered at origin with radius $L_t$, $L_t$ is the Lipschtiz constant of $\Lmcal^{t}$, $\zeta_t$ is some positive constant, and $\grad \tilde\Lmcal^{(t)}(\Wbold_k)$ is some approximation of $\grad \Lmcal^{(t)}(\Wbold_k)$.  One can show that $\norm{d_k^{(t)} - \grad \Lmcal^{(t)}(\Wbold_k)}\to 0$ as $k\to\infty$, hence achieving the variance reduction.

To conclude this section, a comprehensive list, to our best knowledge, to include all existing optimization methods in \textbf{\S}~\ref{subsec:scalar} \& \textbf{\S}~\ref{subsec:multi-obj-opt}, is summarized in Table \ref{tab:optimization}.
\input{table_files/optimization}

\begin{applebox}{Remarks}
    \begin{enumerate}[leftmargin=0.4cm, label=(\roman*)]
    \item The MOO approach, though it generally requires extra efforts to find the common descent directions, provides a solid framework to conduct convergence analysis.
    \item The MOO approach helps explore more diversified solutions over the Pareto front, whereas the scalarization approach cannot find the solutions on the concave part of the Pareto front. Obtaining diversified solutions helps users to understand trade-offs among a set of objectives.
    \item MOO approach gives the flexibility to incorporate user preferences in the solutions and does not require laborious tuning on task weights.
    \end{enumerate}
\end{applebox}

\subsubsection{Adversarial training} 
\label{sec:adversarial}
In the era of DL, joint task modeling has shown promising success by employing feature propagation or task balancing. However, it is important to acknowledge that task-specific features do not consistently result in mutual benefits, and learning multiple loosely connected tasks simultaneously introduces irrelevant noise. While task balancing helps alleviate the negative impact of transfer learning, it neglects the information exchange between tasks, often leading to suboptimal solutions. To address this issue, adversarial training \citep{adhikarla2022memory}, as an optimization approach, can effectively disentangle the space between task-shared and -specific features by inherently preventing feature interference. This approach involves introducing a task discriminator, which distinguishes features or gradients learned from different tasks. The discriminator is trained along with a shared feature extractor to converge to a saddle point where the discriminator is unable to differentiate features or gradients learned from different tasks. Research in this field can be categorized into two main approaches based on the type of information utilized for adversarial training: representation-based and gradient-based. ASP-MTL (aka AdvMTL)~\citep{liu2017adversarial} first proposes an adversarial MTL framework to learn task-shared and -specific features independently and introduces adversarial training to make shared features invariant to the involved tasks. MTA$_{(\text{adv})}$N~\citep{liu2018multi} presents an adversarial MTL framework in the image generation tasks, where multiple existing factors for image generation are considered as tasks and disentangled in an adversarial way with the training of shared encoder. RD4MTL~\citep{meng2019representation} employs adversarial training to encourage the features from different tasks to be disentangled and the features of irrelevant tasks to be minimally informative. GREAT4MTL~\citep{sinha2018gradient} and AAMTRL~\citep{mao2020adaptive} utilize the gradients derived from different tasks and disentangle the space using gradient reversal procedure~\citep{ganin2016domain}.

\underline{Representation-Based.} Adversarial Shared-Private Multi-Task Learning (ASP-MTL, aka AdvMTL)~\citep{liu2017adversarial} first proposes an adversarial MTL framework to alleviate the interference of shared and specific feature spaces among involved tasks. The underlying observation is the fact that the same word in a sentence may indicate different sentiments in different tasks, e.g. the "\textit{infantile}" in product reviews "\textit{The infantile cart is simple and easy to use.}" and product review "\textit{This kind of humor is infantile and boring.}". "\textit{infantile}" is a potential backdoor word encoded in the shared feature space as it expresses a neutral attitude in the product review while it conveys a negative attitude in the movie review. ASP-MTL addresses this issue by dividing the feature space into shared and specific (private) space in a parallel manner, as shown in {Fig.}~\ref{asp-mtl}, and disentangles them using orthogonality constraints and adversarial losses. Let $\Xmcal^{(t)}_s$ and $\Xmcal^{(t)}_p$ denote the representations of shared and private layers for the $t$-th task, respectively. The adversarial training process alternates between the shared feature generator $G$ (parametrized by $\Wbold_s$) and the task discriminator $D$ (parametrized by $\Wbold_d$) through a minimax optimization:
\begin{equation}
    \label{classic_advmtl}
    \Lmcal_{adv} = \min\nolimits_{\Wbold_s}\max\nolimits_{\Wbold_d} - \Lmcal_{CE}[D_{\Wbold_d}(\Xmcal^{(t)}_s), \tbold^{(t)}], t = 1, \cdots, T,
\end{equation}
where $\tbold^{(t)}$ denotes the ground-truth label to indicate the task type, and $\Lmcal_{CE}$ means the use of Cross Entropy loss in practice. To further extract task invariant features from the shared layers, ASP-MTL introduces the orthogonality constraint as follows to disentangle the shared and private feature space.
\begin{equation}
    \Lmcal_{orth}^{(t)} = \|\text{vec}(\Xmcal^{(t)}_s)^\top\text{vec}(\Xmcal^{(t)}_p)\|_F^2, t = 1, \cdots, T,
\end{equation}
where we abuse the vectorization vec$(\cdot)$ to preserve the sample dimension of the output feature tensors. The final learning objective function consists of three components as below:
\begin{equation}
    \Lmcal_{total} = \sum\nolimits_{t=1}^T(\Lmcal_{spec}^{(t)} + \lambda\Lmcal_{adv}^{(t)} + \gamma\Lmcal_{orth}^{(t)}),
\end{equation}
where $\Lmcal_{spec}^{(t)}$ is the task-specific objective for the $t$-th task, and $\lambda$ and $\gamma$ are hyper-parameters to balance the learning terms. This total objective is trained with backpropagation via the advantage of gradient reversal layer (GRL)~\citep{ganin2015unsupervised}.

Multi-Task Adversarial Network (MTA$_{(\text{adv})}$N)~\citep{liu2018multi} targets the problem of multiple factors existing in image generation. The architecture of MTA$_{(\text{adv})}$N is shown in {Fig.~\ref{mtadvn}}, where the shared encoder $E$ extracts the features that are disentangled across style factors for the use of content classification (discriminator $D_C$) and generation (generator $G$). Let the original image and the corresponding content label be represented by $\Xmcal$ and $\ybold$, respectively. Then the training of the generation task entails the updation of shared feature extractor $E$ and generator $G$:
\begin{equation}
    \Lmcal_G = \min\nolimits_{E,G}\sum\nolimits_{(i, j): \ybold_j = \ybold_i, \zbold_j = \zbold_i^\prime} \|G(E(\Xmcal_i), \zbold_i^\prime) - \Xmcal_j\|^2_2,
    \label{mtan_generator_loss}
\end{equation}
where $i \text{and} j$ are data indices. $\zbold^\prime$ is sampled from the style label codebook $\Zmcal$. {Eq.}~ (\ref{mtan_generator_loss}) means that the generator $G$ tries to reconstruct the data $\Xmcal_i$ itself if $\zbold^\prime_i = \zbold_i$ and tries to minimize the distance between the style-transferred $\Xmcal_i$ and any sample $\Xmcal_j$ with the same content and style labels (i.e. $\ybold_j = \ybold_i, \zbold_j = \zbold_i^\prime$) otherwise.

The key adversarial training of style labels is defined using Earth Mover’s Distance (EMD) loss~\citep{arjovsky2017wasserstein} as follows:
\begin{equation}
    \Lmcal_S = \min\nolimits_{E}\max\nolimits_{D_S} \sum\nolimits_i-\Lmcal_{EMD}(\xbold_i, \zbold_i) - \lambda \Omega_{GP}(D_S),
\end{equation}
where $\Omega_{GP}(D_S)$ is a gradient penalty term~\citep{gulrajani2017improved} for the purpose of training stability and $\lambda$ serves as a trade-off hyper-parameter.
To add the classification of content factor, the total training objective is formulated as follows: 
\begin{equation}
    \Lmcal_{total} = \min\nolimits_{E,G} \Lmcal_G + \alpha \min\nolimits_{E}\max\nolimits_{D_S} \Lmcal_S + \beta \min\nolimits_{E,D_C} \Lmcal_C,
\end{equation}
where $\Lmcal_C$ denotes the Cross-Entropy loss of the content classification task, $\alpha$ and $\beta$ both are the hyper-parameters.

Representation Disentanglement for Multi-Task Learning (RD4MTL)~\citep{meng2019representation} aims to disentangle the indiscriminate mixing of properties in medical image analysis. As depicted in {Fig.}~\ref{rd4mtl}, an adversarial training process encourages the features from different tasks to be disentangled and minimally informative. Let $\Zbold^{(t)}$ represent the latent features extracted by the specific encoder $E_{\theta^{(t)}}$ from the original image $\Xmcal$, then $\Lmcal_{cls}^{(t)}$ as the $t$-th task-specific classification loss can be calculated as follows:
\begin{equation}
    \Zbold^{(t)} = E_{\theta^{(t)}}(\Xmcal), \Lmcal_{cls}^{(t)} = \Lmcal_{CE}(D_{\phi^{(t)}}(\Zbold^{(t)}), \ybold^{(t)}), t = 1, \cdots, T,
\end{equation}
where $\ybold^{(t)}$ is the ground truth label of the $t$-th task, and $\Lmcal_{CE}$ is the Cross Entropy loss in practice. Furthermore, the adversarial regularization uses a minimax competition process as below:
\begin{equation}
    \Lmcal_{adv}^{(t)} = \min\nolimits_{\{\theta^{(s)}, \phi^{(s)}\}_{s\neq t}^T}\max\nolimits_{\psi^{(t)}} \sum\nolimits_{s\neq t}^{T} - \Lmcal_{CE}(D_{\psi^{(t)}}(\Zbold^{(s)}), \ybold^{(s)}), t = 1, \cdots, T,
\end{equation}
then the total training objective can be formulated as follows:
\begin{equation}
    \Lmcal_{total} = \sum\nolimits_{t=1}^T (\Lmcal_{cls}^{(t)} + \lambda\Lmcal_{adv}^{(t)}),
\end{equation}
where $\lambda$ balances the two loss terms.

Adaptive Adversarial Multi-Task Representation Learning (AAMTRL)~\citep{mao2020adaptive} investigates the theoretical mechanism of adversarial MTL via using Lagrangian duality, and further proposes the AAMTRL that can improve the performance of classical adversarial MTL (aka AMTRL methods in~\citep{mao2020adaptive}). For simplicity, if the shared and -private features for the $t$-th are represented by $\Xmcal_s^{(t)}$ and $\Xmcal_p^{(t)}$, aligning with the formalization in {Eq.}~(\ref{classic_advmtl}). Assume the shared feature extractor $E$ (parametrized by $\Wbold_s$) and task discriminator $D$ (parametrized by ${\Wbold_d}$) to be Bayes-optimal, AAMTRL introduces the matrix $\Rbold$ to measure the task relatedness, where
\begin{equation}
    r_{i,j} = \frac{D_j(\Xmcal_s^{(i)}) + D_i(\Xmcal_s^{(j)})}{D_i(\Xmcal_s^{(i)}) + D_j(\Xmcal_s^{(j)})},
\end{equation}
where $r_{i,j}$ is the $(i,j)$-th entry of the matrix $\Rbold$, and $D_i$ represents the probability that the discriminator $D$ classify the input representations as $i$-th task type. In AAMTRL, the adaptation is realized by the weighting strategy of task-specific objectives $\{\Lmcal^{(t)}_{spec}\}_{t=1}^T$:
\begin{equation}
    \Lmcal_{spec} = \sum\nolimits_{t=1}^T \alpha_t\Lmcal^{(t)}_{spec}, \alpha_t = \boldsymbol{1}\Rbold/(\boldsymbol{1}\Rbold\boldsymbol{1}^\top).
\end{equation}

The classic adversarial MTRL problem can be regard as the Lagrangian dual function of the following equality-constrained optimization problem:
\begin{equation}
    \min\nolimits_{\{\Wbold_s, \Wbold_d\}} \Lmcal_{spec},\quad\quad\text{s.t.}\quad\Lmcal_{adv} = 0.
\end{equation}
To avoid the sub-optimal solution of the traditional Lagrangian duality in solving the problem above, an augmented Lagrangian with a quadratic form is proposed as follows:
\begin{equation}
    \min\nolimits_{\{\Wbold_s, \Wbold_d\}} \Lmcal_{spec} + \lambda\Lmcal_{adv} + r/2 \Lmcal^2_{adv},
\end{equation}
where $\lambda$ is the Lagrangian multiplier, and $ r$ is the penalty hyper-parameter that can balance the duality gap. By using Lagrangian duality, AAMTRL can have an exact generalization error bound that is minimally investigated in the classic AMTRL.

\underline{\emph{Gradient-Based.}}
GRadiEnt Adversarial Training for MTL (GREAT4MTL)~\citep{sinha2018gradient} is one of the scenarios of GRadiEnt Adversarial Training (GREAT) that tries to make the gradients indistinguishable across involved tasks. As depicted in {Fig.}~\ref{great4mtl}, the encoder $E_{\theta}$ extracts shared features for multiple tasks, and the decoders $\{D_{\phi^{(t)}}\}_{t=1}^T$ are used to perform $T$ involved tasks. Thus, the basic learning objectives for specific tasks are:
\begin{equation}
    \Lmcal_{spec}^{(t)} = \min\nolimits_{\theta, \{\phi^{(t)}\}_{t=1}^T} \Lmcal^{(t)}(D_{\phi^{(t)}}(E_{\theta}(\Xmcal^{(t)})), \ybold^{(t)}), t = 1, \cdots, T,
\end{equation}
where $\{(\Xmcal^{(t)}, \ybold^{(t)})\}_{t=1}^T$ is the total dataset containing $T$ tasks, and $\Lmcal^{(t)}$ is dependent on the task type. In GREAT4MTL, the Gradient-Alignment Layer (GAL) $G_{\psi}$ is placed after the shared encoder and before the task-specific decoders to perform task discrimination. Unlike representation-based methods that attend to the features, $G_{\psi}$ is trained using gradients from different tasks as inputs:
\begin{equation}
    \Lmcal_{adv} = \min\nolimits_{\{\phi^{(t)}\}_{t=1}^{T}}\max\nolimits_{\psi} \sum\nolimits_{t=1}^T - \Lmcal_{CE}(G_{\psi}(\grad_{E_{\theta}(\Xmcal^{(t)})} D_{\phi^{(t)}} \Lmcal_{spec}^{(t)}, \ybold^{(t)}), \tbold^{(t)}),
\end{equation}
where $\tbold$ is the ground truth label to indicate the task type, and the Cross-Entropy loss $\Lmcal_{CE}$ is used to calculate the task classification error. Then the total training objective function is:
\begin{equation}
\label{great_total_loss}
    \Lmcal_{total} = \sum\nolimits_{t=1}^T \Lmcal_{spec}^{(t)} + \Lmcal_{adv}.
\end{equation}
The GRL is inserted before the GAL to streamline the minimax optimization process above. The trade-off hyper-parameter is eliminated in {Eq.}~(\ref{great_total_loss}) by using different learning rates during the training process of $\Lmcal_{spec}$ and $\Lmcal_{adv}$.

ASTMT~\citep{maninis2019attentive} also employs the GREAT strategy to effectively disentangle the task-shared and task-specific features acquired from the shared backbone and single-tasking components, as illustrated in the right portion of {Fig.}~\ref{astmt}. It highlights the compatibility of GREAT to be seamlessly integrated with other frameworks.

\begin{applebox}{Remarks}
\begin{enumerate}[leftmargin=0.4cm, label=(\roman*)]
    \item Adversarial training effectively disentangles the feature space into shared and task-specific components, ensuring that shared features remain indistinguishable across multiple tasks, while specific features retain their distinctiveness.
    \item Adversarial training can sometimes lead to unstable training dynamics, especially if the adversarial and task-specific components are not well-balanced. This can manifest as oscillations in learning or difficulty in achieving convergence.
\end{enumerate}
\end{applebox}


\subsubsection{Mixture of Experts (MoE)} 
\label{sec:moe}
Deep neural-based architectures have been extensively utilized in real-world MTL problems. However, the challenge of scaling high-capacity deep neural networks to adapt to multi-task settings remains conceptually appealing. The MoE~\citep{jacobs1991adaptive} framework inherently incorporates multiple expert networks, each of which can be selected for learning different tasks. The modern MoE layer~\citep{eigen2013learning, shazeer2017} has transformed the MoE module into a universally adaptable component that seamlessly integrates into various systems, including CNNs, RNNs, and Transformers, enabling plug-and-play functionality. The MoE layer, as depicted in {Fig.}~\ref{moe}, generally comprises a set of $N$ expert networks $\{E_n\}_{n=1}^N$ and a gating network $G$, whose output depends on the input data $\Xmcal$. This gating network generates a sparse $N$-dimensional vector that selects the necessary expert networks to compute the final prediction as follows:
\begin{equation}
    \tilde{\ybold} = \sum\nolimits_{n=1}^N G(\Xmcal)_n E_n(\Xmcal),
\end{equation}
where $G(\Xmcal)_n\in\{0, 1\}$ is the $n$-th entry of the sparse vector generated by the gating network $G$, and $\tilde{\ybold}$ represents the.
Beyond MoE for STL, Multi-gate Mixture-of-Experts (MMoE)~\citep{ma2018modeling} explicitly introduces multiple gates/routers ($\{G_t\}_{t=1}^T$) for each task, as shown in Fig.~\ref{multi-moe}. The final prediction for the $t$-th task is calculated as
\begin{equation}
    \label{eq:MMoE}
    \ybold^{(t)} = \sum\nolimits_{n=1}^N G_t(\Xmcal^{(t)})_n E_n(\Xmcal^{(t)}), t = 1, \cdots, T,
\end{equation}
where $(\Xmcal^{(t)}, \ybold^{(t)})$ represents the sampled data from $t$-th task.
This prior research has inspired the development and utilization of multi-router MoE for MTL. It includes DSelect-k that selects top $k$ experts for each task, MT-Tag~\citep{gupta2022sparsely}, demonstrating the robustness of Multi-Router MoE to the loosely related tasks, CmoIE~\citep{wang2022multi}, which constructs more insightful experts instead of incompetent ones, Mod-Squad~\citep{chen2023mod}, specializing experts for specific tasks by measuring the mutual information (MI) between tasks and experts, and SummaReranker~\citep{ravaut2022summareranker}, performing re-ranking on a set of summary candidates to select the best one. On the other hand, task-conditioned routing with a shared router/gate is another variant where task-dependent representations are fed into the only existing router, making their expert selections, as depicted in Fig.~\ref{single-moe} for comparison. The shared-router MoE is discussed separately from the Multi-router MoE in M$^3$ViT\citep{fan2022m3vit}. Task-level MoE~\citep{ye2022eliciting} designs different router architectures with varying complexities under shared-router settings, including MLP, LSTM, and Transformer. In both ways, task relationships are captured in different mixture patterns of experts assembling.

\input{tex_files/02-2/MoE_framework}
\underline{\emph{Multi-Router MoE.}}
Multi-gate Mixture of Experts (MMoE)~\citep{ma2018modeling} replaces the shared layers in the hard parameter architecture with multiple MoE layers and retains individual routers for each task, resembling the soft parameter architecture. The computational process of predicting $t$-th task is shown in {Eq.}~(\ref{eq:MMoE}). The router networks of MMoE is the softmax of the linear transformations of the input data representation:

\begin{minipage}{0.6\textwidth}
\begin{align}
    G_t(\Xbold^{(t)}) = \text{softmax}(\Wbold_t\Xbold^{(t)}), t = 1, \cdots, T,
\end{align}
\end{minipage}

where $\Wbold_t\in\Rmbb^{N\times D}$, and $N, D$ is the number of experts and the number of features. In comparison to the soft parameter sharing architecture, MMoE features routers solely for each task, resulting in a lighter size and enhanced scalability with an increasing number of tasks. In addition, the conditional computation~\citep{bengio2013estimating, shazeer2017} of the MoE layer requires the activation of only specific parts of the experts on a per-example basis. While \citet{shazeer2017} offers a top-$k$ gating function by adding tunable Gaussian noise, the theoretically scary discontinuities can lead to convergence issues if learning via gradient-based optimization. 

Differentiable Selection of top-$k$ experts(DSelect-$k$)~\citep{hazimeh2021dselect} bridges this gap by proposing a continuously differentiable and sparse gate in the context of MMoE. Obviously, the direct cardinality constraint ($\ell_0$ norm) on the output vector of the gate function is not amenable to SGD. To address this issue, a binary encoding scheme is introduced to realize top-$k$ selection via unconstrained minimization. Let $\Zbold\in\Rmbb^{k\times m}$ denote a matrix that selects the top-$k$ experts, whose $i$-th row $\zbold_i$ is a $m$-dimensional binary encoding of the index of any single expert, where $m = \log_2 N$ and $N$ is the number of total experts. The gate output vector $\qbold$ is defined as follows:
\begin{equation}
    \label{eq:unconstained_gate_output}
    \qbold_{\boldsymbol{\alpha}, \Zbold} = \sum\nolimits_{i=1}^k \sigma(\boldsymbol{\alpha})_i r(\zbold_i),
\end{equation}
where $\boldsymbol{\alpha}\in\Rmbb^k$ is a learnable vector to control the importance of the final selected top-$k$ experts, and $r(\zbold_i)\in\Rmbb^{N}$ defines the single expert selector that returns a one-hot encoding of the index of some selected expert. It is noticeable that $\|q(\alpha, \Zbold)\|_0 \leq k$ and $\sum\nolimits_{i=1}^N q(\alpha, \Zbold)_i = 1$, which realize the similar property for the gate output without any constraint involved. Furthermore, DSelect-k using a element-wise smoothing function $S:\Rmbb\rightarrow\Rmbb$ to relax every binary variable in $\Zbold$ to be continuous in the range $(-\infty, +\infty)$:
\begin{equation}
    \label{eq:unconstained_gate_output_smoothed}
    \tilde{\qbold}_{\boldsymbol{\alpha}, \Zbold} \approx \qbold_{\boldsymbol{\alpha}, S(\tilde{\Zbold})}, S(z) = 
    \begin{cases}
        0, & \text{if}~z \leq -\gamma/2,\\
        (-2/\gamma^3) z^3 + (3/(2\gamma)) z + 1/2, & \text{if}~-\gamma/2 \leq z \leq \gamma/2,\\
        1, & \text{if}~z \geq \gamma/2,
    \end{cases}
\end{equation}
where $\gamma$ is a hyper-parameter that controls the width of the fractional region. Eqs. (\ref{eq:unconstained_gate_output}) and (\ref{eq:unconstained_gate_output_smoothed}) transform the top-$k$ selection to be unconstrained and first-order differentiable.

Multi-Task Task-aware Gating (MT-TaG)~\citep{gupta2022sparsely} designs the task-aware sparse gating function to route expert selection for each task. The incorporation of task-conditioned information into the routing mechanism is realized by constraining each embedding to only the top-$1$ expert selection. Let $\xbold_{i}^{(t)}$ be the token/embedding representation in the $i$-th position of the input sequence for the $t$-th task. A linear mapping process is first applied to obtain the touting logits below:
\begin{equation}
    \tilde{\xbold}_{i}^{(t)} = \Wbold^{(t)} \xbold_{i}^{(t)}, t = 1, \cdots, T,
\end{equation}
then the only expert routing is as follows through a softmax process:
\begin{equation}
    h(\xbold_{i}^{(t)}) = \max\nolimits_j (\frac{e^{\tilde{\xbold}_{i}^{(t)}}}{\boldsymbol{1}^\top e^{\tilde{\xbold}_{i}^{(t)}}})_j \cdot E_j(\xbold_{i}^{(t)}), t = 1, \cdots, T,
\end{equation}
where $h$ denotes the task-conditioned representation calculated by the selected experts. Noticeably, the task relationship is implicitly encompassed within the variable $h$, thereby remaining independent of the experts involved. 
SummaReranker~\citep{ravaut2022summareranker} targets only the abstractive summarization task but utilizes different metrics to measure it. The re-ranking on a set of summary candidates generated by MMoE can consistently promote the base model.

However, the promise of MMoE has been validated in MTL with the explicit task relationship backups. Calibrated Mixture of Insightful Experts(CMoIE)~\citep{wang2022multi} investigates the negative transfer in MMoE caused by incompetent experts in certain applications. Specifically, a conflict resolution module between each pair of experts and the expert communication among the layers of different experts are introduced to advocate the diversity and capacity of experts. Additionally, a mixture
calibration structure employed in the routing networks encourages the expert responsibilities to handle more tasks without losing their specialty. For any input data $\Xmcal$, the conflict resolution employs the Euclidean distance to measure the outputs from each pair of experts:
\begin{equation}
    \Dbold_{i,j} = \ell_2(E_i(\Xmcal), E_j(\Xmcal)), i, j = 1, \cdots, N,
\end{equation}
where $N$ is the number of total experts, and $\Dbold\in\Rmbb^{N\times N}$ denotes the distance matrix between each pair of experts. Based on the max-margin
$t$-distribution, the corresponding conflict attention matrix for each pair of experts is calculated to highlight the excessively similar expert pairs:
\begin{equation}
    \Abold_{i,j} = 1/(1 + \max(0, \Dbold_{i,j} - R_i)), i, j = 1, \cdots, N,
\end{equation}
where $R_i$ is the conflict radius of the expert $E_i$ that defines the upper quartile of $\{\Dbold_{i,j}\}_{j=1}^N$. Furthermore, the conflict loss is proposed as follows:
\begin{equation}
    \Lmcal_{conflict} = - \sum\nolimits_{i=1}^N\sum\nolimits_{j=1}^N (\Abold\odot\Dbold)_{i,j},
\end{equation}
where $\Lmcal_{conflict}$ is combined with multi-task loss in an end-to-end training process. To capture implicit task relationships by constructing task-aware representations, the fusion matrix $\Fbold\in\Rmbb^{N\times N}$ is defined using multilinear
map as follows:
\begin{equation}
    \label{eq:gate_layer_fusion}
    \Fbold_l^{(t)} = G^{(t)}(\Xmcal^{(t)})\otimes H_l(\Xmcal^{(t)}), t = 1, \cdots, T,
\end{equation}
where $G^{(t)}$ and $H_l$ are the routing networks and another hidden-layer gating network before the $l$-th layer for the experts. Let the hidden representations at the $l$-th layer of $E_n$ denote by $\zbold_l^n, n = 1, \cdots, N$, and then stack all of them by the way of $[\zbold_l^1, \cdots, \zbold_l^n]^\top$ to be the hidden representation matrix $\Zbold_l$. Through the fusion process defined in {Eq.}~(\ref{eq:gate_layer_fusion}), the input of $(l+1)$-th layer of $\{E_n\}_{n=1}^N$ is diffused by:
\begin{equation}
    \tilde{\Zbold}_{l+1}^{(t)} = \Fbold_l^{(t)}\Zbold_l^{(t)} + \Zbold_l^{(t)}, t = 1, \cdots, T,
\end{equation}
where the representation is tailored by the task-specific fusion matrix. The residual block ($+ \Zbold_l^{(t)}$) above can suppress the individuality ruin of experts during the fusion process. To further enhance the specialization and concentration of experts on specific tasks, the mixture calibration introduces a dynamic temperature $\tau^{(t)}$ to control the logits for each routing network:
\begin{equation}
    G^{(t)}(\Xmcal^{(t)}) = \text{softmax}(g^{(t)}(\Xmcal^{(t)})/\tau^{(t)}), t = 1, \cdots, T,
\end{equation}
where the temperature parameters are progressively decreased from $1$ during the training process. 

Mod-Squad~\citep{chen2023mod} also allows cooperation and specialization in the process of matching experts and tasks. To make the experts dependent on tasks, the mutual information between them is first measured as below:
\begin{equation}
    \Imcal(\Tmcal; E) = \sum\nolimits_{t=1}^T\sum\nolimits_{n=1}^N P(\Tmcal_t, E_n)\log\frac{P(\Tmcal_t, E_n)}{P(\Tmcal_t) P(E_n)},
\end{equation}
where the joint probability will be decided by the number of data that are routed inside a task to the target expert. Then the total loss can be formulated as follows:
\begin{equation}
    \Lmcal_{total} = \sum_{t=1}^T\lambda_t\Lmcal_{\Tmcal_t} - \gamma\sum\nolimits_{\forall \text{~MoE~layers~} l} I(T; E_l),
\end{equation}
where $\lambda_t$ is the hyper parameter to control the $t$-th task-specific loss $\Tmcal_t$, and $\gamma$ balances the multi-task loss term and mutual information term.

\underline{Shared-Router (Task-Conditioned) MoE.} Task-Level MoE~\citep{ye2022eliciting} first uses a shared router that takes the task representation as input, which is selected from a look-up embedding table. Moreover, Task-Level MoE first investigates the combinations of different backbone (MLP, LSTM, and Transformer) and softmax (softmax,  Gumbel-Softmax, and ST Gumbel-Softmax~\citep{jang2016categorical}) variations of routers. M$^3$ViT~\citep{fan2022m3vit} customizes MoE into a ViT backbone, which compares the multi-router MoE and shared-router MoE. ViT-based MMoE can feature hardware memory efficiency, as certified in Edge-MoE~\citep{sarkar2023edge}.

To circumvent the limitations associated with a fixed single expert, the AdaMV-MoE~\citep{chen2023adamv}, denoted as the Adaptive Mixture of Experts framework for Multi-task Vision Recognition, possesses the capacity to autonomously ascertain the number of sparsely activated MoE based on input token embeddings. Task-specific router networks are employed to select the most relevant experts for individual tasks. This process can be mathematically expressed as:
\begin{equation}
    G_t(\xbold^{(t)}) = \sum_{n=1}^{N_t}\Rmcal_t(\xbold)\cdot E_n(\xbold), t = 1, \cdots, T,
\end{equation}
where $\Rmcal_t$ is the router for $t$-th task. It should be noted that the number of experts ($N_t$) engaged is not predefined. AdaMV-MoE incorporates an adaptive mechanism, specifically the Adaptive Expert Selection (AES) technique, to dynamically adjust this quantity based on task-specific loss values observed during validation on datasets ($\Lmcal_{\text{val}}^{(t)}$). If $\Lmcal_{\text{val}}^{(t)}$ exhibits no signs of decline over several iterations, the number of experts ($N_t$) should be augmented by 1. In contrast, if it exceeds the best loss value above, the number of experts should be reduced. Ultimately, after numerous iterations, the number of experts can be stabilized.

\begin{applebox}{Remarks}
\begin{enumerate}[leftmargin=0.4cm, label=(\roman*)]
    \item MoE seamlessly accommodates tasks with different backbones, making it well-suited for scenarios with diverse task requirements and complexities.
    \item MoE efficiently allocates resources by assigning different experts to different tasks, avoiding redundancy and optimizing computational resources.
    \item MoE exhibits remarkable scalability, rendering it highly suitable for large-scale industry applications on the ground.
\end{enumerate}
\end{applebox}

\subsubsection{Graph based}
\label{sec:graph}
Graphs have been widely used in data mining and machine learning due to their unique representation of objects and their interactions. Graph neural networks~(GNNs)~\citep{sperduti1997supervised, gori2005new, scarselli2008graph, wu2020comprehensive}, which leverage nodes and edges among their connected nodes in graphs to conduct inference, have gained applause with impressing performance in capturing the inter-nodes relations on graphs. It is natural to consider the tasks and corresponding data samples in MTL as nodes and their relations as the edges to construct a graph for MTL~\citep{alon2017graph}. 
Via conducting graph mining on such graphs, relations among tasks or data samples in MTL can be better understood so as to assist the final MTL model in conducting inference~\citep{chen2019multi,cao2022relational,liu2020asymmetric,liu2022structured}



\input{tex_files/02-2/multikernel}


MultiKernel~\citep{widmer2010leveraging} conducts MTL over a series of classification tasks with predefined hierarchical relations, which is often the case for biological problems. Notably, it constructs a tree that reflects the hierarchical relations between tasks and domains, where leaf nodes are the tasks it studies (\eg, dog), whose parent and ancestors (non-leaf nodes) are the corresponding biological domains (\eg, mammals and animals). 

For a queried task $\boldsymbol{x}$, MultiKernel classifies it over every task $t$'s predictor $f_t$ by 
\begin{equation}
    f_t(\boldsymbol{x})=(\boldsymbol{u}_t+\sum\nolimits_{r\in\{t\text{'s ancestors}\}}\lambda_{t,r}\boldsymbol{u}_r)^{\top}\boldsymbol{x}+b_t,
\end{equation}
where $\lambda_{t,r}$ is a pre-calculated constant inversely related to the distance between task $t$ and its ancestors $r$. $\boldsymbol{u}_t$ is the representation of task $t$. The representations of nodes within the predefined tree are learned by minimizing the task error. $b_t$ is a learnable variable.

ML-GCN~\citep{chen2019multi} is a graph convolutional network (GCN)-based MTL model for capturing the label correlations in multi-label image recognition. Specifically, different from traditional MTL, ML-GCN pre-constructs a correlation matrix that reflects labels' co-occurrence patterns within datasets. This matrix enables the system to build a label graph, where each node represents a label, and whose feature is the corresponding word embedding. 


On retrieving the label graph, ML-GCN jointly trains a CNN and a GCN for the MTL. The CNN learns from image datasets to retrieve image representations, and the GCN learns from the label graph 
to generate label representations. ML-GCN retrieves multi-label prediction $\hat{\boldsymbol{y}}$ for an input image $\boldsymbol{x}$ by computing dot products between image representations and label representations as $\hat{\boldsymbol{y}}=\boldsymbol{W}\cdot f(\boldsymbol{x};\theta)$,
where $f(\cdot)$ and $\theta$ are the CNN model and its parameters respectively. $\boldsymbol{W}=\{\boldsymbol{w}^{(i)}\}_{i=0}^{C}$ is the set of label representations output the GCN.

ML-GCN resorts to the traditional multi-label classification loss for training. 
The entire construction of ML-GCN is shown in {Fig.}~\ref{ml-gcn}.

MetaLink~\citep{cao2022relational} assumes that, for a given data point, at the inference time, the multi-task model has access to its labels from auxiliary tasks. Based on this assumption, MetaLink leverages labels from other tasks to improve the predictive performance. Particularly, MetaLink constructs a knowledge graph to capture not only the task-task relations as in ML-GCN but also the inter- and intra-relations between tasks and data.

The knowledge graph consists of two types of nodes: (1) data nodes, whose features are embeddings computed by the neural networks, and (2) task nodes, whose features are the last layer weights of the corresponding task-specific neural networks. Whenever a data sample belongs to a task, an edge is connected between these two nodes, and the label of the edge describes how the data point is classified in the particular task. In this way, MetaLink transfers the traditional MTL to a link prediction task between data nodes and task nodes, as shown in {Fig.}~\ref{metalink}.

In terms of updating the entire model, MetaLink does not specify the criterion or introduce any particular regularizing terms. 

\begin{applebox}{Remarks}
\begin{enumerate}[leftmargin=0.4cm, label=(\roman*)]
    \item Graph-based representations allow tasks to be modeled as nodes and their relationships as edges. This enables the capturing of intricate dependencies and relationships among tasks, providing a more nuanced understanding compared to simplistic task relationship learning.
    \item GCN exhibits scalability in handling MTL scenarios with large number of tasks.
    \item GCN excels in information propagation across tasks within a graph structure. The interconnected nature of tasks in a graph allows for the sharing of relevant information, fostering collaborative learning.
\end{enumerate}
\end{applebox}


\subsubsection{Neural Architecture Search (NAS)}
\label{sec:nas}
NAS is a popular method in designing deep neural networks automatically, which has the potential to revolutionize the way neural networks are designed and used in many different fields, including MTL. NAS in MTL refers to the use of NAS to design neural networks that can perform multiple tasks simultaneously. This is different from traditional neural network design, where a separate network is typically trained for each task. In MTL, the goal is to learn a shared representation that can be used to perform multiple tasks effectively. Conventional architecture realizes multi-tasking by hard-parameter sharing that trains multiple task heads that share shallow feature extractors, e.g., TCDCN~\citep{zhang2014facial} and Fast RCNN~\citep{girshick2014rich, girshick2015fast}, or by training separate neural network to perform all each task with the shared trunk, e.g., Cross-Stitch Networks~\citep{misra2016cross} and NDDR-CNN~\citep{gao2019nddr}. However, the potential design space for deep multi-task neural architectures grows exponentially with the depth, and incorporating more tasks significantly expands the range of optimal solutions. 

NAS can be used as an automatic approach to search for the optimal architecture for an MTL system. This involves defining a search space that includes a range of possible architectures and using a search algorithm to explore this space and identify the best-performing architecture. The search algorithm can be based on techniques such as reinforcement learning, evolutionary algorithms, or gradient-based optimization. There are several benefits to using NAS in multi-task learning. For example, it can reduce the need for manual design of the network architecture, improve the performance of the multi-task system, and reduce the amount of data and computation required to train the network. It can also be used to identify architectures that are more efficient and easier to implement in practice.

Fully-Adaptive Feature Sharing (FAFS)~\citep{lu2017fully} is the earliest method that trains networks with an adaptive widening process. The initial network is a slimmed-down version from reducing the number of convolutional filters in CNN or neurons in MLP. It gradually expands through a multi-round widening and training procedure, facilitated by a top-down splitting algorithm. In practice, the original active layer, depicted as the $L$-th layer in Fig.~\ref{fafs}, consists of numerous branches. These branches are then grouped together in the lower $(L-1)$-th layer. Subsequently, the $(L-1)$-th layer becomes the new active layer, and this iterative process continues from the top layers until the convergence. 


Branched Multi-Task Networks (BMTN)~\citep{DBLP:conf/bmvc/VandenhendeGGB20} argues that learning layer sharing level in the early soft parameter sharing methods suffer from sub-optimal solutions, and relying solely on NAS to design the MTL architecture is significantly cumbersome. By leveraging the affinities of involved multiple tasks using Representation Similarity Analysis (RSA)~\citep{dwivedi2019representation}, BMTN can automatically cluster the tasks at shared locations, in which bottom layers are task-agnostic and top layers gradually grow to be task-specific. For each task, as depicted in Fig.~\ref{bmtn}, BMTN initially computes the representation dissimilarity matrices (RDMs) between $K$ images at $D$ locations. The RDMs are defined as $1-\rho$, where $\rho$ represents the Pearson correlation coefficient~\citep{pearson1895vii}. Subsequently, the task affinity tensor $\mathcal{A}\in\mathbb{R}^{D\times T\times T}$ is established based on the RDMs of all tasks using the Spearman's correlation coefficient~\citep{spearman1961proof}. Finally, BMTN is established by minimizing the sum of these task dissimilarity scores (i.e. $1 - \mathcal{A}_{d,i,j}$) between each pair of tasks $i$ and $j$ at every location $d$, $i,j=1, \cdots, T, d = 1, \cdots, D$.



Multi-Task Learning by Neural Architecture Search (MTL-NAS)~\citep{gao2020mtl} is a method to search cross-task edges into fixed single-task network backbones. The framework is shown in {Fig.}~\ref{nddr-nas-module}.
It involves a single-shot gradient-based search algorithm that can optimize the architecture weights overall legal connections defined by the search space. Specifically, this search algorithm contains the continuous relaxation and the discretization procedures.
This novel search algorithm is able to close the performance gap between search and evaluation and also generalizes the popular single-shot gradient-based methods such as DARTS~\citep{liu2018darts} and SNAS~\citep{xie2018snas}.

\begin{applebox}{Remarks}
\begin{enumerate}[leftmargin=0.4cm, label=(\roman*)]
    \item NAS facilitates the automatic and adaptive discovery of task-specific neural network architectures, departing from conventional hard or soft parameter sharing stereotypes.
    \item NAS not only searches for architectures but can also optimize hyperparameters during the search process. This automatic tuning ensures that the MTL model is configured with optimal settings for each task, reducing the need for manual fine-tuning.
    \item NAS can discover architectures that capture these dependencies effectively, allowing tasks to share information efficiently. This adaptability is crucial in scenarios where where tasks have a significant influence on each other.
\end{enumerate}
\end{applebox}

%% file: table_files/deep_summary.tex
\begin{table*}[t]
    \centering
    \tiny
    \caption{Summary of deep MTL models.} 
    \label{tab:deep-sum}
    \midsepremove
    \scalebox{0.5}{
    \begin{threeparttable}
    \begin{tabular}{lllllllllll}
    
    \rowcolor{gray!40}
    \toprule
        Model Name & Origin & Year & MTL Strategy & Backbone & Sharing & Modality & Task & Measurement & Loss Function & Availability\tnote{1}\\
    \toprule 
         \multirow{3}{*}{TCDCN} & \multirow{3}{*}{ECCV} & \multirow{3}{*}{\citeyear{zhang2014facial}} & \multirow{3}{*}{Early stopping} & \multirow{3}{*}{CNN} & \multirow{3}{*}{Hard} & \multirow{3}{*}{Image} & Facial landmark detection/head pose estimation/& \multirow{1.8}{*}{Mean error (mErr)~\citep{burgos2013robust},}& \multirow{1.8}{*}{Mean squared error (MSE),} & \\
         
         & & & & & & & gender classification/age estimation/expression & \multirow{1.8}{*}{failure rate~\citep{dantone2012real}} & \multirow{1.8}{*}{cross-entropy (CE) loss}& \href{http://mmlab.ie.cuhk.edu.hk/projects/TCDCN.html}{\textcolor{citegreen}{Official}}\\
          
         & & & & & & & recognition/facial attribute inference & & & \\   
         
    \midrule 
    \rowcolor{gray!20}
         & ACL- & & & & & & & & &  \\ 
    \rowcolor{gray!20}
         \multirow{-2}{*}{MTL-ML} & IJCNLP & \multirow{-2}{*}{\citeyear{dong2015multi}} & \multirow{-2}{*}{---} & \multirow{-2}{*}{RNN} & \multirow{-2}{*}{Hard} & \multirow{-2}{*}{Text} & \multirow{-2}{*}{Multiple-target language translation} & \multirow{-2}{*}{BLEU-4~\citep{papineni2002bleu}, Delta} & \multirow{-2}{*}{CE loss} & \multirow{-2}{*}{---}\\

    \midrule 
         Vanilla & & & & & & & Part-Of-Speech (POS)/Chunking/Combinatory & & &  \\
         
         Cascading & \multirow{-2}{*}{ACL} & \multirow{-2}{*}{\citeyear{sogaard-goldberg-2016-deep}} & \multirow{-2}{*}{Cascading} & \multirow{-2}{*}{LSTM} & \multirow{-2}{*}{Hard} & \multirow{-2}{*}{Text} & Categorical Grammar (CCG) Supertagging & \multirow{-2}{*}{F1 score, Micro-F1 score} & \multirow{-2}{*}{CE loss} & \multirow{-2}{*}{---}\\

    \rowcolor{gray!20}
    \midrule 
         &  &  & & & & & Surface normals estimation (normals)/semantic & mErr/median error (medErr)/within $t^{\circ}$ in angular & & \\
    \rowcolor{gray!20}
        \multirow{-1.8}{*}{Cross-stitch} & & & & & & & segmentation (semseg), object detection/attribute & distance (within $t^{\circ}$), pixel accuracy (pixacc), & & \\
    \rowcolor{gray!20}
        \multirow{-1.8}{*}{networks} & \multirow{-3}{*}{CVPR} & \multirow{-3}{*}{\citeyear{misra2016cross}} & \multirow{-3}{*}{---} & \multirow{-3}{*}{CNN} & \multirow{-3}{*}{Soft} & \multirow{-3}{*}{Image} & prediction & mIoU, fwIU, mAP & \multirow{-3}{*}{CE loss} & \multirow{-3}{*}{\href{https://github.com/helloyide/Cross-stitch-Networks-for-Multi-task-Learning}{\textcolor{reforange}{Unofficial}}} \\

    \midrule 
        ASP-MTL (aka & & & & & Hard \& & & & & CE loss, adversarial loss, & \\
       AdvMTL) & \multirow{-2}{*}{arXiv} & \multirow{-2}{*}{\citeyear{liu2017adversarial}} & \multirow{-2}{*}{Adversarial training} & \multirow{-2}{*}{LSTM} & Soft & \multirow{-2}{*}{Text} & \multirow{-2}{*}{Text classifications} & \multirow{-2}{*}{Error rate} & orthogonality constraint & \multirow{-2}{*}{\href{http://pfliu.com/paper/adv-mtl.html}{\textcolor{citegreen}{Official}}}\\

    \midrule 
    \rowcolor{gray!20}
        & & & Cascading, adding & & & & Part-Of-Speech (POS) tagging/chunking/parsing/ & Accuracy (acc), F1, MSE, unlabeled attachment & CE loss, softmax loss, & \\
    \rowcolor{gray!20}
        \multirow{-2}{*}{JMT} & \multirow{-2}{*}{EMNLP} & \multirow{-2}{*}{\citeyear{hashimoto-jmt:2017:EMNLP2017}} & constraints & \multirow{-2}{*}{LSTM} & \multirow{-2}{*}{Soft} & \multirow{-2}{*}{Text} & semantic relatedness/textual entailment & score (UAS)/labeled attachment score (LAS) & KL-divergence & \multirow{-2}{*}{\href{https://github.com/rubythonode/joint-many-task-model}{\textcolor{reforange}{Unofficial}}}\\
        
    \midrule 
        & & & & & & & Object detection/mask estimation/object  & & Mask regression loss, & \\
        \multirow{-2}{*}{MNCs} & \multirow{-2}{*}{CVPR} & \multirow{-2}{*}{\citeyear{dai2016instance}} & \multirow{-2}{*}{Cascading} & \multirow{-2}{*}{CNN} & \multirow{-2}{*}{Hard} & \multirow{-2}{*}{Image} & categorization & \multirow{-2}{*}{mAP$@$IoU} &  softmax loss & \multirow{-2}{*}{\href{https://github.com/daijifeng001/MNC}{\textcolor{citegreen}{Official}}}\\

        

    \midrule 
    \rowcolor{gray!20}
        FAFS & CVPR & \citeyear{lu2017fully} & NAS & CNN & Hard & Image & person attribute classification & Acc/recall & CE loss &\href{https://github.com/luyongxi/deep_share}{\textcolor{citegreen}{Official}} \\

    \midrule 
        & & & & & Hard \& & & & & & \\
        \multirow{-2}{*}{MRN} & \multirow{-2}{*}{NeurIPS} & \multirow{-2}{*}{\citeyear{long2017learning}} & \multirow{-2}{*}{Task conditioning} & \multirow{-2}{*}{CNN} & Soft & \multirow{-2}{*}{Image} & \multirow{-2}{*}{classifications on different domains} & \multirow{-2}{*}{Acc} & \multirow{-2}{*}{CE loss} & \multirow{-2}{*}{\href{https://github.com/thuml/MTlearn}{\textcolor{citegreen}{Official}}}\\

        

        

    \midrule 
    \rowcolor{gray!20}
        &  &  &  &  &  &  & Depth/scene parsing/contour & rel~\citep{eigen2014depth}/RMSE/log10 mErr/ & CE loss, softmax loss & \\
    \rowcolor{gray!20}
        \multirow{-2}{*}{PAD-Net} & \multirow{-2}{*}{CVPR} & \multirow{-2}{*}{\citeyear{xu2018pad}} & \multirow{-2}{*}{Mutual distillation} & \multirow{-2}{*}{CNN} & \multirow{-2}{*}{Hard} & \multirow{-2}{*}{Image} & prediction/normals & acc with threshold $\delta$ (acc-$\delta$), IoU/acc & Euclidean loss & \multirow{-2}{*}{---}\\

    \midrule 
        MT$\text{A}_{(\text{adv})}$N & CVPR & \citeyear{liu2018multi} & Adversarial training & CNN & Hard & Image & font/glyph, identity/pose/illumination & Recognition rate & CE loss, adversarial loss & ---\\

    \midrule 
    \rowcolor{gray!20}
        & & & cross-task & & & & & rel/ & berHu loss~\citep{laina2016deeper}, & \\ 
        
    \rowcolor{gray!20}
        \multirow{-2}{*}{TRL} & \multirow{-2}{*}{ECCV} & \multirow{-2}{*}{\citeyear{zhang2018joint}} & attention & \multirow{-2}{*}{CNN} & \multirow{-2}{*}{Hard} & \multirow{-2}{*}{Image} & \multirow{-2}{*}{Depth estimation (depth)/semseg} & RMSE/acc-$\delta$, pixacc/mean acc/mIoU & CE loss, uncertainty loss & \multirow{-2}{*}{---}\\



    \midrule 
        \multirow{2}{*}{MMoE} & \multirow{2}{*}{KDD} & \multirow{2}{*}{\citeyear{ma2018modeling}} & MoE & \multirow{2}{*}{MLP} & Hard \& & Tabular & Income/education/marriage prediction,  & \multirow{2}{*}{Area Under the Curve (AUC)} & \multirow{2}{*}{CE loss} & \multirow{2}{*}{\href{https://github.com/shenweichen/DeepCTR}{\textcolor{reforange}{Unofficial}}}\\
        
        & & & & & soft & data & engagement/satisfaction in recommendation & & & \\

    \midrule 
    \rowcolor{gray!20}
        & & & & & & Tabular & & & & \\
        
    \rowcolor{gray!20}
        \multirow{-2}{*}{Soft Order} & \multirow{-2}{*}{ICLR} & \multirow{-2}{*}{\citeyear{meyerson2018beyond}} & \multirow{-2}{*}{feature fusion} & \multirow{-2}{*}{CNN, MLP} & \multirow{-2}{*}{Soft} & data, image & \multirow{-2}{*}{Classification, attribute recognition}  & \multirow{-2}{*}{mErr} & \multirow{-2}{*}{CE loss} & \multirow{-2}{*}{---}\\

    \midrule 
         & & & & & &  & classification/colorization/edge/denoised & & & \\
        
        \multirow{-2}{*}{GREAT4MTL} & \multirow{-2}{*}{arXiv} & \multirow{-2}{*}{\citeyear{sinha2018gradient}} & \multirow{-2}{*}{adversarial training} & \multirow{-2}{*}{CNN} & \multirow{-2}{*}{Hard} & \multirow{-2}{*}{Image} & reconstruction, depth/normal/keypoint & \multirow{-2}{*}{Err, RMSE, $1 - |\text{cos}(\cdot, \cdot)|$} & \multirow{-2}{*}{CE loss} & \multirow{-2}{*}{---}\\

    \midrule 
    \rowcolor{gray!20}
        Sluice &  &  & Adding constraints, &  & Hard \& &  &  Chunking/entity recognition (NER)/semantic  &  &  & \\
    \rowcolor{gray!20}
        networks & \multirow{-2}{*}{AAAI} & \multirow{-2}{*}{\citeyear{ruder2019latent}} & early stopping  & \multirow{-2}{*}{LSTM} & Soft & \multirow{-2}{*}{Text} & role labeling (SRL)/POS tagging & \multirow{-2}{*}{Acc} & \multirow{-2}{*}{CE loss} & \multirow{-2}{*}{\href{https://github.com/sebastianruder/sluice-networks}{\textcolor{citegreen}{Official}}} \\

    \midrule 
         & & & & CNN, & & & NER/Entity Mention Detection (EMD)/Relation & F1 score/precision/recall, MUC/B3/CEAFe & & \\
         \multirow{-2}{*}{HMTL} & \multirow{-2}{*}{AAAI} & \multirow{-2}{*}{\citeyear{sanh2019hierarchical}} & \multirow{-2}{*}{cascading} & LSTM & \multirow{-2}{*}{Hard} & \multirow{-2}{*}{Text} & Extraction (RE)/Coreference Resolution (CR) & \citep{moosavi2016coreference} & \multirow{-2}{*}{CE loss} & \multirow{-2}{*}{\href{https://github.com/huggingface/hmtl}{\textcolor{reforange}{Unofficial}}} \\

    \midrule 
    \rowcolor{gray!20} 
         & & & & CNN, & & & Segment labeling/Named Entity Labeling & & CRF loss, CE loss, ranking & \\
    \rowcolor{gray!20}    
         \multirow{-2}{*}{DCMTL} & \multirow{-2}{*}{AAAI} & \multirow{-2}{*}{\citeyear{gong2019deep}} & \multirow{-2}{*}{cascading} & LSTM & \multirow{-2}{*}{Hard} & \multirow{-2}{*}{Text} & (NEL)/slot filling & \multirow{-2}{*}{F1 score/precision/recall} & loss~\citep{vu2016bi} & \multirow{-2}{*}{\href{https://github.com/gy910210/DCMTL}{\textcolor{citegreen}{Official}}} \\

    \midrule 
         & & & & & & & Normals/semseg, age estimation/gender & mErr/medErr/within $t^{\circ}$, mIoU, pixacc, mean/ & & \\
         \multirow{-2}{*}{NDDR-CNN} & \multirow{-2}{*}{CVPR} & \multirow{-2}{*}{\citeyear{gao2019nddr}} & \multirow{-2}{*}{feature fusion} & \multirow{-2}{*}{CNN} & \multirow{-2}{*}{Soft} & \multirow{-2}{*}{Image} & classification & median absolute error (absErr), acc & \multirow{-2}{*}{CE loss} & \multirow{-2}{*}{\href{https://github.com/ethanygao/NDDR-CNN}{\textcolor{citegreen}{Official}}} \\

    \midrule 
    \rowcolor{gray!20}
        &  &  & cross-task &  &  &  &  & RMSE/rel/acc with $t$, mErr/medError/within & CE loss, $\ell_1$ loss, berHu loss & \\
    \rowcolor{gray!20}
        \multirow{-2}{*}{PAP} & \multirow{-2}{*}{CVPR} & \multirow{-2}{*}{\citeyear{zhang2019pattern}} & attention & \multirow{-2}{*}{CNN} & \multirow{-2}{*}{Hard} & \multirow{-2}{*}{Image} & \multirow{-2}{*}{Semseg/depth/normals} & $t^{\circ}$, mIoU/mean accuracy (mAcc)/pixacc & affinity loss~\citep{zhang2019pattern}  & \multirow{-2}{*}{---} \\

    \midrule 
        MT$\text{A}_{(\text{atten})}$N & & & & & Hard \& & & Semseg/depth/normals, 10 classifications & mIoU/pixacc, mErr/medErr/within $t^{\circ}$, & &  \\ 

        (\& DWA) & \multirow{-2}{*}{CVPR} & \multirow{-2}{*}{\citeyear{liu2019end}} & \multirow{-2}{*}{Adaptive weighting} & \multirow{-2}{*}{CNN} & Soft & \multirow{-2}{*}{Image} & (visual domain decathlon\tnote{2}\,\,\,) & absErr/real error, accuracy & \multirow{-2}{*}{CE loss, $\ell_1$ loss, dot product} & \multirow{-2}{*}{\href{https://github.com/lorenmt/mtan}{\textcolor{citegreen}{Official}}} \\

        

    \midrule 
    \rowcolor{gray!20}
        & & & & & & & Semseg/depth/edge/normals/human parts/ & mIoU/osdF/mErr/maximum F-measure (maxF)/ &  & \\
    \rowcolor{gray!20}
        \multirow{-2}{*}{ASTMT} & \multirow{-2}{*}{CVPR} & \multirow{-2}{*}{\citeyear{maninis2019attentive}} & \multirow{-2}{*}{attention, single-tasking} & \multirow{-2}{*}{CNN} & \multirow{-2}{*}{Hard} & \multirow{-2}{*}{Image} & saliency estimation/albedo & RMSE/$\Delta_m$ & \multirow{-2}{*}{CE loss, $\ell_1$ loss} & \multirow{-2}{*}{\href{https://github.com/facebookresearch/astmt}{\textcolor{citegreen}{Official}}}\\

    \midrule 
        {ML-GCN} & CVPR & \citeyear{chen2019multi} & Graph based & CNN, GCN & Hard & Image & Multi-label recognition & precision, recall, F1 & CE loss & \href{https://github.com/megvii-research/ML-GCN}{\textcolor{citegreen}{Official}} \\

    \midrule 
    \rowcolor{gray!20}
        RD4MTL & arXiv & \citeyear{meng2019representation} & Adversarial training & CNN & Hard & Image & Classifications & Acc & CE loss, adversarial loss & \href{https://github.com/qmeng99/Multi-task-Representation-Disentanglement}{\textcolor{citegreen}{Official}} \\

    \midrule 
        \multirow{2}{*}{MTL-NAS} & \multirow{2}{*}{CVPR} & \multirow{2}{*}{\citeyear{gao2020mtl}} & \multirow{2}{*}{NAS} & \multirow{2}{*}{CNN} & \multirow{2}{*}{Adaptive} & \multirow{2}{*}{Image} & Semseg/normals, object classification/scene & mErr/medErr/Within $t^\circ$, mIoU/pixacc, & \multirow{2}{*}{CE loss, $\ell_2$ loss} & \multirow{2}{*}{\href{https://github.com/bhpfelix/MTLNAS}{\textcolor{citegreen}{Official}}}\\
        & & & & & & & classification & Acc &  & \\
        
    \midrule 
    \rowcolor{gray!20}
         & & & & & & & Semseg/edge/depth/keypoint detection (point), & & & \\
    \rowcolor{gray!20} 
         \multirow{-2}{*}{BMTN} & \multirow{-2}{*}{BMVC} & \multirow{-2}{*}{\citeyear{vandenhende2019branched}} & \multirow{-2}{*}{NAS} & \multirow{-2}{*}{CNN} & \multirow{-2}{*}{Adaptive} & \multirow{-2}{*}{Image} & attribute classification &  \multirow{-2}{*}{mIoU, pixacc, $\ell_1$, Acc} & \multirow{-2}{*}{CE loss, $\ell_1$ loss, $\Delta_m$} & \multirow{-2}{*}{\href{https://github.com/hurricane2018/Branched-Multi-Task-Networks-Deciding-What-Layers-To-Share}{\textcolor{citegreen}{Official}}}\\

    \midrule 
        \multirow{2}{*}{PSD} & \multirow{2}{*}{CVPR} & \multirow{2}{*}{\citeyear{zhou2020pattern}} & \multirow{2}{*}{Distillation}& \multirow{2}{*}{CNN} & \multirow{2}{*}{Hard \& soft} & \multirow{2}{*}{Image} & \multirow{2}{*}{Semseg/depth/normals} & RMSE/rel/acc with $t$, mIoU/mean accuracy/ & \multirow{2}{*}{CE loss, $\ell_1$ loss, berHu loss} & \multirow{2}{*}{---}\\

        & & & & & & & & pixacc, mErr/medErr/within $t^{\circ}$  & & \\

    \midrule 
    \rowcolor{gray!20}
        & ECCV & & distillation & & Hard \& & & & mIoU/pixacc, absErr/rel, mErr/medErr/within & & \\
    \rowcolor{gray!20}
        \multirow{-2}{*}{KD4MTL} & Workshop & \multirow{-2}{*}{\citeyear{li2020knowledge}} & knowledge & \multirow{-2}{*}{CNN} & soft & \multirow{-2}{*}{Image} & \multirow{-2}{*}{Semseg/depth/normals, classification} & $t^\circ$, Acc & \multirow{-2}{*}{CE loss, $\ell_1$ loss, dot product} & \multirow{-2}{*}{\href{https://github.com/VICO-UoE/KD4MTL}{\textcolor{citegreen}{Official}}} \\

    \midrule 
        \multirow{2}{*}{MTI-Net} & \multirow{2}{*}{ECCV} & \multirow{2}{*}{\citeyear{vandenhende2020mti}} & multi-task & \multirow{2}{*}{CNN} & Hard \& & \multirow{2}{*}{Image} & Semseg/depth/edges detection (edges)/normals/ & mIoU, RMSE, mErr, optimal dataset-scale F- & \multirow{2}{*}{CE loss, $\ell_1$ loss} & \multirow{2}{*}{\href{https://github.com/SimonVandenhende/Multi-Task-Learning-PyTorch}{\textcolor{citegreen}{Official}}}\\

        & & & distillation & & Soft & & saliency estimation/human parts  & measure (odsF)~\citep{martin2004learning}, $\Delta_m$ & & \\

    \midrule 
    \rowcolor{gray!20}
        & & & NAS, & & & & Regression, face attribute prediction, semseg/ & & & \\
    \rowcolor{gray!20}
        \multirow{-2}{*}{LTB} & \multirow{-2}{*}{ICML} & \multirow{-2}{*}{\citeyear{guo2020learning}} & task grouping & \multirow{-2}{*}{CNN} & \multirow{-2}{*}{Soft} & \multirow{-2}{*}{Image} & normals/depth/keypoints/edges & \multirow{-2}{*}{Acc, CE, cos, mean absErr}  & \multirow{-2}{*}{CE loss, $\ell_1$ loss, cosine loss} & \multirow{-2}{*}{---} \\

    \midrule 
        & & & & CNN \& & & &  & & & \\
        \multirow{-2}{*}{AAMTRL} & \multirow{-2}{*}{ICML} & \multirow{-2}{*}{\citeyear{mao2020adaptive}} & \multirow{-2}{*}{adversarial training} & LSTM & \multirow{-2}{*}{Hard} & \multirow{-2}{*}{Text} & \multirow{-2}{*}{Classifications} & \multirow{-2}{*}{Relatedness evolution, acc, influence of \#task}  & \multirow{-2}{*}{Any 1-Lipschitz loss} & \multirow{-2}{*}{---} \\

        

        

        

    \midrule 
    \rowcolor{gray!20}
        & & & & & Hard \& & Tabular & Sub-tasks in the recommendation systems, & & & \\
    \rowcolor{gray!20} 
        \multirow{-2}{*}{CGC \& PLE} & \multirow{-2}{*}{RecSys} & \multirow{-2}{*}{\citeyear{tang2020progressive}} & \multirow{-2}{*}{MoE} & \multirow{-2}{*}{MLP} & soft & data & income/education/marriage prediction & \multirow{-2}{*}{AUC/MSE, MTL gain} & \multirow{-2}{*}{CE loss, $\ell_2$ loss} & \multirow{-2}{*}{\href{https://github.com/shenweichen/DeepCTR}{\textcolor{reforange}{Unofficial}}}\\





        

    \midrule 
        \multirow{2}{*}{TSNs} & \multirow{2}{*}{ICCV} & \multirow{2}{*}{\citeyear{sun2021task}} & task relationship learning, & \multirow{2}{*}{CNN} & \multirow{2}{*}{Hard} & \multirow{2}{*}{Image} & Semseg/depth/edges/normals/ & \multirow{2}{*}{mIoU, RMSE, mErr, odsF, $\Delta_m$} & \multirow{2}{*}{CE loss, $\ell_1$ loss} & \multirow{2}{*}{\href{https://github.com/GuoleiSun/TSNs}{\textcolor{citegreen}{Official}}}\\

        & & & task conditioning & & & & saliency estimation/human parts & & & \\

    \midrule 
    \rowcolor{gray!20}
        & & & knowledge distillation, & & & & Classification/detection/semseg/depth/ & & & \\
    \rowcolor{gray!20}
        \multirow{-2}{*}{MuST} & \multirow{-2}{*}{ICCV} & \multirow{-2}{*}{\citeyear{ghiasi2021multi}} & task conditioning & \multirow{-2}{*}{CNN} & \multirow{-2}{*}{Hard} & \multirow{-2}{*}{Image} & normals & \multirow{-2}{*}{Acc, mIoU, RMSE, odsF} & \multirow{-2}{*}{CE loss, $\ell_1$ loss} & \multirow{-2}{*}{---}\\ 

    \midrule 
        \multirow{2}{*}{AuxSegNet} & \multirow{2}{*}{ICCV} & \multirow{2}{*}{\citeyear{xu2021leveraging}} & cross-task & \multirow{2}{*}{CNN} & Hard \& & \multirow{2}{*}{Image} & \multirow{2}{*}{Semseg/classification/saliency detection} & \multirow{2}{*}{mIoU/precision/recall} & Multi-label softmax & \multirow{2}{*}{\href{https://github.com/xulianuwa/AuxSegNet}{\textcolor{citegreen}{Official}}}\\
        
        & & & attention  & & Soft & & & & loss, CE loss& \\

    \midrule 
    \rowcolor{gray!20}
        & & & cross-task & & Hard \& & & Semseg/depth estimation/edges/normals/ & & & \\
        
    \rowcolor{gray!20}
        \multirow{-2}{*}{ATRC} & \multirow{-2}{*}{ICCV} & \multirow{-2}{*}{\citeyear{bruggemann2021exploring}} & attention & \multirow{-2}{*}{CNN} & soft & \multirow{-2}{*}{Image} & saliency estimation/human parts & \multirow{-2}{*}{mIoU, RMSE, mErr, odsF, maxF, $\Delta_m$} & \multirow{-2}{*}{CE loss, $\ell_1$ loss} & \multirow{-2}{*}{\href{https://github.com/brdav/atrc}{\textcolor{citegreen}{Official}}}\\

        

    \midrule 
        \multirow{2}{*}{DSelect-k} & \multirow{2}{*}{NeurIPS} & \multirow{2}{*}{\citeyear{hazimeh2021dselect}} & \multirow{2}{*}{MoE} & \multirow{2}{*}{MLP, CNN} & Hard \& & Tabular & \multirow{2}{*}{engagement/satisfaction task, classification} & \multirow{2}{*}{Total loss, Acc, AUC/RMSE, \#expert}  & \multirow{2}{*}{CE loss, $\ell_2$ loss} & \multirow{2}{*}{\href{https://github.com/google-research/google-research/tree/master/dselect_k_moe}{\textcolor{citegreen}{Official}}}\\
        
        & & & & & soft & data, Image &  &  & & \\

    \midrule 
    \rowcolor{gray!20}
        & & & & & Hard \& & & 16 Language understanding tasks, e.g. textual & Acc, Spearman correlation~\citeyear{spearman1961proof}, Matthews & & \\
        
    \rowcolor{gray!20}
        \multirow{-2}{*}{MT-TaG} & \multirow{-2}{*}{ArXiv} & \multirow{-2}{*}{\citeyear{gupta2022sparsely}} & \multirow{-2}{*}{MoE} & \multirow{-2}{*}{Transformer} & soft & \multirow{-2}{*}{Text} & entailment, sentiment classification, etc. & correlation coefficient~\citep{matthews1975comparison}  & \multirow{-2}{*}{CE loss, MSE} & \multirow{-2}{*}{---}\\

    \midrule 
    \rowcolor{gray!20}
        & & & & & Hard \& & Tabular & &  & & \\
        
    \rowcolor{gray!20}
        \multirow{-2}{*}{CrossDistil} & \multirow{-2}{*}{AAAI} & \multirow{-2}{*}{\citeyear{yang2022cross}} & \multirow{-2}{*}{distillation} & \multirow{-2}{*}{MLP} & soft & data & \multirow{-2}{*}{Finish watching/like} & \multirow{-2}{*}{AUC, multi-AUC~\citep{hand2001simple}}  & \multirow{-2}{*}{CE loss} & \multirow{-2}{*}{---}\\

        

    \midrule 
        \multirow{2}{*}{MulT} & \multirow{2}{*}{CVPR} & \multirow{2}{*}{\citeyear{bhattacharjee2022mult}} & cross-task & CNN \& & \multirow{2}{*}{Hard} & \multirow{2}{*}{Image} & Semseg/depth/reshading/normals/ & MTL gain, mErr of & CE loss, $\ell_1$ loss, & \multirow{2}{*}{\href{https://ivrl.github.io/MulT/}{\textcolor{citegreen}{Official}}}\\
        
        & & & attention & Transformer & & & keypoints/edges & domain generalization & rotate loss~\citep{zamir2018taskonomy} & \\

    \midrule 

    \rowcolor{gray!20}
        & & & cross-task attention, & & & & & & CE loss, $\ell_1$ loss, cross- & \\

    \rowcolor{gray!20}
        & & & task balancing (\textit{spec.}, & & \multirow{-1.8}{*}{Hard \&} & & \multirow{-1.8}{*}{Semseg/depth/saliency detection/} & & task contrastive loss, & \\
        
    \rowcolor{gray!20}
        \multirow{-3}{*}{MTFormer} & \multirow{-3}{*}{ECCV} & \multirow{-3}{*}{\citeyear{xu2022mtformer}} & \citet{kendall2018multi}) & \multirow{-3}{*}{Transformer} & \multirow{-1.8}{*}{soft} & \multirow{-3}{*}{Image} & \multirow{-1.8}{*}{human parts} & \multirow{-3}{*}{mIoU, RMSE, $\Delta_m$} & uncertainty loss & \multirow{-3}{*}{---}\\

    \midrule 
        \multirow{2}{*}{MQTransformer} & \multirow{2}{*}{arXiv} & \multirow{2}{*}{\citeyear{xu2022multi}} & cross-task & \multirow{2}{*}{Transformer} & Hard \& & \multirow{2}{*}{Image} & Semseg/depth/edges/normals/saliency & \multirow{2}{*}{mIoU, RMSE, mErr, odsF, maxF} & \multirow{2}{*}{CE loss, $\ell_1$ loss} & \multirow{2}{*}{---}\\
        
        & & & attention & & Soft & &  estimation/human parts & & & \\


        

    \midrule 
    \rowcolor{gray!20}
        & & & & & & Image, & & & & \\
    \rowcolor{gray!20}
        \multirow{-2}{*}{MetaLink} & \multirow{-2}{*}{ICLR} & \multirow{-2}{*}{\citeyear{cao2022relational}} & \multirow{-2}{*}{Graph based} & \multirow{-2}{*}{MLP, GNN} & \multirow{-2}{*}{Hard} & Graph & \multirow{-2}{*}{Classification} & \multirow{-2}{*}{mAP, ROC AUC} & \multirow{-2}{*}{CE loss} & \multirow{-2}{*}{\href{https://github.com/snap-stanford/GraphGym}{\textcolor{citegreen}{Official}}}\\ 

    \midrule 
        \multirow{2}{*}{DeMT} & \multirow{2}{*}{AAAI} & \multirow{2}{*}{\citeyear{zhang2023demt}} & cross-task & \multirow{1}{*}{CNN \&} & Hard \& & \multirow{2}{*}{Image} & Semseg/depth/edges/normals/saliency  & \multirow{1}{*}{mIoU, RMSE, mErr, odsF, maxF,} & \multirow{2}{*}{CE loss, $\ell_1$ loss} & \multirow{2}{*}{\href{https://github.com/yangyangxu0/DeMT}{\textcolor{citegreen}{Official}}} \\
        
        & & & attention & \multirow{1}{*}{Transformer} & Soft & & estimation/human parts & \multirow{1}{*}{$\Delta_m$} & &  \\

    \midrule 
    \rowcolor{gray!20}
        & & & cross-task & & Hard \& & & & & CE loss, berHu loss, cosine & \\
    \rowcolor{gray!20}
        \multirow{-2}{*}{mTEB} & \multirow{-2}{*}{WACV} & \multirow{-2}{*}{\citeyear{lopes2023cross}} & attention & \multirow{-2}{*}{CNN} & soft & \multirow{-2}{*}{Image} & \multirow{-2}{*}{Semseg/depth/normals/edges} & \multirow{-2}{*}{$\Delta_m$, mIoU, RMSE, mErr, F1} & loss~\citep{guizilini2021geometric} & \multirow{-2}{*}{\href{https://github.com/astra-vision/DenseMTL}{\textcolor{citegreen}{Official}}}\\

    \midrule 
        \multirow{2}{*}{OKD-MTL} & \multirow{2}{*}{WACV} & \multirow{2}{*}{\citeyear{jacob2023online}} & distillation, task & \multirow{2}{*}{Transformer} & Hard \& & \multirow{2}{*}{Image} & \multirow{2}{*}{Semseg/depth/normals} & $\Delta_p$, mIoU/pixacc, absErr/rel, mErr/medErr/ & Adaptive feature distillation loss, & \multirow{2}{*}{---}\\
        
        & & & weighting & & Soft & & & within $t^\circ$ & CE loss, $\ell_1$ loss, cosine loss & \\

    \midrule 
    \rowcolor{gray!20}
        & & & & & Hard \& & & & & & \\
        
    \rowcolor{gray!20}
        \multirow{-2}{*}{AdaMV-MoE} & \multirow{-2}{*}{ICCV} & \multirow{-2}{*}{\citeyear{chen2023adamv}} & \multirow{-2}{*}{MoE} & \multirow{-2}{*}{Transformer} & soft & \multirow{-2}{*}{Image} & \multirow{-2}{*}{classification/detection/Seg} & \multirow{-2}{*}{Acc, Average Precision (AP)} & \multirow{-2}{*}{CE loss}& \multirow{-2}{*}{\href{https://github.com/google-research/google-research/tree/master/moe_mtl}{\textcolor{citegreen}{Official}}}\\

        

        

        

    \bottomrule
    \end{tabular}
    \begin{tablenotes}
    \footnotesize
    \item[1] This column provides the link to the implementation or execution. Click on "\textcolor{citegreen}{Official}" or "\textcolor{reforange}{Unofficial}" to access the website.
    \item[2] Part of PASCAL in Detail Workshop Challenge, CVPR 2017, July 26th, Honolulu, Hawaii, USA. \href{https://www.robots.ox.ac.uk/~vgg/decathlon/}{https://www.robots.ox.ac.uk/$\sim$vgg/decathlon}.
    \item[3] We use ``state" here to represent the domain of reinforcement learning, including the observations of states of environment, the positions of object, the actions made by agent, etc.
    \item[4] The average rank of MTL on all different tasks. MR = 1 if a method ranks first across all tasks.
    \end{tablenotes}
    \end{threeparttable}
    }
\end{table*}


%% file: tex_files/02-2/all_archi.tex
\begin{figure}[h]
    \centering

    \begin{subfigure}{0.3\textwidth}
        \includegraphics[width=1\textwidth]{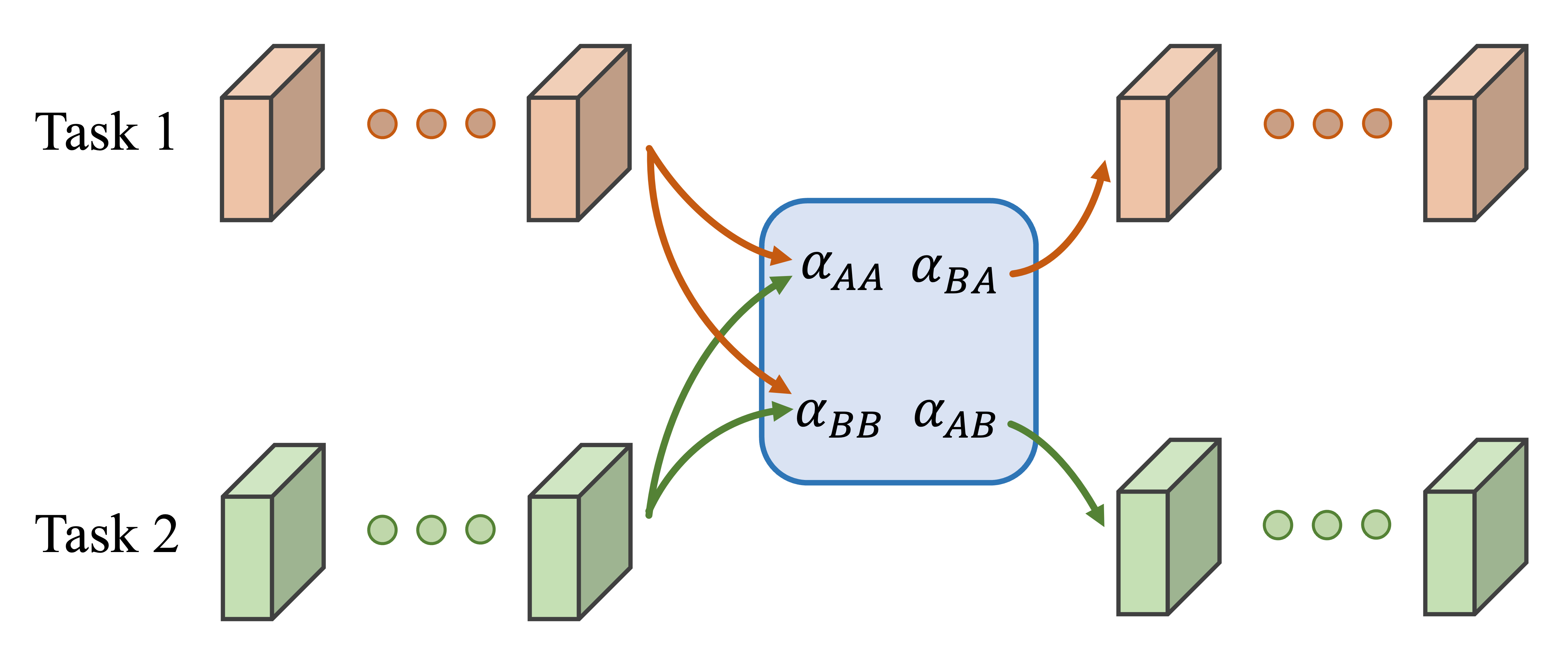}
        \caption{Cross-stitch unit.}
        \label{cross-stitch-unit}
    \end{subfigure} \quad\quad
    \begin{subfigure}{0.3\textwidth}
        \includegraphics[width=1\textwidth]{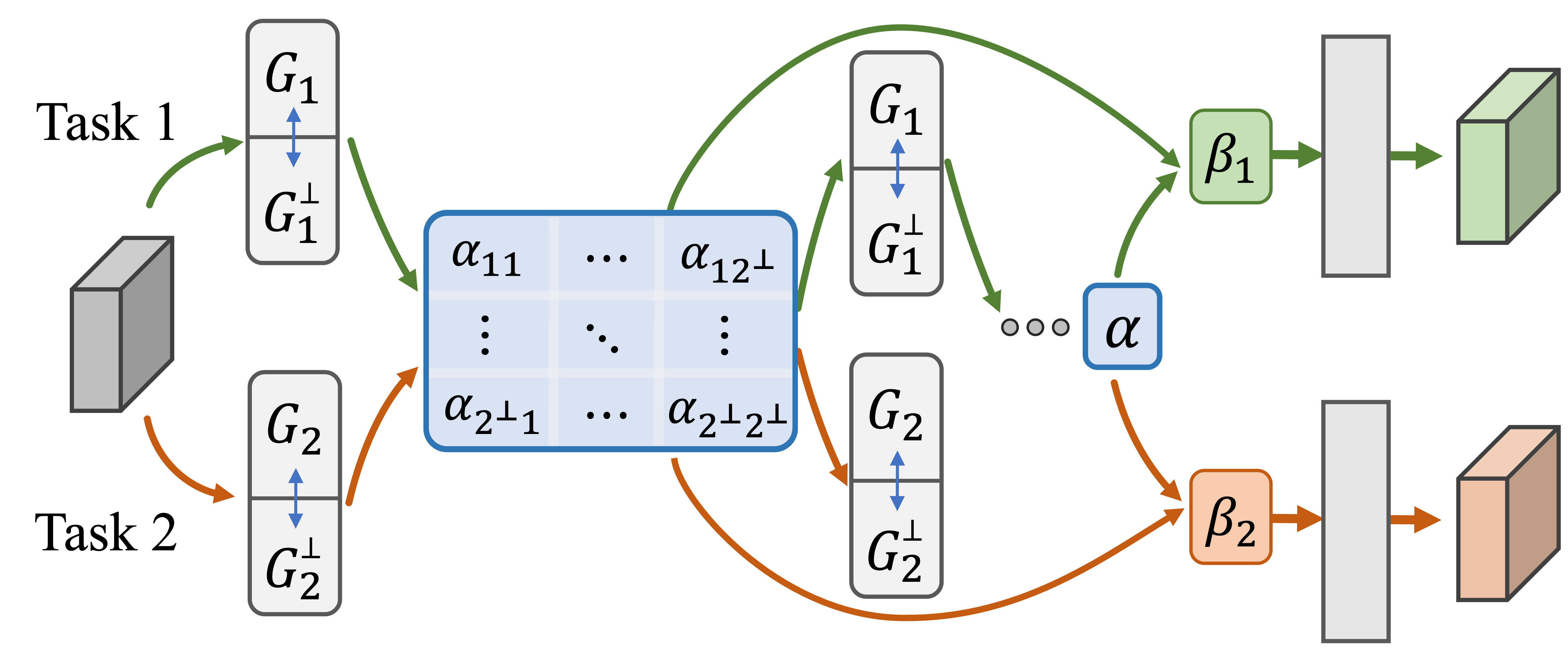}
        \caption{Sluice block.}
        \label{sluice-block}
    \end{subfigure}\quad\quad
    \begin{subfigure}{0.3\textwidth}
        \includegraphics[width=1\textwidth]{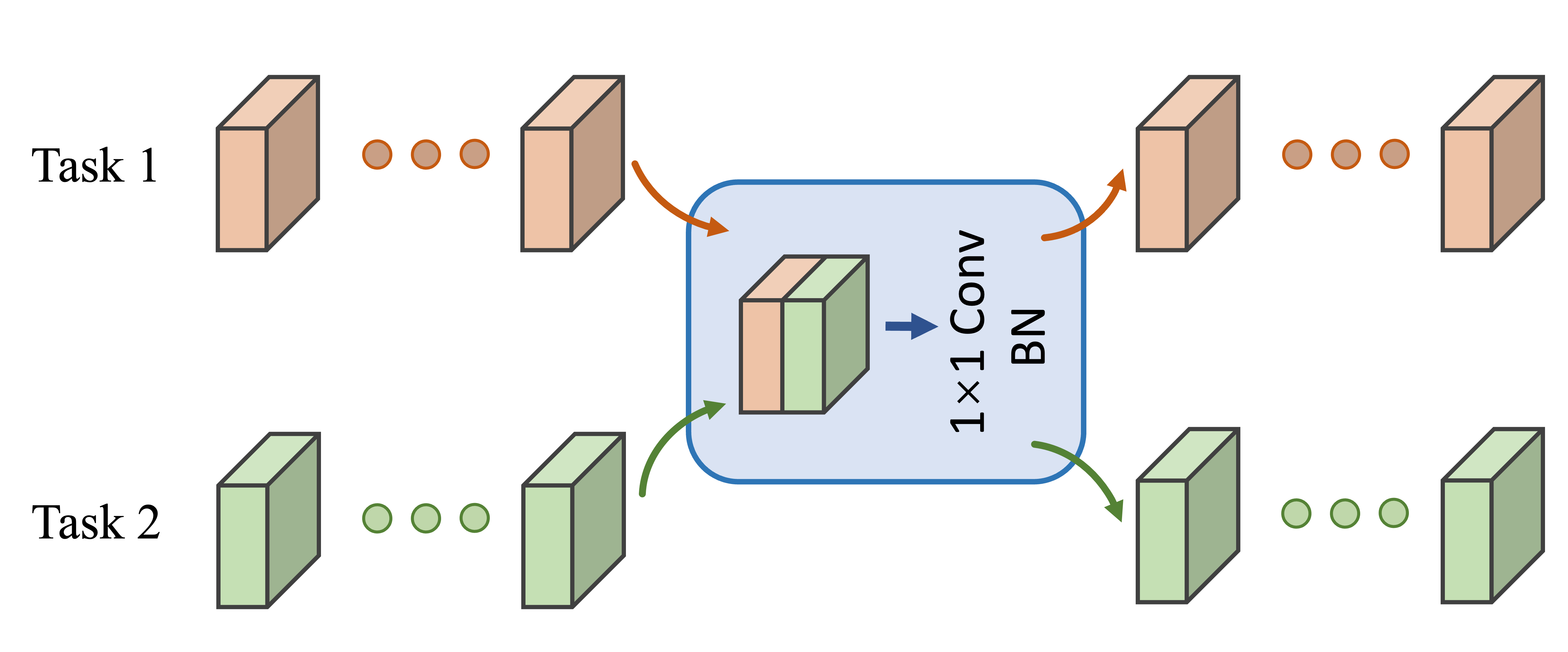}
        \caption{NDDR unit.}
        \label{nddr-unit}
    \end{subfigure}
    
    \begin{subfigure}{0.3\textwidth}
        \includegraphics[width=1\textwidth]{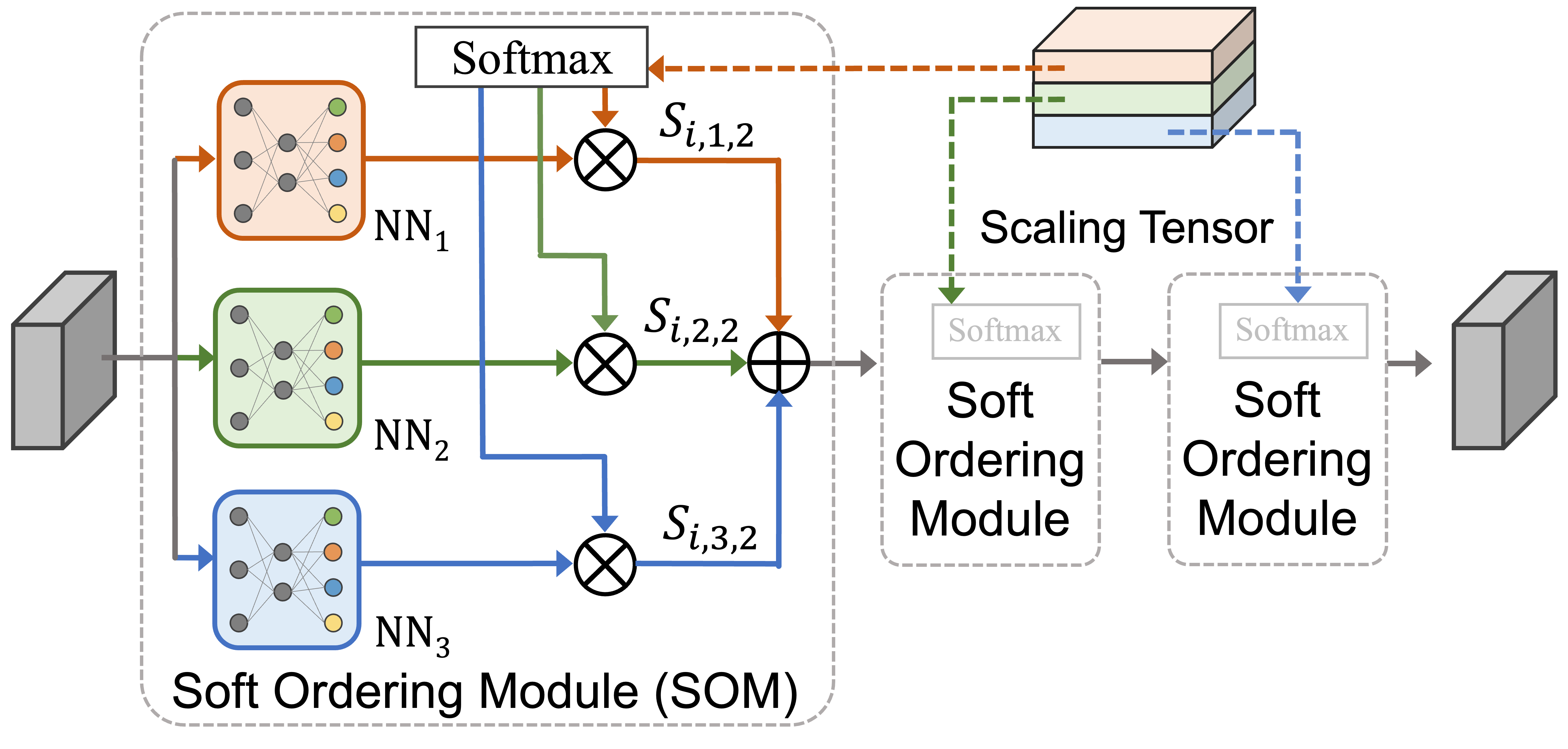}
        \caption{Soft Order.}
        \label{soft-order}
    \end{subfigure}\quad\quad
    \begin{subfigure}{0.3\textwidth}
        \includegraphics[width=1\textwidth]{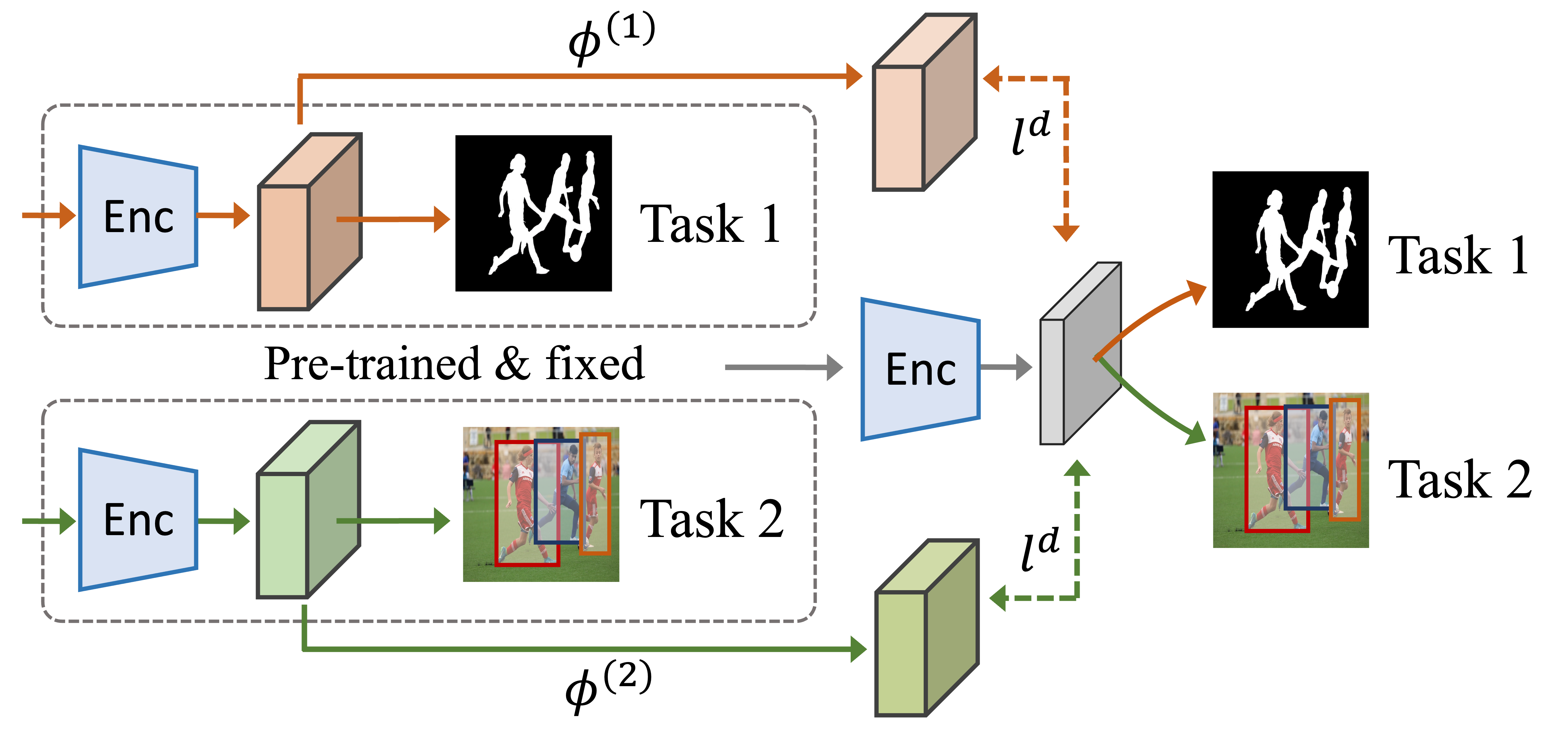}
        \caption{KD4MTL pipeline.}
        \label{kd4mtl-pipeline}
    \end{subfigure}\quad\quad
    \begin{subfigure}{0.3\textwidth}
        \includegraphics[width=1\textwidth]{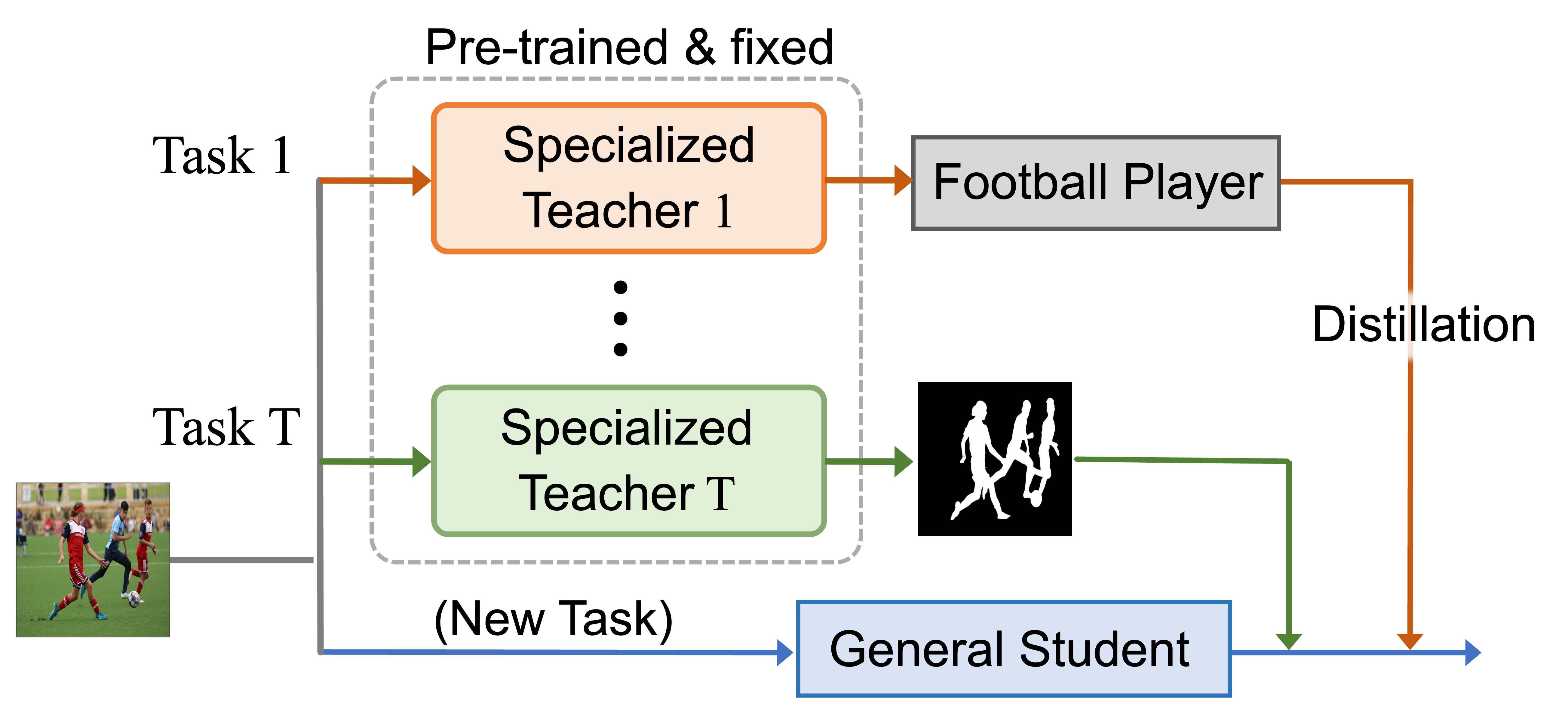}
        \caption{MuST pipeline.}
        \label{must-pipeline}
    \end{subfigure}

    \begin{subfigure}{0.32\textwidth}
        \includegraphics[width=1\textwidth]{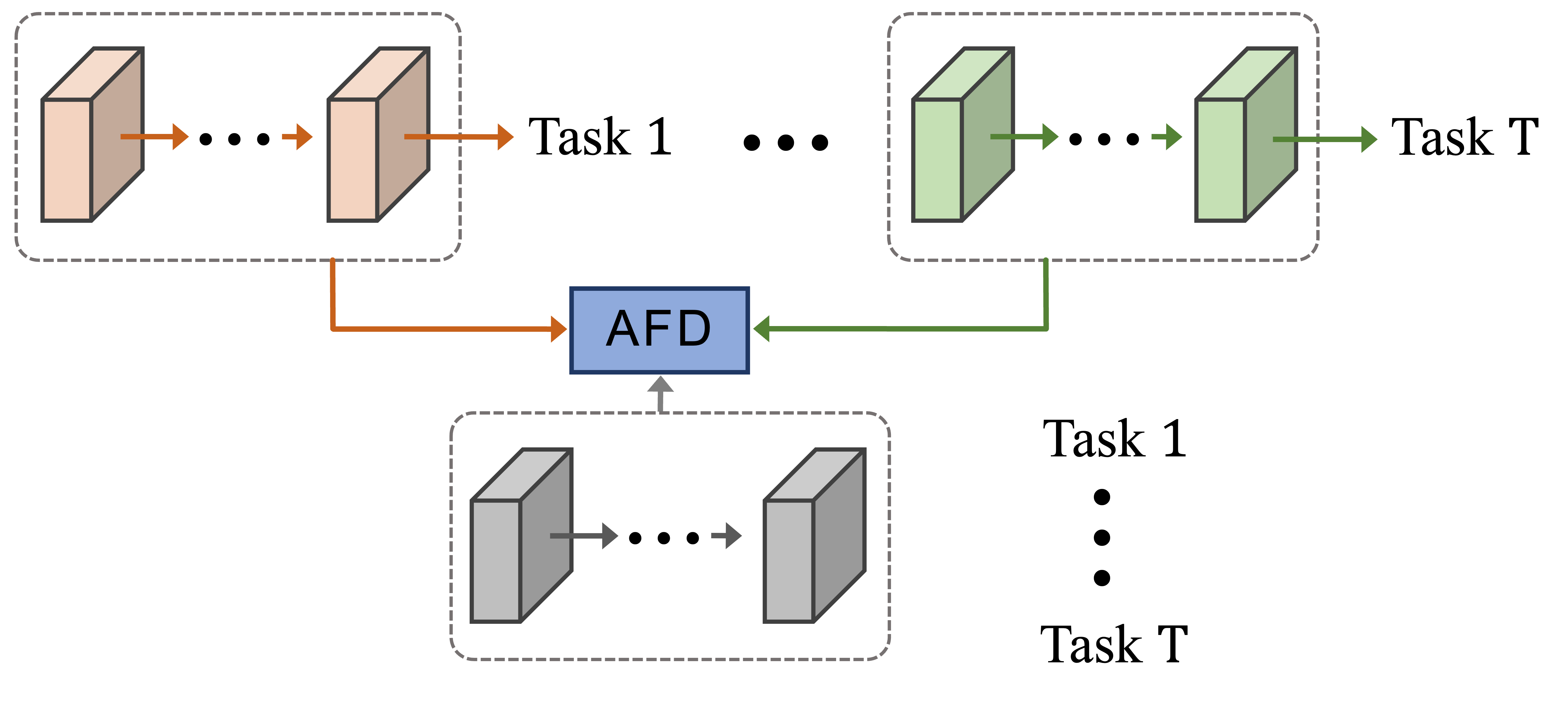}
        \caption{OKD-MTL pipeline.}
        \label{okd-pipeline}
    \end{subfigure}\quad\quad\quad\quad
    \begin{subfigure}{0.39\textwidth}
        \includegraphics[width=1\textwidth]{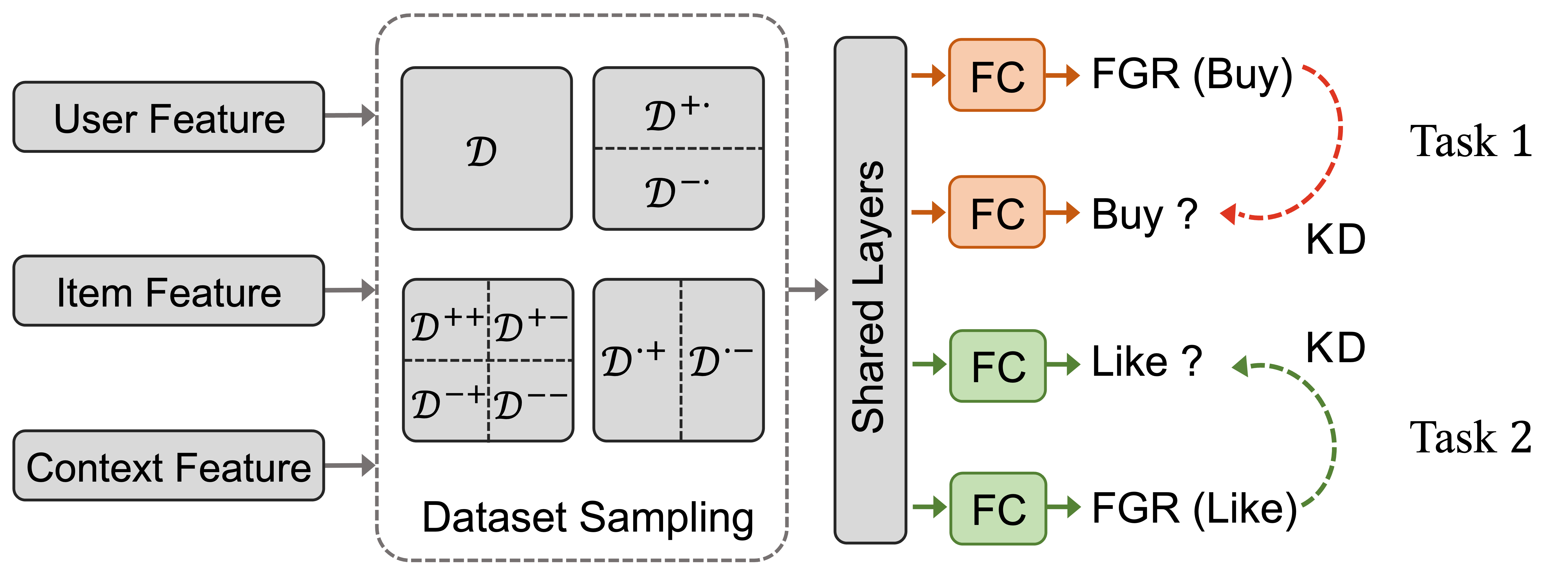}
        \caption{CrossDistil pipeline.}
        \label{crossdistil}
    \end{subfigure}
    
    \begin{subfigure}{0.32\textwidth}
        \includegraphics[width=1\textwidth]{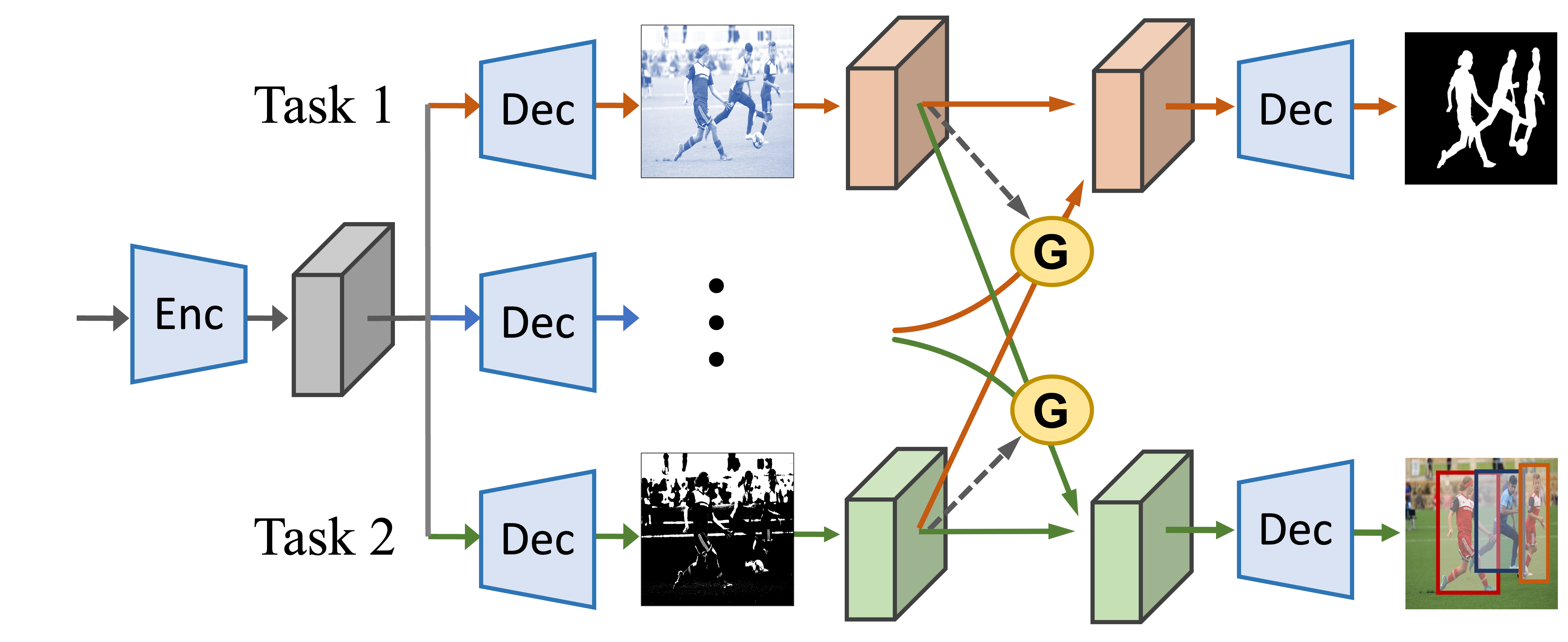}
        \caption{PAD module.}
        \label{pad-module}
    \end{subfigure}\quad\quad\quad\quad
    \begin{subfigure}{0.32\textwidth}
        \includegraphics[width=1\textwidth]{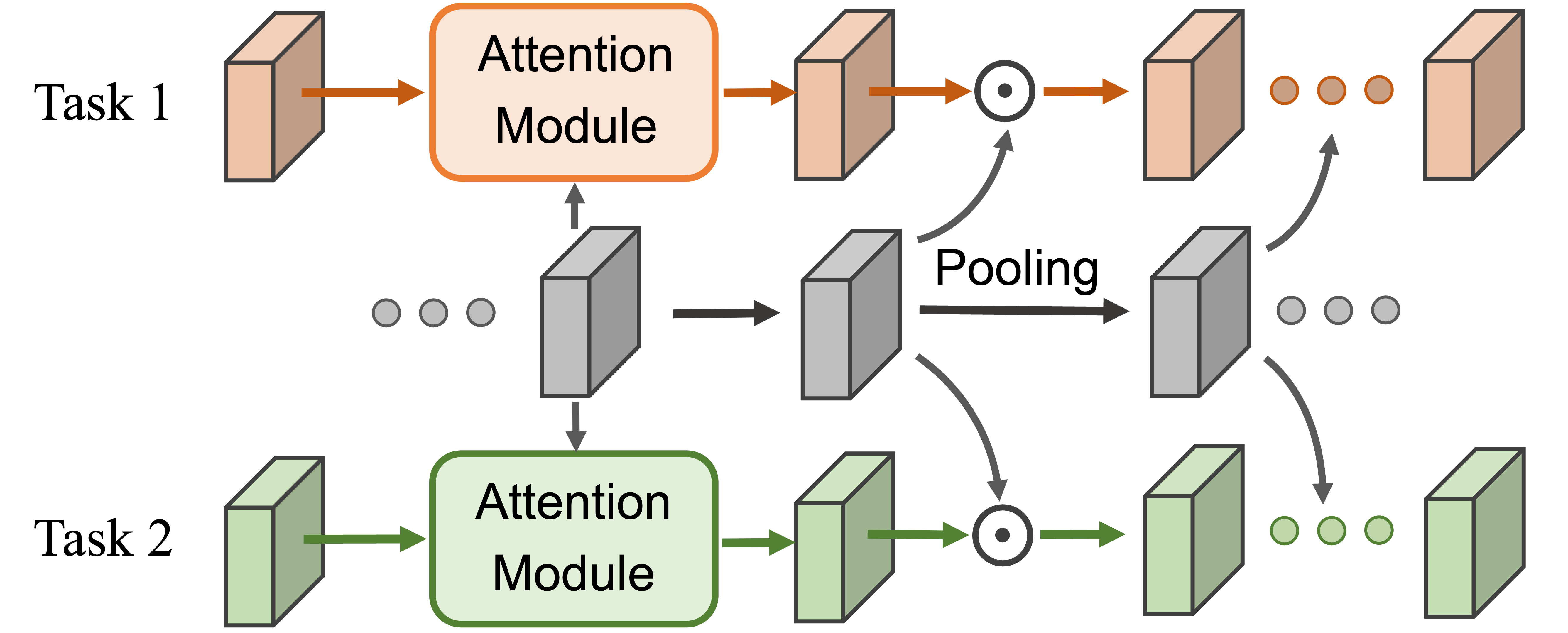}
        \caption{MTAN module.}
        \label{mtan-module}
    \end{subfigure}

    \begin{subfigure}{0.35\textwidth}
        \includegraphics[width=1\textwidth]{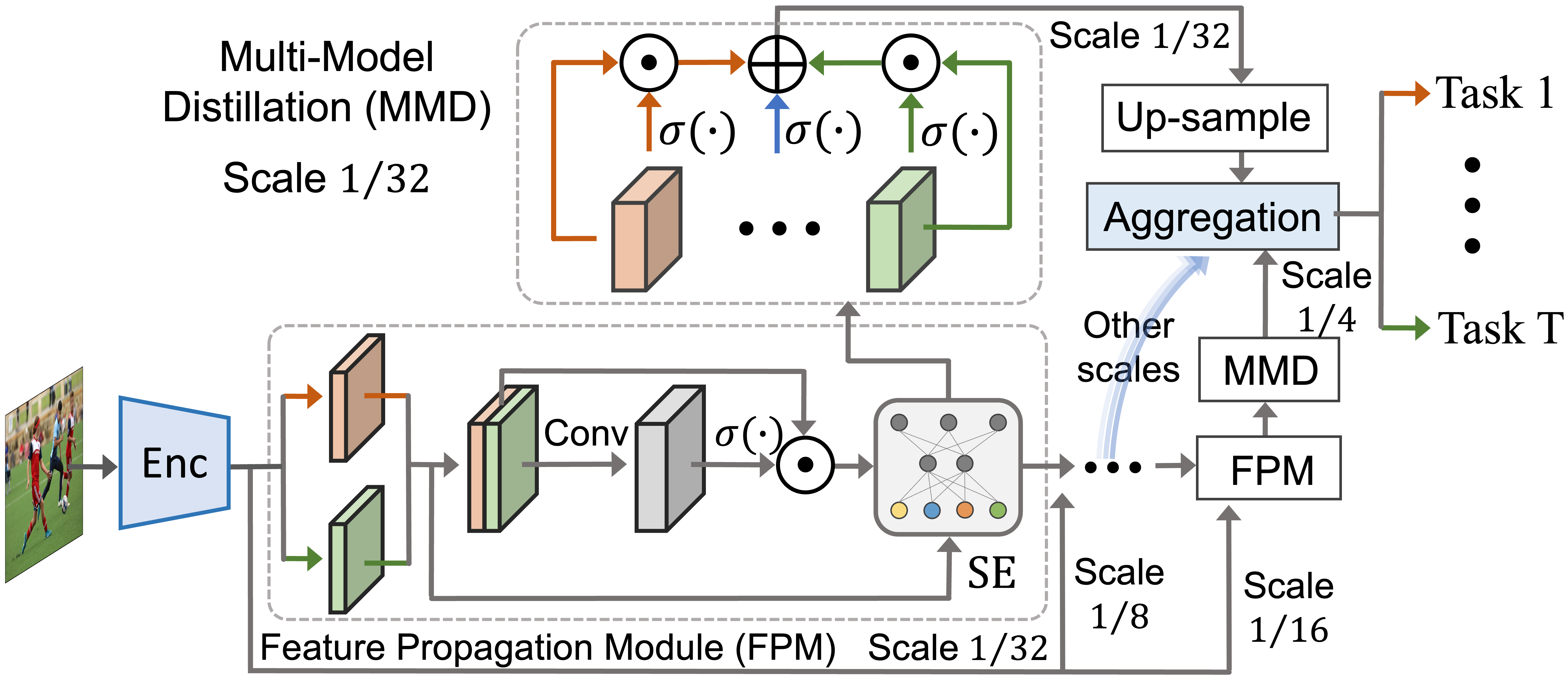}
        \caption{MTI-Net.}
        \label{mti-net}
    \end{subfigure}\quad\quad\quad\quad
    \begin{subfigure}{0.35\textwidth}
        \includegraphics[width=1\textwidth]{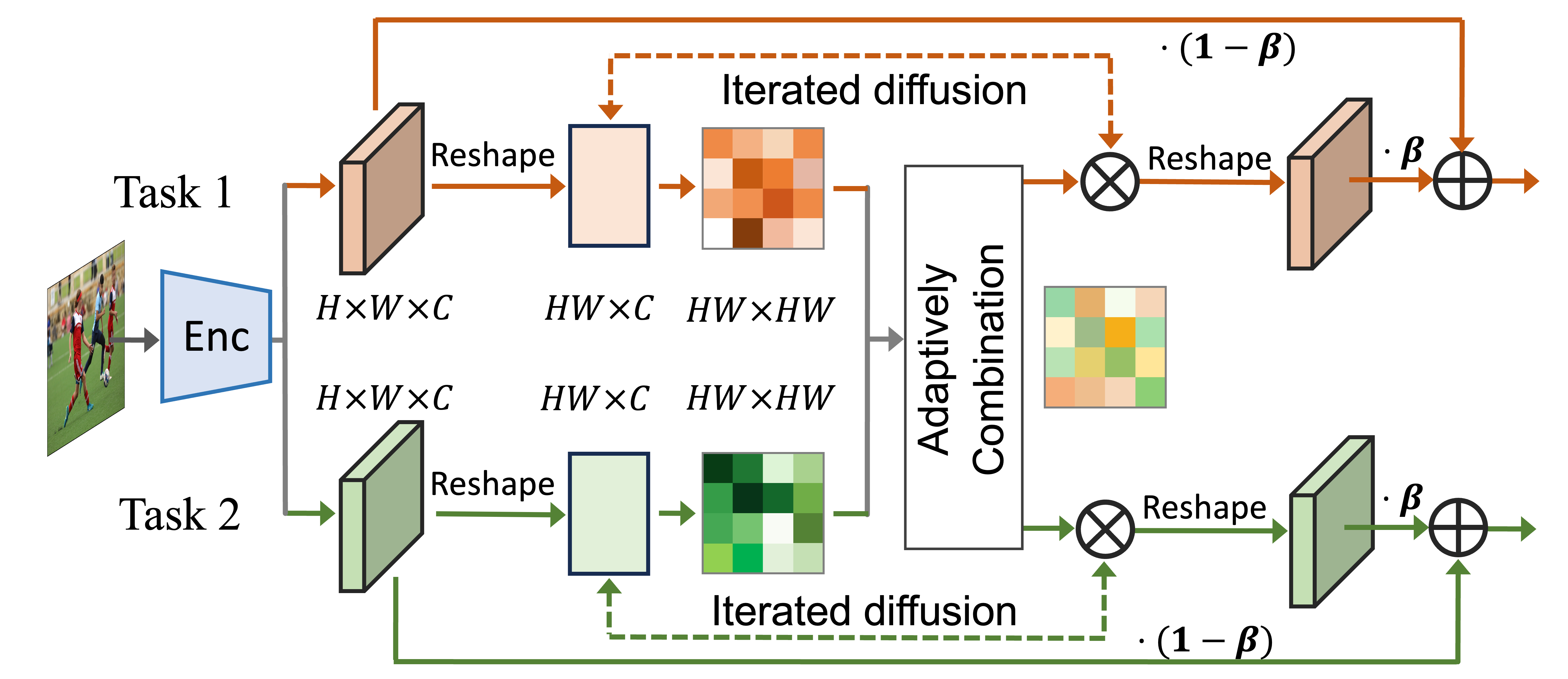}
        \caption{PAP module.}
        \label{pap-module}
    \end{subfigure}
    
\end{figure}
\clearpage
\begin{figure}[h]\ContinuedFloat
    \centering
    
    \begin{subfigure}{0.3\textwidth}
        \includegraphics[width=1\textwidth]{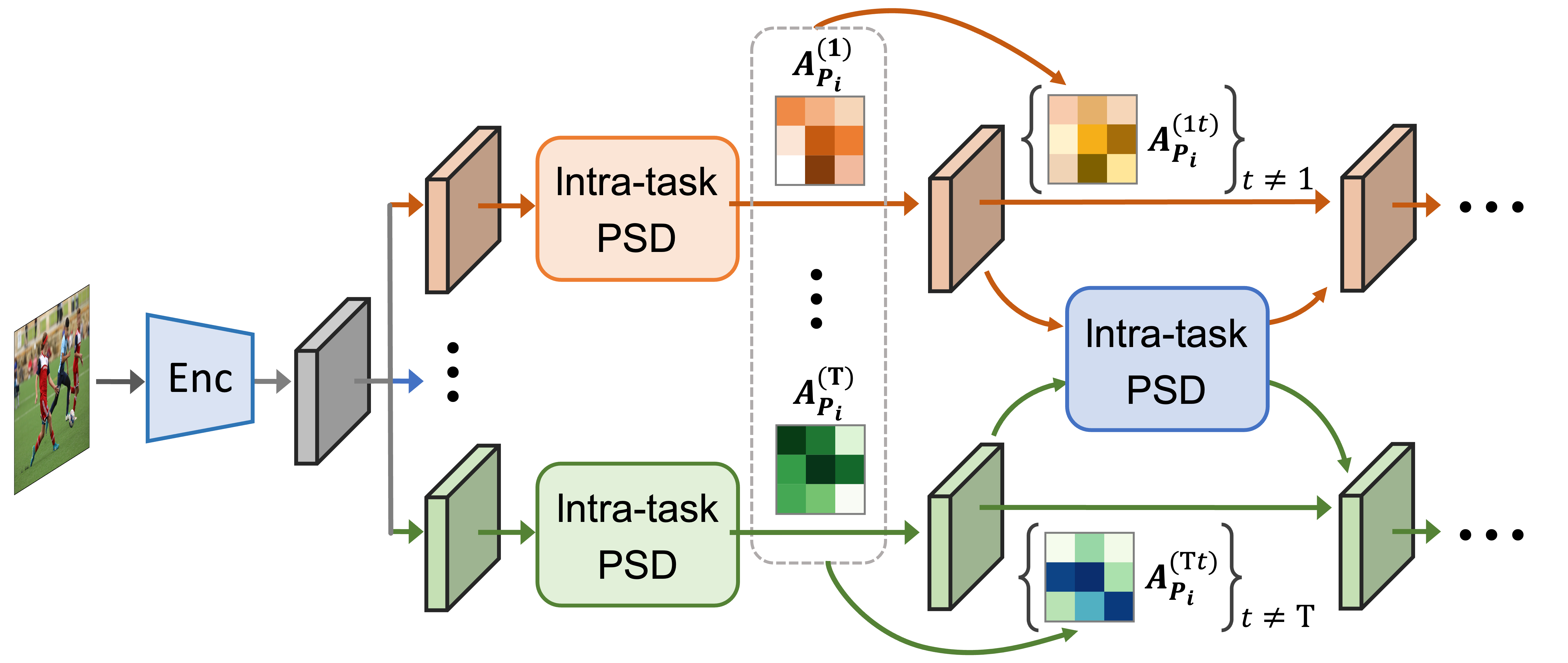}
        \caption{PSD module.}
        \label{psd-module}
    \end{subfigure}\quad
    \begin{subfigure}{0.3\textwidth}
        \includegraphics[width=1\textwidth]{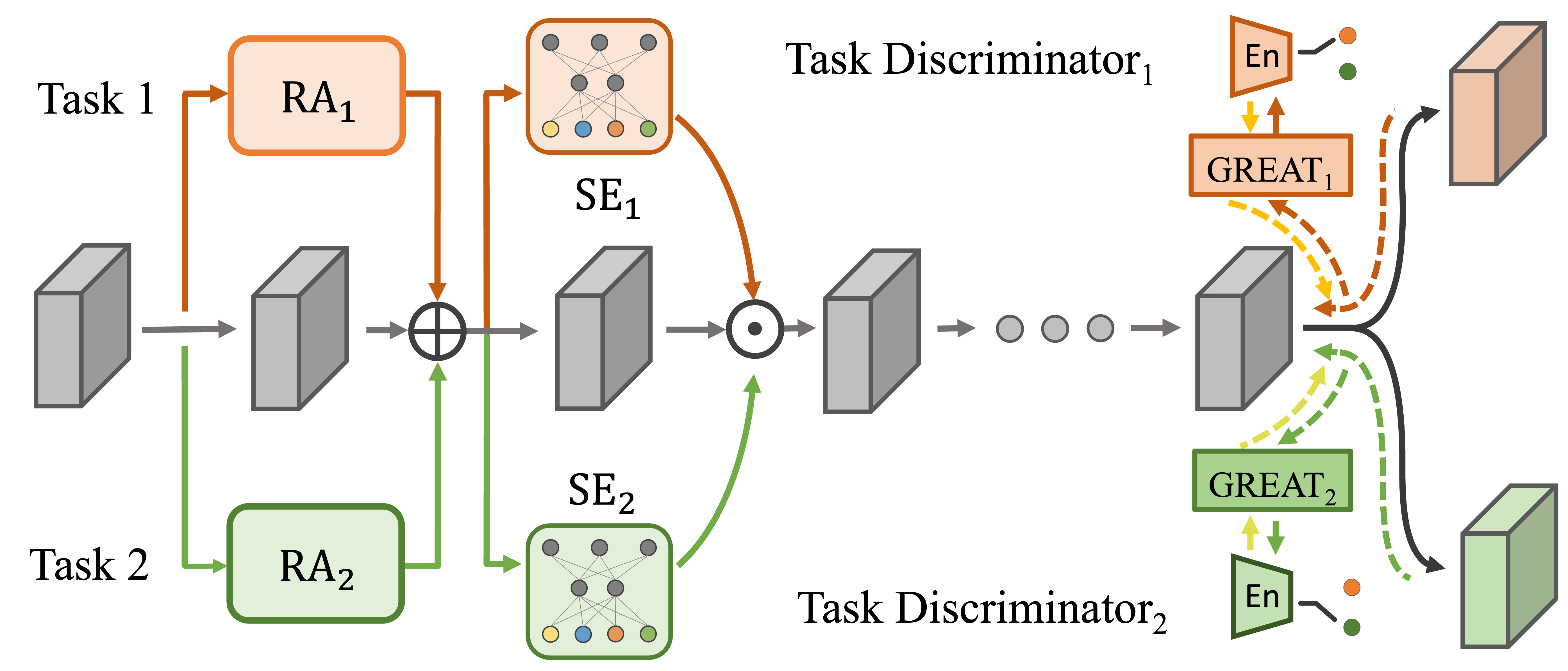}
        \caption{ASTMT.}
        \label{astmt}
    \end{subfigure}\quad
    \begin{subfigure}{0.26\textwidth}
        \includegraphics[width=1\textwidth]{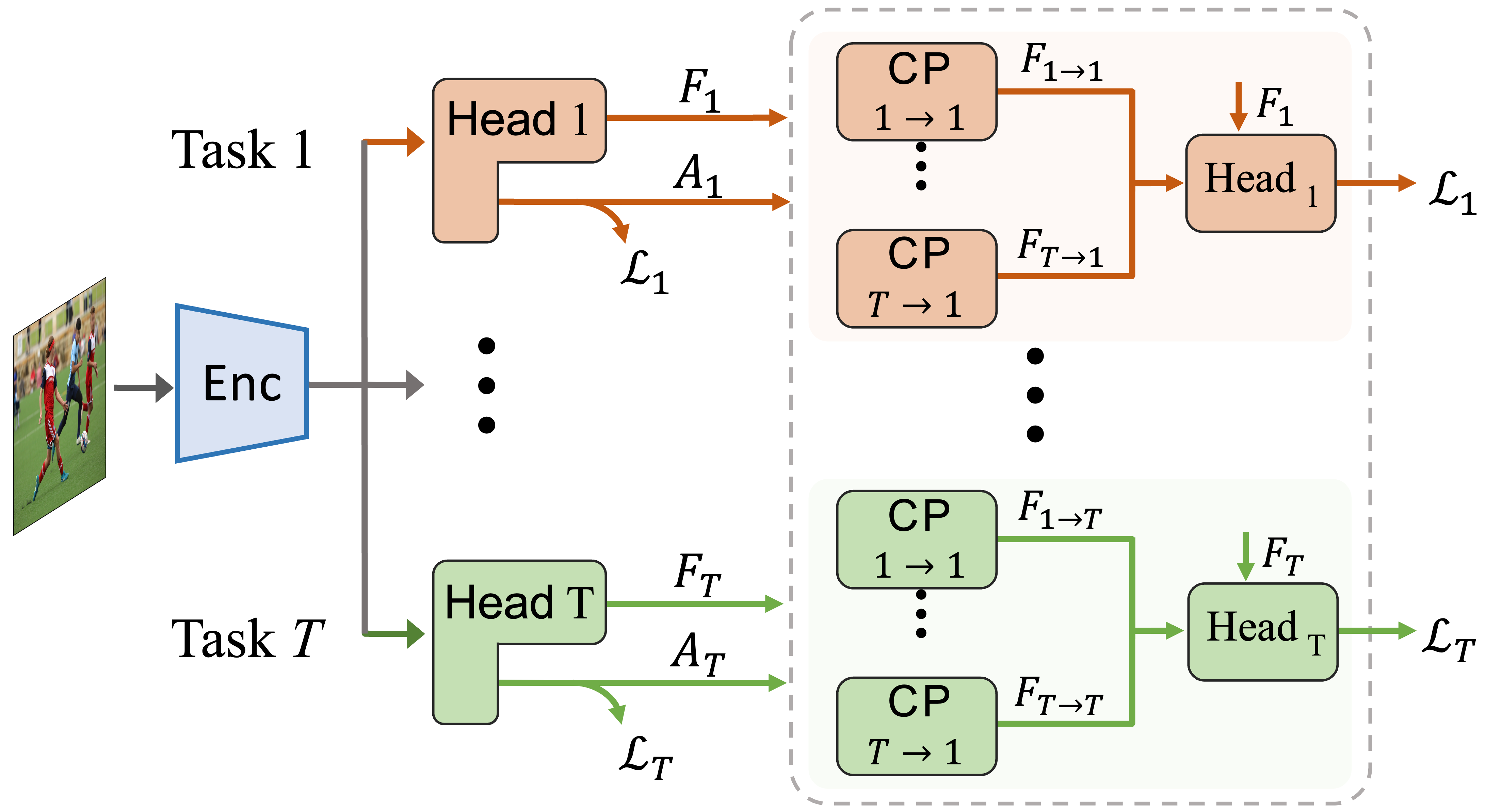}
        \caption{ATRC module.}
        \label{atrc-module}
    \end{subfigure}
    
    \begin{subfigure}{0.33\textwidth}
        \includegraphics[width=1\textwidth]{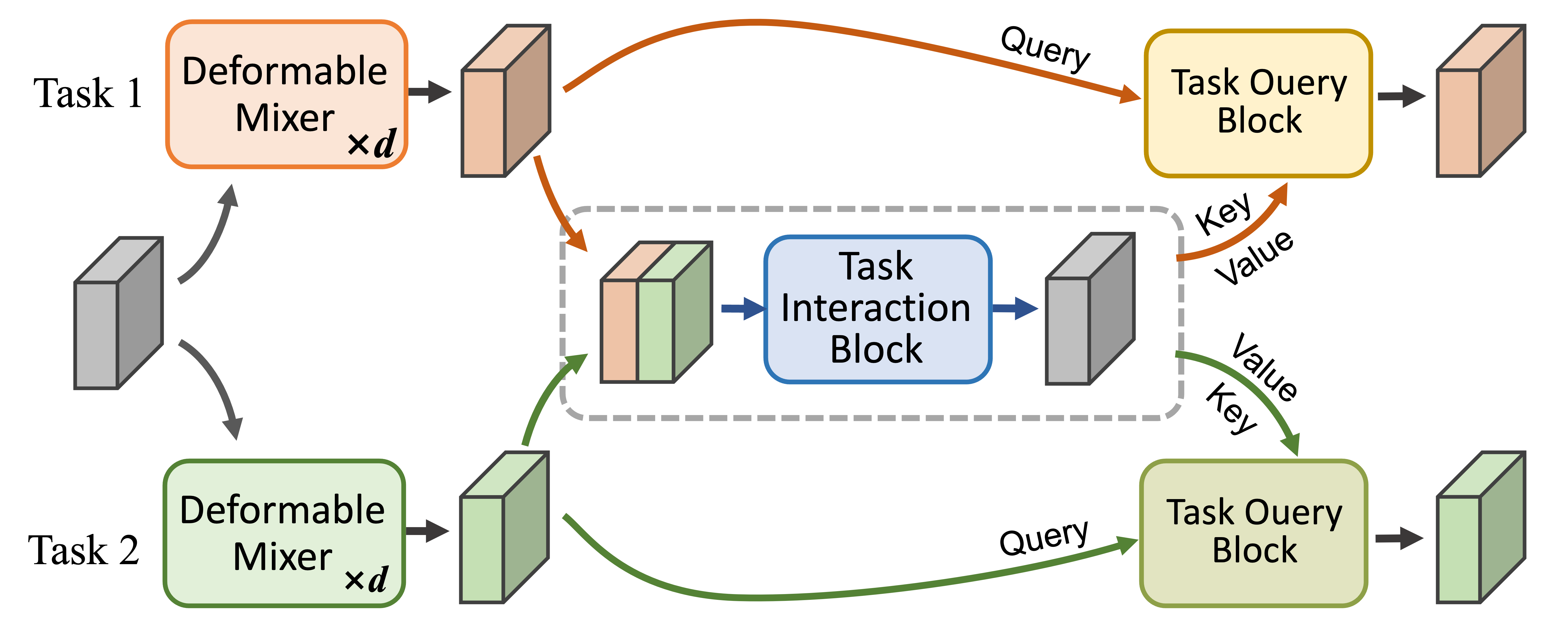}
        \caption{DeMT block.}
        \label{demt-block}
    \end{subfigure}\quad\quad\quad\quad
    \begin{subfigure}{0.33\textwidth}
        \includegraphics[width=1\textwidth]{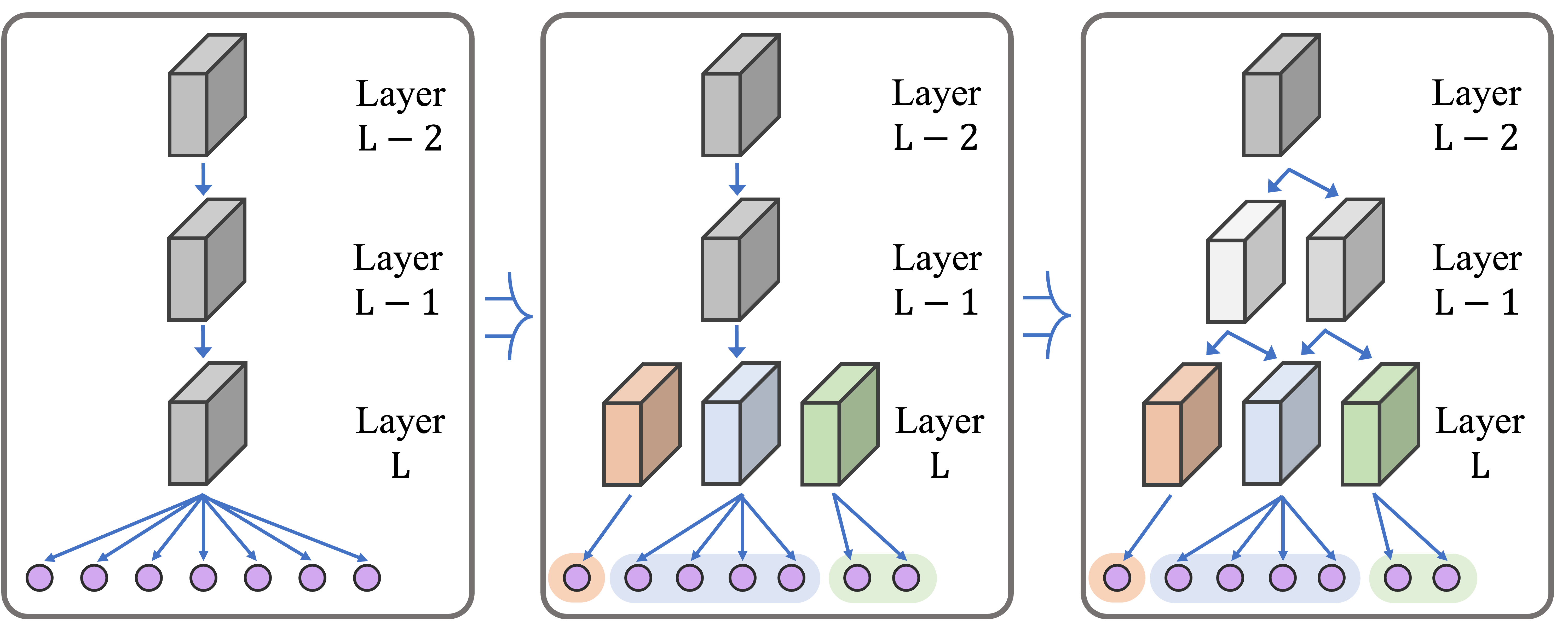}
        \caption{FAFS.}
        \label{fafs}
    \end{subfigure}
    
    \begin{subfigure}{0.33\textwidth}
        \includegraphics[width=1\textwidth]{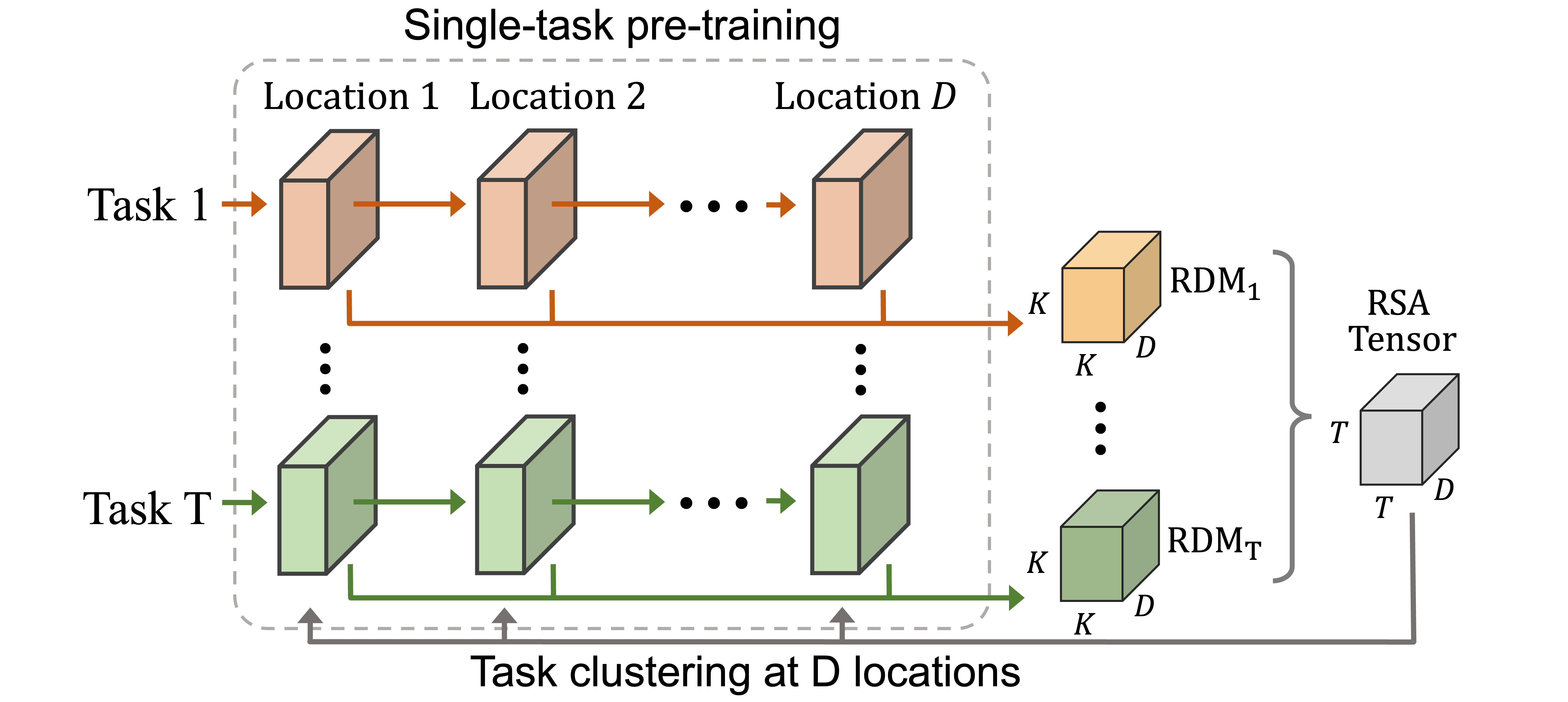}
        \caption{BMTN.}
        \label{bmtn}
    \end{subfigure}\quad\quad\quad\quad
    \begin{subfigure}{0.33\textwidth}
        \includegraphics[width=1\textwidth]{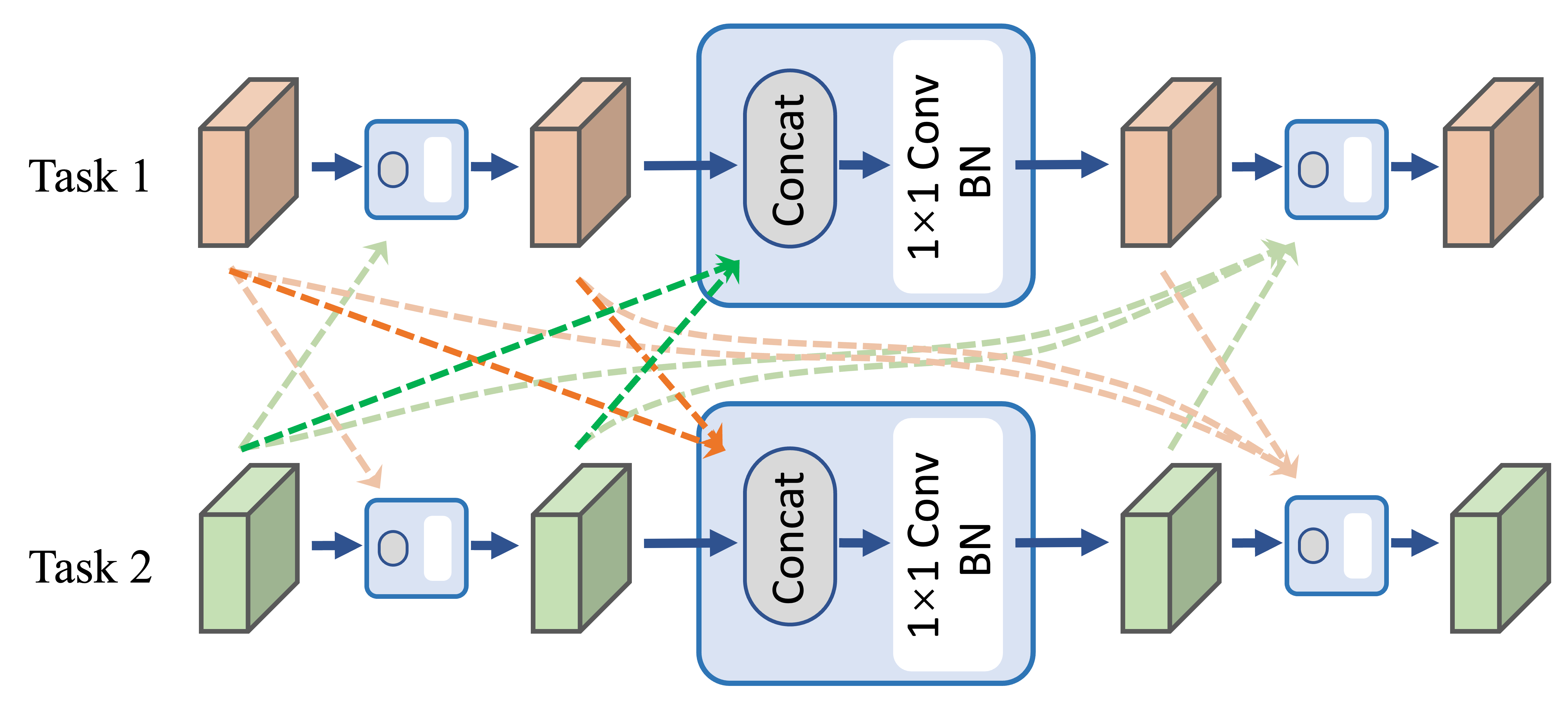}
        \caption{MTL-NAS module.}
        \label{nddr-nas-module}
    \end{subfigure}
    
    \begin{subfigure}{0.22\textwidth}
        \includegraphics[width=1\textwidth]{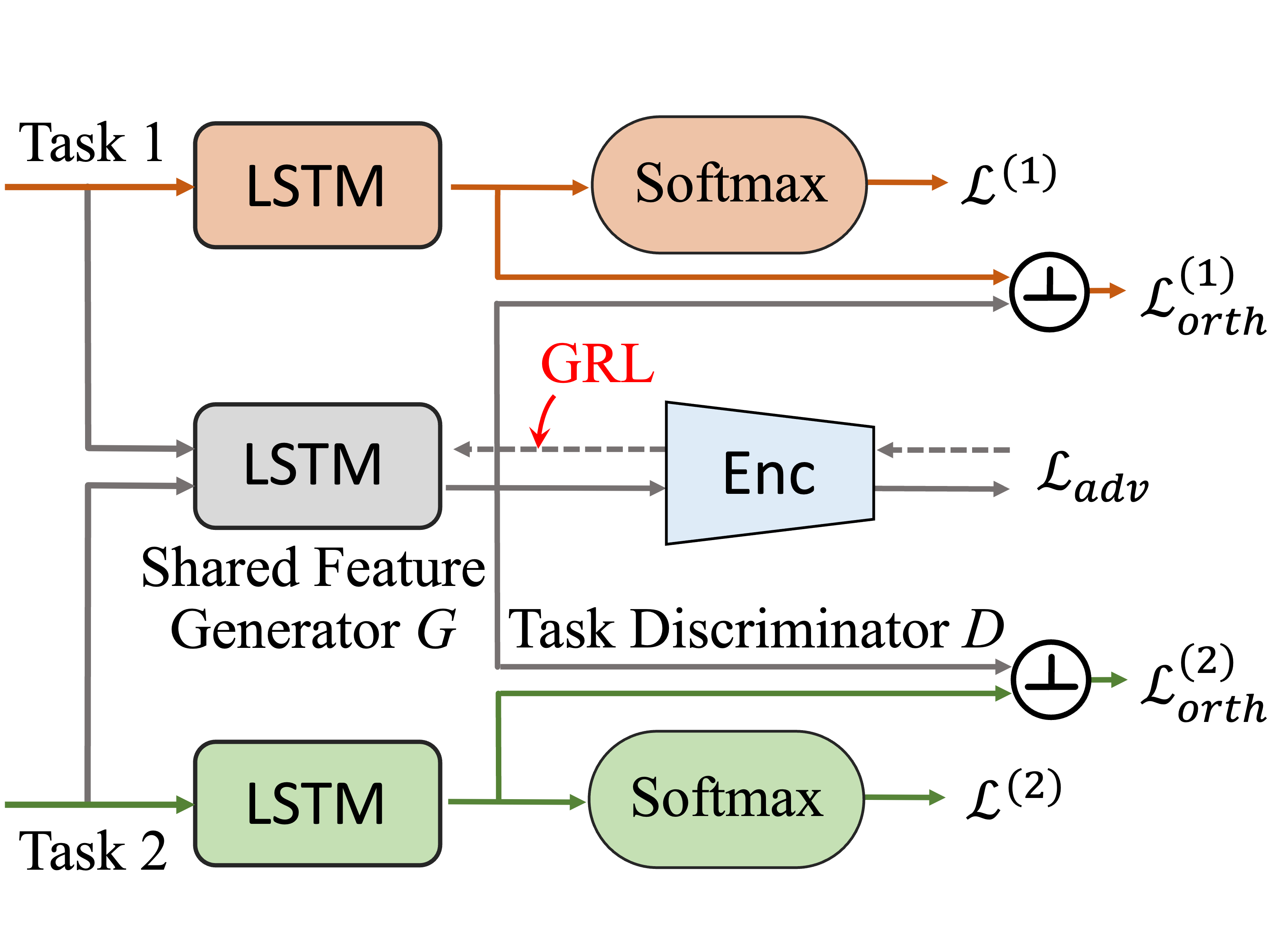}
        \caption{ASP-MTL.}
        \label{asp-mtl}
    \end{subfigure}
    \begin{subfigure}{0.22\textwidth}
        \includegraphics[width=1\textwidth]{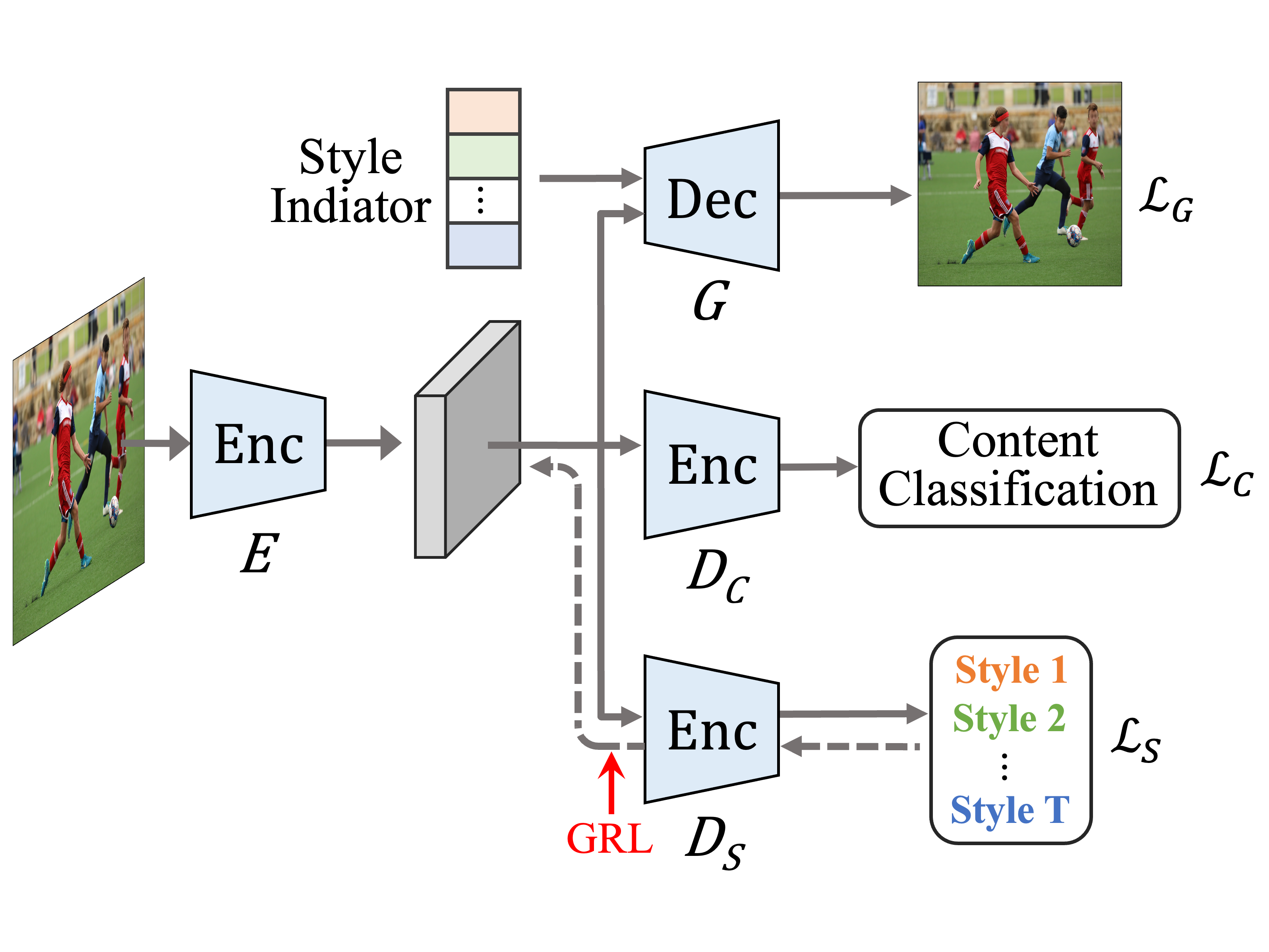}
        \caption{MTA$_{\text{(adv)}}$N.}
        \label{mtadvn}
    \end{subfigure}
    \begin{subfigure}{0.22\textwidth}
        \includegraphics[width=1\textwidth]{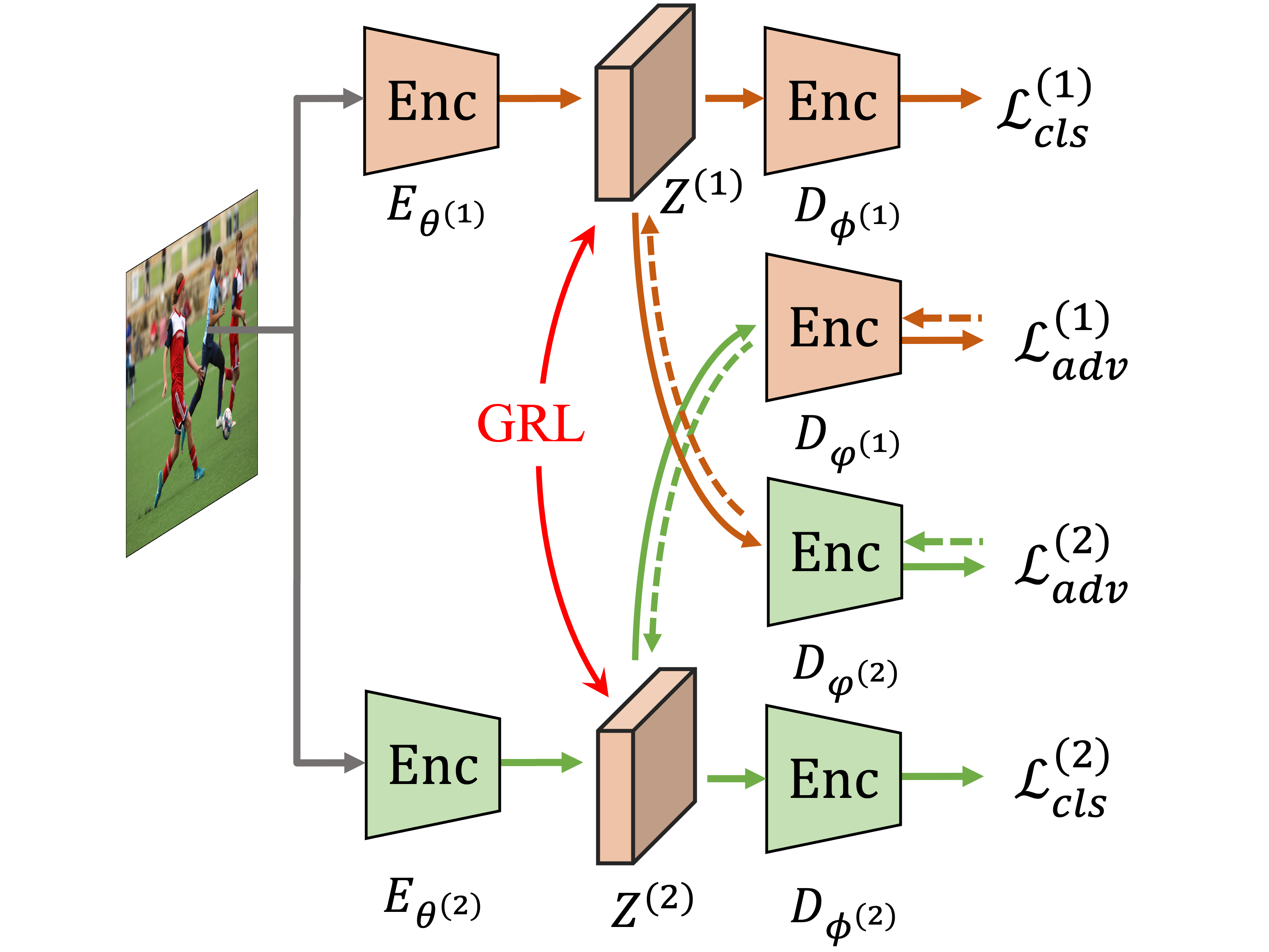}
        \caption{RD4MTL.}
        \label{rd4mtl}
    \end{subfigure}
    \begin{subfigure}{0.22\textwidth}
        \includegraphics[width=1\textwidth]{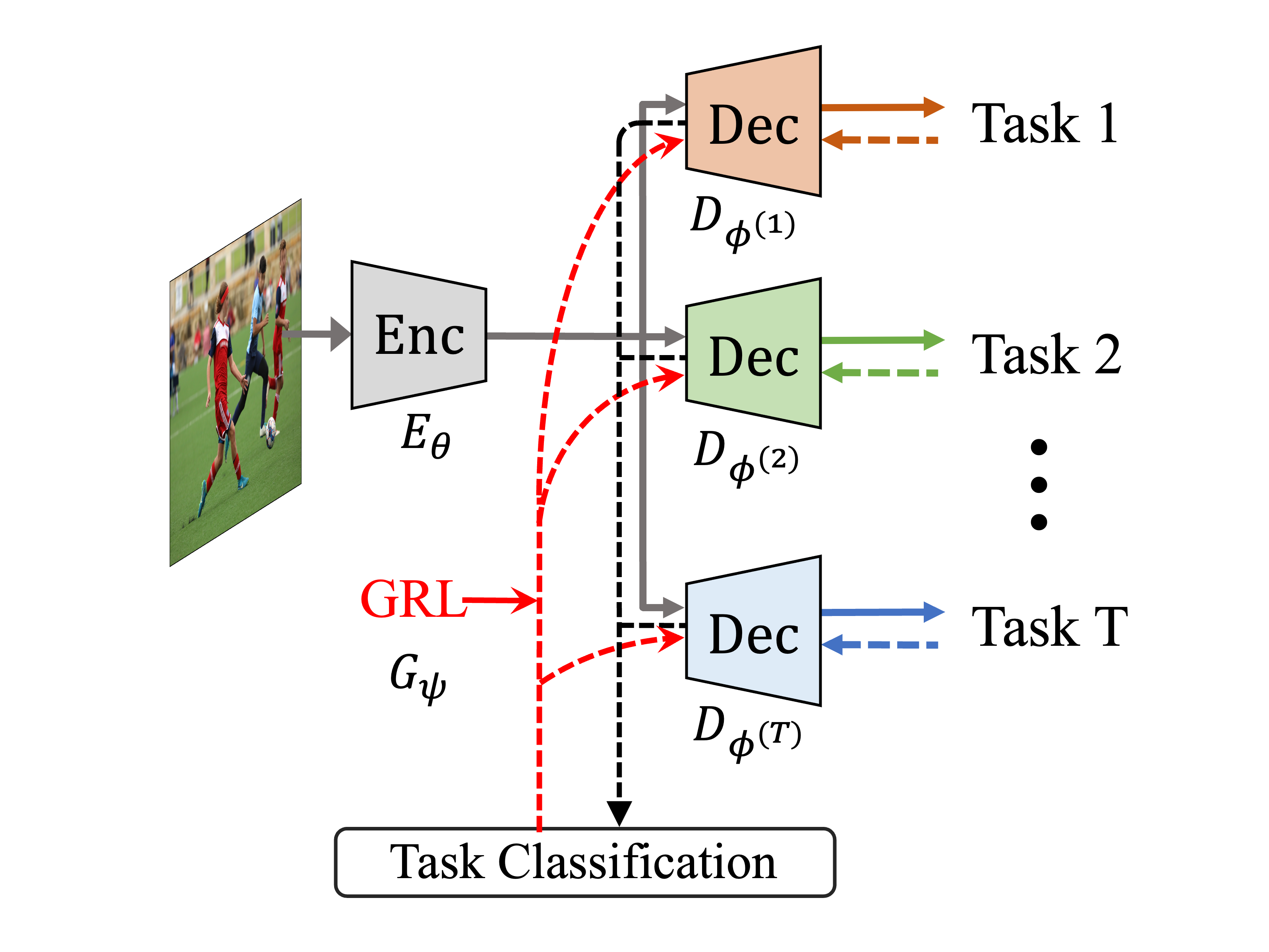}
        \caption{GREAT4MTL.}
        \label{great4mtl}
    \end{subfigure}
    
    \begin{subfigure}{0.38\textwidth}
        \includegraphics[width=1\textwidth]{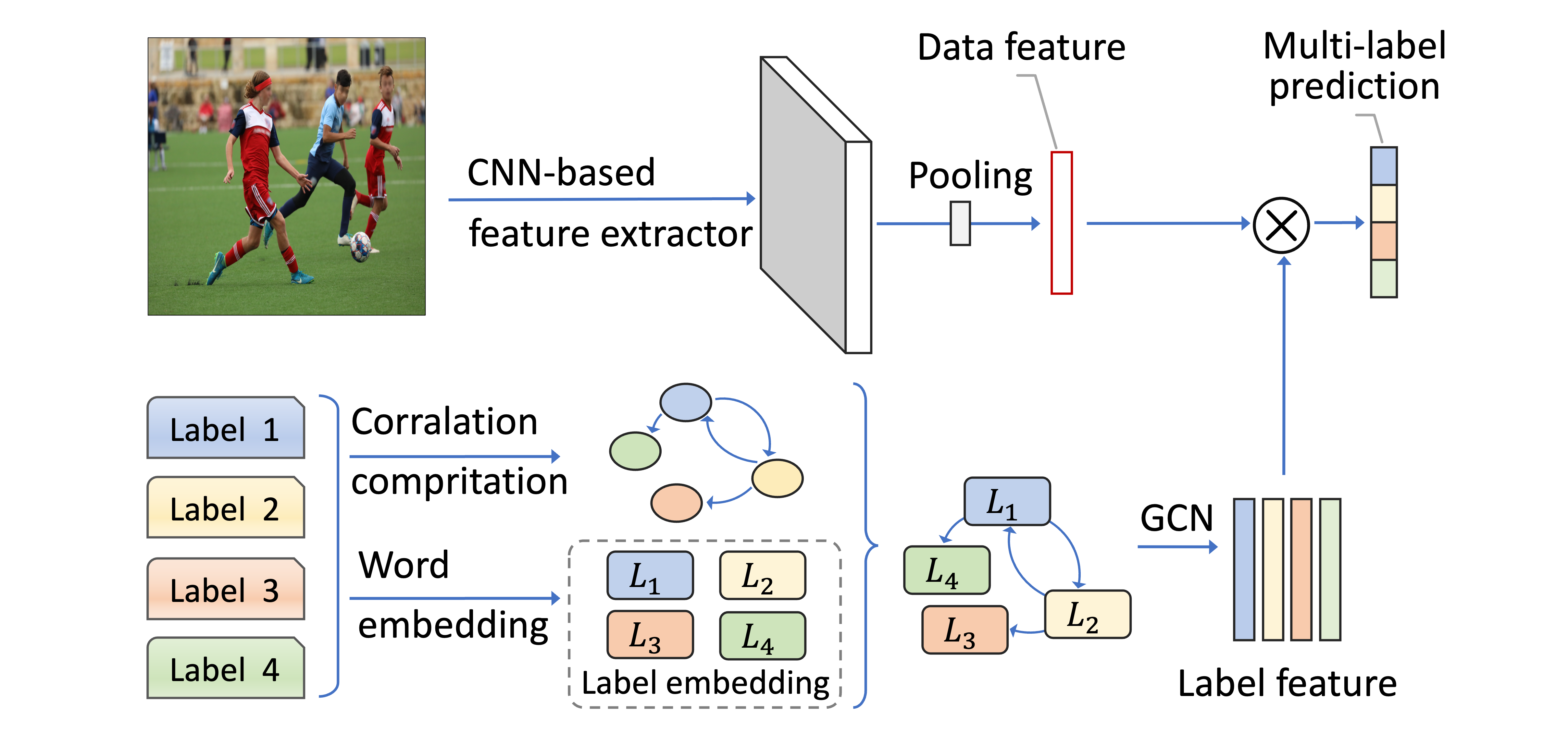}
        \caption{ML-GCN.}
        \label{ml-gcn}
    \end{subfigure}\quad\quad
    \begin{subfigure}{0.4\textwidth}
        \includegraphics[width=1\textwidth]{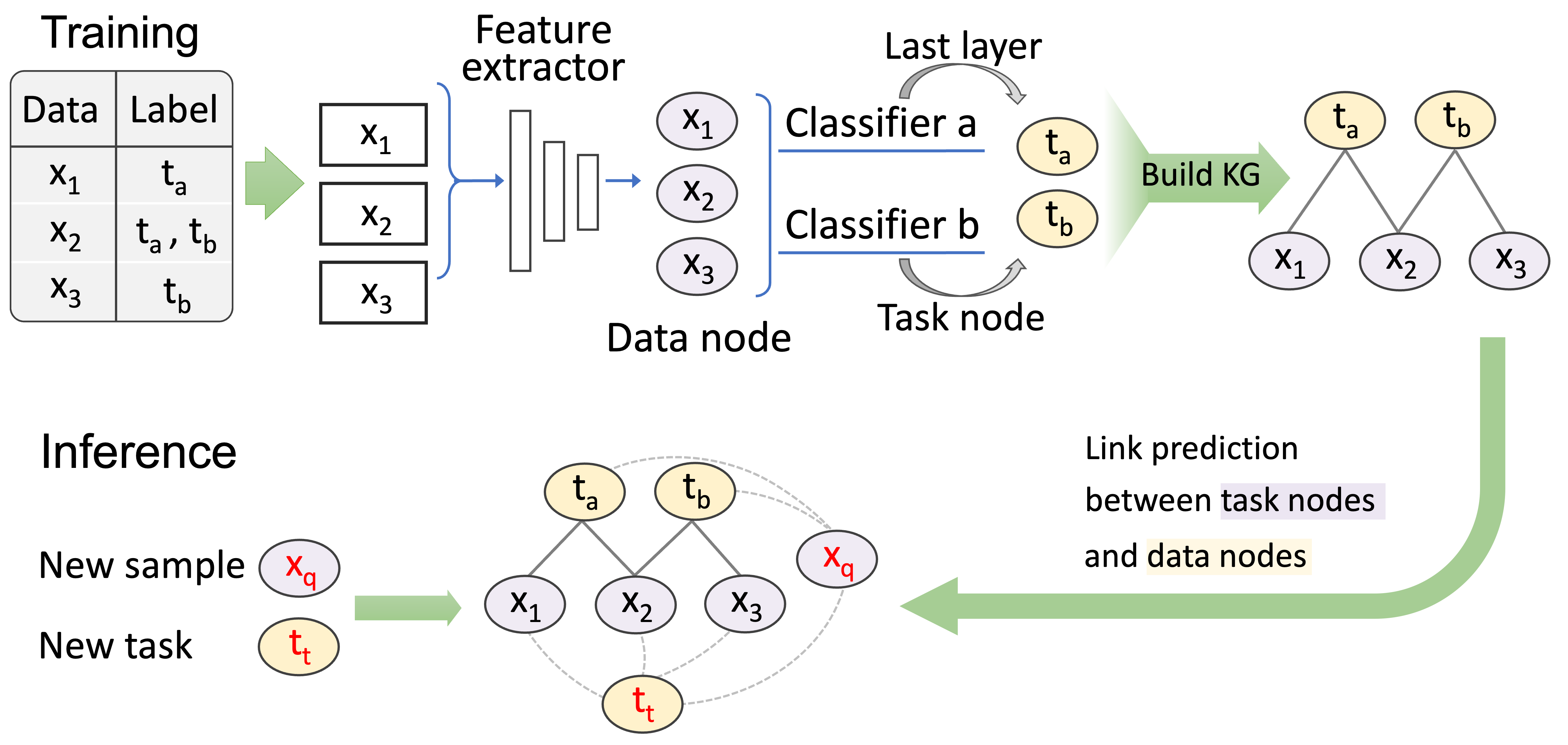}
        \caption{MetaLink.}
        \label{metalink}
    \end{subfigure}
    
  \caption{Frameworks of deep learning techniques used in MTL. (a--d) Feature fusion: cross-stitch networks, Sluice Network, NDDR-CNN, and Soft Order. (e--h) Knowledge distillation: KD4MTL, MuST, OKD-MTL, and CrossDistill. (i--p) Attention: PAD, MTAN, MTI-Net, PAP, PSD, ASTMT, ATRC, and DeMT. (q--s) NAS: FAFS, BMTN, and MTL-NAS. (t-w) Adversarial MTL: ASP-MTL, MTAN, RD4MTL, and AAMTRL. (x-y) Graph: ML-GCN and MetaLink.}
  \label{fig:deep_all}
\end{figure}


%% file: tex_files/02-2/hard-soft.tex
\begin{figure*}[t]
    \centering

    \begin{subfigure}{0.3\textwidth}
        \includegraphics[width=1\textwidth]{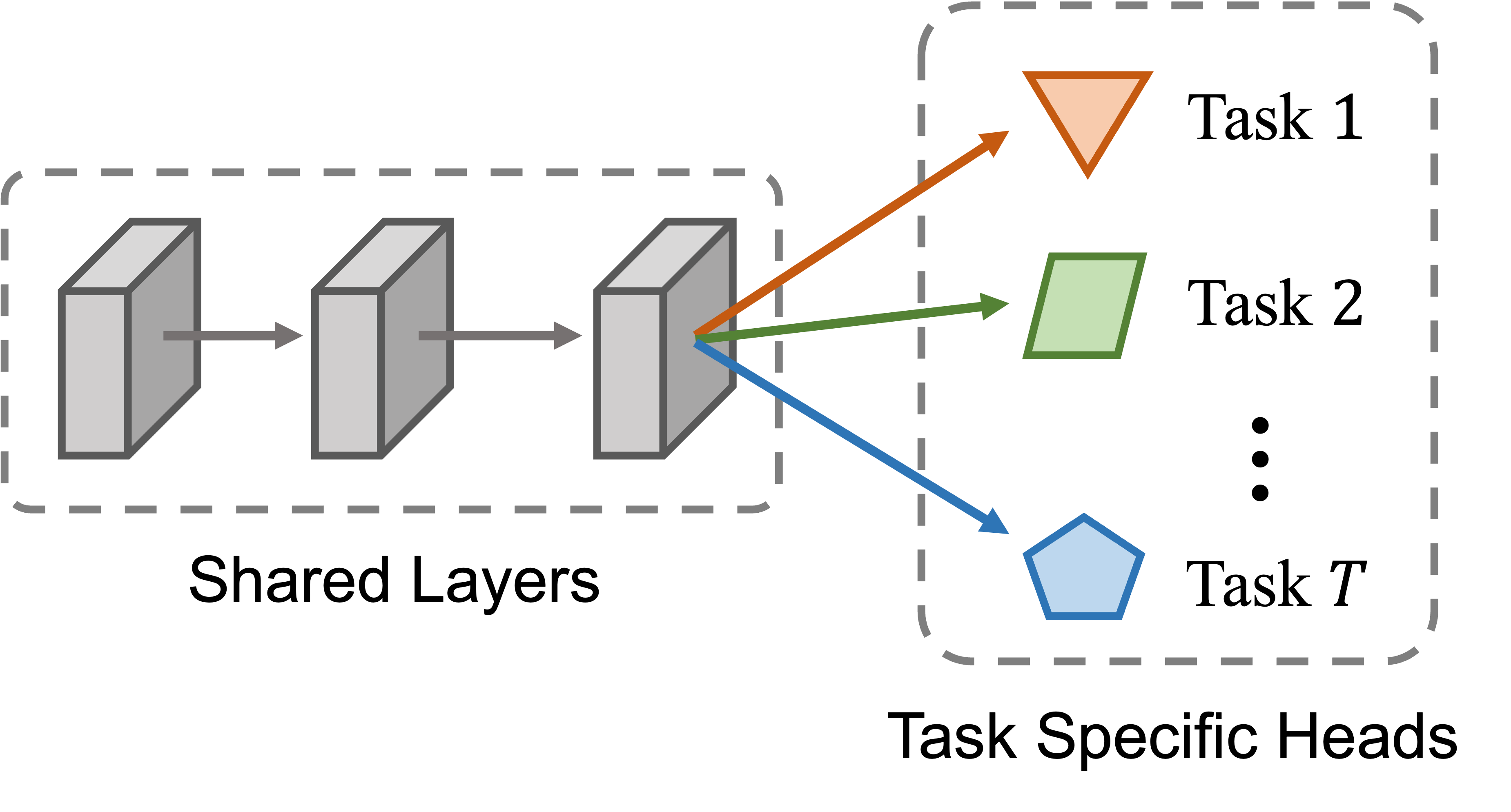}
        \caption{Hard sharing.}
        \label{hard}
    \end{subfigure}\quad
    \begin{subfigure}{0.3\textwidth}
        \includegraphics[width=1\textwidth]{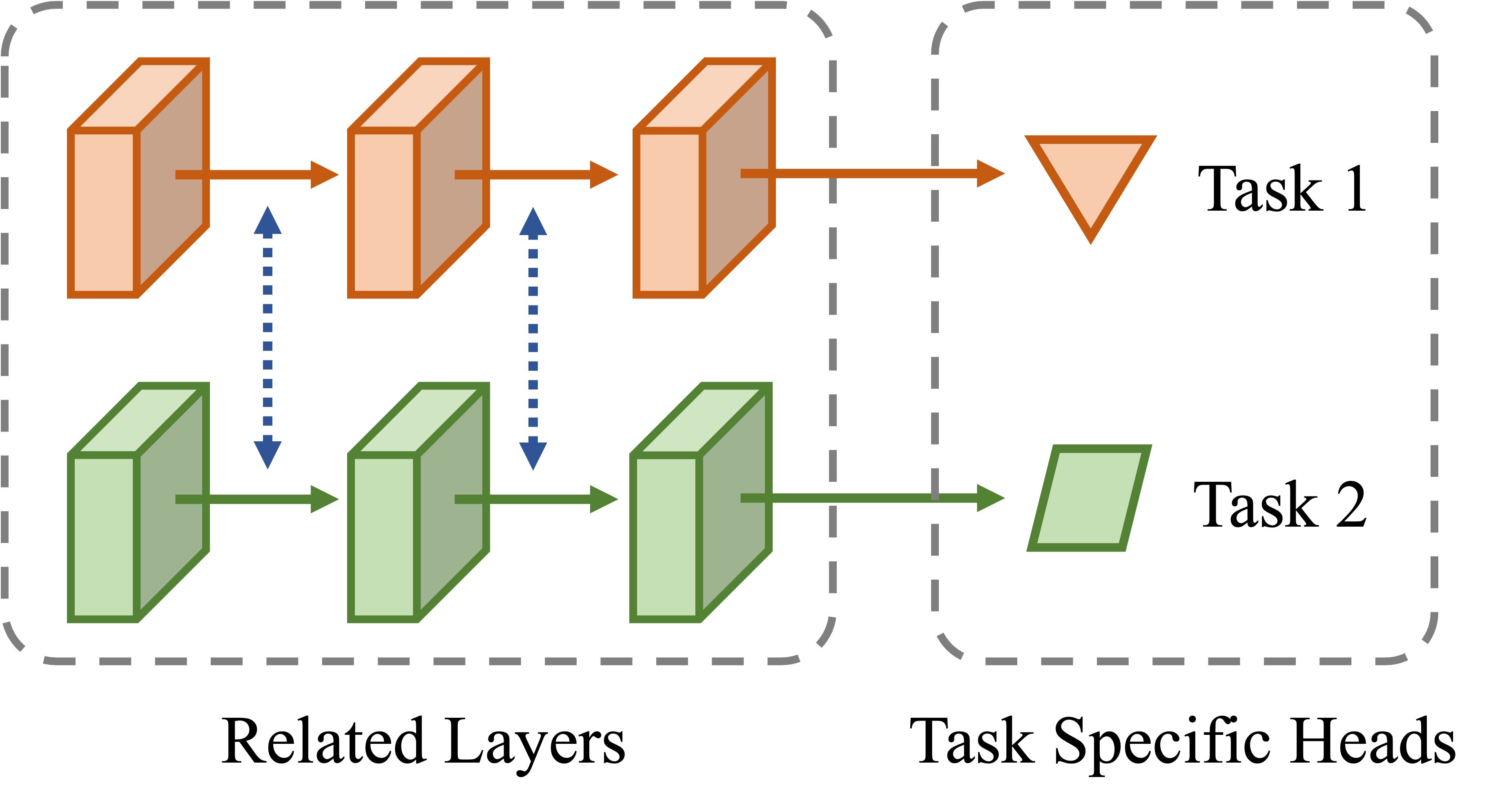}
        \caption{Soft sharing.}
        \label{soft}
    \end{subfigure}\quad
    \begin{subfigure}{0.3\textwidth}
        \includegraphics[width=1\textwidth]{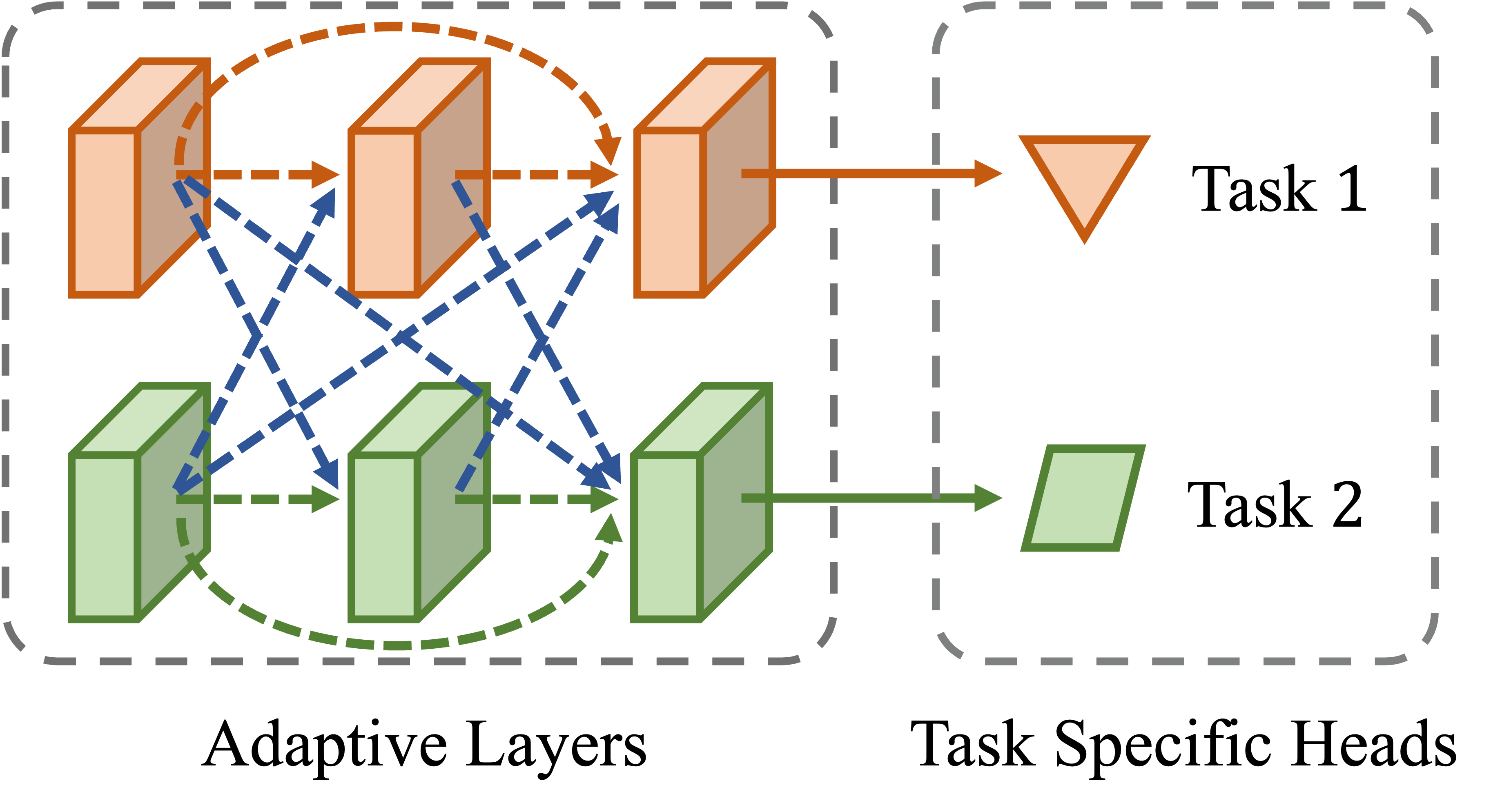}
        \caption{Adaptive sharing.}
        \label{adaptive}
    \end{subfigure}
    \caption{Architecture taxonomy proposed by~\citet{ruder2017overview} for deep multi-task sharing: (a) Hard parameter sharing, (b) soft parameter sharing, and (c) adaptive sharing. The 1D arrows indicate computations within the neural networks involving learnable parameters. The 2D shapes and 3D cubes represent the final responses and extracted features, respectively.}
    \label{hard_soft}
\end{figure*}


%% file: table_files/notation_deep.tex
\begin{table*}[t]
    \centering
    \tiny
    \caption{Summary of notations used in Sec.~\myref{deep-era}.}
    \midsepremove
\scalebox{0.75}{
    \begin{tabular}{ll}
    \toprule
    \rowcolor{gray!40}
        Notation & Description \\

    \midrule
        $b, B$ & Batch size. \\

\rowcolor{gray!20}
    \midrule
        $lr$ & Learning rate. \\
        
    \midrule
        $\Xmcal_l^t\in \mathbb{R}^{(B\times )H\times W\times C}$ & Feature maps output from $l$-th layer of $t$-th task, where $(B, )H, W, C$ are (batch size,) \#height, \#width, and \#channel. \\

\rowcolor{gray!20}
    \midrule
        $\Wmcal\in \mathbb{R}^{S\times S\times C_{\text{in}}\times C_{\text{out}}}$ & Convolution filter, where $S$ denotes the size of filter, and $C_{\text{in}}, C_{\text{out}}$ denote the number of input and output channels, respectively.\\

     \midrule
        $\text{exp}(\cdot)$ & Exponential function. \\

\rowcolor{gray!20}
    \midrule
        $\sigma(\cdot)$ & Sigmoid function, where $\sigma(x) = 1/(1+\text{exp}(-x))$. \\

    \midrule
        $\text{softmax}(\cdot)$ & Softmax function, where $[\text{softmax}(\xbold)]_j=\text{exp}(x_j)/\sum_i\text{exp}(x_i)$ for any entry index $j$. \\

\rowcolor{gray!20}
    \midrule
        $\text{sim}(\cdot,\cdot)$ & An arbitrary similarity function, e.g. cosine similarity cos$(\cdot,\cdot)$.\\
        
    \midrule
        $\odot$ & The element-wise dot product. \\

\rowcolor{gray!20}
    \midrule
        $LN(\cdot)$ & Layer norm. \\

    \midrule
        $MHSA(q, k, v)$ & Multi-head self-attention operator. \\

\rowcolor{gray!20}
    \midrule
        $CONV_{\Wmcal}(\cdot)$ & Convolution operation parametrized by $\Wmcal$. \\
        
    \midrule
        $RESHAPE(\cdot)$ & Reshape operation to rearrange the original feature maps in $\Rmbb^{H\times W\times C}$ space into a new $\Rmbb^{HW\times C}$ space.\\

    \bottomrule
    \end{tabular}
}
    \label{tab:deep_notation}
\end{table*}

%% file: tex_files/02-2/hard-example.tex
\begin{figure*}[t]
    \centering

    \begin{subfigure}{0.46\textwidth}
        \includegraphics[width=1\textwidth]{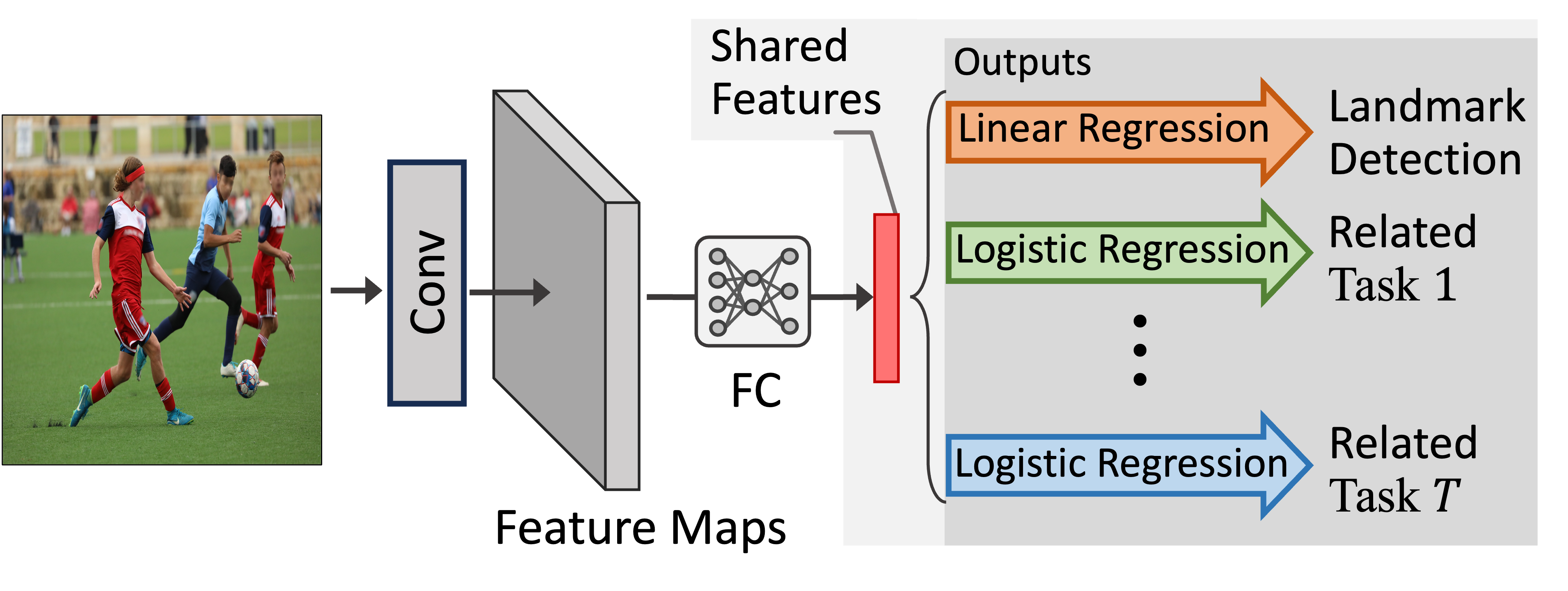}
        \caption{TCDCN.}
        \label{tcdcn}
    \end{subfigure}\quad\quad
    \begin{subfigure}{0.46\textwidth}
        \includegraphics[width=1\textwidth]{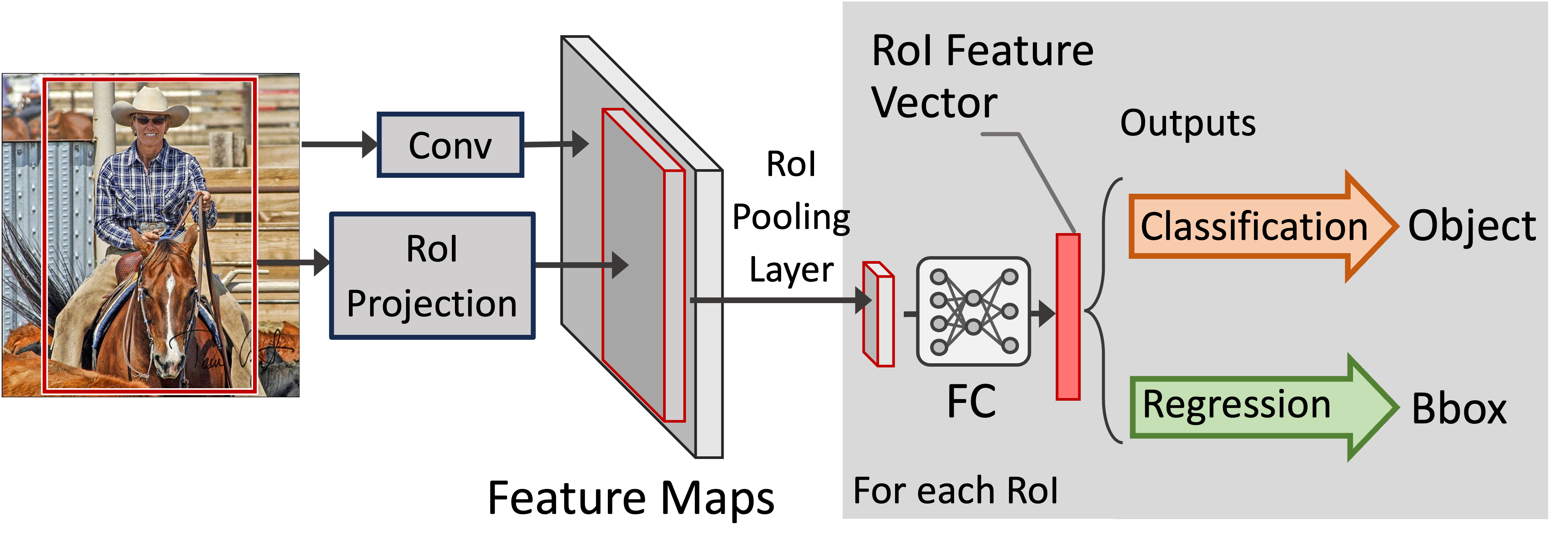}
        \caption{Fast RCNN.}
        \label{fast-rcnn}
    \end{subfigure}
    
    \caption{Two of the earliest applications of hard-parameter sharing in CNNs: (A) the Tasks-Constrained Deep Convolutional Network (TCDCN), which jointly extracts common features from human faces for multiple tasks such as landmark detection, head pose estimation, and facial attribute inference; and (B) the Fast Region-based Convolutional Network method (Fast R-CNN), where each region of interest (RoI) is projected into a fixed-size feature map first and then mapped to a feature vector used for both object probability prediction and bounding-box offsets regression.}
    \label{fig:hard-example}
\end{figure*}

%% file: tex_files/02-2/cascading_framework.tex
\begin{figure*}[t]
    \centering

    \begin{subfigure}{0.475\textwidth}
        \includegraphics[width=1\textwidth]{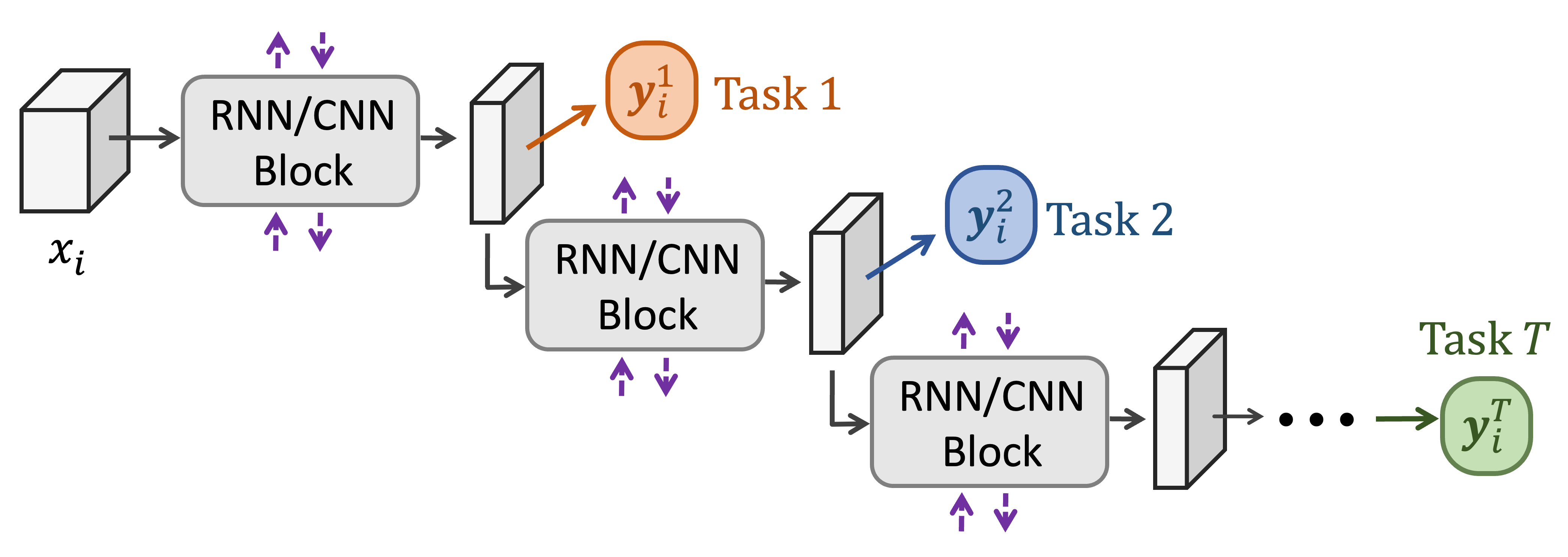}
        \caption{Vanilla Cascading.}
        \label{cascade_1}
    \end{subfigure}\quad
    \begin{subfigure}{0.47\textwidth}
        \includegraphics[width=1\textwidth]{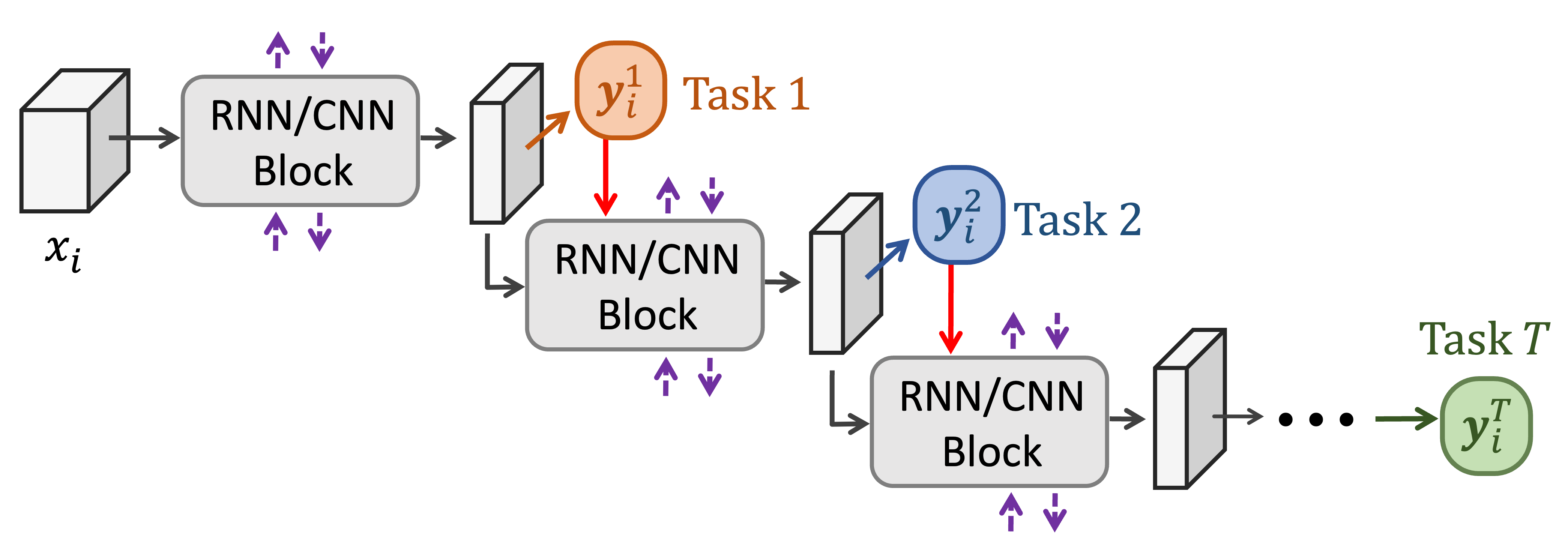}
        \caption{Cascading with Prediction Shortcuts.}
        \label{cascade_2}
    \end{subfigure}

    \begin{subfigure}{0.47\textwidth}
        \includegraphics[width=1\textwidth]{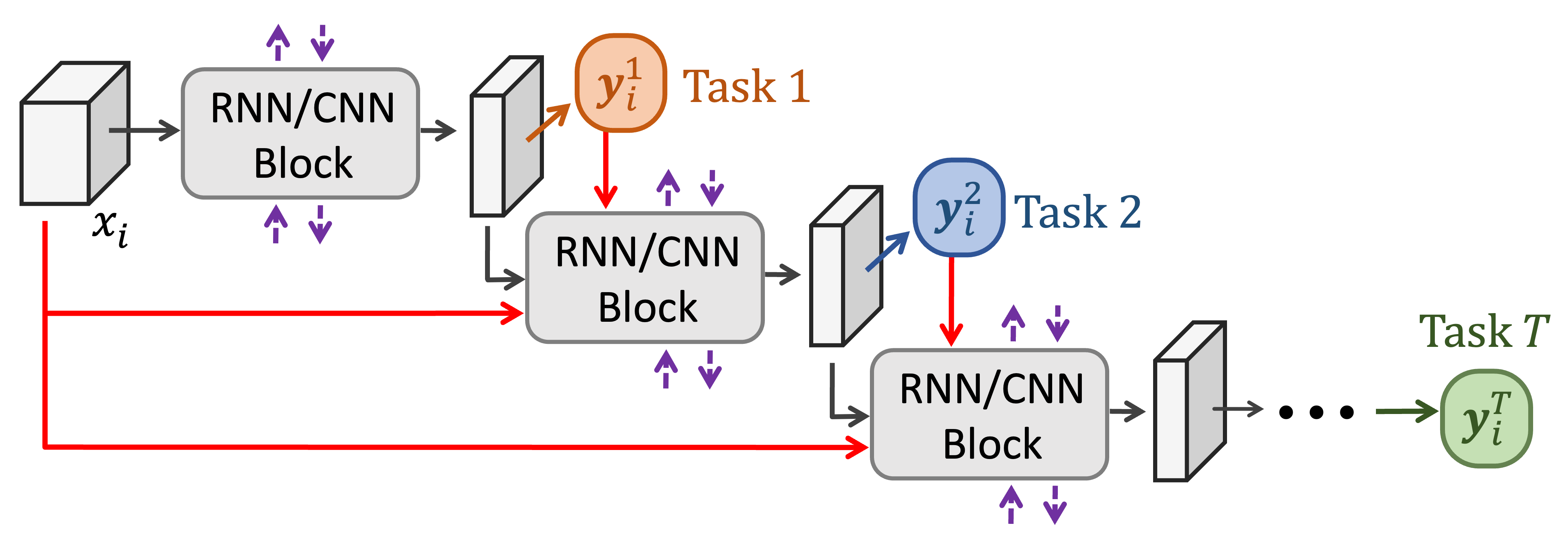}
        \caption{Cascading with Prediction and Feature Shortcuts.}
        \label{cascade_3}
    \end{subfigure}\quad
    \begin{subfigure}{0.47\textwidth}
        \includegraphics[width=1\textwidth]{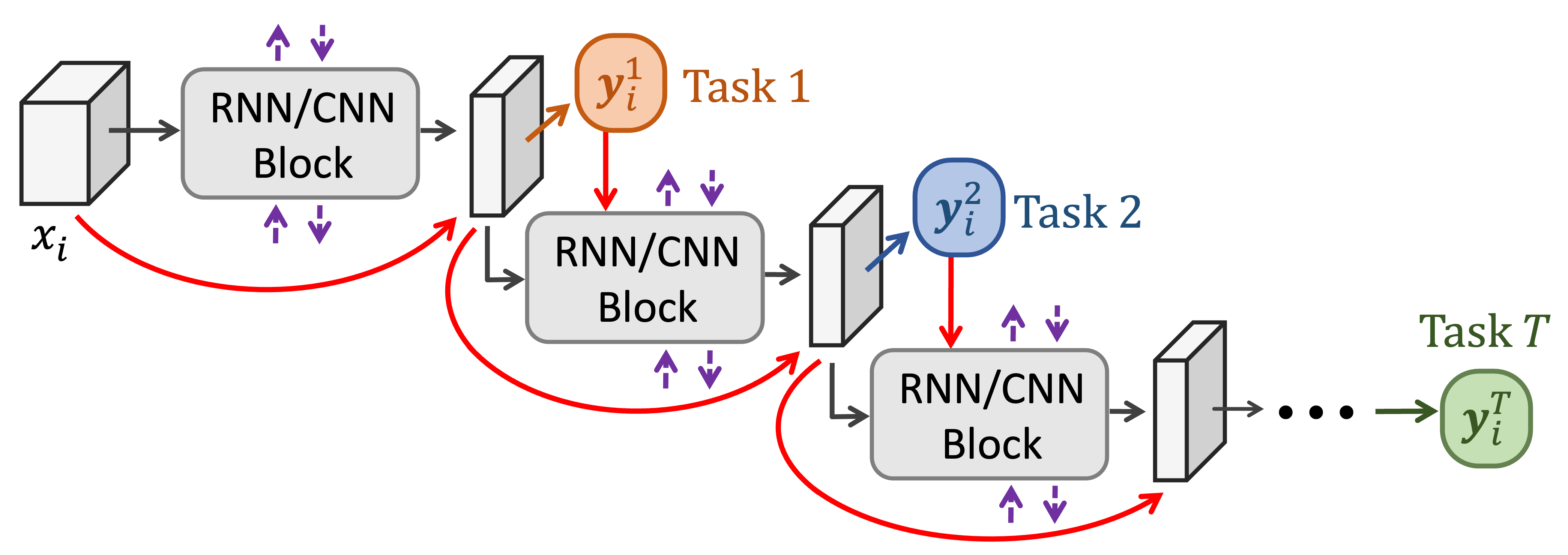}
        \caption{Cascading with Prediction and Residual Shortcuts.}
        \label{cascade_4}
    \end{subfigure}
    
    \caption{The taxonomy of cascading structures into four categories: (A) the vanilla cascading structure, (B) the cascading structure with prediction shortcuts, (C) the cascading structure with prediction and feature shortcuts, and (D) the cascading structure with prediction and residual shortcuts.}
    \label{cascading_taxonomy}
\end{figure*}

%% file: tex_files/02-2/context-pooling.tex
\begin{wrapfigure}[14]{r}{4cm}
    \centering
    \includegraphics[width=1\linewidth]{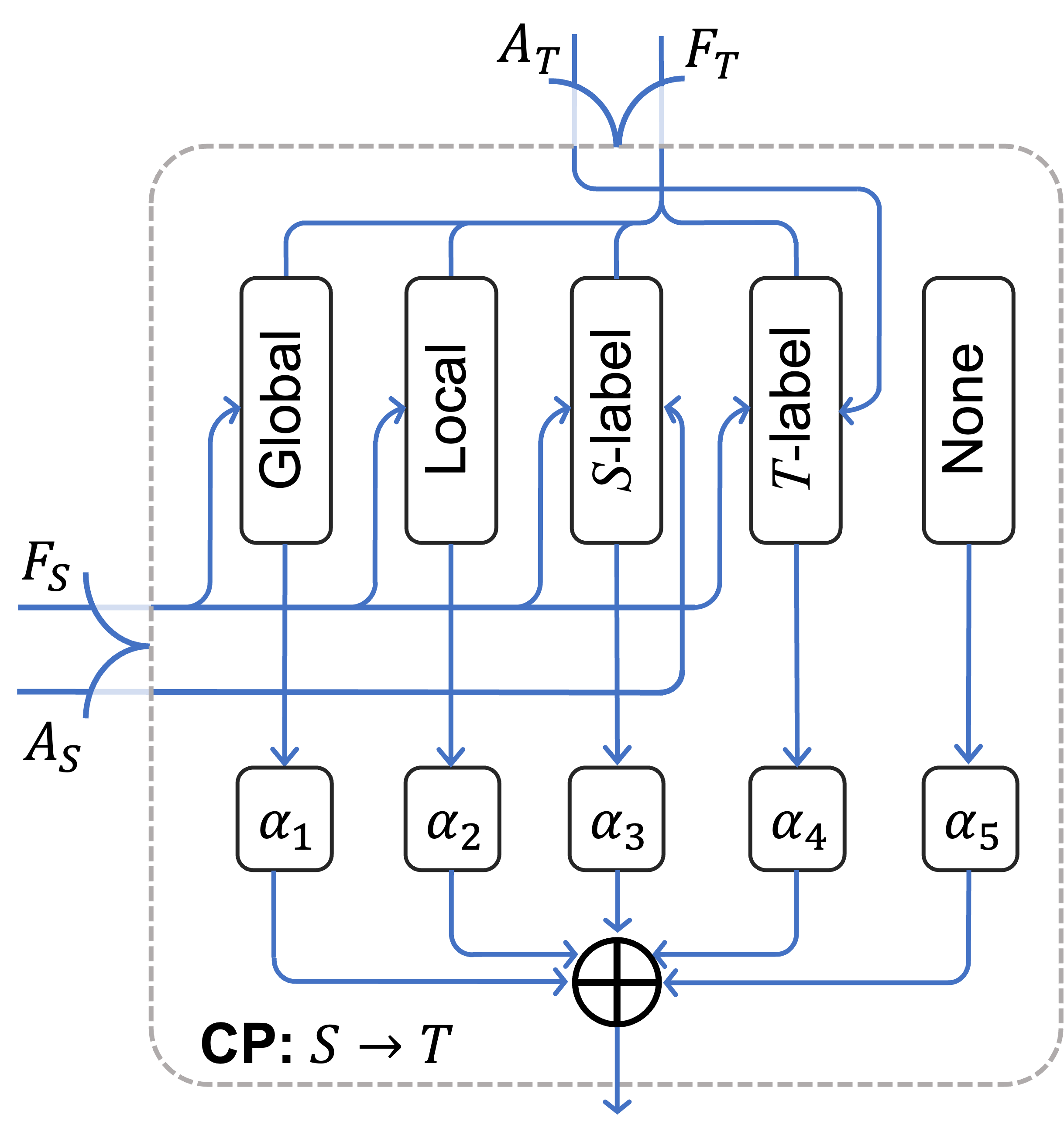}
    \caption{The computational details of Context Pooling (CP).}
    \label{context_pooling}
\end{wrapfigure}

%% file: tex_files/02-2/gradients_issues.tex
\begin{figure}[!htb]
    \centering
    \begin{subfigure}{0.3\textwidth}
        \includegraphics[width=1\textwidth]{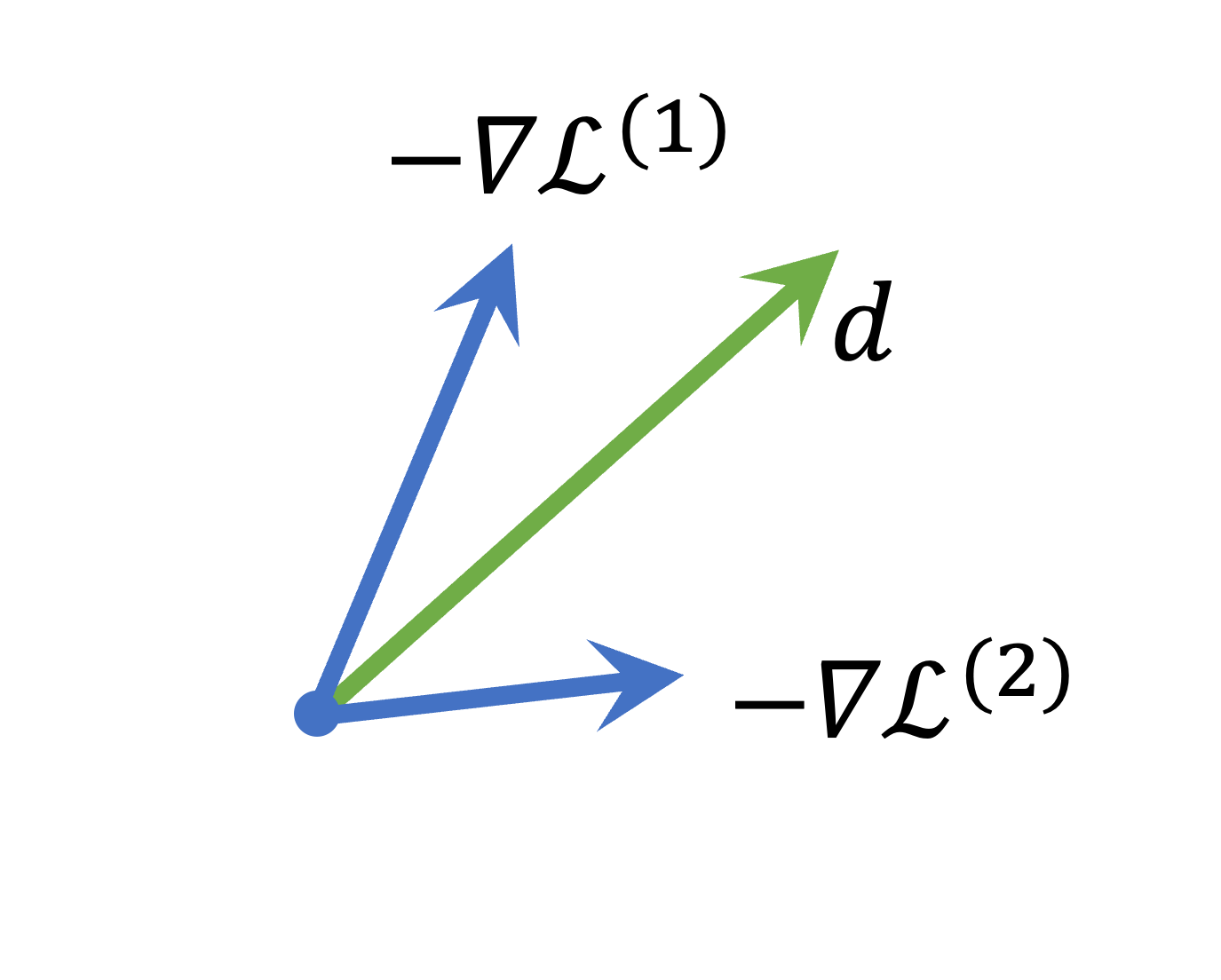}
        \caption{Dominant Gradients Issue.}
        \label{conflicting}
    \end{subfigure} \quad\quad\quad\quad
    \begin{subfigure}{0.28\textwidth}
        \includegraphics[width=1\textwidth]{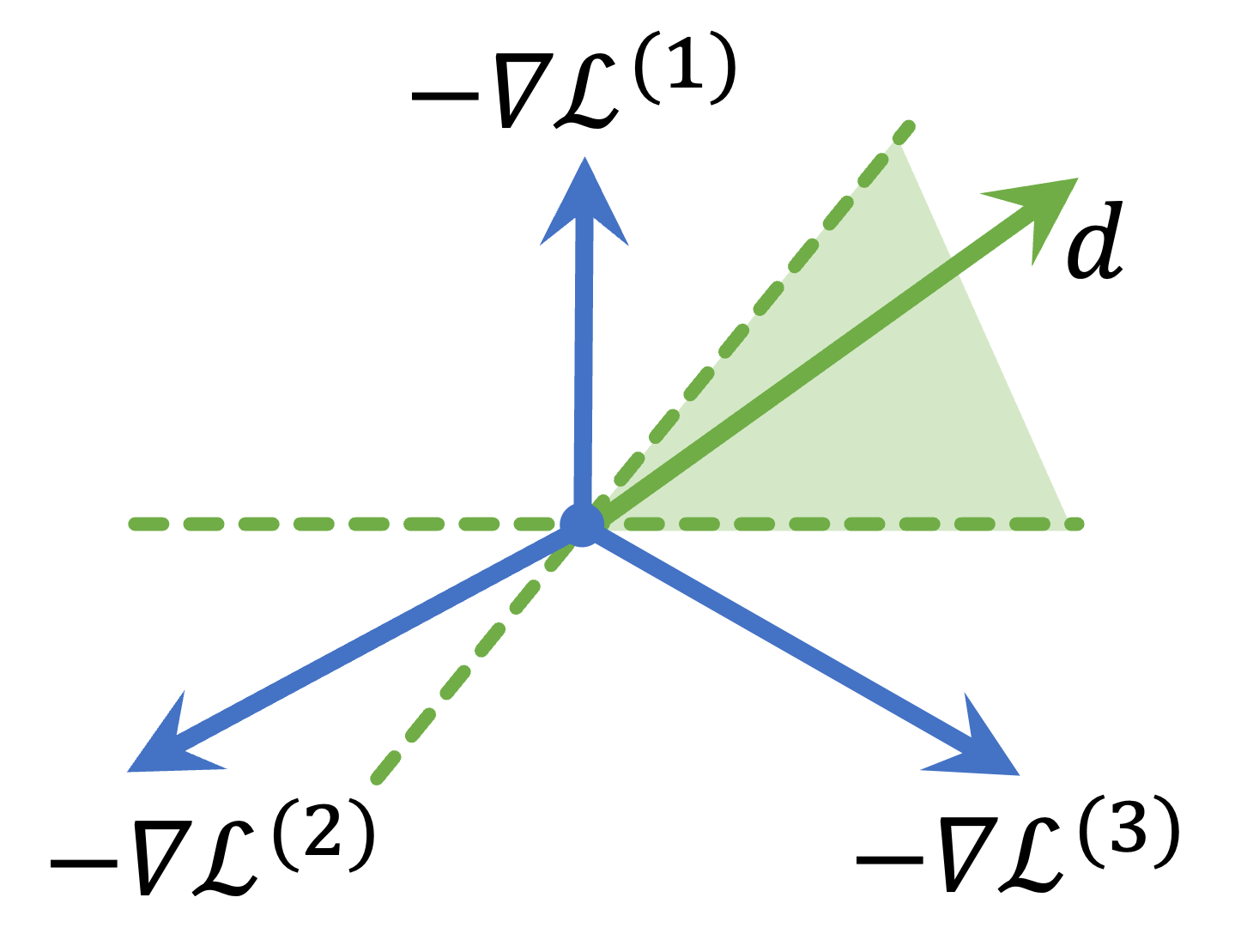}
        \caption{Conflicting Gradients Issue.}
        \label{non-conflicting}
    \end{subfigure}
    \caption{(a)\text{dominant gradients} issue. The update direction $\dbold$ is dominated by the negative gradient of the loss of task 1. (b)\text{conflicting gradients} issue. When $\{\alpha^{(t)}\}_{t=1}^{3}$ are not properly set, the update direction $\dbold$ can decreases the loss of task 1 and 3 while increases the loss of task 2. Therefore, the pefromance on the task 2 is compromised.}
    \label{fig:mtl.gradient.issues}
\end{figure}

%% file: tex_files/02-2/PCGrad.tex
\begin{figure}[ht]
    \centering
    \begin{subfigure}{0.23\textwidth}
        \includegraphics[width=1\textwidth]{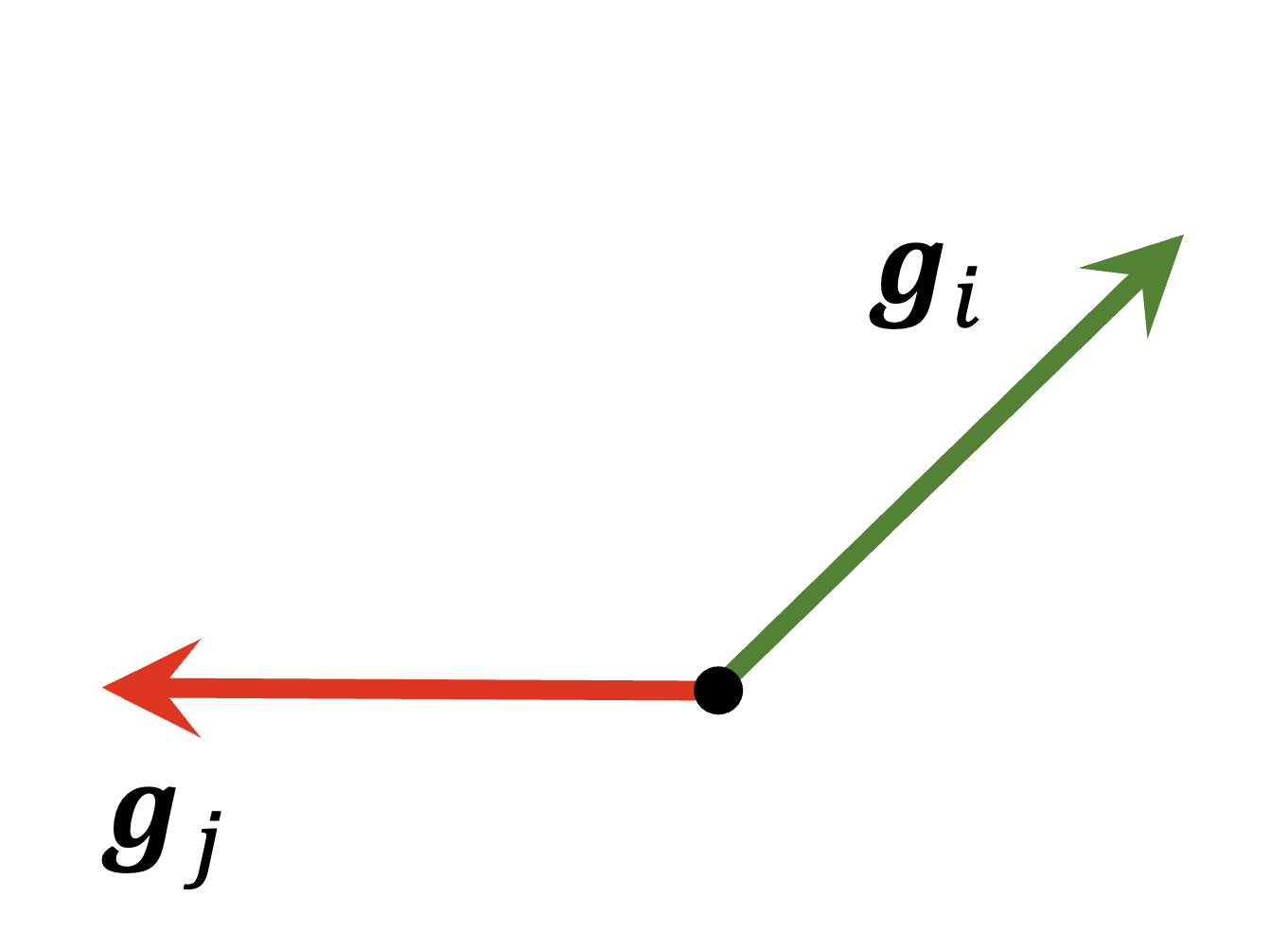}
        \caption{Conflicting.}
        \label{conflicting}
    \end{subfigure}
    \begin{subfigure}{0.23\textwidth}
        \includegraphics[width=1\textwidth]{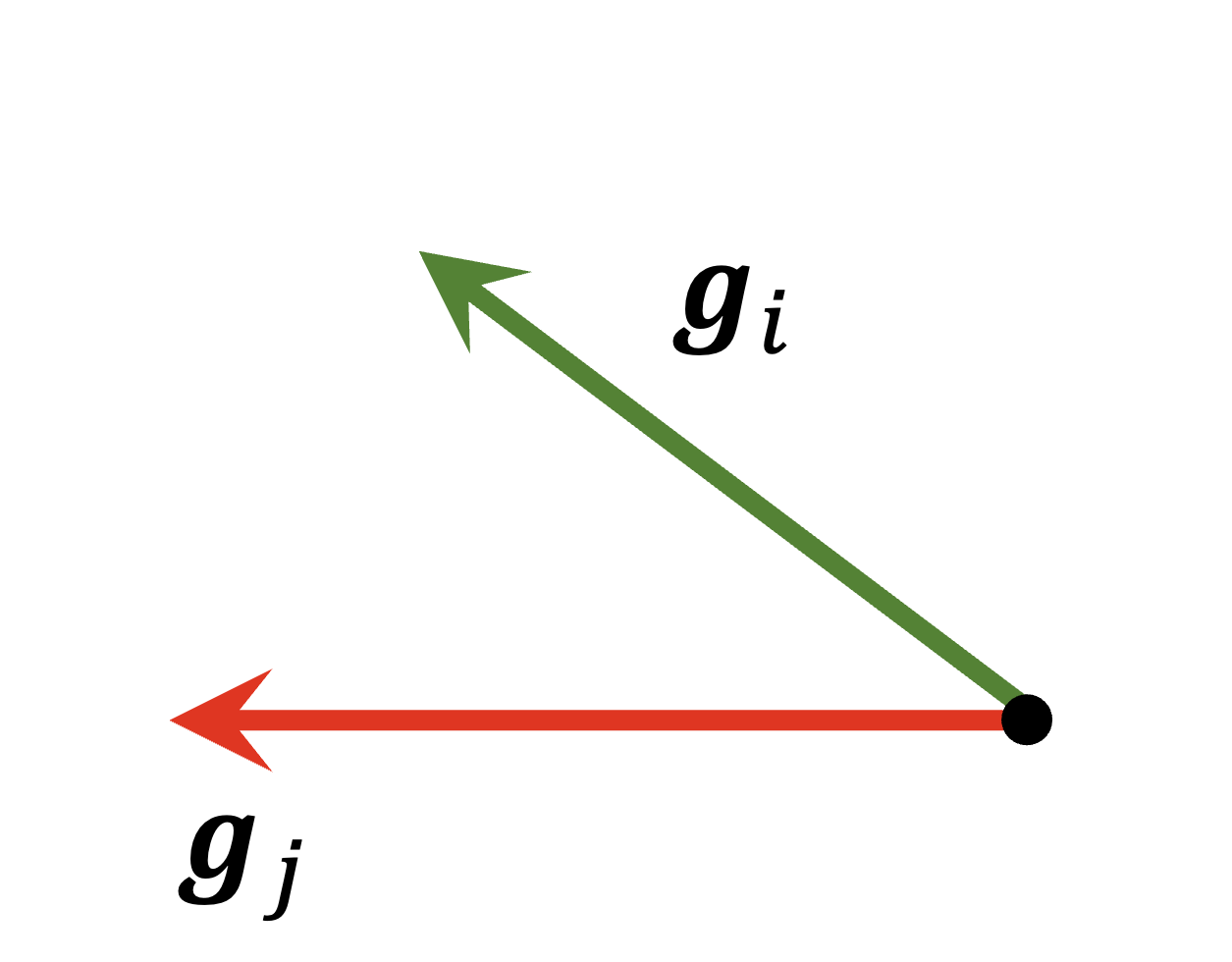}
        \caption{Non-conflicting.}
        \label{non-conflicting}
    \end{subfigure}
    \begin{subfigure}{0.22\textwidth}
        \includegraphics[width=1\textwidth]{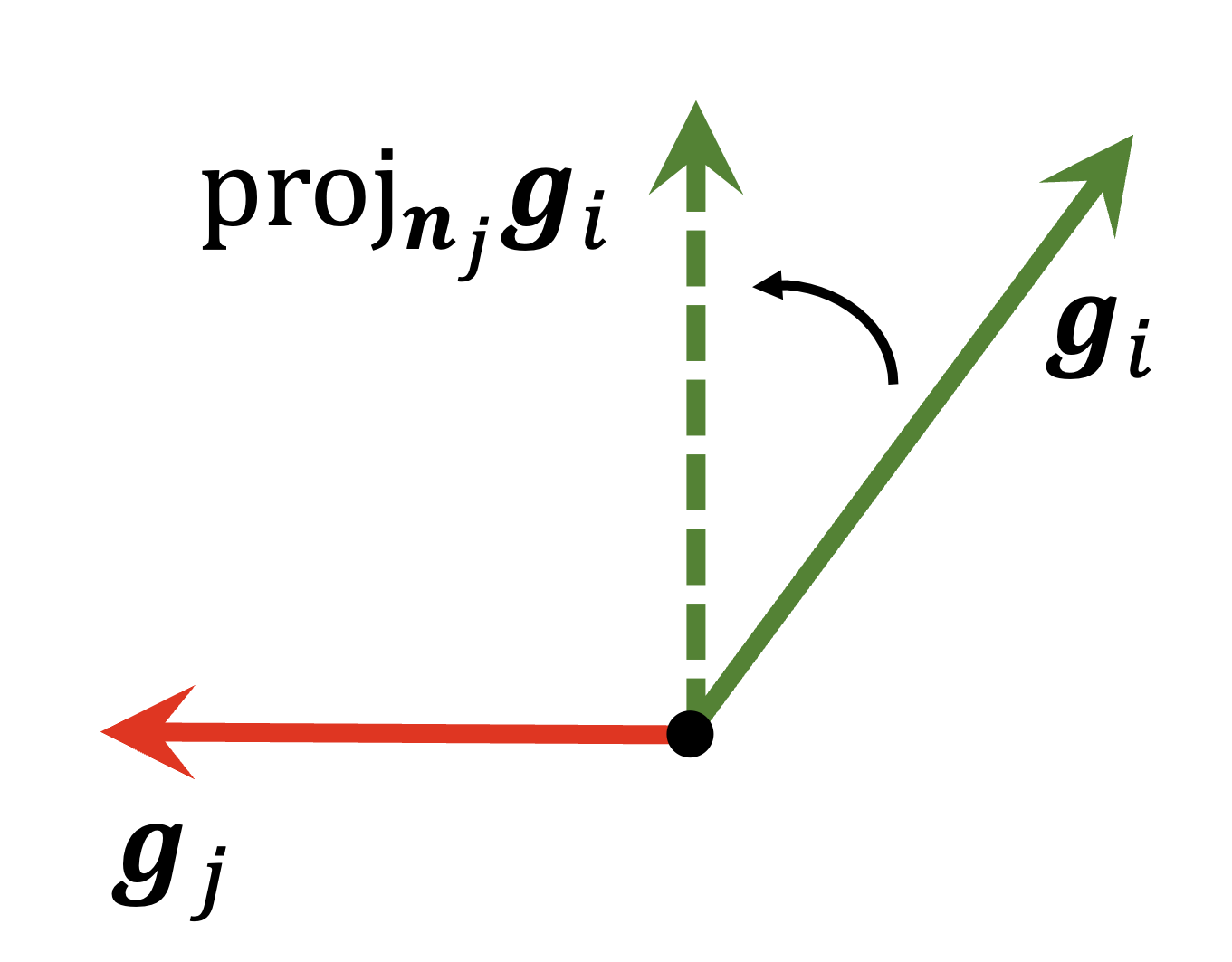}
        \caption{Projecting $\gbold_i$ to $\nbold_j$.}
        \label{i2j}
    \end{subfigure}
    \begin{subfigure}{0.23\textwidth}
        \includegraphics[width=1\textwidth]{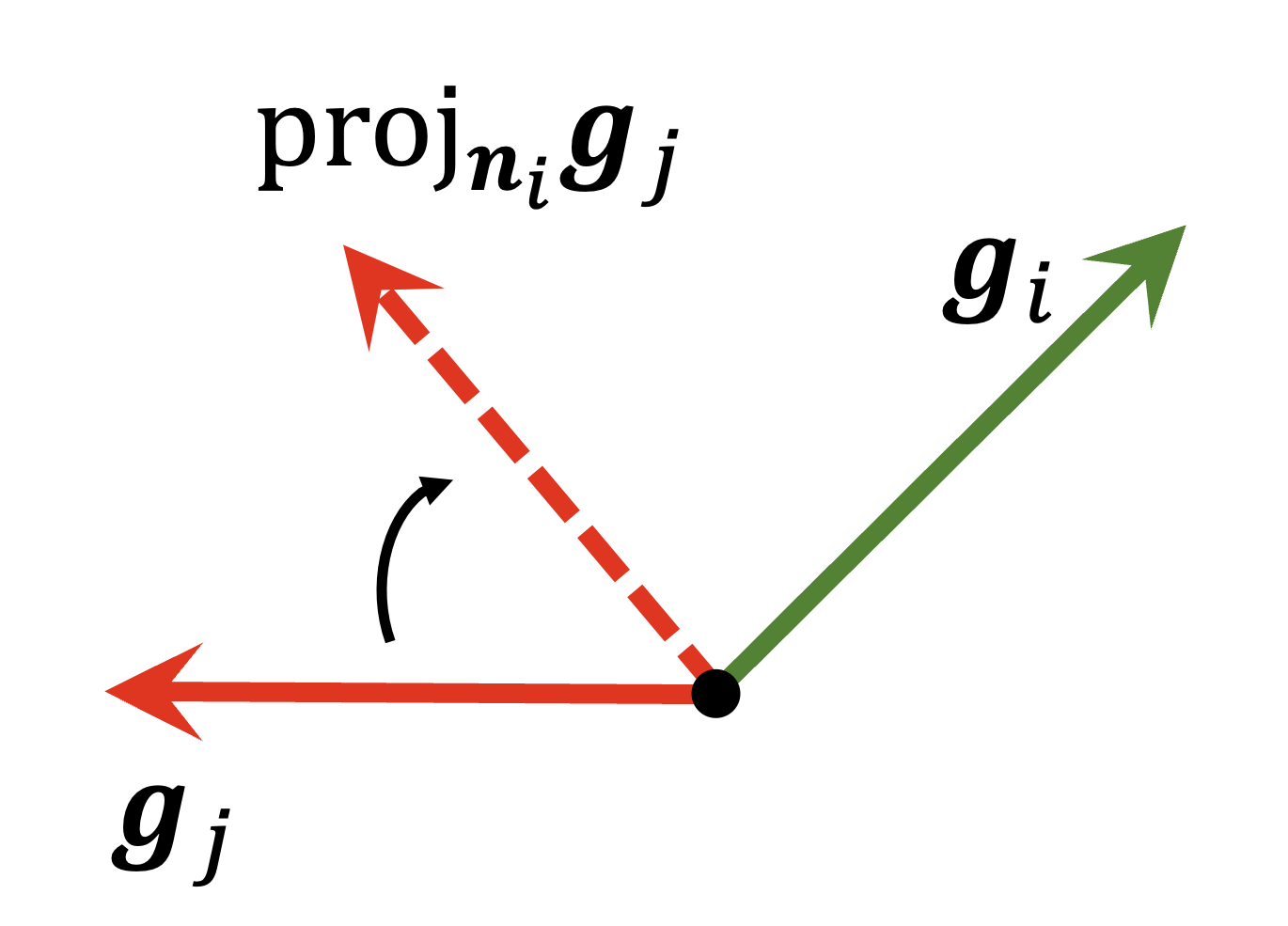}
        \caption{Projecting $\gbold_j$ to $\nbold_i$.}
        \label{j2i}
    \end{subfigure}
    \caption{Demonstration of gradient projection technique used in \citet{yu2020gradient}.}
    \label{fig:PCGrad}
\end{figure}

%% file: tex_files/02-2/pareto.tex
\begin{wrapfigure}[22]{r}{6.6cm}
    \centering
    \includegraphics[scale=0.45]{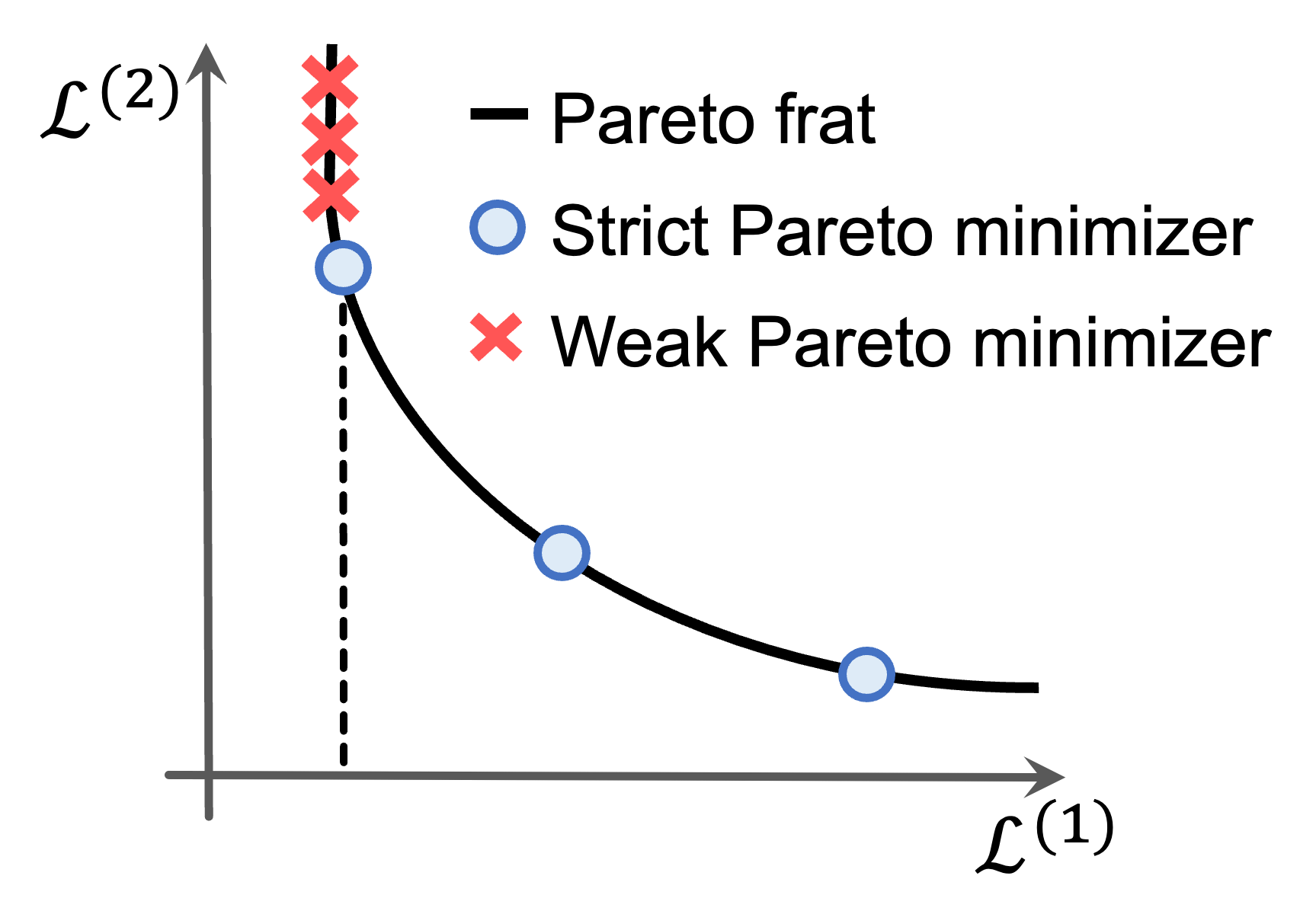}
    \caption{An illustration of weak and strict Pareto minimizers and Pareto front. We emphasize that the circles and crosses on the curve are NOT weak and strict Pareto minimizers. Instead, those $\Wbold^*$s that generate circles and crosses are weak and strict Pareto minimizers, respectively. In this figure, all circles and crosses are Pareto stationary points. We remark that in this example, the Pareto front is convex and continuous. The Pareto front can also be non-convex and/or discontinued, for example, see \citet[Page 12]{liu2020review}.}
    \label{fig:Pareto}
\end{wrapfigure}

%% file: table_files/optimization.tex
\begin{table}[!h]
\caption{Algorithms for the MTL as a multi-objective optimization.}
\label{tab:optimization}
\midsepremove
\scalebox{0.4}{
\begin{tabular}{lllllll}
\toprule
\rowcolor{gray!40}        Algorithm &
  Venue &
  Year &
  Method &
  Convergence &
  Highlight &
  Availability\tnote{1} \\ \midrule
Uncertainty Weighting &
  CVPR &
  \citeyear{kendall2018multi} &
  Dynamic Weighting &
  --- &
  Optimize $\{\alpha^{(t)}\}_{t=1}^{T}$ and $\Wbold$ simultaneously. &
  \href{https://github.com/yaringal/multi-task-learning-example}{\textcolor{citegreen}{Official}} \\ \midrule
\rowcolor{gray!20}        GradNorm &
  ICML &
  \citeyear{chen2018gradnorm} &
  Dynamic Weighting &
  --- &
  \begin{tabular}[c]{@{}l@{}}Adjust $\{\alpha^{(t)}\}$ is based on the average gradient norm of each tasks\\  and the relative progress achieved for each tasks.\end{tabular} &
  \href{https://github.com/hosseinshn/GradNorm}{\textcolor{reforange}{Unofficial}} \\ \midrule
MGDA-MTL &
  NeurIPS &
  \citeyear{sener2018multi} &
  Multi-Objective Opt. &
  Asymptotic Convergence &
  \begin{tabular}[c]{@{}l@{}}Seminal work, which proposes to use MOO to solve deep MTL problems\\ based on multi-gradient  descent algorithm.\end{tabular} &
  \href{https://github.com/isl-org/MultiObjectiveOptimization}{\textcolor{citegreen}{Official}} \\ \midrule
\rowcolor{gray!20}        RMTL &
  Thesis &
  \citeyear{liu2018exploration} &
  Dynamic Weighting &
  --- &
  Adjust $\{\alpha^{(t)}\}$ is based on the relative progress achieved for each tasks. &
  \href{https://github.com/gitter-lab/pria_lifechem}{\textcolor{citegreen}{Official}} \\ \midrule
LBTW &
  AAAI &
  \citeyear{liu2019loss} &
  Dynamic Weighting &
  --- &
  Adjust $\{\alpha^{(t)}\}$ using the reinforcement learning strategy. &
  \href{https://github.com/chao1224/Loss-Balanced-Task-Weighting}{\textcolor{citegreen}{Official}} \\ \midrule
\rowcolor{gray!20}        DWA &
  CVPR &
  \citeyear{liu2019end} &
  Dynamic Weighting &
  --- &
  $\{\alpha^{(t)}\}$ is adapted to both samples and tasks. &
  \href{https://github.com/lorenmt/mtan}{\textcolor{citegreen}{Official}} \\ \midrule
MLDT &
  CVPR &
  \citeyear{zheng2019pyramidal} &
  Dynamic Weighting &
  --- &
  $\{\alpha^{(t)}\}$ is adapted to the likelihood of a loss reduction. &
  \href{https://github.com/TencentYoutuResearch/PersonReID-Pyramid}{\textcolor{citegreen}{Official}} \\ \midrule
\rowcolor{gray!20}  
Pareto MTL &
  NeurIPS &
  \citeyear{lin2019pareto} &
  Multi-Objective Opt. &
  Asymptotic Convergence &
  Attemp to incorporate user's preference into the solution. &
  \href{https://github.com/Xi-L/ParetoMTL}{\textcolor{citegreen}{Official}} \\ \midrule
Controllable Pareto MTL &
  arXiv &
  \citeyear{lin2020controllable} &
  Multi-Objective Opt. &
  --- &
  Use a hypernetwork to learn the entire Pareto front. &
  \href{https://exeter-ecml.github.io/papers/0038-controllable-pareto-multitask-learning}{\textcolor{citegreen}{Official}} \\ \midrule
\rowcolor{gray!20} PCGrad &
  NeurIPS &
  \citeyear{yu2020gradient} &
  Gradient Correction &
  --- &
  Projecting onto orthogonal subspace to mitigate the gradient conflicts. &
  \href{https://github.com/tianheyu927/PCGrad}{\textcolor{citegreen}{Official}} \\ \midrule
        GradDrop &
  NeurIPS &
  \citeyear{chen2020just} &
  Gradient Correction &
  --- &
  Only keep gradients are consistent in signs in each update. &
  \href{https://github.com/tensorflow/lingvo/blob/master/lingvo/core/graddrop.py}{\textcolor{citegreen}{Official}} \\ \midrule
\rowcolor{gray!20} Continuous Pareto MTL &
  ICML &
  \citeyear{ma2020efficient} &
  Multi-Objective Opt. &
  --- &
  Construct a continuous, frst-order approximation of the local Pareto set. &
  \href{https://github.com/mit-gfx/ContinuousParetoMTL}{\textcolor{citegreen}{Official}} \\ \midrule
        EPO Search &
  ICML &
  \citeyear{mahapatra2020multi} &
  Multi-Objective Opt. &
  --- &
  \begin{tabular}[c]{@{}l@{}}Find a Pareto stationary solution to exactly match a user's preference.\\ Require losses to be non-negative.\end{tabular} &
  \href{https://github.com/dbmptr/EPOSearch}{\textcolor{citegreen}{Official}} \\ \midrule
\rowcolor{gray!20}  
AuxiLearn &
  ICLR &
  \citeyear{navon2021auxiliary} &
  Bi-level Opt. &
  --- &
  Learn to combine losses in a nonlinear fashion. &
  \href{https://github.com/AvivNavon/AuxiLearn}{\textcolor{citegreen}{Official}} \\ \midrule
       IMTL &
  ICLR &
  \citeyear{liu2021towards} &
  Gradient Correction &
   &
  \begin{tabular}[c]{@{}l@{}}Find $\{\alpha^{(t)}\}_{t=1}^{T}$ such that the aggregated gradient $\sum_{t=1}^{T}\alpha^{(t)}\grad \Lmcal^{(t)}(\Wbold)$ has equal \\ projection onto the raw gradients of individual tasks.\end{tabular} &
  \href{https://github.com/JohnLaMaster/Impartial-Multi-Task-Learning}{\textcolor{reforange}{Unofficial}} \\ \midrule
\rowcolor{gray!20}   
GradVac &
  ICLR &
  \citeyear{wang2021gradient} &
  Dynamic Weighting &
  --- &
  Encourage more geometrically aligned parameter updates for close tasks. &
  \href{https://github.com/chenllliang/Gradient-Vaccine}{\textcolor{reforange}{Unofficial}} \\ \midrule
     PHN &
  ICLR &
  \citeyear{navon2021learning} &
  Multi-Objective Opt. &
  --- &
  Use a hypernetwork to learn the entire Pareto front. &
  \href{https://github.com/AvivNavon/pareto-hypernetworks}{\textcolor{citegreen}{Official}} \\ \midrule
  
\rowcolor{gray!20}    CAGrad &
  NeurIPS &
  \citeyear{liu2021conflictaverse} &
  Gradient Correction &
  Asymptotic Convergence &
  The search direction is find by solving a subproblem that is similar to MGDA. &
  \href{https://github.com/Cranial-XIX/CAGrad}{\textcolor{citegreen}{Official}} \\ \midrule
     SVGD &
  NeurIPS &
  \citeyear{liu2021profiling} &
  Multi-Objective Opt. &
  \begin{tabular}[c]{@{}l@{}}Convergence rate for strongly convex\\ and third-order continuously differentiable functions\end{tabular} &
  \begin{tabular}[c]{@{}l@{}}
  Integrate MGDA with Stein variational gradient descent and Langevin dynamics \\to obtain diverse solutions.\end{tabular} &
  \href{https://github.com/gnobitab/MultiObjectiveSampling}{\textcolor{citegreen}{Official}} \\ \midrule
\rowcolor{gray!20}   COSMOS &
  ICDM &
  \citeyear{ruchte2021scalable} &
  --- &
  --- &
  \begin{tabular}[c]{@{}l@{}} A single optimization run to approximate the full set of the Pareto front by combining \\ preferences vectors sampled from Dirichlet distribution and training data.\end{tabular} &
  \href{https://github.com/ruchtem/cosmos}{\textcolor{citegreen}{Official}} \\ \midrule
        HV Maximization &
  arXiv &
  \citeyear{deist2021multi} &
  --- &
  --- &
  Utilize hyper-volume to approximate sample level Pareto front. &
  \href{https://github.com/timodeist/multi_objective_learning}{\textcolor{citegreen}{Official}} \\ \midrule
\rowcolor{gray!20}        PNG &
  UAI &
  \citeyear{ye2022optimization} &
  Multi-Objective Opt. &
  Convergence rate for convex losses &
  \begin{tabular}[c]{@{}l@{}}Minimize preference loss over the Pareto front (manifold optimization)\\ while only using the first order information.\end{tabular} &
  --- \\ \midrule
RLW \& RGW &
  TMLR &
  \citeyear{lin2022reasonable} &
  Dynamic Weighting &
  \begin{tabular}[c]{@{}l@{}}Converge to a neighborhood\\ of the optimal solution under\\ strongly convex assumption.\end{tabular} &
  \begin{tabular}[c]{@{}l@{}}Sample the weights $\{\alpha^{(t)}\}_{t=1}^{T}$ from a given distribution at each step.\end{tabular} &
  \href{https://github.com/median-research-group/libmtl}{\textcolor{reforange}{Unofficial}} \\ \midrule
\rowcolor{gray!20}        Nash-MTL &
  ICML &
  \citeyear{navon2022multi} &
  Multi-Objective Opt. &
  Asymptotic Convergence &
  \begin{tabular}[c]{@{}l@{}}Formulate the problem of finding a common descent direction as a bargaining game.\end{tabular} &
  \href{https://github.com/AvivNavon/nash-mtl}{\textcolor{citegreen}{Official}} \\ \midrule
(X)WC-MGDA &
  ICML &
  \citeyear{momma2022multi} &
  Dynamic Weighting &
  --- &
  Lift the restriction of non-negativity requirement on losses in EPO search. &
  --- \\ \midrule
\rowcolor{gray!20}        Rotograd &
  ICLR &
  \citeyear{javaloy2022rotograd} &
  \begin{tabular}[c]{@{}l@{}}Dynamic Weighting +\\ Gradient Correction\end{tabular} &
  --- &
  \begin{tabular}[c]{@{}l@{}}Dynamic Weighting via gradient norm\\ Gradient Correction via rotating the feature-space\end{tabular} &
  \href{https://github.com/adrianjav/rotograd}{\textcolor{citegreen}{Official}} \\ \midrule
MoCo~\citeyear{fernando2023mitigating} &
  ICLR &
  2023 &
  Multi-Objective Opt. &
  \begin{tabular}[c]{@{}l@{}}Convergence rates for \\ convex \& nonconvex losses\end{tabular} &
  Stochastic Gradient \& Variance Reduction &
  \href{https://github.com/heshandevaka/Trade-Off-MOL}{\textcolor{citegreen}{Official}}\\ \midrule
\rowcolor{gray!20}        Recon &
  ICLR &
  \citeyear{shi2023recon} &
  Gradient Correction &
  --- &
  \begin{tabular}[c]{@{}l@{}}Turn shared parameters that most likely to cause \\ gradient conflicts into task specific parameters.\end{tabular} &
  \href{https://github.com/moukamisama/Recon}{\textcolor{citegreen}{Official}} \\ \midrule
Aligned-MTL &
  CVPR &
  \citeyear{Senushkin_2023_CVPR} &
  Gradient Correction &
  --- &
  \begin{tabular}[c]{@{}l@{}}Use the condition number of the linear system to measure\\ the severity of gradient dominance and conflicting issues.\end{tabular} &
  \href{https://github.com/SamsungLabs/MTL}{\textcolor{citegreen}{Official}} \\  \midrule

\rowcolor{gray!20}        Achievement-based MTL &
  ICCV &
  \citeyear{yun2023achievement} &
  Dynamic Weighting &
  --- &
  \begin{tabular}[c]{@{}l@{}} Use training progress to dynamically weight tasks and use geometric mean to average loss from tasks\end{tabular} &
  \href{https://github.com/samsung/Achievement-based-MTL}{\textcolor{citegreen}{Official}} \\ \midrule

  FULLER &
  ICCV &
  \citeyear{huang2023fuller} &
  Dynamic Weighting &
  --- &
  \begin{tabular}[c]{@{}l@{}} Use gradient norm of different tasks to adjust the weights for tasks.\end{tabular} &
  --- \\
  
 \bottomrule
\end{tabular}
}
\end{table}

%% file: tex_files/02-2/MoE_framework.tex
\begin{wrapfigure}[25]{r}{5.2cm}
    \centering
    \begin{subfigure}{0.38\textwidth}
        \includegraphics[width=1\textwidth]{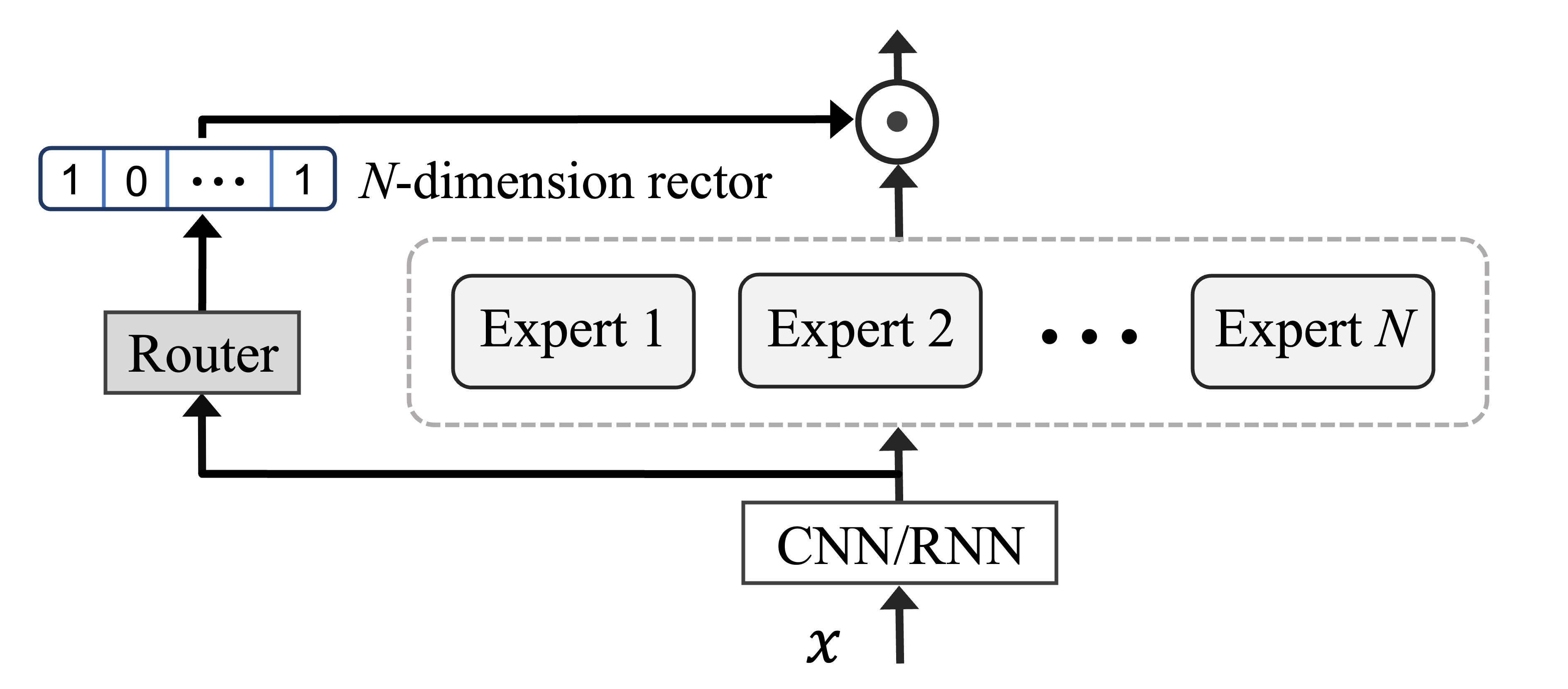}
        \caption{MoE.}
        \label{moe}
    \end{subfigure}
    \begin{subfigure}{0.38\textwidth}
        \includegraphics[width=1\textwidth]{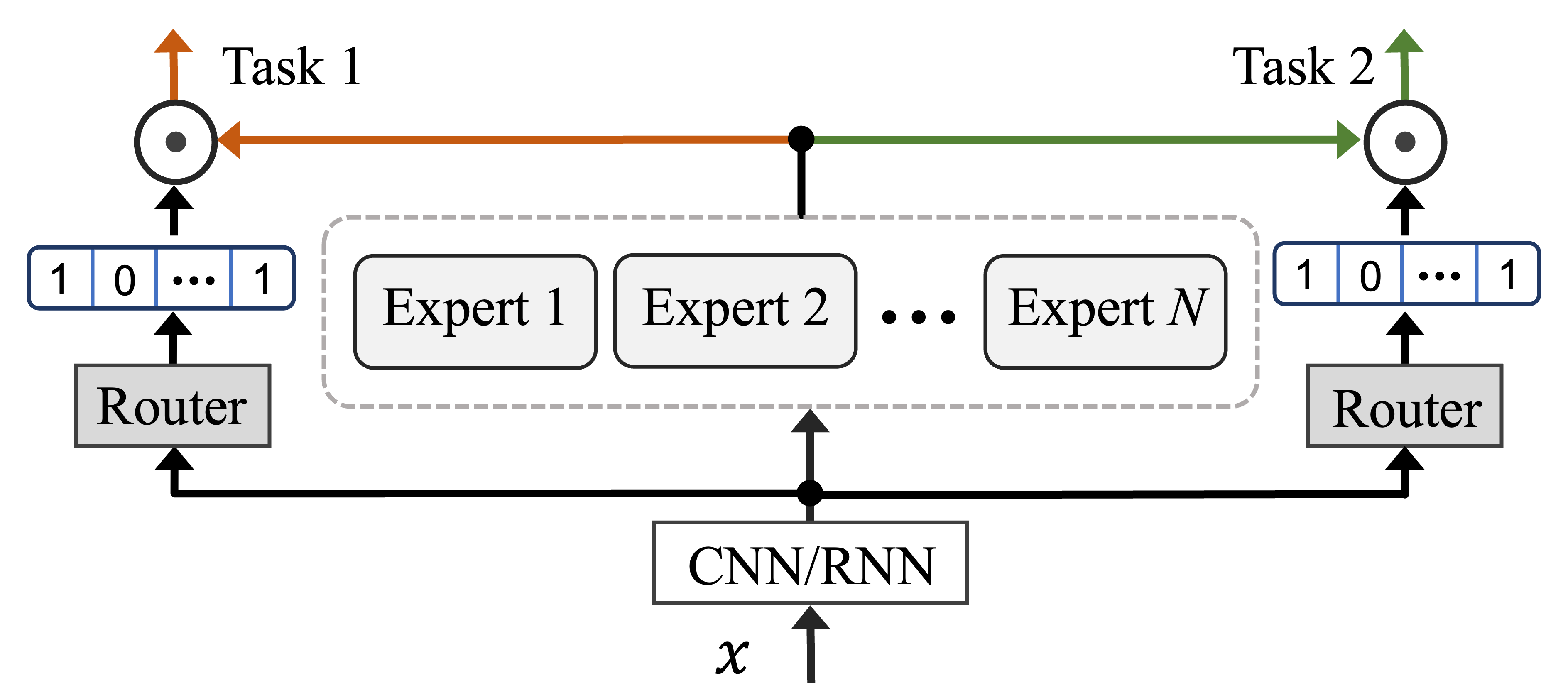}
        \caption{Multi-Router MoE.}
        \label{multi-moe}
    \end{subfigure}
    \begin{subfigure}{0.38\textwidth}
        \includegraphics[width=1\textwidth]{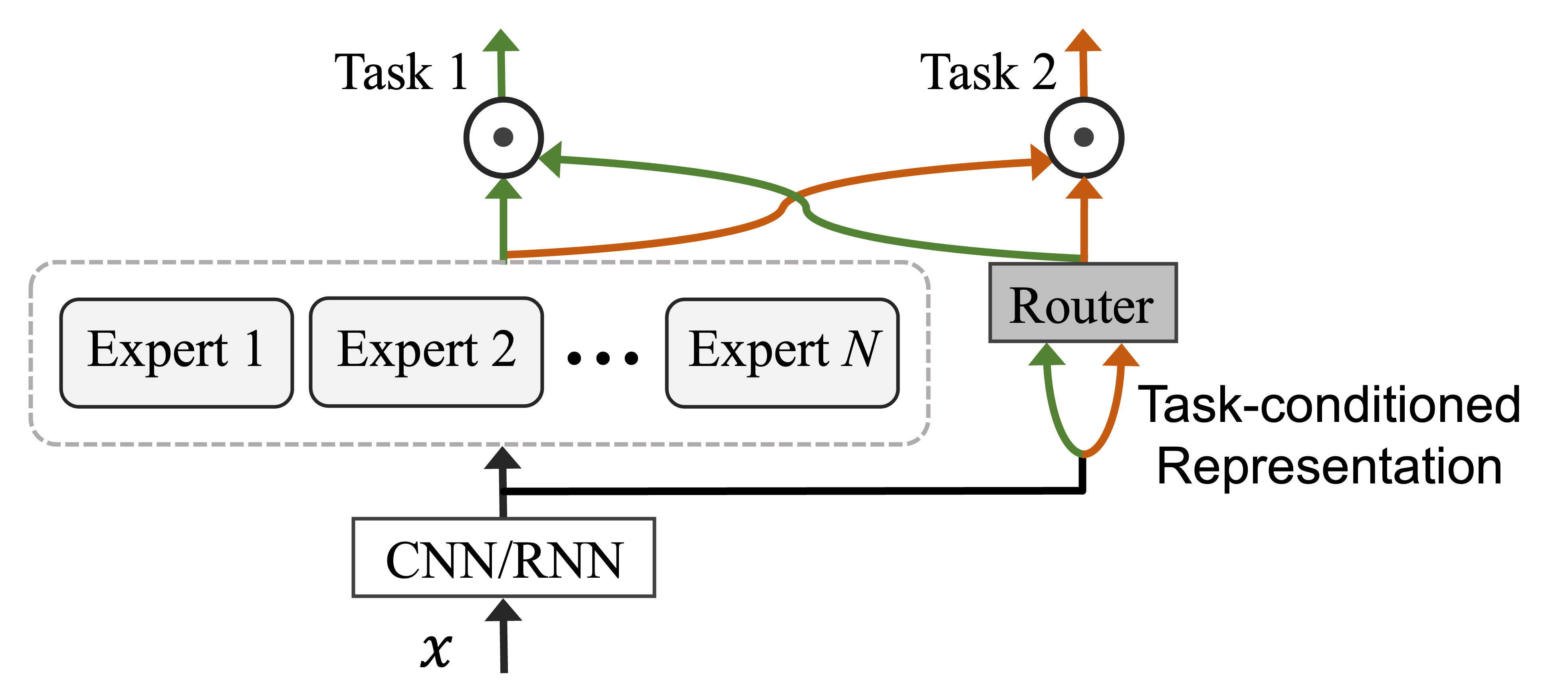}
        \caption{Single-Router MoE.}
        \label{single-moe}
    \end{subfigure}
    \caption{The taxonomy of (a) MoE into two categories: (b) Multi-Router MoE (c) Single-Router MoE.}
    \label{moe-taxonomy}
\end{wrapfigure}

%% file: tex_files/02-2/multikernel.tex
\begin{wrapfigure}[11]{r}{8cm}
    \centering
    \includegraphics[width=0.85\linewidth]{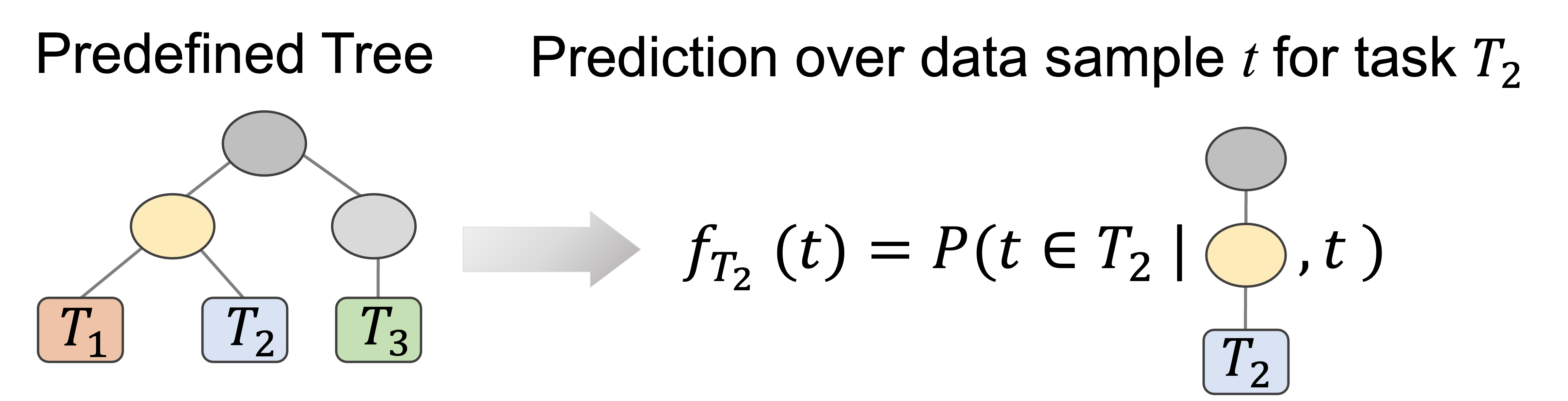}
      \caption{An example of MultiKernel predicting the probability of a data sample $t$ belonging to task $T_2$. MultiKernel conducts the prediction based on $T_2$ and its domains, whose hierarchical relation is extracted from the predefined tree. Specifically, ellipses are domains, and squares are tasks.}
    \label{fig:multikernel}
\end{wrapfigure}

%% file: 02-3_method_pfm.tex

\input{tex_files/02-3/PFM_framework}
\subsection{Foundation Model Era: Towards Unified and Versatile}
\label{fm-era}
{~~}\vspace{2pt}\\
AI models are shifting their focus from deeper networks (e.g., ConvNets~\citep{fukushima1980neocognitron, lecun1998gradient, he2016deep, liu2022convnet}, GANs~\citep{goodfellow2020generative}, CapsNets~\citep{sabour2017dynamic}, RNNs~\citep{rumelhart1986learning, hochreiter1997long}) to foundation (e.g., BERT~\citep{devlin2018bert}, GPT-4\footnote{\url{https://openai.com/research/gpt-4}}~\citep{openai2023gpt4}, SAM~\citep{kirillov2023segment}, DALL$\cdot$E 3\footnote{\url{https://openai.com/dall-e-3}}~\citep{ramesh2021zero}). Such foundation models leverage (usually in self-supervised, unsupervised, and assisted-manual ways) web-scale pretraining data in the wild  and then adapt their backbones to different downstream tasks~\citep{bommasani2021opportunities, zhou2023comprehensive}, thus inherently non-conflict towards MTL. In light of recent development of scalable learners, particularly Transformers, foundation models evolve from parameter-based transfer learning with new emergent capabilities. They facilitate the integration of multiple tasks into a pretrained backbone, achieved through only fine-tuning or even zero-shot learning (ZSL). In this context, the emergent properties in foundation models extend MTL from a fixed set of tasks (where training and test tasks are identical) to handling unknown tasks. When viewed from a task-oriented perspective, MTL, empowered by foundation models, can be categorized into three distinct types: 
\begin{enumerate}
    \item \emph{(Downstream) Task-Generalizable Fine-tuning.} This category involves the uni-modal learning of inclusive representations in semi-supervised, self-supervised, and unsupervised learning manners. Notable examples include BiGAN~\citep{donahue2016adversarial, donahue2019large}, BERT~\citep{devlin2018bert}, MoCo~\citep{he2020momentum, chen2020improved, chen2021empirical}, , SimCLR~\citep{chen2020simple, chen2020big}, MAE~\citep{he2022masked}, and GPT~\citep{radford2018improving, radford2019language, brown2020language, openai2023gpt4}. The learned encoders should be transferable to a variety of downstream supervised tasks, thereby enabling them to be multi-task learners.
    \item \emph{Task-Promptable Engineering.} In this category, the original inputs are modified through task-specific prompts (e.g., SAM~\citep{kirillov2023segment}) during the pretraining stage. Prompt engineering can affect the representation of data and facilitate the learners with few-shot and even zero-shot abilities toward new tasks.
    \item \emph{Task-Agnostic Unification.} This category highlights that the representations remain unbiased toward specific tasks and data modalities via employing a unified serialization/sequence of data tokens, including Pix2Seq~\citep{chen2022pixseq, chen2022unified}, UniTAB~\citep{yang2022unitab}, Unified-IO~\citep{lu2022unified}, Uni-Perceiver~\citep{nips_zhu2022uni, cvpr_zhu2022uni, li2023uni}, OFA~\citep{wang2022ofa, bai2022ofasys}, Gato~\citep{reed2022generalist}, UnIVAL~\citep{shukor2023unified}, etc. As a result, multi-modal learners can obtain the generalizability from existing tasks to new ones, even those involving diverse data modalities.
\end{enumerate}

\subsubsection{Downstream Task Fine-Tuning} 
At the moment of Pretrained Foundation Models (PFMs)~\citep{zhou2023comprehensive} inception, the terminology ``pre-training'' remained somewhat ambiguous within the field of DL research. This practice involves the initial learning of model backbones on a general dataset, e.g., ImageNet~\citep{deng2009imagenet, russakovsky2015imagenet}, followed by their transfer to other tasks that commence fine-tuning with a warm-up initialization. Consequently, a similar process of ``fine-tuning'' before PFMs pertains to the fine-tuning of model backbones. In our context, fine-tuning with the changes of backbone parameters refers to model tuning, unless otherwise specified. It matters since PFMs are costly to backpropagate, and the ability to generalize large frozen backbone to multiple downstream tasks referred to as downstream fine-tuning, can ease this burden. By confining our discussion to the context of downstream fine-tuning within the frozen model, we can extend the previous definition of MTL (refer to Definition~\ref{def:MTL}). In this context, a single model can effectively handle a set of tasks. This approach also facilitates a clear separation from the domain of (parameter-based) TL.

In the context of fine-tuning for downstream tasks facilitated by PFMs, the process typically begins with the pre-training of a backbone foundation model on large data in the wild. This pre-training phase often employs unsupervised or self-supervised methods. Subsequently, the pretrained backbone is fine-tuned using task-specific domain datasets, as illustrated in Fig.~\ref{fine-tuning}. Leveraging the task-unbiased representations acquired from the frozen backbone, fine-tuning of task-specific heads (e.g., simple MLPs for classification tasks or mask decoders for dense prediction tasks) frequently yields competitive or even superior results when compared to prior supervised outcomes across a spectrum of diverse downstream tasks.

Nonetheless, it is important to note that the pre-training phase tends to restrict data modality due to the constraints of self-supervised techniques, which are inherently data-specific. For instance, methodologies like masked image modeling (MIM) in MAE are suitable for image data, while masked language modeling (MLM) in BERT is tailored for text data. Subsequent review provides an in-depth exploration of downstream task fine-tuning methods categorized by data modality. Specifically, we will discuss these methods these methods within the domains of vision, language, and vision-language tasks.

\underline{Vision Tasks.} Early pre-training techniques in computer vision primarily focus on learning from pretext tasks. Exemplar CNN~\citep{dosovitskiy2014discriminative, alexey2016discriminative}, for instance, initially pretrains backbone models by discriminating various patches within unlabeled data. 
In the case of {Inpainting}~\citep{pathak2016context}, the pretext task involves predicting the masked central parts of images. 
{Colorization}~\citep{zhang2016colorful}, on the other hand, establishes mappings from grayscale images to their colored versions. 
{Split-Brain Autoencoders}~\citep{zhang2017split} forces the network to split into two disjoint sub-networks, each processing one-half of the input images while predicting the corresponding missing parts from the other sub-network. Recently, {BEiT}~\citep{bao2021beit, peng2022beit} and {MAE}~\citep{he2022masked} simply reconstruct the random mask patches of the images to pretrain the backbones, i.e., masked image modeling (MIM). Other MIM methods contain {iBOT}~\citep{zhou2021ibot}, {CAE}~\citep{chen2023context}, {SimMIM}~\citep{xie2022simmim}, {BEVT}~\citep{wang2022bevt}, {ConMIM}~\citep{yi2022masked}, {VideoMAE}~\citep{tong2022videomae, wang2023videomae}, to name a few.
{Jigsaw}~\citep{noroozi2016unsupervised} and {Completing Damaged Jigsaw Puzzles (CDJP)}~\citep{kim2018learning} employ Jigsaw puzzles as pretext tasks during model pre-training. {Counting}~\citep{noroozi2017representation} can also serve as a pretext task to facilitate representation learning. 
{Noise As Targets (NAT)}~\citep{bojanowski2017unsupervised} focuses on learning representations by aligning the deep features of the backbone with predefined targets in a low-dimensional space.
{RotNet}~\citep{gidaris2018unsupervised}, however, is designed for predicting different image rotations. Notably, such early pre-training techniques of pretext tasks typically do not require manual annotations, allowing for fast training without the necessity of developing new loss functions. Downstream multiple tasks commonly include classification, object detection, and segmentation. Thus, parameter-efficient training (PEFT) of MTL models becomes challenging since the model must adapt to the needs of multiple tasks simultaneously. MTLoRA~\citep{agiza2024mtlora} is the first to address this problem and dominates other SOTA PEFT methods.

An alternative line of research aims to design a general representation learning algorithm that is unbiased to the pretext tasks, often referred to as contrastive self-supervised learning (SSL)~\citep{jaiswal2020survey, liu2021self}. This method unlocks the potential of representations by introducing a novel loss function that hinges on the concept of ``contrast.'' If we denote the sets of samples that are similar and dissimilar to $\Xmcal$ as $\Xmcal^+$ and $\Xmcal^-$ respectively, the Noise Contrastive Estimation (NCE) loss~\citep{gutmann2010noise} can be defined as
\begin{equation}
    \Lmcal_{\text{NCE}} = \Embb_{\Xmcal, \Xmcal^+, \Xmcal^-}\left[-\log(e^{f(\Xmcal)^\top f(\Xmcal^+)})/[e^{f(\Xmcal)^\top f(\Xmcal^+)} + e^{f(\Xmcal)^\top f(\Xmcal^-)}]\right],
    \label{NCE_loss}
\end{equation}
where the function $f(\cdot)$ represents the encoder function used to learn image embedding. It is worth noting that the cosine-based similarity measurement mentioned above can be customized to suit various scenarios. Additionally, the InfoNCE loss~\citep{oord2018representation} extends this concept by incorporating a more extensive set of dissimilar pairs as
\begin{equation}
     \Lmcal_{\text{InfoNCE}} = \Embb_{\Xmcal, \Xmcal^+, \Xmcal^b}\left[-\log(e^{f(\Xmcal)^\top f(\Xmcal^+)})/[e^{f(\Xmcal)^\top f(\Xmcal^+)} + \sum_{b=1}^{B-1} e^{f(\Xmcal)^\top f(\Xmcal^b)}]\right],
     \label{infoNCE_loss}
\end{equation}
where $B$ represents the batch size, comprising $B-1$ negative pairs $\{(\Xmcal,\Xmcal^b)\}_{b=1}^{B-1}$ and one positive pair $(\Xmcal, \Xmcal^+)$. These loss functions are closely linked to the maximization of mutual information (MI) between the encoded representations.

Many contrastive SSL methods draw from the loss functions (\ref{NCE_loss}) and (\ref{infoNCE_loss}) to acquire task-invariant representations. {Non-parametric instance discrimination (NPID)}~\citep{wu2018unsupervised} can capture apparent similarity among instances using NCE. In contrast, 
{contrastive predictive coding (CPC)}~\citep{oord2018representation, henaff2020data} first introduces the InfoNCE loss for the pre-training of RNN in an autoregressive manner. {Deep InfoMax (DIM)}~\citep{hjelm2018learning}, Deep Graph InfoMax (DGI)~\citep{velivckovic2018deep}, and Augmented Multiscale DIM (AMDIM)~\citep{bachman2019learning} take a direct approach by maximizing the MI between representations. Contrastive multiview coding (CMC)~\citep{tian2020contrastive} extends the concept of MI maximization to incorporate more than two views, {MoCo}~\citep{he2020momentum, chen2020improved, chen2021empirical} employs InfoNCE but introduces the momentum contrast based on a memory bank used in~\citep{wu2018unsupervised}. {SimCLR}~\citep{chen2020simple, chen2020big} proposes a novel contrastive loss known as the normalized temperature-scaled cross-entropy loss (NT-Xent) for representation learning. Bootstrap Your Own Latent (BYOL)~\citep{grill2020bootstrap}, conversely, takes a different approach by obviating the need for negative pairs. On the other hand, several other methods~\citep{caron2018deep, caron2020unsupervised, goyal2021self, li2020prototypical} endeavor to employ clustering algorithms that contrast data representations based on class prototypes.

\underline{Language Tasks.} In the domain of language, initial pre-training approaches utilizing word embeddings~\citep{mikolov2013distributed, pennington2014glove} to predict subsequent tokens for a warm start have shown potential in enhancing the performance of downstream NLP tasks~\citep{dai2015semi, mccann2017learned}. Nonetheless, these methods often rely on a limited dataset for pre-training, which restricts their effectiveness and prevents consistently satisfactory outcomes across the spectrum of downstream NLP tasks. Current Transformer-based Pre-trained Foundations Models (PFMs) in natural language processing can be broadly classified into three types~\citep{wang2022pre}: \textit{encoder-only}, \textit{decoder-only}, and \textit{encoder-decoder} architectures. Encoder-only architectures employ a bidirectional Transformer encoder designed to reconstruct masked tokens. Decoder-only models utilize a unidirectional Transformer decoder that predicts tokens in a left-to-right autoregressive fashion. Encoder-decoder models are crafted for sequence-to-sequence (seq2seq) generation tasks, pretrained by masking tokens in the source sequence and predicting them in the target sequence. 

This taxonomy aligns with the constraints in terms of tasks. Since the encoder-only architectures, e.g., BERT~\citep{devlin2018bert}, ERNIE 1.0/2.0~\citep{sun2019ernie, sun2020ernie}, SpanBERT~\citep{joshi2020spanbert}, DeBERTa~\citep{he2020deberta}, and GLaM~\citep{du2022glam}, are pretrained to predict masked tokens based on the bidirectional context, they are better suited for understanding tasks rather than generation tasks. They are adept at tasks like document classification, named entity recognition, and question answering where the full context is available and the task is to understand or extract information rather than generate it. Encoder-only models often have a fixed maximum sequence length, which limits their ability to handle very long documents directly. They are not designed for incremental token-by-token generation and thus are inefficient for tasks that require such predictions, like text completion or interactive text generation. Conversely, decoder-only architectures, e.g., GPT-3~\citep{brown2020language}, PanGu-$\alpha$~\citep{zeng2021pangu}, Turing-NLG, HyperCLOVA~\citep{kim2021changes}, Gopher~\citep{rae2021scaling}, LaMDA~\citep{thoppilan2022lamda}, PaLM~\citep{chowdhery2022palm}, Open Pre-trained Transformers (OPT)~\citep{zhang2022opt}, LLaMA~\citep{touvron2023llama, touvron2023llama}, PanGu-$\Sigma$~\citep{ren2023pangu} and PaLM-2~\citep{anil2023palm}, are pre-trained in a unidirectional context, making them well-suited for generative tasks such as language modeling and text generation. However, this unidirectional training means they may be less effective for tasks that require understanding the full context of the input, as they can only condition on the left context. These models generate one-text token at a time, which can be slower compared to models that handle the entire input at once, and they might struggle with tasks requiring bidirectional context.
Encoder-decoder Architectures, e.g., T5~\citep{raffel2020exploring}, BART~\citep{lewis-etal-2020-bart}, ERNIE 3.0~\citep{sun2021ernie}, Switch Transformers~\citep{fedus2022switch} and Flan-T5~\citep{chung2022scaling}, are more flexible as they can handle both understanding and generation tasks. While they offer considerable advantages in terms of their adaptability to various tasks, they come with trade-offs in terms of model complexity, resource requirements, and potential issues with error propagation.

\underline{Vision-Language Tasks.}
PFMs effectively manage multiple tasks without requiring model tuning. However, the aforementioned methods remain constrained to a unimodal context. In real-world scenarios, there is a natural requirement for multimodal or cross-modal intelligence. Such intelligence should handle multiple tasks across diverse modalities and domains. Vision-Language (VL), as its name implies, bridges CV and NLP. It was among the first areas to be extensively explored by the research community for multi-modal learning in recent years. Given the intricacy and scope of VL tasks, foundation models employing vision-language pre-training (VLP) have rapidly gained prominence, showcasing notable performance. Initial VLP approaches~\citep{su2019vl, li2019visualbert, tan2019lxmert, chen2020uniter, kim2021vilt, li2021align}
centered on task-specific tasks such as visual question answering (VQA), image captioning, visual grounding, etc.

The advent of the contrastive language-image pre-training (CLIP)~\citep{radford2021learning}, however, marks a significant leap forward in multiple downstream tasks, as it jointly refines dual encoders to align (image, text) pairs within latent embedding space, showcasing learning SOTA multimodal representations from unstructured image-text data. The general representations by cross-modal contrastive learning validate stellar performance in zero-shot transfer across various vision-language (VL) tasks. In a similar trajectory, the Large-scale Image and Noisy-text embedding (ALIGN)~\citep{jia2021scaling} method leverages uncurated data, amplifying the efficacy of VLP in downstream cross-modal retrieval tasks. Other contrastive VLP methods contain ALBEF~\citep{li2021align}, WenLan~\citep{huo2021wenlan}, triple contrastive learning (TCL)~\citep{yang2022vision}, and BLIP~\citep{li2022blip, li2023blip}. 
All these methods contribute to the learning of general-purpose visual and linguistic representations, seamlessly adapting to a variety of downstream tasks ranging from cross-modal reasoning (e.g., VQA) and cross-modal matching (e.g., Image Text Retrieval and Visual Referring Expression), to vision and language generation tasks.
Notably, DALL$\cdot$E~\citep{ramesh2021zero} stands out in its remarkable capability to perform text-to-image generation tasks in a zero-shot manner, meeting commercial application standards. This underscores the potential and versatility of VLP in facilitating generalist applications.


\begin{applebox}{Remarks}
\begin{enumerate}[leftmargin=0.4cm, label=(\roman*)]
    \item Downstream fine-tuning reduces the data requirements for downstream tasks and also the training (fine-tuning) time and resources. 
    \item Downstream fine-tuning eases the intensive training burden and enhances the accessibility of PFMs, rendering them a practical solution available to anyone.
    \item Downstream fine-tuning necessitates that the data modalities for downstream tasks remain consistent with those pretrained in pretext tasks.
    \item Due to PFMs containing pretext task biases, the full potential of multi-task performance remains unrealized.
\end{enumerate}
\end{applebox}




\subsubsection{Task Prompting}

As the evolution of PFMs advances, the incorporation of prompting into the tuning process of frozen PFMs for downstream tasks has initially become widely recognized through the name of ``prompt design''~\citep{brown2020language} and subsequently carried forward through the practice of ``prompt tuning.''~\citep{lester2021power}
Conceptually, prompts serve as carriers of task-descriptive information, enabling the adaptation of PFMs to various tasks in a manner that can be either manually crafted or automatically generated, as illustrated in Fig.~\ref{task-prompting}. The primary use of prompts lies in their built-in ability to significantly alleviate the demands of task-specific fine-tuning through freezing backbone parameters of PFMs and only learning task-indicating prompts, ultimately leading to enhanced few-shot or even zero-shot generalizability, all while requiring augmenting inputs and maintaining minimal to no parameter updates. A comprehensive examination of prompt taxonomy exceeds the scope of this section. Consequently, we adopt the notion of task prompting to encompass all prompt engineering methodologies within the framework of task adaptation and generalization. 

The additional task-specific prompts augmented with the model can be hard and soft~\citep{gu2023systematic}. The hard prompts contain task instructions or hints from human-interpretable natural language, including human instructions~\citep{radford2019language, efrat2020turking} in the early stage and more advanced In-Context Learning (ICL)~\citep{dong2022survey} and chain-of-thought (CoT)~\citep{yu2023towards, chu2023survey}. The soft prompts are also referred to as continuous prompting or prompt tuning that optimizes prompts implicitly in the embedding space, which can be learned/propagated to align with specific tasks.

\underline{Hard Prompt Engineering.} Large Language Models (LLMs), via making predictions based on a few examples in the context, i.e. ICL, can finally perform different tasks. This learning from demonstration and analogy are also presented as emergent abilities~\citep{wei2022emergent} in LLMs. GPT-3~\citep{brown2020language} first verified that LLMs are few-shot learners and that different tasks can be performed given a few examples in the form of demonstration context. InstructGPT~\citep{ouyang2022training} further aligned LLMs with user intent using reinforcement learning from human feedback (RLHF). The developments in ICL contain strategies both in training stage~\citep{wei2021finetuned, chen-etal-2022-improving, min-etal-2022-metaicl, wang2022super, iyer2022opt, wei2023symbol, gu2023pre} and inference stage~\citep{liu2021makes, rubin-etal-2022-learning, gonen2022demystifying, sorensen-etal-2022-information, zhang2022active, li2023finding, lu-etal-2022-fantastically, honovich2022instruction, zhou2022least, hao2022structured, xu2023small, xu2023k}. FLAN~\citep{wei2021finetuned} tuned LLMs via natural language instruction templates over 60 NLP tasks and surpassed zero-shot GPT-3 on some of the datasets. MetaICL~\citep{min-etal-2022-metaicl} introduced meta-training for ICL on a more broad spectrum (100-level) of NLP tasks. Sup-NatInst~\citep{wang2022super} presented a benchmark of 1000-level NLP tasks and proposed T$k$-Instruct that can outperform InstructGPT with fewer parameters. OPT-IML~\citep{iyer2022opt} Scales LLMs instruction meta-learning to 2000 NLP tasks through the lens of generalization. Symbol Tuning~\citep{wei2023symbol} targets the situation when instructions or natural language are insignificant in predicting the task. PICL~\citep{gu2023pre} enhanced the ICL ability for LLMs by pre-training to maintain task generalization, while previous investigations are how to select in-context examples for better few-shot capabilities during the testing stage~\citep{liu2021makes}. Other methods~\citep{gonen2022demystifying, sorensen-etal-2022-information, zhang2022active, li2023finding, lu-etal-2022-fantastically, honovich2022instruction, zhou2022least, hao2022structured, xu2023small, xu2023k} tried to understand why the performance varifies from different prompts and how to pick better prompts from different angles.
After prompt retriever~\citep{rubin-etal-2022-learning} is verified efficient for ICL, many efforts used the prompt pool as a tool to support retrieval-based prompting, where relevant prompts or context are retrived for ICL~\citep{rubin2021learning, li2023unified, ye2023compositional, zhang2023makes}. 

Furthermore, chain-of-thought (CoT) prompts are a series of instructions with progressive orders, which can help LLMs perform complex reasoning tasks step by step~\citep{wei2022chain, kojima2022large, zhang2022automatic, fu2022complexity, ho2022large, trivedi2022interleaving, chen2022program}. Manual-CoT~\citep{wei2022chain} first explores how to improve the ability of LLM by generating CoT. Zero-Shot-CoT~\citep{kojima2022large} proposes a single task-agnostic zero-shot prompt to surpass ICL even without input-output demonstrations. Complex-CoT~\citep{fu2022complexity} shows that complex reasoning chains excel simple chains. Auto-CoT~\citep{zhang2022automatic} mitigates the mistakes that could happen in precious manual ways by automatically constructing demonstrations for different questions. Fine-tune-CoT~\citep{ho2022large} can use teacher-generated reasoning to fine-tune smaller models. IRCoT~\citep{trivedi2022interleaving} interleaves retrieval with steps and, in turn, improves the ability of CoT by retrieved results. PoT~\citep{chen2022program} uses programming language statements to delegate math computations.

\underline{Soft Prompt Tuning.} In comparison, soft prompt tuning can backpropagate prompt vectors using gradient descent.~\citet{lester2021power} introduces the concept of ``prompt tuning'' and distinguishes it from previous model tuning and prompt design methods. During the training, prompt tuning can refine the prompts to improve learning performance on specific tasks. Thus, the multi-task setting can be realized by simply mixing training data across different tasks. Soft Prompt Transfer (SPoT)~\citep{vu2021spot} pioneers the demonstration that prompt tuning can efficiently transfer from source to target tasks, offering a parameter-efficient approach to prompt-based transfer learning across diverse tasks. 
P-Tuning~\citep{liu2022p} empirically optimizes prompt tuning to be universally effective across a wide range of tasks. 
ATTEntional Mixtures of Prompt Tuning (ATTEMPT)~\citep{asai2022attempt} exemplifies this concept by combining multiple prompts trained on large-scale source tasks, generalizing instance-wise prompts on target tasks while keeping model parameters and source prompts frozen. Multi-task Pre-trained Modular Prompt (MP$^2$)~\citep{sun-etal-2023-multitask} enhances FSL for prompt tuning in multi-task settings. \citet{10.1145/3583780.3614913} is the first to showcase that prompt learning achieves SOTA performance for MTL in FSL settings, even surpassing ChatGPT.
Hierarchical Prompt (HiPro) learning~\citep{liu2023hierarchical} evaluates prompt tuning on standard MTL datasets and outperforms SOTA MTL methodologies by learning task-shared and task-individual prompts. Multitask Vision-Language Prompt Tuning (MVLPT)~\citep{shen2024multitask} incorporates cross-task knowledge into learning a single transferable prompt for vision-language models (VLMs). Prompt Guided Transformer (PGT)~\citep{lu2024prompt} introduces a prompt-conditioned Transformer block, integrating task-specific prompts into the self-attention mechanism, achieving global dependency modeling and parameter-efficient feature adaptation across multiple tasks. PromptonomyViT (PViT) model, as introduced in \citet{herzig2024promptonomyvit}, leverages prompts to capture task-specific information in video Transformers. 

Prefix-tuning~\citet{li2021prefix} is another lightweight alternative to fine-tune LLMs for different tasks while also keeping model parameters frozen. Prefix-tuning learns a continuous task-specific vector prefixed to the subsequent tokens. It can obtain comparable performance in the full data setting and outperform fine-tuning in low-data settings.~\citet{chen2022unisumm} proposes a Unified few-shot Summarization (UniSumm) model pretrained on multiple text summarization tasks, which exhibits the capability to generalize to different few-shot tasks through the utilization of prefix-tuning. \citet{chong2023leveraging} trains a prefix transfer module to selectively leverage the knowledge from various prefixes according to the input text. Collaborative domain-Prefix tuning for cross-domain NER (CP-NER)~\citep{chen2023one} utilizes text-to-text generation, grounding domain-related instructions to transfer knowledge to new domain tasks. Prefix-tuning approaches highlight the importance of leveraging prefixes and domain-specific information for improving performance in multiple tasks.

\begin{applebox}{Remarks}
\begin{enumerate}[leftmargin=0.4cm, label=(\roman*)]
    \item Task prompting stands out as highly parameter-efficient, demanding fewer than $0.01\%$ of task-specific parameters even for models exceeding a billion parameters~\citep{lester2021power}.
    \item Task-specific prompts exhibit a remarkable degree of adaptability, affording the capacity for on-the-fly customization to accommodate a diverse set of tasks, thus enhancing the flexibility in managing a multitude of heterogeneous tasks simultaneously.
    \item Task prompting facilitates the achievement of few-shot and even zero-shot learning, empowering PFMs to effectively perform tasks with minimal to no examples.
    \item Researchers/practitioners can have fine-grained control over how the model performs different tasks, as prompts can be customized to guide the model behavior precisely.
    \item The prompt itself is not transferable across different PFMs, thus leading to the limitations of scalability and reusability of prompt designs.
    \item Human involvement in prompting, e.g., crafting prompts or selecting appropriate templates, is time-consuming and bias-inducing.
\end{enumerate}
\end{applebox}

\subsubsection{Unified Generalist Models}
The ambitious aspiration, shared by both research communities and industries, has always been to transition from specialization to unification, thereby constructing an ideal generalist model capable of addressing a diverse set of tasks with varying modalities. The advent of large language models (LLMs)

The blueprint of designing general-purpose multimodal foundation models aligns with the recent unified models such as Gato~\citep{reed2022generalist}, Unified-IO~\citep{lu2022unified}, and OFA~\citep{wang2022ofa}, Uni-Perceiver~\citep{zhu2022uni, li2023uni}, etc. These methods can perform a variety of tasks spanning from CV to NLP, without modality limitations. Please see Fig.~\ref{fig:unified-mtl} as an illustration.
\input{tex_files/02-3/unified-mtl}
To pretrain via a Transformer backbone for the general MTL usage, we need to tokenize the input multi-modal data. For images, the commmon practice should obey the sequencing of non-overlapping $16\times 16$ patches in raster order in ViT~\citep{dosovitskiy2020image}, with the size of $256/16$ for each patch. Typically, the bounding boxes of objects in region-based tasks are represented by the quantization scheme of Pix2Seq~\citep{chen2022pixseq}. In the text preprocessing, the OFA framework adopts the exact same BPE Tokenizer~\citep{sennrich2015neural} used in BART~\citep{lewis-etal-2020-bart}, and its tokens are originally ordered along with the raw input text. Based on this prepossessing, it is possible to build a unified vocabulary for all visual, linguistic, and multi-modal tokens. After that, suppose we are given a sequence of tokens $\mathbf{x}_{i,b}$ as input, where $i=1,\cdots,I$ indexes the tokens in a data sample and $b=1,\cdots,B$ indexes a sample in a training batch. The architecture for a unified model is parametrized by $\theta$. Then we are able to autoregressively train the model via the chain rule as follows:
\begin{align}
    \mathcal{L}_\theta(\mathbf{x}_{1,1},\cdots,\mathbf{x}_{i,b}) = \sum_{b=1}^B\log\prod_{i=1}^{I}p_{\theta}(\mathbf{x}_{i,b}|\mathbf{x}_{1,b},\cdots,\mathbf{x}_{i-1,b}) = \sum_{b=1}^B\sum_{i=1}^{I}\log p_{\theta}(\mathbf{x}_{i,b}|\mathbf{x}_{<i,b})
\end{align}

\begin{applebox}{Remarks}
\begin{enumerate}[leftmargin=0.4cm, label=(\roman*)]
    \item The unified generalist model allows for modality-agnostic and task-agnostic learning, overcoming the limitations inherent to specific tasks. This implies that any task can be modeled into an omnivorous model.
    \item The unified generalist model achieves parameter efficiency and saves storage space in terms of many tasks.
    \item The unified generalist model is pre-trained using multimodal data all at once but possesses enduring utility.
\end{enumerate}
\end{applebox}


The concept of a unified architecture for multi-modal MTL can be traced back to OmniNet~\citep{pramanik2019omninet}, taking insights from the potentials of Transformers such as, \citet{pramanik2019omninet} propose a single model in their work to support tasks with multiple input modalities as well as asynchronous MTL. 
\citet{lu202012} investigates the relationships between vision-language (VL) tasks, and proposes a single model targeting 12 datasets simultaneously. \citet{li2021towards} introduces the concept of unified foundation models by jointly pre-training Transformers on unpaired images and text data. Unified Transformer (UniT) model~\citep{hu2021unit} is a realization of this concept. It first features separate encoders for different input modalities and a shared decoder over the encoded input representations. Each task is associated with specific heads in the shared decoder.
Unified Foundation Model
\citet{wang2022ofa, bai2022ofasys} proposes One-for-All (OFA) as a task-agnostic and modality-agnostic framework. OFA aims to unify task-specific layers for downstream tasks, providing a versatile solution. However, it is important to note that OFA currently lacks support for video data and necessitates fine-tuning for downstream tasks. Uni-Perceiver~\citep{zhu2022uni} is a unified architecture for generic perception for zero-shot and few-shot tasks, which includes a video tokenizer with temporal positional embeddings. Uni-Perceiver v2~\citep{li2023uni} further introduces task-balanced gradient normalization to ensure stable MTL, which enables larger batch-size training for various tasks. More importantly, unlike OFA~\citep{wang2022ofa}, Uni-Perceiver v2 requires no task-specific adaptation. Mask DETR with Improved deNoising anchOr boxes (Mask DINO)~\citep{li2023mask} is a unified framework designed for object detection and segmentation. Mask DINO uses an additional mask prediction branch to unify the query selection for masks. All-in-one Transformer~\citep{wang2023all} unifies video and text encoders via introducing a token rolling operation to encode temporal representations from videos. Omnivorous Masked Auto-Encoder(OmniMAE)~\citep{girdhar2023omnimae} shows that MAE can be used to pretrain a ViT on images and videos without any human labels. OmniVec~\citep{srivastava2024omnivec} also pretrains a unified architecture from self-supervised masked data, including visual, audio, text, and 3D, which realizes the cross-modal task generalization.





%% file: tex_files/02-3/PFM_framework.tex
\begin{figure*}[t]
    \centering
    \begin{subfigure}{0.3\textwidth}
        \includegraphics[height=2.3\textwidth]{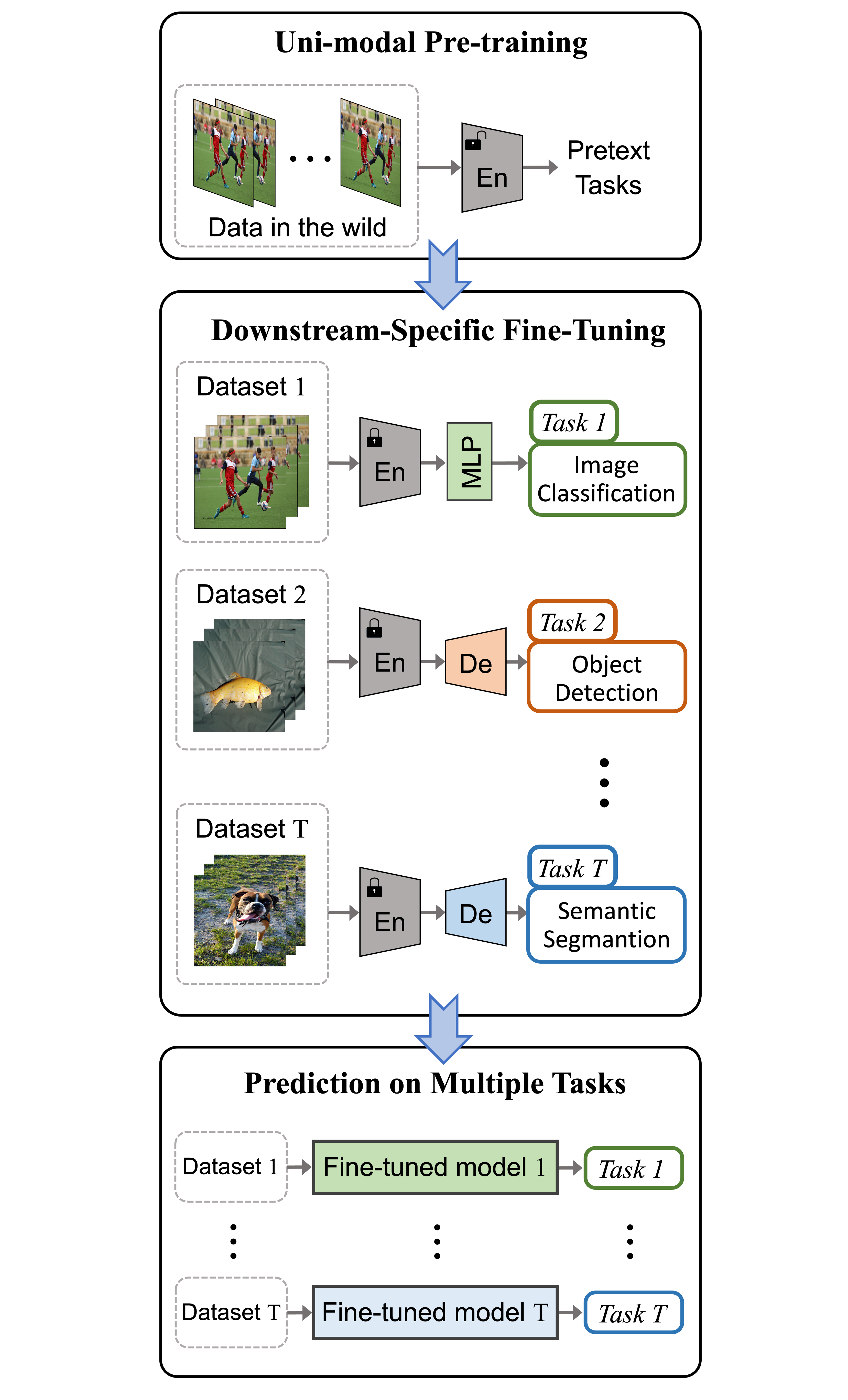}
        \caption{Downstream Fine-Tuning.}
        \label{fine-tuning}
    \end{subfigure}\quad\quad\quad
    \begin{subfigure}{0.3\textwidth}
        \includegraphics[height=2.3\textwidth]{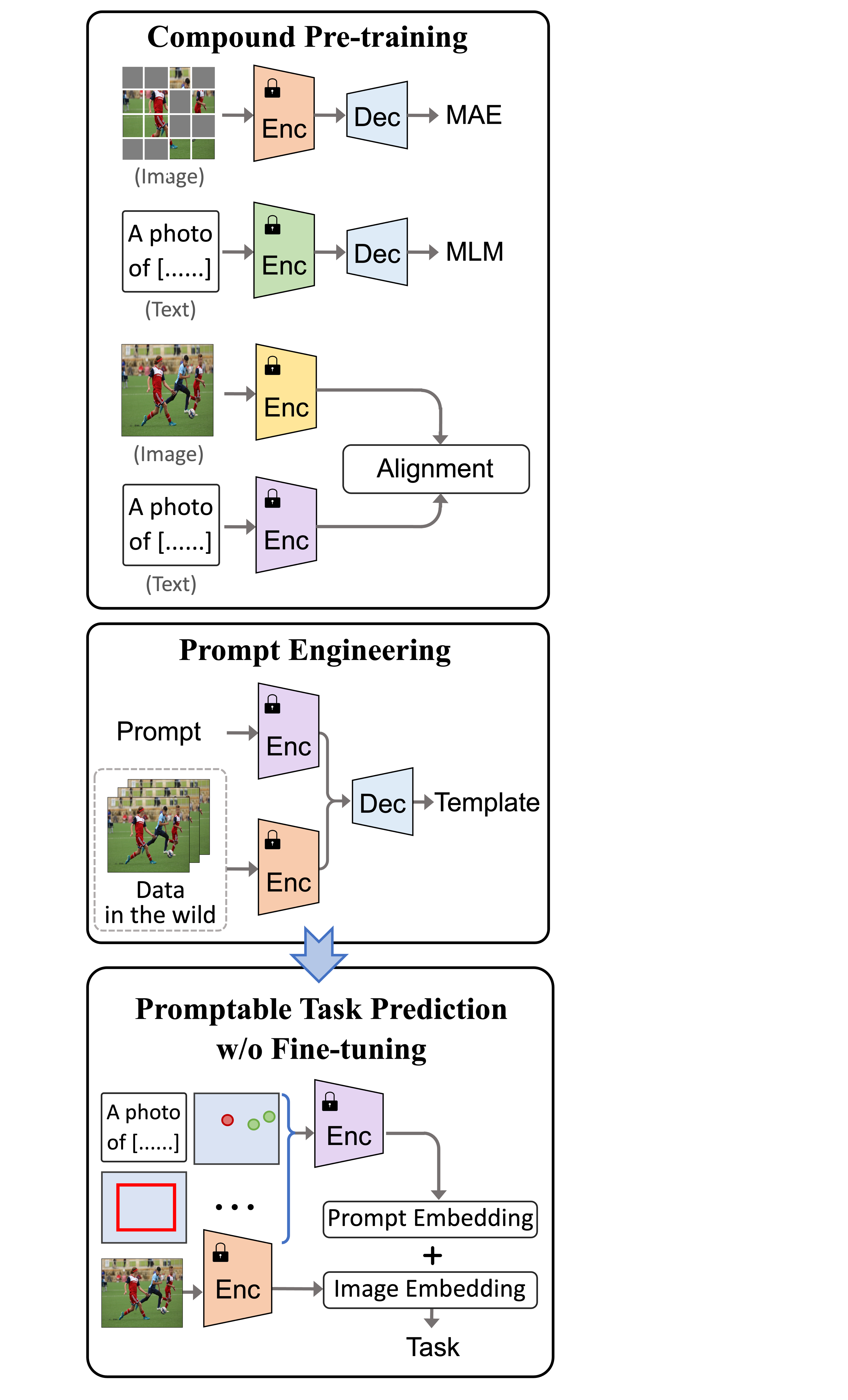}
        \caption{Task Prompting.}
        \label{task-prompting}
    \end{subfigure}
    \begin{subfigure}{0.3\textwidth}
        \includegraphics[height=2.3\textwidth]{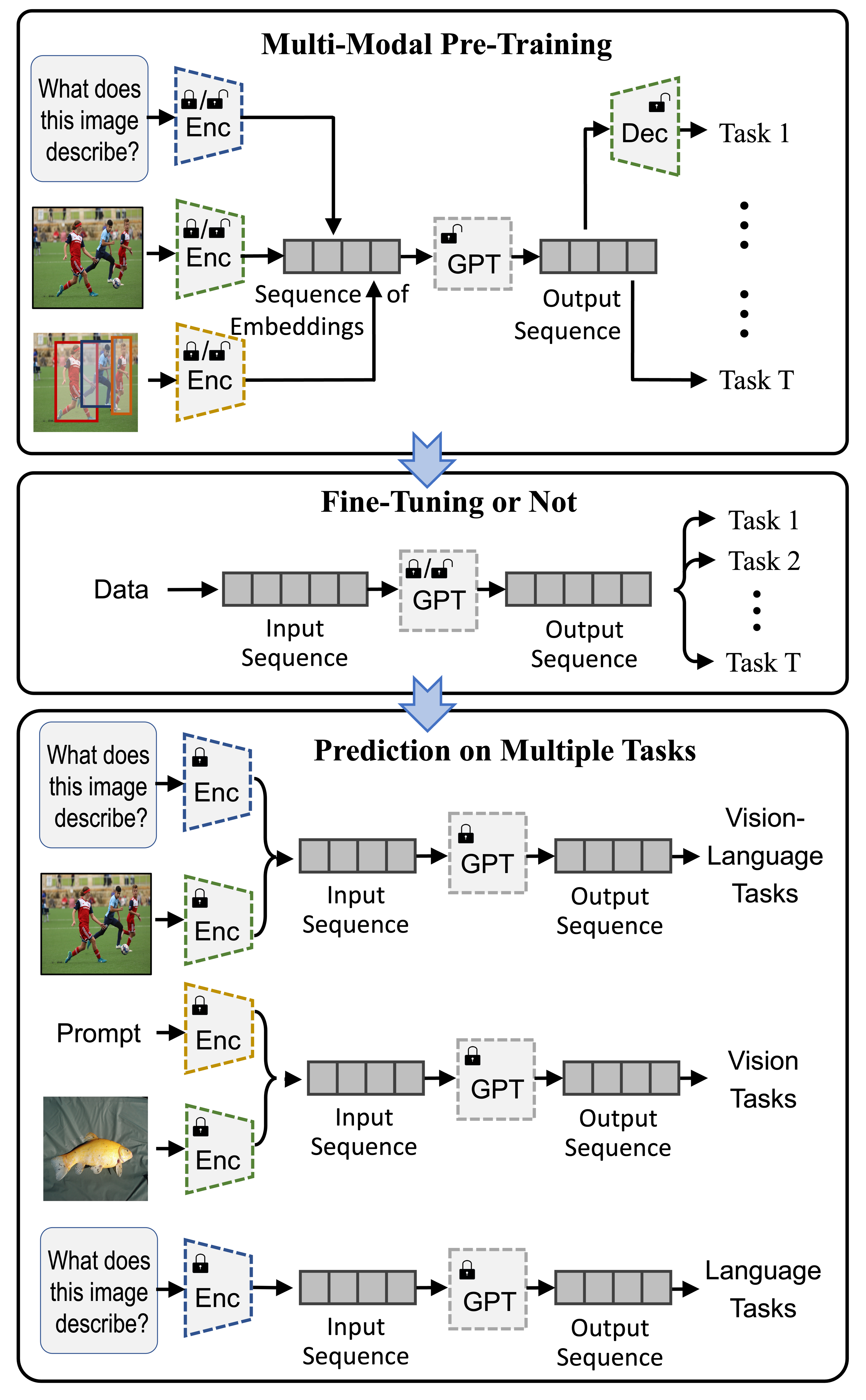}
        \caption{Unified Generalist Model.}
        \label{unified-generalist-model}
    \end{subfigure}
    \caption{The taxonomy of PFMs of MTL into three categories: (A) Downstream Task Fine-Tuning (B) Task Prompting (C) Unified Generalist Model.}
    \label{PFM-taxonomy}
\end{figure*}

%% file: tex_files/02-3/unified-mtl.tex
\begin{figure*}[t]
    \centering
    \includegraphics[width=0.85\linewidth]{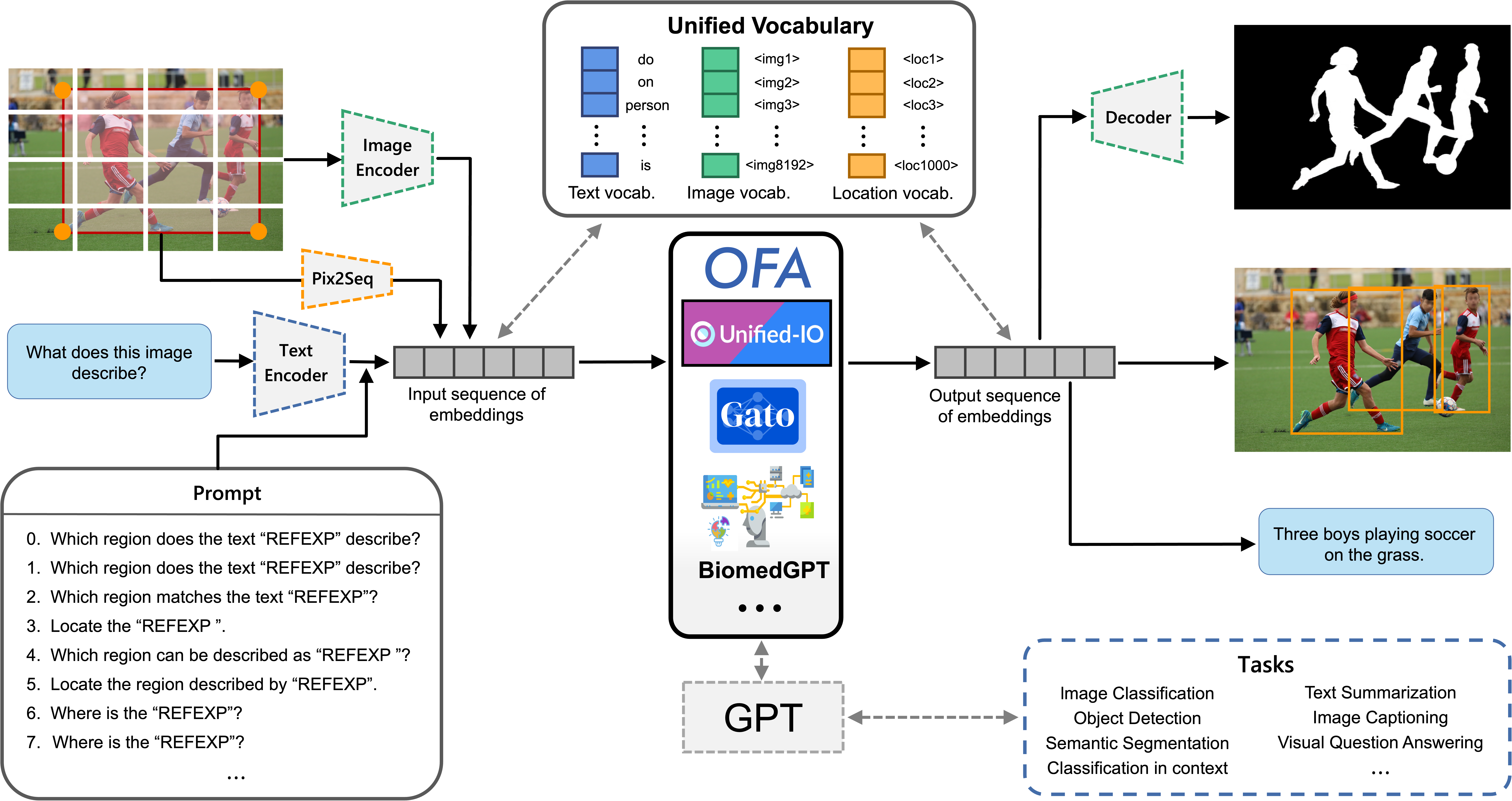}
      \caption{The Framework of unified generalist model, which can unify architectures, tasks, and modalities through a simple seq2seq learning architecture.}
    \label{fig:unified-mtl}
\end{figure*}

%% file: 03_miscellaneous.tex

\section{Miscellaneous}
\label{sec:miscellaneous}

\subsection{Fairness and Bias in MTL}
\label{subsec:fairness}
{~~}\vspace{2pt}\\
While most of the existing research about bias and fairness implications primarily focuses on STL~\citep{mehrabi2021survey}, \citet{wang2021understanding} pioneer the exploration of the fairness-accuracy trade-off within the MTL setting. The challenge of unaligned fairness goals arises in MTL models that optimize accuracy for all tasks. The introduction of novel multi-task fairness metrics, such as average relative fairness gap and average relative error, aids in quantifying this trade-off in MTL applications. \citet{li2023fairness} emphasize that misspecification of majority and minority groups in involved tasks disproportionately affects minority tasks, and they propose over-parameterization as a viable solution to achieve fairness by covering all tasks. \citet{hu2023fairness} extend the definition of Strong Demographic Parity~\citep{agarwal2019fair, jiang2020wasserstein} to MTL using multi-marginal Wasserstein barycenters~\citep{chzhen2020fair}, providing an optimal fair multi-task solution to the fairness-accuracy trade-off. Additionally, \citet{roy2022learning} further demonstrates that improving fairness can positively impact accuracy performance. Learning to Teach Fair Multi-Tasking (L2T-FMT)~\citep{roy2022learning} introduces a teacher-student network to address fair MTL problems. In this framework, the teacher guides the student in selecting fairness or accuracy objectives during training, offering a dynamic approach to balancing these objectives.
Drawing an analogy, \citet{roy2023fairbranch} liken the negative impact of task-specific fairness to negative transfer and introduces FairBranch, a method that groups related tasks to mitigate this negative transfer through fairness loss gradient conflict correction. In recent years, prioritizing fair MTL to mitigate biases arising from negative transfer has emerged as a promising direction. This approach can ensure that models treat all tasks fairly, avoiding disproportionate impacts on specific groups or tasks. By preventing biased outcomes, fair MTL contributes to averting potential societal harm.
\subsection{Security and Privacy in MTL}
\label{subsec:mtlsp}
{~~}\vspace{2pt}\\
\emph{Attack and Defense.}
MTL is an impactful technique employed to bolster attacks in diverse sectors. It notably expedites the creation of adversarial examples for numerous tasks simultaneously through the exploitation of task-shared knowledge \citep{guo2020multi}. In the field of automatic speaker verification, multi-task learning strategies have been utilized to identify replay attack spoofing and to classify different types of replay noise \citep{shim2018replay}. With regard to reinforcement learning, the vulnerability of multi-task federated reinforcement learning algorithms to adversarial attacks has been examined, resulting in the development of an adaptable attack method and a refined federated reinforcement learning algorithm \citep{anwar2021multi}. Additionally, within the realm of deep reinforcement learning, a multi-objective strategy for developing attack policies has been suggested, considering both the performance degradation and the cost related to the attack \citep{garcia2020learning}.
Conversely, MTL can also serve as a means to heighten the model's resilience, leading to an improved defense against a wide array of malicious attacks. For instance, the robustness of models to adversarial attacks on individual tasks has been shown to increase when models are trained on multiple tasks concurrently \citep{mao2020multitask, guo2020multi}. Likewise, multi-task learning has been employed for adversarial defense \citep{naseer2022stylized}, using supplementary data from the feature space to design more formidable adversaries and boost the model's resilience. Through the utilization of multi-task objectives, such as cross-entropy loss, feature-scattering, and margin losses, more powerful perturbations can be devised for adversarial training. This technique has been used in several domains, such as computer vision and speech recognition, and has demonstrated enhanced adversarial accuracy and resilience \citep{pal2021adversarial, chan2021multiple}.


\emph{Privacy-preserving.}
Privacy-preserving multi-task learning (PP-MTL)~\citep{liu2018privacy} aims to ensure the confidentiality of sensitive data and boost learning outcomes by facilitating knowledge transfer across related tasks. PP-MTL algorithms employ cryptographic mechanisms to safeguard data residing across various locations or nodes, using these to relay cumulative data - for instance, gradients or supports - to a centralized server where the aggregated data is processed to create the desired models. Existing strategies cannot deliver a demonstrable or verifiable security assurance for the transferred cumulative data. To tackle this shortcoming, various innovative PP-MTL protocols have been suggested, leveraging cutting-edge cryptographic methods to deliver the strongest possible security assurance \citep{liu2018privacy}. Furthermore, differential private stochastic gradient descent algorithms have been employed to optimize the comprehensive multi-task model and safeguard the privacy of training data by introducing appropriately calibrated noise to the gradient of loss functions \citep{zhang2020privacy}. To maintain the privacy of distributed data, privacy-preserving distributed MTL frameworks have been introduced, incorporating a privacy-preserving proximal gradient algorithm. This algorithm updates models asynchronously and offers guaranteed differential privacy \citep{xie2017privacy}.

\emph{Federated Learning.}
Federated Multi-task Learning (FMTL)~\citep{smith2017federated} represents a platform for training machine learning models over distributed device networks. By personalizing models for individual clients, it successfully navigates the statistical complexities posed by federated learning, given the heterogeneity of local data distributions \citep{smith2017federated}. It effectively manages high communication overhead, lags, and reliability in distributed multi-task learning \citep{marfoq2021federated}. The efficacy of FMTL has been demonstrated on real-world federated datasets, even with non-convex models \citep{sarcheshmehpour2021networked}. It can be utilized in both a central server-client and a fully decentralized structure and provides the capacity to serve personalized models to clients unseen during training \citep{corinzia2019variational}. Furthermore, the over-the-air computation can be integrated within FMTL to enhance system efficiency, reducing channel usage without a substantial drop in learning performance \citep{ma2022over}.

\subsection{Distribution Shifts in MTL}
{~~}\vspace{2pt}\\
\label{sec:distri_shift}
While Multi-Task Learning (MTL) excels at leveraging shared information to boost individual task performance (\myref{motiv}), its real-world applicability often hinges on its ability to adapt to unforeseen data distributions. Distribution shifts, where the data encountered during deployment deviates from the training distribution, are omnipresent challenges that can significantly degrade MTL performance, especially on new tasks or domains. Recognizing and mitigating these shifts is crucial not just for maintaining the generalizability and resilience of MTL models but also for unlocking their full potential in real-world applications.

Recent research offers a diverse arsenal of approaches to tackle distribution shifts in MTL. Vision Transformer Adapters (ViTA) \citep{bhattacharjee2023vision} introduce dedicated modules within the model architecture that enhance adaptability to diverse tasks and data distributions. Techniques like regularizing spurious correlations \citep{hu2022improving} target misleading associations between tasks, reducing their influence on the overall model performance. Scalarization methods provide a scalable framework for handling the complexities of multi-task and multi-domain learning while facing distribution shifts \citep{royer2023scalarization}. Multi-objective learning strategies, exemplified by approaches addressing catastrophic forgetting in time-series applications \citep{10.1145/3502728}, strive to mitigate the issue of forgetting previously learned skills when encountering new data. Finally, techniques like reward modeling \citep{faal2023reward} demonstrate their versatility in addressing distribution shifts, as seen in mitigating toxicity issues in transformer-based language models. This array of advancements underscores the ongoing efforts to equip MTL models with enhanced adaptability and resilience to varying task distributions, ultimately paving the way for their reliable and widespread real-world application.

Looking ahead, the evolving landscape of MTL research envisions models that not only react to distribution shifts but proactively anticipate and address them. As highlighted in a recent comprehensive study \citep{adhikarla2023robust}, understanding and mitigating distribution shifts are becoming paramount for MTL's success. The ability to navigate diverse and dynamic data distributions is crucial for the broader deployment of MTL in complex, real-world scenarios. By advancing techniques that enhance adaptability and robustness, researchers are striving to empower MTL models to excel in the face of evolving task and domain landscapes, unlocking their potential to revolutionize a wide array of applications.


\subsection{Non-supervised MTL}
{~~}\vspace{2pt}\\
 \emph{semi-supervised learning.}
 Supervised learning has been a fundamental technique in machine learning in recent years. However, it faces the limitation of requiring a substantial amount of labeled data to yield promising results, a process that is both time-consuming and costly. To mitigate this, semi-supervised learning has been introduced, leveraging the diverse array of unlabeled datasets to reduce the dependence on labeled data. Previous existing semi-supervised algorithms are not often amenable to MTL, for instance, \citep{liu2007semi} introduces a semi-supervised multitask learning (MTL) framework, featuring $M$ parameterized classifiers. Each classifier is associated with a partially labeled data manifold and is jointly learned under a soft-sharing prior that influences their parameters. This approach effectively utilizes unlabeled data by basing the learning of classifiers on neighborhood structures. Besides, \citep{augenstein2018multi} presents a method that models the relationship between labels by inducing a joint label embedding space for multi-task learning and proposes a $Tranfer Network$ which learns to transfer labels between tasks and uses semi-supervised learning to leverage them for training. In real-world applications, multi-task regression is a prevalent challenge. \citep{zhang2009semi} proposes the SMTR method, which is grounded in Gaussian Processes (GP). This method operates under the assumption that the kernel parameters for all tasks share a common prior. To enhance SMTR, the approach incorporates unlabeled data by modifying the GP prior's kernel function into a data-dependent one. This modification leads to a semi-supervised extension of the original SMTR method, aptly named SSMTR. Additionally, \citep{chen2020multi} introduces a multi-task mean teacher model for semi-supervised shadow detection, effectively utilizing unlabeled data and simultaneously learning multiple aspects of shadows. Specifically, they construct a multi-task baseline model designed to detect shadow regions, edges, and count, leveraging the complementary information of these elements. This baseline model is then implemented in both student and teacher networks. The approach further involves aligning the predictions from the three tasks across these networks, using this alignment to compute a consistency loss on unlabeled data. This loss is combined with the supervised loss from labeled data based on the predictions of the multi-task baseline model, thereby enhancing the model's learning effectiveness. \citep{nguyen2019multi} proposed a network employing a multi-task learning approach to detect manipulated images and videos and to identify the manipulated regions within each query. To enhance the network's generalizability, a semi-supervised learning approach is integrated in which the architecture comprises an encoder and a Y-shaped decoder. The activation of encoded features facilitates binary classification. Meanwhile, the outputs of the decoder's branches serve distinct purposes: one for segmenting the manipulated regions and the other for reconstructing the input. This dual functionality significantly contributes to the improvement of the overall performance of the network. Semi-supervised multitask learning (MTL) has emerged as a popular field, with various preceding studies, as mentioned above, that propose different mechanisms that integrate semi-supervised concepts. These studies have demonstrated their effectiveness through numerous experimental results. Despite these advancements, there remains a substantial scope for further research in this subfield. Continued exploration in semi-supervised MTL promises to yield many more valuable insights and findings.
 
\emph{unsupervised learning.}
Moving beyond the realm of semi-supervised learning, the real-world often presents scenarios where obtaining labeled data of all tasks in MTL learning is not feasible, underscoring the significance of unsupervised learning in the field of multitask learning (MTL). OpenAI, in their groundbreaking study by \citep{radford2019language}, introduced the widely acclaimed GPT model, demonstrating a significant advancement in multitask learning (MTL) within the field of natural language processing. Their research showed that language models begin to autonomously learn a variety of MTL tasks - including question answering, machine translation, reading comprehension, and summarization - without the need for explicit supervision. This capability was notably observed when the GPT model was trained on $WebText$, a vast new dataset comprising millions of webpages. This development highlights a major stride in the field, showcasing the potential of large language models to adapt to a wide array of tasks through extensive unsupervised learning. Besides, to alleviate the limitation of existing clustering approaches that neglect the underlying relationship and treat these clustering tasks either individually or simply together, the study by \citep{5360241} introduces an innovative clustering approach called $Multi-task  clustering$, which conducts several related clustering tasks concurrently and leverages the relationships between these tasks to improve clustering performance. This approach comprises two key components: (1) Within-task clustering, which involves clustering the data for each task individually within its own input space, and (2) Cross-task clustering, where the shared subspace is learned simultaneously, and the data from all tasks are clustered together. This dual-faceted strategy optimizes the clustering results by combining individual task insights with cross-task synergies. Another notable example is in the context of point cloud tasks,  where \citep{hassani2019unsupervised} introduces an unsupervised multi-task model. This model is designed to concurrently learn point and shape features. It incorporates three unsupervised tasks: clustering, reconstruction, and self-supervised classification. These tasks are used to train a multi-scale graph-based encoder. Beyond, \citep{argyriou2006multi} introduces a method for learning a low-dimensional representation shared across multiple related tasks. This method extends the well-known 1-norm regularization problem by incorporating a novel regularizer that controls the number of features common to all tasks. The authors demonstrate that this approach can be formulated as a convex optimization problem and develop an iterative algorithm to solve it. The algorithm operates in a dual-step manner: it alternates between a supervised step and an unsupervised step. In the unsupervised step, it learns representations common across tasks, while in the supervised step, it utilizes these common representations to learn task-specific functions. This approach effectively combines supervised and unsupervised learning techniques to enhance multi-task learning.
\subsection{Others}




\subsubsection{Applications of MTL}
In the DL era, the advancement of multimodal analysis and MTL paradigms has brought challenges and also opened up fantastic probabilities to the realm of MTL. In addition to the applications investigated in the paper, MTL plays an important role in many different fields such as visual assessment \citep{yu2019towards, zhang2023blind}, healthcare\citep{zhang2023knowledge, zhao2023multi, zhang2023biomedgpt}, transportation\citep{wang2023multi, feng2023forecast}, language models \citep{liu2020multi, hu2021unit} and recommender systems\citep{zhang2023advances, deng2023unified}. Briefly,\citet{zhang2023blind} develop a general and automated multitask learning scheme for image quality assessment by blind individuals. \citet{zeng2023new} combine MTL algorithms with a deep belief network for the diagnosis of Alzheimer’s disease. Wang \textit{et al}. \citep{wang2023multi} propose a multi-task Weakly supervised learning framework to infer transition probability between road segments. \citet{gao2023enhanced} utilize the relation-aware GCNs to fully capture the multi-relation neighborhood features.

Despite the achievements in recent years, many outstanding MTL approaches still suffer from limitations that restrict their application to certain real-world scenarios. For example, it is difficult to capture the complex inter-scenario correlations with multiple tasks. Besides, in large-scale tasks, it remains a challenge to design scalable models and deal with the parameter explosion issue. Therefore, the scalability of MTL models is still a direction worth exploring \citep{zhang2023advances}.



\subsubsection{MTL+X}

\emph{MTL + Continual Learning.} Biased forgetting of previous knowledge caused by new tasks remains challenging in continual learning. \citet{lyu2021multi} propose Multi-Domain Multi-Task (MDMT) rehearsal to train the old tasks and new tasks together while keeping tasks from isolation. \citet{he2019task} utilize meta-learning to achieve task-agnostic continual learning. MTL is a promising technique to mitigate catastrophic forgetting via learning task-relatedness.

\emph{Multi-Task Reinforcement Learning (MTRL).} MTRL~\citep{vithayathil2020survey} holds promise in the context of Reinforcement Learning (RL), given the natural presence of diverse tasks like reach, push and pick in robotic manipulation. In the early stage, \citet{wilson2007multi} approaches it as the solution to a sequence of Markov Decision Processes (MDPs) and employs a hierarchical Bayesian framework to infer the characteristics of new environments based on knowledge gained from previous environments. \citet{hessel2019multi} introduce a method to automatically adjust the contribution of each task to the updates of a single agent. This ensures that all tasks exert a similar impact on the learning dynamics. \citet{taiga2022investigating} investigates multi-task pretraining and generalization in RL.
\citet{cheng2023multi} propose an attention-based multi-task reinforcement learning approach to learn a compositional policy for each task.

%% file: 04-resource.tex
\section{Resources}
\label{benchmark}
In this section, we offer useful tools and resources that can help researchers and practitioners implement MTL models.

\input{table_files/dataset}
\subsection{Dataset}
{~~}\vspace{2pt}\\
In this section, we introduce benchmark datasets for MTL from a taxonomic perspective. Specifically, based on the different datasets spawning a series of typical data-driven models, we classify many MTL datasets into three categories: regression task, classification task, and dense prediction task.

\subsubsection{Regression task}
\vspace{5mm}
\emph{Synthetic Data.} This dataset is often artificially defined by researchers, thus different from one another, e.g.~\citet{caruana1997multitask, bakker2003task, evgeniou2004regularized, argyriou2008convex, jalali2010dirty, zhou2011clustered, titsias2011spike, zhang2012convex, maurer2013sparse, han2016multi, parra2017spectral, nie2018calibrated, ma2018modeling}, to name a few. The features are often generated via drawing random variables from a shared distribution and adding irrelevant variants from other distributions, and the corresponding responses are produced by a specific computational method. In such a manner, data in different tasks would contain both the task-specific and -shared features that contribute to the learning for estimation.

\emph{School Data.}
~\citet{mortimore1988school} comes from the Inner London Education Authority (ILEA) and contains $15,362$ records of student examination, which are described by $27$ student- and school-specific features from $139$ secondary schools. The goal is to predict exam scores from $27$ features, and the prediction in $139$ schools would be generally handled as $139$ tasks.

\emph{SARCOS Data.}\footnote{2000. SARCOS Data. \url{gaussianprocess.org/gpml/data}} This dataset is in humanoid robotics consists of $44,484$ training examples and $4,449$ test examples. The goal of learning is to estimate the inverse dynamics model of a $7$ degrees-of-freedom (DOF) SARCOS anthropomorphic robot arm, each of which corresponds to a task and contains 21 features---7 joint positions, 7 joint velocities, and 7 joint accelerations.
\emph{Computer Survey Data.}~\citet{lenk1996hierarchical} is from a survey on the likelihood (11-point scale from 0 to 10) of purchasing personal computers. There are $20$ computer models as examples, each of which contains 13 computer descriptions (e.g., price, CPU speed, and screen size) and 6 subject-level covariates (e.g., gender, computer knowledge, and work experience) as features and ratings of $179$ subjects as targets, i.e., tasks.
\emph{Climate Dataset.}\footnote{2017-now. Climate Dataset. \url{www.cambermet.co.uk}} This real-time dataset is collected from a sensor network (e.g., anemometer, thermistor, and pressure transducer) of four climate stations---Cambermet, Chimet, Sotonmet and Bramblemet---in the south on England, which can represent $4$ tasks as needed. The archived data are reported in 5-minute intervals, including $\sim10$ climate signals (e.g., wind speed, wave period, barometric pressure, and water temperature). Generally, air temperature is considered as the dependent variable and others as independent~\citep{parra2017spectral, zhao2019multiple}.

\subsubsection{Classification task}


\emph{20 Newsgroups.}
~\citet{Lang95} is a collection of approximately $19,000$ netnews articles, organized into $20$ hierarchical newsgroups according to the topic, such as root categories (e.g., comp, rec, sci, and talk) and sub-categories (e.g., comp.graphics, sci.electronics, and talk.politics.guns). Users can design different combinations as multiple text classifications tasks~\citep{he2011graphbased, tan2015transitive, zhang2018multi, mao2020adaptive, xiao2020efficient}.

\emph{Reuters-21578 Collection.}\footnote{1996. Reuters-21578 Collection. 
\url{www.daviddlewis.com/resources/testcollections/reuters21578/}} This text collection contains 21578 documents from Reuters newswire dating back to 1987. These documents were assembled and indexed with more than 90 correlated categories---5 top categories (i.e., exchanges, orgs, people, place, topic), and each of them includes variable sub-categories. Users can independently define the related multiple tasks by choosing different combinations of categories, e.g., \citet{zheng2020multi, xiao2021new} provide more detailed descriptions.

\emph{CelebA Dataset.}
CelebFaces Attributes Dataset (CelebA)~\citep{liu2018large} is a large-scale face attributes dataset with more than 200K celebrity images, each with 40 attribute annotations. The images in this dataset cover large pose variations and background clutter. CelebA has large diversities, large quantities, and rich annotations, including 10,177 identities, 202,599 face images, and 5 landmark locations, 40 binary attribute annotations per image. The dataset can be employed as the training and test sets for the following computer vision tasks: face attribute recognition, face recognition, face detection, landmark (or facial part) localization, and face editing \& synthesis.
\emph{MultiMNIST Dataset.}
This dataset originated from validating a capsule system~\citep{sabour2017dynamic}, but it is also a MTL version of MNIST dataset
~\citep{lecun1998gradient}. By overlaying multiple images together, traditional digit classification is converted to an MTL problem, where classifying the digits in different positions is considered as distinctive task. \citet{sener2018multi} contributes a standard construction for the research community.
\emph{ImageCLEF-2014 Dataset.}\footnote{2014. ImageCLEF-2014. \url{www.imageclef.org/2014/adaptation}} This dataset is a benchmark for domain adaptation challenge, which contains $2,400$ images of 12 common categories selected from 4 domains: Caltech 256, ImageNet 2012, Pascal VOC 2012, and Bing. These 4 domains are commonly considered as different tasks in MTL.

\emph{Office-Caltech Dataset.}
~\citet{gong2012geodesic} is a standard benchmark for domain adaption in computer vision, consisting of real-world images of 10 common categories from the Office dataset and Caltech-256 dataset. There are $2,533$ images from 4 distinct domains/tasks: Amazon, DSLR, Webcam, and Caltech.

\emph{Office-31 Dataset.}
~\citet{saenko2010adapting} consists of 4,110 images from 31 object categories across 3 domains/tasks: Amazon, DSLR, and Webcam. 

\emph{Office-Home Dataset.}
~\citet{venkateswara2017deep} is collected for object recognition to validate domain adaptation models in the era of DL, which includes $15,588$ images in office and home settings (e.g., alarm clock, chair, eraser, keyboard, telephone, etc.) organized into 4 domains/tasks: Art (paintings, sketches and artistic depictions), Clipart (clipart images), Product (product images from \url{www.amazon.com}), and Real-World (real-world objects captured with a regular camera).

\emph{DomainNet Dataset.}
~\citet{peng2019moment} is annotated for the purpose of multi-source unsupervised domain adaptation (UDA) research. It contains $\sim 0.6$ million images from 345 categories across 6 distinct domains, e.g., sketch, infograph, quickdraw, real, etc.

\emph{SYNTHIA Dataset.}
~\citet{ros2016synthia} is a synthetic dataset created to address the need for a large and diverse collection of images with pixel-level annotations for vision-based semantic segmentation in urban scenarios, particularly for autonomous driving applications. It consists of precise pixel-level semantic annotations for 13 classes, including sky, building, road, sidewalk, fence, vegetation, lane-marking, pole, car, traffic signs, pedestrians, cyclists, and miscellaneous objects.

\emph{SVHN Dataset.}
Street View House Numbers (SVHN)~\citep{yang2021few} is a digit classification benchmark dataset that contains 600,000 32×32 RGB images of printed digits (from 0 to 9) cropped from pictures of house number plates. The cropped images are centered in the digit of interest, but nearby digits and other distractors are kept in the image. SVHN has three sets: training, testing sets and an extra set with 530,000 images that are less difficult and can be used for helping with the training process.

\emph{Deepfashion Dataset.}
DeepFashion~\citep{liu2016deepfashion} is a large-scale clothes dataset with comprehensive annotations. It contains over 800,000 images, which are richly annotated with massive attributes, clothing landmarks, and correspondence of images taken under different scenarios including store, street snapshot, and consumer.

\emph{ACE05 Dataset.}\footnote{2005. ACE05 Dataset. \url{catalog.ldc.upenn.edu/LDC2006T06}} The ACE 2005 Multilingual Training Corpus comprises the comprehensive collection of training data in English, Arabic, and Chinese for the 2005 Automatic Content Extraction (ACE) technology evaluation. The corpus includes diverse data types that have been annotated for entities, relations, and events. The Linguistic Data Consortium (LDC), with support from the ACE Program and additional assistance from LDC, carried out the annotation of this dataset.

\emph{ATIS Dataset.}
The ATIS (Airline Travel Information Systems) dataset ~\citep{hemphill-etal-1990-atis} comprises audio recordings along with corresponding manual transcripts of human interactions with automated airline travel inquiry systems. These interactions involve individuals seeking flight-related information. The dataset includes 17 distinct intent categories representing different user intents. In the original data split, the training set contains 4,478 intent-labeled reference utterances, the development set contains 500 utterances, and the test set contains 893 utterances.


\subsubsection{Dense prediction task}
\emph{CityScapes Dataset.}
~\citet{cordts2016cityscapes} consists of 5,000 images with high-quality annotations and 20,000 images with coarse annotations from 50 different cities, which contains 19 classes for semantic urban scene understanding. Specifically, pixel-wise semantic and instance segmentation together with ground truth inverse depth labels are often used as three different tasks~\citep{kendall2018multi, liu2019end} in MTL.
\emph{NYU-Depth Dataset V2.}
~\citet{silberman2012indoor} is comprised of 1,449 images from 464 indoor scenes across 3 cities, which contains 35,064 distinct objects of 894 different classes. The dense per-pixel labels of class, instance, and depth are used in many computer vision tasks, e.g., semantic segmentation, depth prediction, and surface normal estimation~\citep{eigen2015predicting}.
\emph{PASCAL VOC Project.}
\footnote{2005. Pascal VOC Project. \url{host.robots.ox.ac.uk/pascal/VOC}} 
This project~\citep{everingham2010pascal} provides standardized image datasets for object class recognition and also has run challenges evaluating performance on object class recognition from 2005 to 2012, where VOC07\footnote{2007. Pascal VOC Challenge 2007. \url{host.robots.ox.ac.uk/pascal/VOC/voc2007/index.html}}, VOC08\footnote{2008. Pascal VOC Challenge 2008. \url{host.robots.ox.ac.uk/pascal/VOC/voc2008/index.html}}, and VOC12\footnote{2012. Pascal VOC Challenge 2012. \url{host.robots.ox.ac.uk/pascal/VOC/voc2012/index.html}} are commonly used for MTL research. The multiple tasks cover classification, detection (e.g., body part, saliency, semantic edge), segmentation, attribute prediction~\citep{farhadi2009describing}, surface normals prediction~\citep{maninis2019attentive}, etc. Many of the annotations are labeled or distilled by the followers~\citep{chen2014detect, maninis2019attentive}. 

\emph{Taskonomy Dataset.}
~\citet{zamir2018taskonomy} is currently the most diverse product for computer vision in MTL, consisting of 4 million samples from 3D scans of $\sim600$ buildings. This product is a dictionary of 26 tasks (e.g., 2D, 2.5D, 3D, semantics, etc.) as a computational taxonomic map for task transfer learning. Accordingly, Tiny-Tasknomy~\citep{standley2020tasks} with 5 sampled dense prediction tasks, e.g., semantic segmentation, surface normal prediction, depth prediction, keypoint detection, and edge detection is considered a commonly used benchmark in MTL.



\subsubsection{Others}

\emph{EMMa Dataset.}
EMMa Dataset~\citep{standley2023extensible} comprises more than 2.8 million objects from Amazon product listings, each annotated with images, listing text, mass, price, product ratings, and its position in Amazon's product-category taxonomy. It includes a comprehensive taxonomy of 182 physical materials, and objects are annotated with one or more materials from this taxonomy. EMMa offers a new benchmark for multi-task learning in computer vision and NLP, allowing for the addition of new tasks and object attributes at scale.

\emph{STREET.}
STREET~\citep{ribeiro2023street} is a multi-task benchmark for structured reasoning and explanations in NLP. It consists of five existing datasets (ARC, SCONE, GSM8K, AQUA-RAT, and AR-LSAT) and introduces a unified reasoning formulation with textual logical units and reasoning graphs. Evaluation metrics and empirical performance analysis using T5-large and GPT-3 models are provided, along with error explanations on a per-dataset basis.

\emph{VKITTI2 Dataset.}
Virtual KITTI~\citep{gaidon2016virtual} is a new video dataset, automatically labeled with accurate ground truth for object detection, tracking, scene and instance segmentation, depth, and optical flow. Virtual KITTI 2~\citep{cabon2020virtual} is a more photo-realistic and better-featured version of the original virtual KITTI dataset. It exploits recent improvements of the Unity game engine and provides new data such as stereo images or scene flow.

\emph{XTREME.} 
The XTREME (Cross-lingual Transfer Evaluation of Multilingual Encoders)~\citep{hu2020xtreme} benchmark is a multi-task evaluation framework to assess the cross-lingual generalization capabilities of multilingual representations across 40 languages and 9 tasks. It highlights the performance disparity between models tested on English, which achieve human-level performance on numerous tasks, and cross-lingually transferred models, which exhibit a significant performance gap, particularly in syntactic and sentence retrieval tasks.

\input{table_files/software_covers.tex}

\subsection{Software Resources}
{~~}\vspace{2pt}\\
To provide playgrounds for researchers to fairly compare different state-of-the-art algorithms in a unified environment, open-source platforms for MTL merge out. Herein we introduce three popular software resources that aim at variant populations in terms of the implementation languages, algorithm comprehensiveness, downstream task realms, and modularization focuses.

\emph{Regularized Multi-Task Learning (RMTL).}\footnote{\url{cran.r-project.org/web/packages/RMTL/index.html}} It is a relatively small yet practical R library for MTL, especially for the ones on biological-related tasks. It includes ten algorithms applicable for regression, classification, joint predictor selection, task clustering, low-rank learning and incorporation of biological networks.

\emph{Multi-tAsk Learning via StructurAl Regularization (MALSAR).}\footnote{\url{github.com/jiayuzhou/MALSAR}} It is a MTL package implemented with Matlab. Compared to RMTL, it does not particularly focus on a certain field yet includes more algorithms. In MALSAR, it implements 14 models with 26 of their variations to test their effectiveness.

\emph{Library for Multi-Task Learning (LibMTL).}\footnote{\url{github.com/median-research-group/LibMTL}} It is a comprehensive open-source Python library built on PyTorch for MTL. There are 104 MTL models combined by 8 architectures and 13 loss weighting strategies in LibMTL. Moreover, it guarantees unified and consistent evaluations among models on three computer vision datasets. Different from the above packages, LibMTL is well-modularized and supports customization over different components such as loss weighting strategies or architectures.

\subsection{Evaluation Metric}
\subsubsection{Single-task Metric} 
In this section, we will introduce some single-task metrics that can be used to evaluate the performance of individual tasks in a multi-task learning (MTL) setup. 

\paragraph{Regression Task Metric} \mbox{}

\emph{Root Mean Squared Error (RMSE)}: RMSE is a commonly used metric to measure the average prediction error in regression tasks. It calculates the square root of the average of squared differences between predicted and true values. RMSE gives higher weights to larger errors, making it sensitive to outliers. It is calculated as:

\[
RMSE = \sqrt{\frac{1}{n} \sum_{i=1}^{n} (\tilde{y}_{i} - y_{i})^2}
\]

where $y_{i}$ represents the true value, $\tilde{y}_{i}$ denotes the predicted value, and $n$ stands for the total number of samples.

\emph{Mean Absolute Percentage Error (MAPE)}: MAPE is a metric used to evaluate the accuracy of predictions in percentage terms. It measures the average percentage difference between predicted and true values. This metric is commonly used in business forecasting tasks. It is calculated as:
\[
MAPE = \frac{1}{n} \sum_{i=1}^{n} \left| \frac{\tilde{y}_{i} - y_{i}}{y_{i}} \right| \times 100
\]

\emph{Symmetric Mean Absolute Percentage Error (SMAPE)}: SMAPE is similar to MAPE but has the advantage of being symmetric, meaning it treats overestimations and underestimations equally. It calculates the average percentage difference between predicted and true values, considering the absolute sum of both. It is calculated as:
\[
SMAPE = \frac{100}{n} \sum_{i=1}^{n} \frac{\left| \tilde{y}_{i} - y_{i} \right|}{(\left| \tilde{y}_{i} \right| + \left| y_{i} \right|)/2}
\]

\emph{Coefficient of Determination \( R^{2} \) (R-squared)}: \( R^{2} \) is a statistical metric that represents the proportion of variance in the dependent variable (the target) that is predictable from the independent variable (the prediction). It indicates how well the predicted values fit the actual data. It is calculated as:
\[
R^{2} = 1 - \frac{\sum_{i=1}^{n} (\tilde{y}_{i} - y_{i})^2}{\sum_{i=1}^{n} (y_{i} - \bar{y})^2}
\]
where \( \bar{y} \) is the mean of the true values \( y_{i} \).

\paragraph{Classification Task Metric} \mbox{}

\emph{Confusion Matrix}: A confusion matrix is a table that allows visualization of the performance of a classification model. It presents the number of true positive (TP), false positive (FP), true negative (TN), and false negative (FN) predictions. The confusion matrix is usually represented as follows:

\begin{equation}
\begin{array}{cc|cc}
& & \text{Predicted Positive} & \text{Predicted Negative} \\
\hline
\text{Actual Positive} & & \text{TP} & \text{FN} \\
\text{Actual Negative} & & \text{FP} & \text{TN}
\end{array}
\end{equation}

\emph{Accuracy}: Accuracy is one of the most straightforward classification metrics, representing the proportion of correctly classified instances over the total number of instances in the dataset. It is calculated as:
\[
Accuracy = \frac{\text{TP} + \text{TN}}{\text{TP} + \text{TN} + \text{FP} + \text{FN}}
\]

\emph{Precision}: Precision is a metric that measures the proportion of true positive predictions (correctly predicted positive instances) over the total number of positive predictions made by the model. It is calculated as:
\[
Precision = \frac{\text{TP}}{\text{TP} + \text{FP}}
\]

\emph{Recall (Sensitivity or True Positive Rate - TPR)}: Recall calculates the proportion of true positive predictions (correctly predicted positive instances) over the total number of actual positive instances in the dataset. It is calculated as:
\[
Recall = \frac{\text{TP}}{\text{TP} + \text{FN}}
\]

\emph{F1 Score}: The F1 score is the harmonic mean of precision and recall. It provides a balance between precision and recall and is especially useful when there is an uneven class distribution. It is calculated as:
\[
F1\_Score = \frac{2 \times \text{Precision} \times \text{Recall}}{\text{Precision} + \text{Recall}}
\]

\emph{Specificity (True Negative Rate)}: Specificity measures the proportion of true negative predictions (correctly predicted negative instances) over the total number of actual negative instances in the dataset. It is calculated as:
\[
Specificity = \frac{\text{TN}}{\text{TN} + \text{FP}}
\]

\emph{Precision-Recall Curve}: The precision-recall curve is a graphical representation of the tradeoff between precision and recall for different classification thresholds. It plots the precision on the y-axis against the recall on the x-axis as the threshold varies.

\emph{Area Under the Receiver Operating Characteristic Curve (AUC-ROC)}: AUC-ROC is a metric that evaluates the performance of a binary classification model across various discrimination thresholds. It represents the area under the ROC curve, where ROC stands for the Receiver Operating Characteristic.

\emph{Formula}: The AUC-ROC is typically computed using various threshold values to calculate the True Positive Rate (TPR) and False Positive Rate (FPR) at each threshold. The AUC-ROC is then obtained by plotting TPR against FPR and calculating the area under the curve.

\paragraph{Object Detection Task Metric} \mbox{}

\emph{Bounding Box}: In object detection, algorithms typically predict bounding boxes and class labels for objects in an image. A bounding box is represented by a set of four coordinates: \((x_{min}, y_{min}, x_{max}, y_{max})\), which define the top-left and bottom-right corners of the box.

\emph{Intersection Over Union (IoU)}: The IoU measures the overlap between the predicted bounding box \(P\) and the ground truth bounding box \(G\). It is defined as:
\[
IoU(P, G) = \frac{Area(P \cap G)}{Area(P \cup G)}
\]

\emph{True Positive (TP), False Positive (FP), and False Negative (FN)}: 
- A detection is considered a TP if the IoU with the ground truth exceeds a given threshold (typically \(0.5\)) and the class label matches.
- A detection is an FP if the IoU is below this threshold, or if there is no corresponding ground truth.
- An FN represents a ground truth box which had no detected box surpassing the IoU threshold.

\emph{Precision}:
\[
Precision = \frac{\text{TP}}{\text{TP} + \text{FP}}
\]

\emph{Recall}:
\[
Recall = \frac{\text{TP}}{\text{TP} + \text{FN}}
\]

\emph{Mean Average Precision (mAP)}: The mAP is a widely-used metric in object detection, averaging the precision values at different recall levels across all classes.

\emph{Precision-Recall Curve for Object Detection}: This curve plots precision against recall values for different IoU thresholds, offering insights into a detection model's performance.

\emph{Average Recall (AR)}: AR averages the recall values obtained at various IoU thresholds.

\paragraph{Image Segmentation Metrics} \mbox{}

\emph{Pixel Accuracy}: Pixel accuracy is a simple metric that measures the proportion of pixels that are correctly classified. For a given image or set of images, it is defined as the ratio of correctly classified pixels to the total number of pixels.

\[
PixelAccuracy = \frac{\text{Number of correctly classified pixels}}{\text{Total number of pixels}}
\]

\emph{Boundary F1 Score (BF)}: The Boundary F1 Score evaluates the accuracy of the boundaries in a segmentation task. Given predicted boundaries \(P\) and ground truth boundaries \(G\), the BF score is the F1 score (harmonic mean of precision and recall) calculated based on the detected boundary pixels.

\[
Precision = \frac{\text{Number of true positive boundary pixels}}{\text{Number of true positive boundary pixels} + \text{Number of false positive boundary pixels}}
\]

\[
Recall = \frac{\text{Number of true positive boundary pixels}}{\text{Number of true positive boundary pixels} + \text{Number of false negative boundary pixels}}
\]

\[
BF = \frac{2 \times Precision \times Recall}{Precision + Recall}
\]

\emph{Panoptic Quality (PQ)}: The Panoptic Quality metric combines segmentation (things and stuff) and detection (things only) into a single score. It is defined as:

\[
PQ = \frac{\sum (p_i \times r_i)}{N_\text{matched regions} + \frac{1}{2} \times N_\text{false positive regions} + \frac{1}{2} \times N_\text{false negative regions}}
\]

Where \(p_i\) is the precision and \(r_i\) is the recall for each matched region \(i\). $N_\text{matched regions}$ is number of matched regions. $N_\text{false positive regions}$ is number of false positive regions. $N_\text{false negative regions}$ is number of false negative regions.

\paragraph{Image Generation Metrics} \mbox{}

\emph{Peak Signal-to-Noise Ratio (PSNR)}: PSNR is a traditional quality metric used to measure the quality of a reconstructed image compared to an original image. Higher values of PSNR indicate better quality. It is defined as:

\[
PSNR = 10 \times \log_{10}\left(\frac{MAX_{I}^2}{MSE}\right)
\]

Where \(MAX_{I}\) is the maximum possible pixel value of the image (often \(255\) for an 8-bit image), and \(MSE\) is the Mean Squared Error between the original and the reconstructed image.

\emph{Structural Similarity Index Measure (SSIM)}: SSIM measures the structural similarity between two images. It provides a more perceptual-based assessment of image quality than PSNR. A value of 1 indicates the images are identical in terms of structural information.

\[
SSIM(x, y) = \frac{(2 \mu_x \mu_y + C_1)(2 \sigma_{xy} + C_2)}{(\mu_x^2 + \mu_y^2 + C_1)(\sigma_x^2 + \sigma_y^2 + C_2)}
\]

Where \(x\) and \(y\) are two images, \(\mu\) represents the mean, \(\sigma\) represents the variance, \(\sigma_{xy}\) is the covariance of \(x\) and \(y\), and \(C_1\) and \(C_2\) are constants to avoid instability when the denominator is close to zero.

\emph{Inception Score (IS)}: The Inception Score is used to evaluate the quality and diversity of generated images in GANs. A higher IS indicates both better image quality and greater diversity. It's calculated using a pre-trained Inception model.

\[
IS = \exp\left(E_{x}[\text{KL}(p(y|x)||p(y))]\right)
\]

Where \(x\) is an image, \(y\) is the label predicted by the Inception model, and \(KL\) is the Kullback-Leibler divergence.

\emph{Fréchet Inception Distance (FID)}: FID measures the similarity between the generated images and real images. It computes the Fréchet distance between two Gaussians fitted to the feature representations of the Inception network for both sets of images. Lower FID scores indicate that the two sets of images are more similar, implying better generation quality.

\[
FID = ||\mu_1 - \mu_2||^2 + \text{Tr}(\Sigma_1 + \Sigma_2 - 2(\Sigma_1\Sigma_2)^{0.5})
\]

Where \(\mu_1, \Sigma_1\) are the mean and covariance of the feature representations for real images and \(\mu_2, \Sigma_2\) are those for generated images.

\paragraph{Text Generation Metrics} \mbox{}

\emph{BLEU (Bilingual Evaluation Understudy)}: BLEU is a metric originally designed for machine translation but is also used in text generation. It measures how many n-grams in the generated text match the n-grams in the reference text(s). The score ranges between 0 and 1, with 1 being a perfect match.

\[
BLEU = \min\left(1, \frac{\text{length of generated text}}{\text{length of reference text}}\right) \times \exp\left(\sum_{n=1}^{N} w_n \log p_n\right)
\]

Where \(w_n\) are the weights for each n-gram (typically \(w_n = \frac{1}{N}\)), \(p_n\) is the precision of n-grams, and \(N\) is the maximum n-gram order.

\emph{ROUGE (Recall-Oriented Understudy for Gisting Evaluation)}: Primarily used for evaluating summary generation, ROUGE measures the overlap between the n-grams in the generated text and the reference text(s).

\[
ROUGE-N = \frac{\sum_{s \in \text{reference summaries}} \sum_{\text{n-gram} \in s} \text{Count}_{\text{match}}(\text{n-gram})}{\sum_{s \in \text{reference summaries}} \sum_{\text{n-gram} \in s} \text{Count}(\text{n-gram})}
\]

Where \(\text{Count}_{\text{match}}\) is the number of matching n-grams between the generated text and reference summary, and \(\text{Count}\) is the number of n-grams in the reference summary.

\emph{Perplexity}: Used for evaluating language models, perplexity measures how well the probability distribution predicted by the model aligns with the true distribution of the words in the text. Lower perplexity values indicate better model performance.

\[
\text{Perplexity} = \exp\left(-\frac{1}{N} \sum_{i=1}^{N} \log p(w_i)\right)
\]

Where \(N\) is the total number of words, and \(p(w_i)\) is the model's predicted probability for word \(w_i\).

\emph{Self-BLEU}: A metric that evaluates the diversity of generated texts. It computes the BLEU score between each generated text and all other generated texts. Lower Self-BLEU scores indicate higher diversity.

\emph{Distinct-N}: Measures the diversity of generated content by computing the ratio of unique n-grams to the total number of generated n-grams. Higher values of Distinct-N indicate greater diversity.

\[
Distinct-N = \frac{\text{Number of unique n-grams}}{\text{Total number of generated n-grams}}
\]

\subsubsection{Multi-task Metric} In this section, we denote by $M^{t}_{MTL}$ and $M^{t}_{STL}$ the STL measurements of MTL method and STL baseline for the $t$-th task, respectively. $M^{t}_{STL}\downarrow$ indicates that a lower value has better performance for the measurement $M_{STL}^t$, and vice versa.

\paragraph{Delta~\citep{dong2015multi}} The performance of MTL method can be simply defined as the difference of the STL measurement between the STL baseline and MTL method:
\begin{equation}
    \text{Delta} = M^{t}_{MTL} - M^{t}_{STL}, t = 1, \cdots, T,
\end{equation}
where $M$ was set to be BLEU-4~\citep{papineni2002bleu} in \citet{dong2015multi},.

\paragraph{MTL gain~\citep{tang2020progressive}} To evaluate the benefit of MTL method over the STL baseline on the $t$-th task, MTL gain is computed as below:
\begin{equation}
    MTL~gain = {(-1)}^{\mathds{1}\{M_{STL}^{t}\downarrow\}}(M_{MTL}^t - M^t_{STL}), t = 1, \cdots, T,
\end{equation}
which is consistent with any positive or negative measurements (c.f. Delta~\citep{dong2015multi}).

\paragraph{$\Delta_m$~\citep{maninis2019attentive}} The performance of MTL method can be quantified by calculating the average per-task drop with respect to the single-task baseline using STL measurements:
\begin{equation}
    \Delta_m = \frac{1}{T}\sum\nolimits_{t=1}^T{(-1)}^{\mathds{1}\{M^{t}_{Baseline}\downarrow\}}(M^{t}_{MTL} - M^{t}_{Baseline})/M^{t}_{Baseline},
\end{equation}

\paragraph{$\Delta_p$~\citep{lin2022reasonable}} Given that many single tasks can be measured by several metrics, e.g. semantic segmentation measured by mIoU and pixacc, by following $\Delta_m$~\citep{maninis2019attentive}, the average of the relative improvement over the MTL method on each metric of each task could be formulated as the MTL performance measurement:
\begin{equation}
    \Delta_p = \frac{1}{T}\sum\nolimits_{t=1}^T \frac{1}{M_t}\sum\nolimits_{m=1}^{M_t} {(-1)}^{\mathds{1}\{M^{t, m}_{Baseline}\uparrow\}}(M^{t,m}_{MTL} - M^{t, m}_{Baseline})/M^{t, m}_{Baseline},
\end{equation}
where $M_t$ is the number of metrics used for the $t$-th task. $M^{t, m}_{Baseline}$ denotes the $m$-th performance measurement of the baseline method, e.g. the STL or vanilla MTL method, for the $t$-th task.



%% file: table_files/dataset.tex
\begin{table*}[t]
    \centering
    \scriptsize
    \caption{Summary of common datasets used in MTL.}
    \scalebox{0.5}{
    \begin{threeparttable}
    \midsepremove
    \begin{tabular}{lllllllll}
    \toprule
    \rowcolor{gray!40}
        Dataset  & Source & Year & Modality &  Task & Synopsis & \#Task & \#Sample & Availability\\
        
    \midrule
        School Data & ILEA & \citeyear{mortimore1988school} & Table & Regression & Predicting student exam scores based on 27 school features. & 139 & 15,362 & \href{http://www.bristol.ac.uk/cmm/learning}{\textcolor{citegreen}{Official}} \\
        
    \midrule
    \rowcolor{gray!20}
        SARCOS Data & Humanoid Robotics & 2000 & Table & Regression & Estimate inverse dynamics model. & 7 & 44,484/4449 & \href{http://gaussianprocess.org/gpml/data}{\textcolor{citegreen}{Official}} \\
        
    \midrule
        Computer Survey Data  & Survey & \citeyear{lenk1996hierarchical} & Table & Regression
     &Likelihood of purchasing personal computers. & 179 & - & - \\
     
     \midrule
     \rowcolor{gray!20}
        Climate Dataset & Sensor network & 2017-now  & Table & Regression & Real-time climate data collected from four climate stations. & 7 & - & \href{https://www.cambermet.co.uk/}{\textcolor{citegreen}{Official}} \\
        
    \midrule
        20 Newsgroups  & Netnews articles & \citeyear{Lang95}  & Text & Classification & Hierarchical text classification. & 20 & 19,000 & \href{http://qwone.com/~jason/20Newsgroups}{\textcolor{citegreen}{Official}} \\
        
    \midrule
    \rowcolor{gray!20}
        Reuters-21578 Collection  & Reuters & 1996  & Text & Classification & Reuters news documents with hierarchical categories. & 90 & 21,578 & \href{http://www.daviddlewis.com/resources/testcollections/reuters21578/}{\textcolor{citegreen}{Official}}\\
        
    \midrule
        MultiMNIST Dataset & MNIST & \citeyear{sabour2017dynamic} & Image & Classification & Classify the digits on the different positions. & 2 & - & \href{http://www.cs.toronto.edu/~tijmen/affNIST}{\textcolor{citegreen}{Official}}\\
        
    \midrule
    \rowcolor{gray!20}
        ImageCLEF-2014 & Caltech, ImageNet, Pascal, Bing & 2014  & Image & Classification & Benchmark dataset for domain adaptation. & 4 & 2,400 & \href{https://www.imageclef.org/2014/adaptation}{\textcolor{citegreen}{Official}} \\
        
    \midrule
        Office-Caltech Dataset & Office, Caltech & \citeyear{gong2012geodesic} & Image & Classification & Benchmark dataset for the annotation and retrieval of images. & 4 & 2,533 & \href{https://www.v7labs.com/open-datasets/office-caltech-10}{\textcolor{citegreen}{Official}}  \\
        
    \midrule
    \rowcolor{gray!20}
        Office-31 Dataset  & Amazon, DSLR, Webcam & \citeyear{saenko2010adapting} & Image & Classification & Objects commonly encountered in office settings. & 3 & 4,110 & \href{https://opendatalab.com/Office-31}{\textcolor{citegreen}{Official}} \\
        
    \midrule
        Office-Home Dataset & Office & \citeyear{venkateswara2017deep}  & Image & Classification & Object recognition and domain adaptation in the era of deep learning. & 4 & 15,588 & \href{https://www.hemanthdv.org/officeHomeDataset.html}{\textcolor{citegreen}{Official}} \\
        
    \midrule
    \rowcolor{gray!20}
        DomainNet Dataset & UDA & \citeyear{peng2019moment} & Image & Classification & Multi-source unsupervised domain adaptation research & 6 & 600,000 & \href{http://ai.bu.edu/M3SDA}{\textcolor{citegreen}{Official}} \\
        
    \midrule
        EMMa Dataset  & Amazon & \citeyear{standley2023extensible} & Image, Text & Classification & Amazon product listings for category prediction & - & 2,800,000 & \href{https://emma.stanford.edu/}{\textcolor{citegreen}{Official}}\\
        
    \midrule
    \rowcolor{gray!20}
        SYNTHIA Dataset  & European Union & \citeyear{ros2016synthia} & Image & Classification &  A synthetic dataset for semantic segmentation. & - & 13,400 & \href{https://synthia-dataset.net/}{\textcolor{citegreen}{Official}}\\
        
    \midrule
        SVHN Dataset  & Stanford & \citeyear{yang2021few} & Image & Classification &  A digit classification benchmark dataset. & - & 600,000 & \href{http://ufldl.stanford.edu/housenumbers/}{\textcolor{citegreen}{Official}}\\
        
    \midrule
    \rowcolor{gray!20}
        CelebA Dataset  & MMLAB & \citeyear{liu2018large} & Image & Classification &  A large-scale face attributes dataset. & 40 & 200,000 & \href{https://mmlab.ie.cuhk.edu.hk/projects/CelebA.html}{\textcolor{citegreen}{Official}}\\
        
    \midrule
        CityScapes Dataset  & Daimler AG & \citeyear{cordts2016cityscapes} & Image &Dense prediction & Semantic urban scene understanding & - & 5,000 & \href{https://www.cityscapes-dataset.com/}{\textcolor{citegreen}{Official}}\\
        
    \midrule
    \rowcolor{gray!20}
       NYU-Depth Dataset V2  & New York University & \citeyear{silberman2012indoor} & Image &Dense prediction  & Indoor scene understanding with per-pixel labels & 3 & 35,064 & \href{https://cs.nyu.edu/~silberman/datasets/nyu_depth_v2.html}{\textcolor{citegreen}{Official}}\\
       
    \midrule
        PASCAL VOC Project  & University of Oxford & \citeyear{everingham2010pascal} & Image &Dense prediction & Object recognition with multiple tasks & - & - & \href{http://host.robots.ox.ac.uk/pascal/VOC}{\textcolor{citegreen}{Official}}\\
        
    \midrule
    \rowcolor{gray!20}
        Taskonomy Dataset  & Standard & \citeyear{zamir2018taskonomy} & Image &Dense prediction & Diverse dataset with 26 tasks for task transfer learning & 26 & 4,000,000 & \href{http://taskonomy.stanford.edu/}{\textcolor{citegreen}{Official}}\\
        
    \midrule
        STREET  & Amazon & \citeyear{ribeiro2023street} & Text &Reasoning & The multi-task structured reasoning and explanation benchmark & - & - & -\\
        
    \midrule
    \rowcolor{gray!20}
        VKITTI2 Dataset & Naver & \citeyear{cabon2020virtual} & Video & Segmentation & A video dataset which is automatically labeled with ground truth & 5 & -  & \href{https://europe.naverlabs.com/research/computer-vision/proxy-virtual-worlds-vkitti-2/}{\textcolor{citegreen}{Official}}\\
        
    \midrule
        XTREME  & Carnegie Mellon & \citeyear{hu2020xtreme} & Text & Translation, QA & A multilingual benchmark for evaluating cross-lingual generalisation & 9 & 400,000 & -\\

    \rowcolor{gray!20}
    \midrule
        Deepfashion Dataset  &  Shopping Websites & \citeyear{liu2016deepfashion} & Image & Classification & A large-scale clothes dataset with comprehensive annotations & 2 & 800,000 & \href{https://mmlab.ie.cuhk.edu.hk/projects/DeepFashion.html}{\textcolor{citegreen}{Official}}\\
        
    \midrule
        ACE05 Dataset  &  News & 2005 & Text & Classification & A large corpus with annotated entities, relations and events & 3 & 52,615 & \href{https://catalog.ldc.upenn.edu/LDC2006T06}{\textcolor{citegreen}{Official}}\\

    \rowcolor{gray!20}
    \midrule
        ATIS Dataset  &  Airline & \citeyear{hemphill-etal-1990-atis} & Text & Classification & A dataset with 17 unique intent categories. & 3 & 5,871 & \href{https://github.com/howl-anderson/ATIS_dataset}{\textcolor{citegreen}{Official}}\\
        
    \bottomrule
    \end{tabular}
    \label{tab:dataset}
    \end{threeparttable}
    }
\end{table*}


%% file: table_files/software_covers.tex
\begin{table*}[t]
    \centering
    \scriptsize
    \caption{Summary of library for MTL.}
    \scalebox{0.65}{
    \begin{threeparttable}
    \midsepremove
    \begin{tabular}{p{0.6in}p{0.4in}p{7.5in}}
    \toprule
    \rowcolor{gray!40}
        Library & Language & Supported Methods\\
    \toprule 
    RMTL & R & Sparse structure learning~\citep{tibshirani1996regression}, multi-task feature selection~\citep{obozinski2006multi}, low rank MTL~\citep{ji2009accelerated, pong2010trace}, graph-based regularised MTL~\citep{widmer2010leveraging}, multi-task clustering~\citep{gu2009learning}\\

\midrule
\rowcolor{gray!20}
    MALSAR & Matlab & Sparse structure learning~\citep{tibshirani1996regression}, regularized MTL~\citep{evgeniou2004regularized}, multi-task feature selection~\citep{obozinski2006multi}, dirty block-sparse model~\citep{jalali2010dirty}, low rank MTL~\citep{ji2009accelerated, pong2010trace}, convex ASO~\citep{chen2009convex}, sparse \& low rank MTL~\citep{chen2012learning}, clustered MTL~\citep{zhou2011clustered}, robust MTL~\citep{chen2011integrating}, robust multi-task feature learning~\citep{gong2012robust}, Temporal group Lasso~\citep{zhou2011multi}, convex fused sparse group Lasso~\citep{zhou2012modeling}, incomplete multi-source feature learning~\citep{yuan2012multi}, multi-stage multi-task feature learning~\citep{gong2012multi}, multi-task clustering~\citep{gu2009learning}\\

\midrule
    LibMTL & Python & Cross-stitch~\citep{misra2016cross}, GradNorm~\citep{chen2018gradnorm}, Uncertainty Weighting~\citep{kendall2018multi}, MGDA-MTL~\citep{sener2018multi}, MMoE~\citep{ma2018modeling}, MultiNet++~\citep{chennupati2019multinet++}, LTB~\citep{guo2020learning}, MTAN \& DWA~\citep{liu2019end}, PCGrad~\citep{yu2020gradient}, GradDrop~\citep{chen2020just}, CGC \& PLE~\citep{tang2020progressive}, IMTL~\citep{liu2021towards}, GradVac~\citep{wang2021gradient}, CAGrad~\citep{liu2021conflictaverse}, DSelect-k~\citep{hazimeh2021dselect}, RLW \& RGW~\citep{lin2022reasonable}, Nash-MTL~\citep{navon2022multi}\\

    \bottomrule
    \end{tabular}
    \label{tab:software}
    \end{threeparttable}
    }
\end{table*}


%% file: 05_discussion.tex
\section{Discussion}
\label{sec:discuss}
In this section, we will discuss several key questions and explore future directions concerning the theories and applications of MTL. 

\emph{Multi-Task Pretraining.} While MTL has demonstrated its remarkable success in real-world scenarios, delving into its underlying mechanisms becomes even more imperative in the era of PFMs. When data in the wild are pre-trained using scalable foundation models to exhibit modality- and task-agnostic characteristics (\textbf{\S}~\ref{fm-era}), an essential question arises: What proportions of different tasks in the pretraining phase can yield best task-generalizable performance?

\emph{Competitive or Collaborative?} While many proposed MTL methods offer benefits to each task under their specific settings, competitive tasks continue to exist in real-world scenarios. Distinguishing between them without human priors before employing MTL remains a challenge. Task prior sharing (\textbf{\S}~\ref{subsec:prior}) and task clustering methods (\textbf{\S}~\ref{sec:clustering}) can play a crucial role, as they can help to know task relations and do not conflict with other multi-task representation learning methods.

\emph{Blessed or Cursed by Large Number of Tasks?} While MTL with a small number of tasks has been proven to outperform STL, and MTL with a large number of tasks has been demonstrated to be learnable, the underlying relationships between these models and the number of tasks raise intriguing questions. The introduction of a new task typically introduces both knowledge and noise to existing tasks. If all tasks are trained equally, (e.g., LLMs), without any selective mechanisms, what are the outcomes for the final learned model concerning each individual task?

\emph{MTL for Other Things.} The pursuit of performance through MTL has been shown to have potential drawbacks in terms of fairness (\textbf{\S}~\ref{subsec:fairness}), security and privacy (\textbf{\S}~\ref{subsec:mtlsp}). However, MTL can also contribute to learning fairness or enhancing security and privacy for involved tasks by incorporating novel metrics. In certain situations, a favorable trade-off between these considerations may exist.

\emph{Illuminating the Unseen with MTL:} To underscore the impactful insights provided by MTL, consider a compelling example where MTL results significantly advanced our understanding of a complex problem. In a medical imaging scenario, MTL was applied to simultaneously predict multiple health-related outcomes, such as disease progression, severity, and patient response to treatment. Unlike STL approaches, MTL unveiled intricate dependencies and interactions between these outcomes, showcasing that certain imaging features played dual roles in influencing multiple health aspects. This holistic perspective allowed researchers to identify subtle correlations and nuanced patterns that were previously obscured by individual task-centric analyses. MTL, in this case, not only improved predictive accuracy but also unraveled hidden intricacies within the data, providing a richer and more comprehensive understanding of the medical conditions under investigation. This example exemplifies how MTL can reveal intricate relationships and enhance interpretability beyond the capabilities of traditional STL methods.

%% file: 06_conclusion.tex
\section{Conclusion}
\label{sec:conclud}
In this survey, we introduce the MTL from rough to precise and review methodologies covering traditional ML, DL, and PFMs era. First, we present the background of MTL, covering the scope, formal definition, comparisons with other paradigms, and motivations behind MTL. After that, we explore how MTL works well and provide the reasons to explain its intrinsic mechanisms. We formalize and illustrate MTL in a framework and further expand the methodology overview based on this MTL framework. Specifically, we summarize the sparse structure learning, feature learning, low-rank learning, and decomposition methods in the traditional learning era. We categorize MTL in DL into feature sharing, task
balancing, and neural architecture search methods; recent task- and modality-agnostic foundation models are also discussed as they can learn universal comprehensiveness across tasks with different data modalities. 

To sum it up, MTL methods in the traditional learning era prefer to "drop" distinctive (task-specific) features to seek consensus. For instance, the classical $\ell_{2,1}$ norm can realize grouped feature selection across tasks to exploit common features that are effective and efficient for joint performance enhancement. Another example is the low-rank learning methods that try to explore common underlying representations via imposing low-dimensional properties for essential factors, where a small set of factors is supposed to govern multiple tasks. However, when it comes to DL models, powerful computational resources make it possible to handle all the features from different tasks, and its hierarchical structure with multiple layers can learn feature interaction across tasks at various levels of abstraction. Accordingly, MTL has been dominated by feature fusing and task-balancing techniques via introducing learnable parameters in the past decade. These learnable parameters play a crucial role in cross-task communication and eavesdropping during the combined training. However, the explanations and mechanisms of these complicated interactions inside the networks still remain poorly understood. More recently, unified foundation models have shown promising results for MTL in real-world scenarios, as data with versatile modalities can be trained simultaneously to learn universal and effective comprehensiveness. 

Overall, we hope this paper provides an extensive review of the research community for a comprehensive understanding of research advances, current and future challenges, and opportunities or prospects for the MTL.

%% file: 07_contribution.tex
\subsection*{Acknowledgments}
This paper is the result of a collaborative effort, with each author contributing significantly to various aspects:
\begin{itemize}
    \item Yutong Dai orchestrated two critical optimizations in MTL, detailed in \textbf{\S}~\ref{subsec:scalar} and \textbf{\S}~\ref{subsec:multi-obj-opt}.
    \item Xiaokang Liu contributed by writing and organizing the section on MTL via low-rank factorization (\textbf{\S}~\ref{subsec:low-rank}).
    \item Jin Huang was responsible for the figure and layout designs, ensuring visual clarity and coherence.
    \item Yishan Shen focused on developing the MTL through prior sharing, as outlined in \textbf{\S}~\ref{subsec:prior}.
    \item Ke Zhang was instrumental in writing and structuring the Graph-based MTL section (\textbf{\S}~\ref{sec:graph}).
    \item Rong Zhou authored the STL metrics section and played a key role in organizing parts of the datasets.\item Eashan Aahikarla delved deeply into the distribution shifts that occur in MTL (\textbf{\S}~\ref{sec:distri_shift}).
    \item Wenxuan Ye took charge of organizing the GitHub website for this project, facilitating broader access and collaboration.
    \item Yixin Liu was pivotal in developing the security and privacy section for the MTL framework, as detailed in \textbf{\S}~\ref{subsec:mtlsp}.
    \item Zhaoming Kong and Kai Zhang were actively involved in discussions about the scope and structure of this survey.
    \item Jun Yu initiated this project in 2021 and managed the contents not specifically mentioned above, providing overall leadership and direction.
    \item Prof. Moore, Prof. Davison, Prof. Namboodiri and Prof. Yin contributed significantly by offering feedback and suggestions during the paper's development.
    \item Prof. Chen finalizes the paper structure, edited different versions of the manuscript, and tailored the materials towards the audiences of the research community.
\end{itemize}
All authors above actively participated in the proofreading and discussion stages of this paper. We extend our sincere gratitude to all for their valuable contributions and collective effort in bringing this research to this final version.